\documentclass[acmtog, authorversion, nonacm]{acmart}

\AtBeginDocument{%
  \providecommand\BibTeX{{%
    \normalfont B\kern-0.5em{\scshape i\kern-0.25em b}\kern-0.8em\TeX}}}

\acmJournal{FACMP}

\usepackage{graphicx}
\usepackage{booktabs}

\usepackage[ruled]{algorithm2e} %
\usepackage{float}
\usepackage{caption}
\usepackage{subcaption}
\usepackage{placeins}

\SetAlFnt{\small}
\SetAlCapFnt{\small}
\SetAlCapNameFnt{\small}
\SetAlCapHSkip{0pt}

\acmJournal{TOG}

\usepackage{amsmath}
\usepackage{comment}
\usepackage{multirow,bigdelim}
\usepackage{lipsum}
\usepackage{array}
\usepackage{wrapfig}

\usepackage{csvsimple}
\usepackage{adjustbox}
\usepackage[percent]{overpic}
\usepackage{makecell}
\usepackage{float}
\usepackage[normalem]{ulem}
\usepackage{blindtext}
\usepackage{xcolor}
\usepackage{soul}
\usepackage{cleveref}
\usepackage{amsfonts}
\usepackage{subcaption}

\makeatletter
\newcommand{\settitle}{\@maketitle}
\makeatother

\newcolumntype{C}[1]{>{\centering\let\newline\\\arraybackslash\hspace{0pt}}m{#1}}

\newif\ifdraft
\drafttrue

\ifdraft
\definecolor{darkpink}{rgb}{0.561, 0.282, 0.427}
\definecolor{darkturquoise}{rgb}{0., 0.81, 0.822}

\newcommand{\dcc}[1]{{\color{red}[\textbf{DC:} #1]}}
\newcommand{\rgc}[1]{{\color{purple}[\textbf{RG:} #1]}}
\newcommand{\opc}[1]{{\color{blue}[\textbf{OP:} #1]}}

\newcommand{\abc}[1]{{\color{green}[\textbf{AB:} #1]}}

\newcommand{\drop}[1]{}

\else
\newcommand{\dcc}[1]{}
\newcommand{\rgc}[1]{}
\newcommand{\opc}[1]{}
\newcommand{\gcc}[1]{}
\newcommand{\hmc}[1]{}
\newcommand{\abc}[1]{}

\fi

\def\naive{na\"{\i}ve\xspace}

\makeatletter
\DeclareRobustCommand\onedot{\futurelet\@let@token\@onedot}
\def\@onedot{\ifx\@let@token.\else.\null\fi\xspace}

\def\eg{\emph{e.g}\onedot}

\def\ie{\emph{i.e}\onedot}

\makeatother

\usepackage[bottom]{footmisc}
\usepackage{arydshln}

\makeatletter
\def\blfootnote{\xdef\@thefnmark{}\@footnotetext}
\makeatother

\begin{document}

\title{LCM-Lookahead for Encoder-based Text-to-Image Personalization}

\author{Rinon Gal}
\affiliation{%
 \institution{Tel Aviv University, NVIDIA}
 \city{Tel Aviv}
 \country{Israel}}
\authornote{
Denotes equal contribution
}
 \author{Or Lichter*}
\affiliation{%
 \institution{Tel Aviv University}
 \city{Tel Aviv}
 \country{Israel}}

\author{Elad Richardson*}
\affiliation{%
 \institution{Tel Aviv University}
 \city{Tel Aviv}
 \country{Israel}}

 \author{Or Patashnik}
\affiliation{%
 \institution{Tel Aviv University}
 \city{Tel Aviv}
 \country{Israel}}

\author{Amit H. Bermano}
\affiliation{%
 \institution{Tel Aviv University}
 \city{Tel Aviv}
 \country{Israel}}
 
\author{Gal Chechik}
\affiliation{%
 \institution{NVIDIA}
 \city{Tel Aviv}
 \country{Israel}}
 
\author{Daniel Cohen-Or}
\affiliation{%
 \institution{Tel Aviv University}
 \city{Tel Aviv}
 \country{Israel}}

\begin{teaserfigure}
    \centering
     \includegraphics[width=0.95\linewidth]{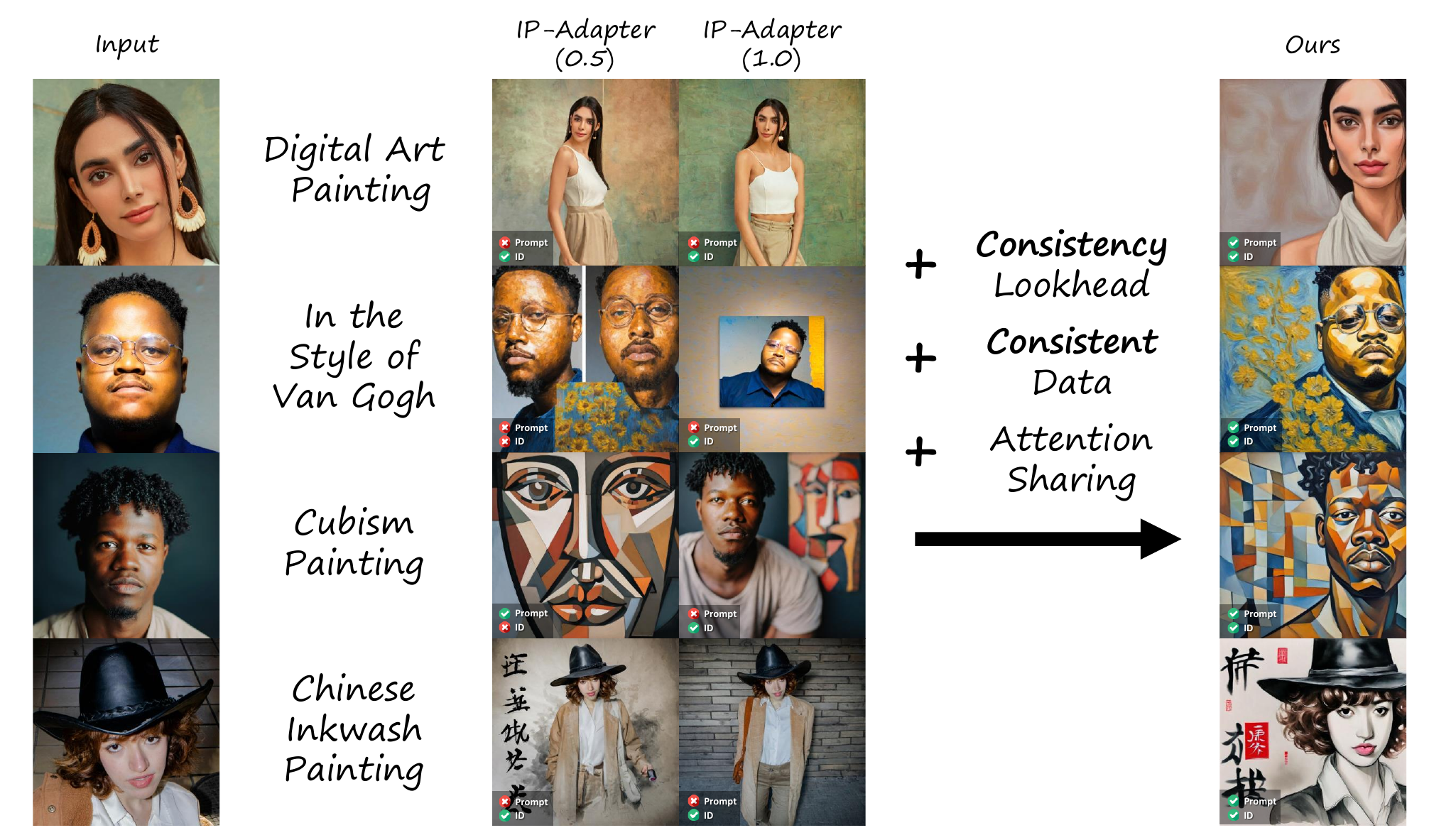}\vspace{-10pt}
    \caption{We introduce a novel LCM-based~\cite{luo2023latent} lookhead mechanism to apply image-space losses to personalization encoder training. These are coupled with consistent data generation and attention sharing techniques to tune existing backbones and improve identity preservation and prompt alignment.}
    \label{fig:teaser}
\end{teaserfigure}

\begin{abstract}
  Recent advancements in diffusion models have introduced fast sampling methods that can effectively produce high-quality images in just one or a few denoising steps. Interestingly, when these are distilled from existing diffusion models, they often maintain alignment with the original model, retaining similar outputs for similar prompts and seeds. These properties present opportunities to leverage fast sampling methods as a shortcut-mechanism, using them to create a preview of denoised outputs through which we can backpropagate image-space losses. In this work, we explore the potential of using such shortcut-mechanisms to guide the personalization of text-to-image models to specific facial identities. We focus on encoder-based personalization approaches, and demonstrate that by tuning them with a lookahead identity loss, we can achieve higher identity fidelity, without sacrificing layout diversity or prompt alignment. We further explore the use of attention sharing mechanisms and consistent data generation for the task of personalization, and find that encoder training can benefit from both.
\keywords{Personalization \and Face Generation \and Consistency Models}

\end{abstract}

\maketitle

\section{Introduction}
\label{sec:intro}
Text-to-image personalization~\cite{gal2022textual,ruiz2022dreambooth} methods enable users to tailor pretrained generative models to their own, \textit{personal} data. Commonly, such methods focus on human data~\cite{gal2023encoder,valevski2023face0,wang2024instantid,yuan2023celebbasis,xiao2023fastcomposer,ruiz2023hyperdreambooth}, where users aim to create novel images of specific individuals which were unseen by the pretrained model. Early works proposed to tackle this task by teaching a model new words that describe the user-provided subjects. They do so by optimizing novel word-embeddings~\cite{gal2022textual,voynov2023p+}, or by fine-tuning the generative model itself~\cite{ruiz2022dreambooth,simoLoRA2023}. However, such approaches require significant per-subject optimization, leading to lengthy personalization times and large compute requirements. More recent lines of work propose to personalize the model using an encoder -- a neural network trained to condition the generative model on user-provided images~\cite{gal2023encoder,Wei_2023_ICCV,shi2023instantbooth,ye2023ipadapter}. These methods can enable inference-time personalization, but they often struggle to maintain a subject's identity, or face difficulties in adapting it to novel styles. 

One manner in which encoder-based methods try to bridge these gaps is by leveraging a pretrained face recognition network as a feature extraction backbone. The intuition here is that the pretraining objective of such networks drives them to encode fine identity details which can later be exploited by the encoder. However, this approach overlooks the training loss itself, which is still driven by only an L2 noise-prediction objective. In the realm of Generative Adversarial Networks (GANs, \cite{goodfellow2014generative}), it was shown that inversion~\cite{bermano2022state,xia2021gan} can be significantly improved by incorporating additional losses that better align with human perception~\cite{richardson2020encoding,nitzan2020face}, \eg, an identity loss. Applying such image-space losses to the personalization process could be beneficial here as well. However, the diffusion training process works on noisy image samples from intermediate diffusion time steps, and it is not clear how these should be passed into a perceptual loss which expects clean, realistic images.

\begin{figure}
    \centering
    \setlength{\tabcolsep}{1.5pt}
    \begin{tabular}{c}
    \begin{tabular}{cccccc}
    & \multicolumn{5}{c}{SDXL} \\
    
    \raisebox{0.053\linewidth}{\rotatebox[origin=t]{90}{\fontsize{8pt}{8pt}\selectfont\begin{tabular}{c@{}c@{}c@{}c@{}} LCM \end{tabular}}} &
        \includegraphics[width=0.08\textwidth,height=0.08\textwidth]{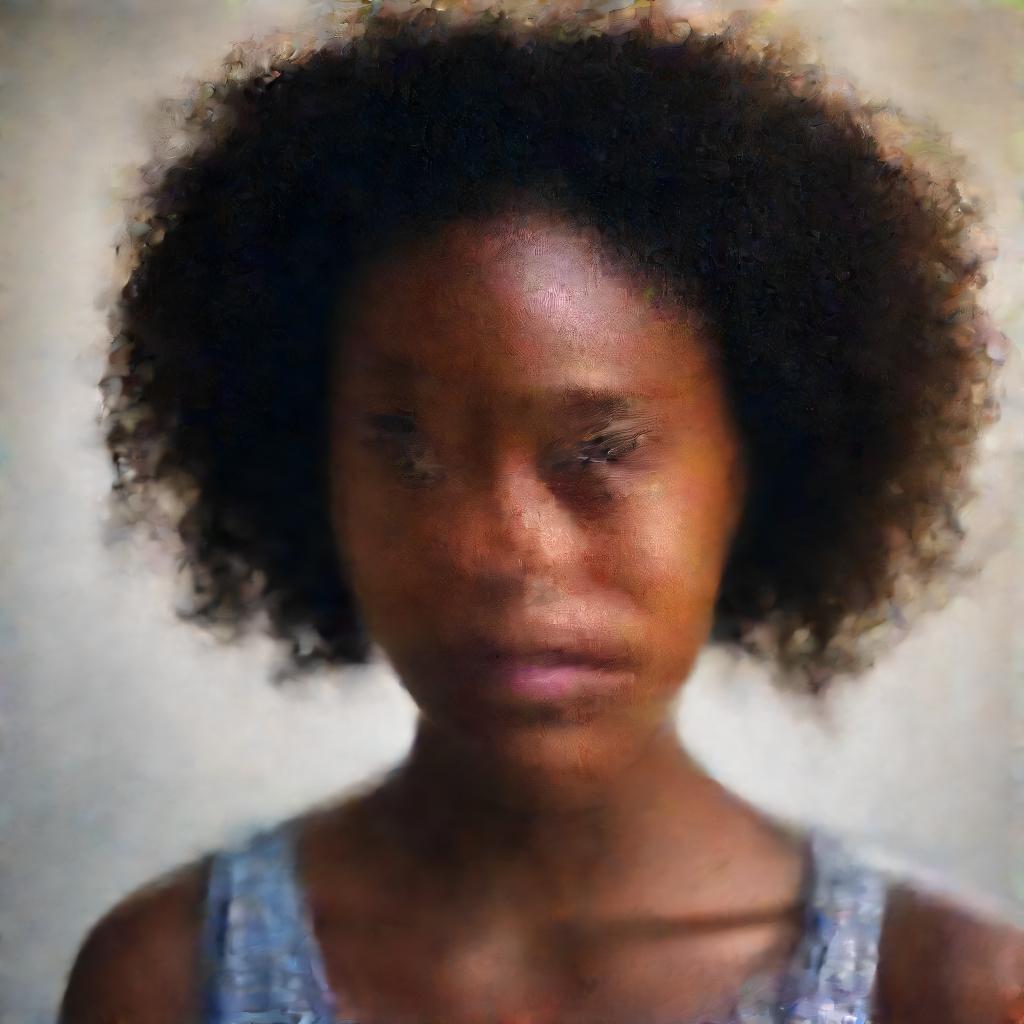} & \includegraphics[width=0.08\textwidth,height=0.08\textwidth]{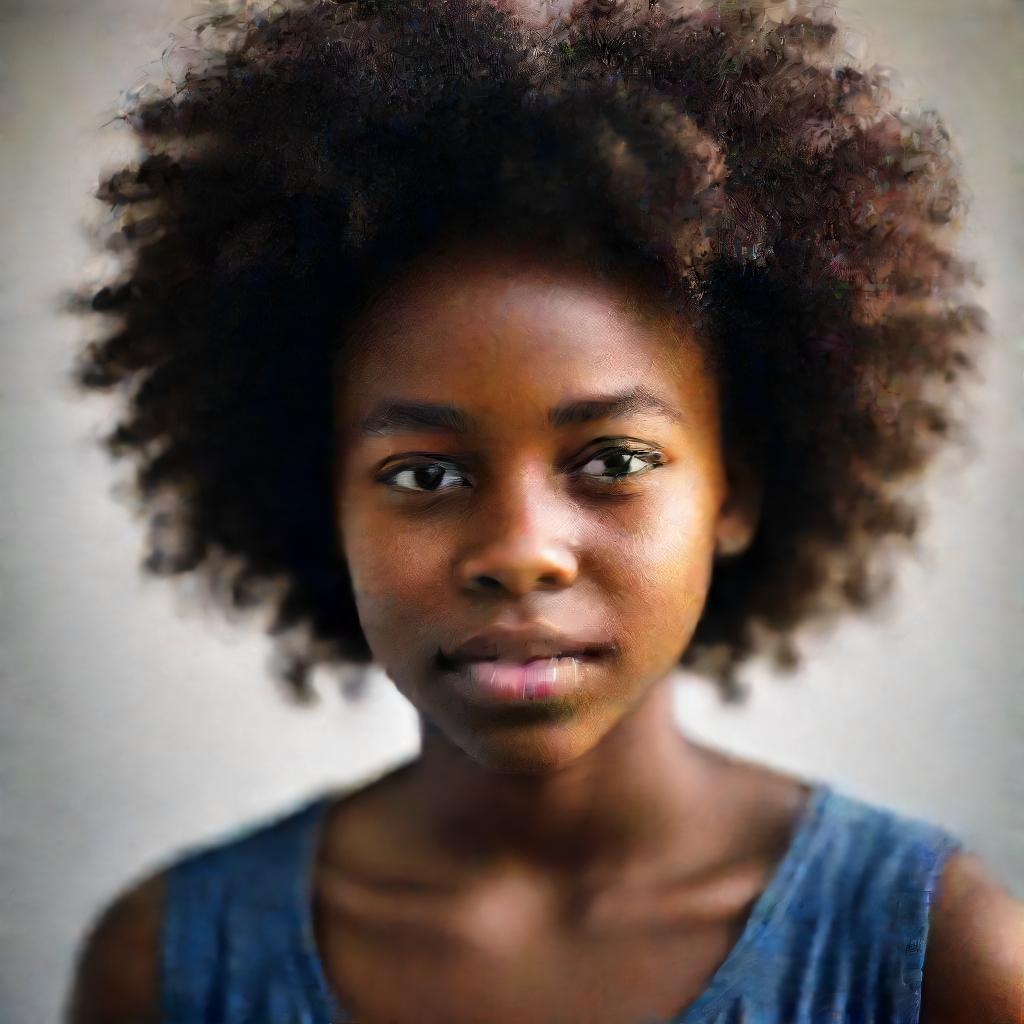} & \includegraphics[width=0.08\textwidth,height=0.08\textwidth]{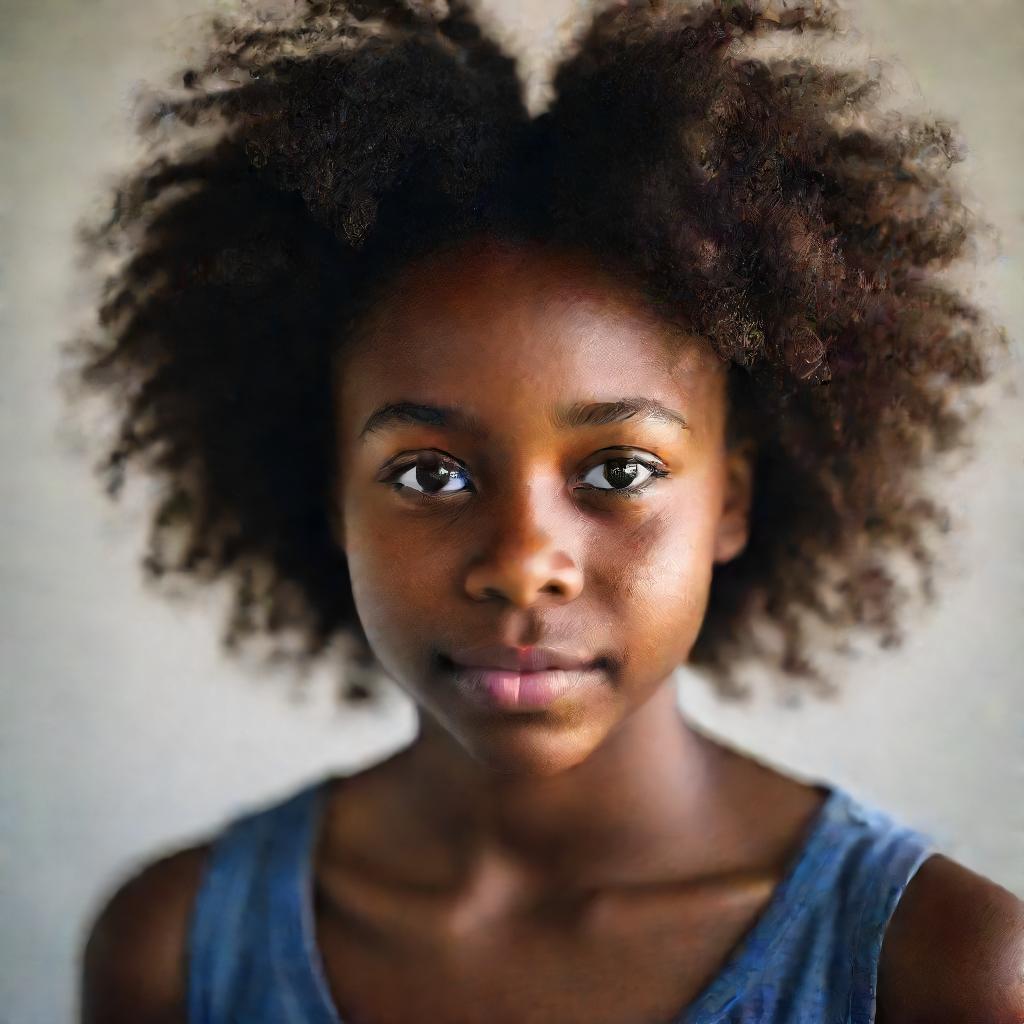} & \includegraphics[width=0.08\textwidth,height=0.08\textwidth]{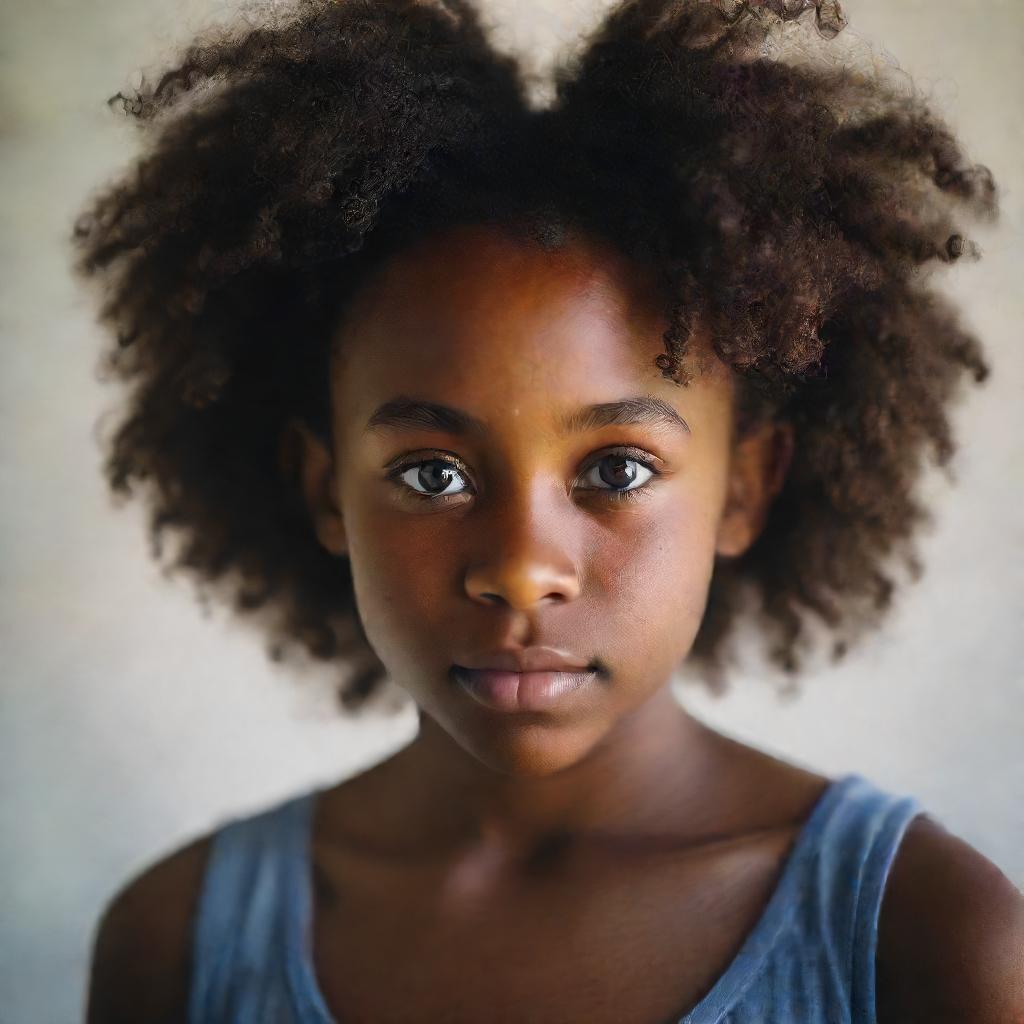} & \includegraphics[width=0.08\textwidth,height=0.08\textwidth]{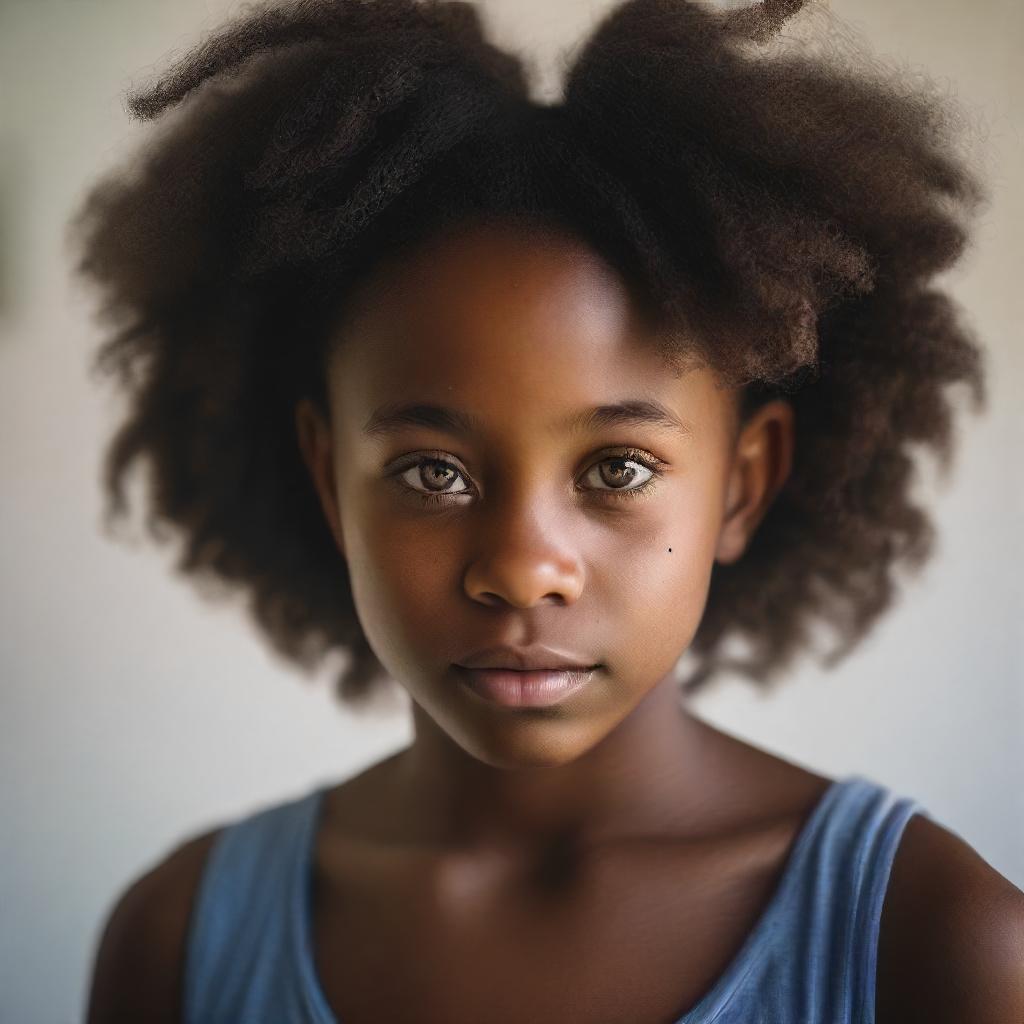}
        \\
        
    \raisebox{0.055\linewidth}{\rotatebox[origin=t]{90}{\fontsize{8pt}{8pt}\selectfont\begin{tabular}{c@{}c@{}c@{}c@{}} DDPM \end{tabular}}} & 
        \includegraphics[width=0.08\textwidth,height=0.08\textwidth]{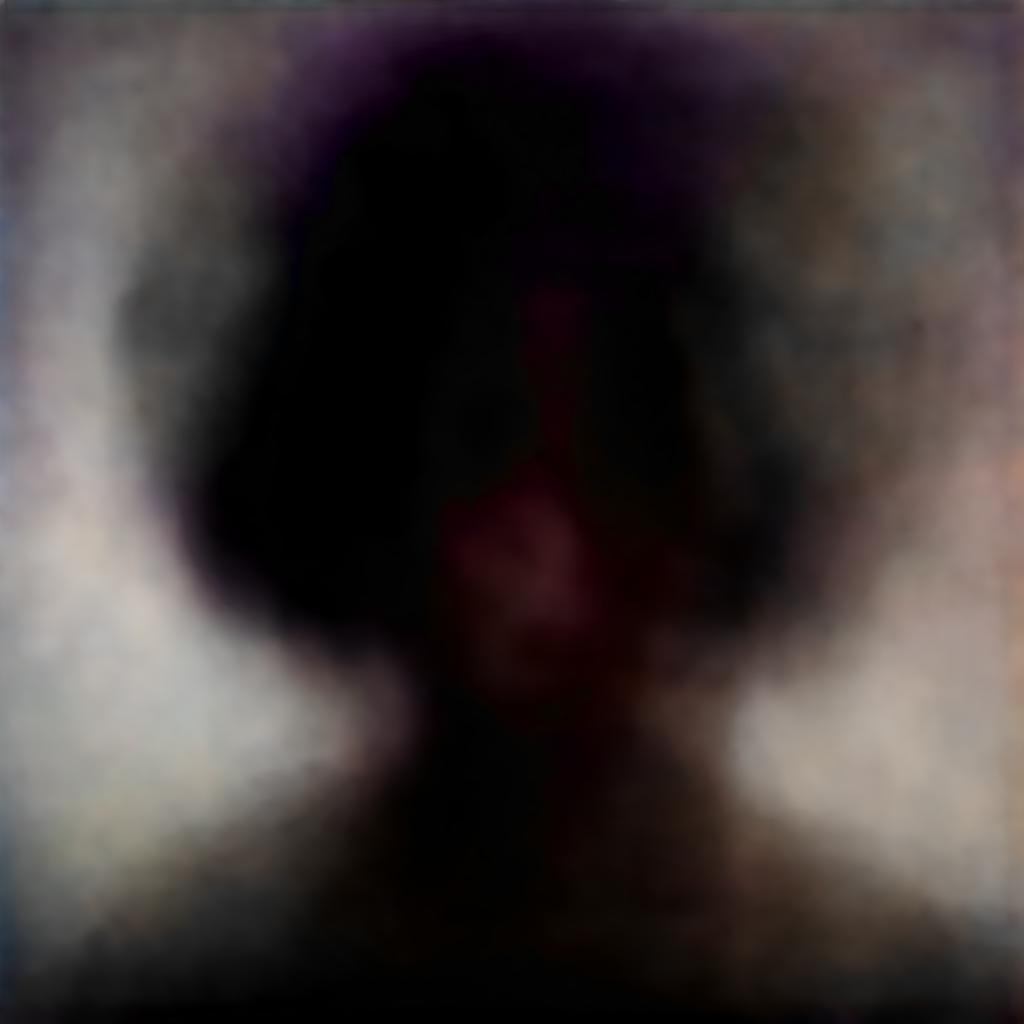} & \includegraphics[width=0.08\textwidth,height=0.08\textwidth]{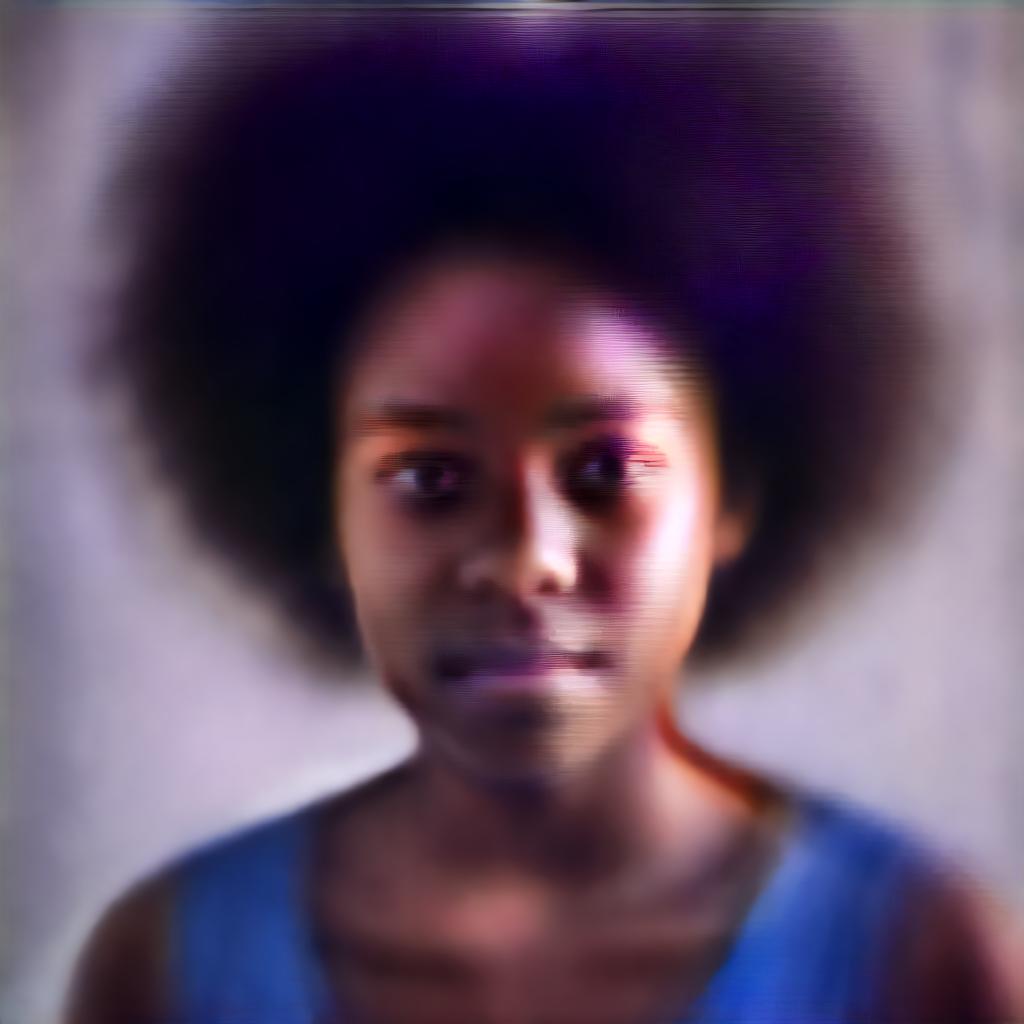} & \includegraphics[width=0.08\textwidth,height=0.08\textwidth]{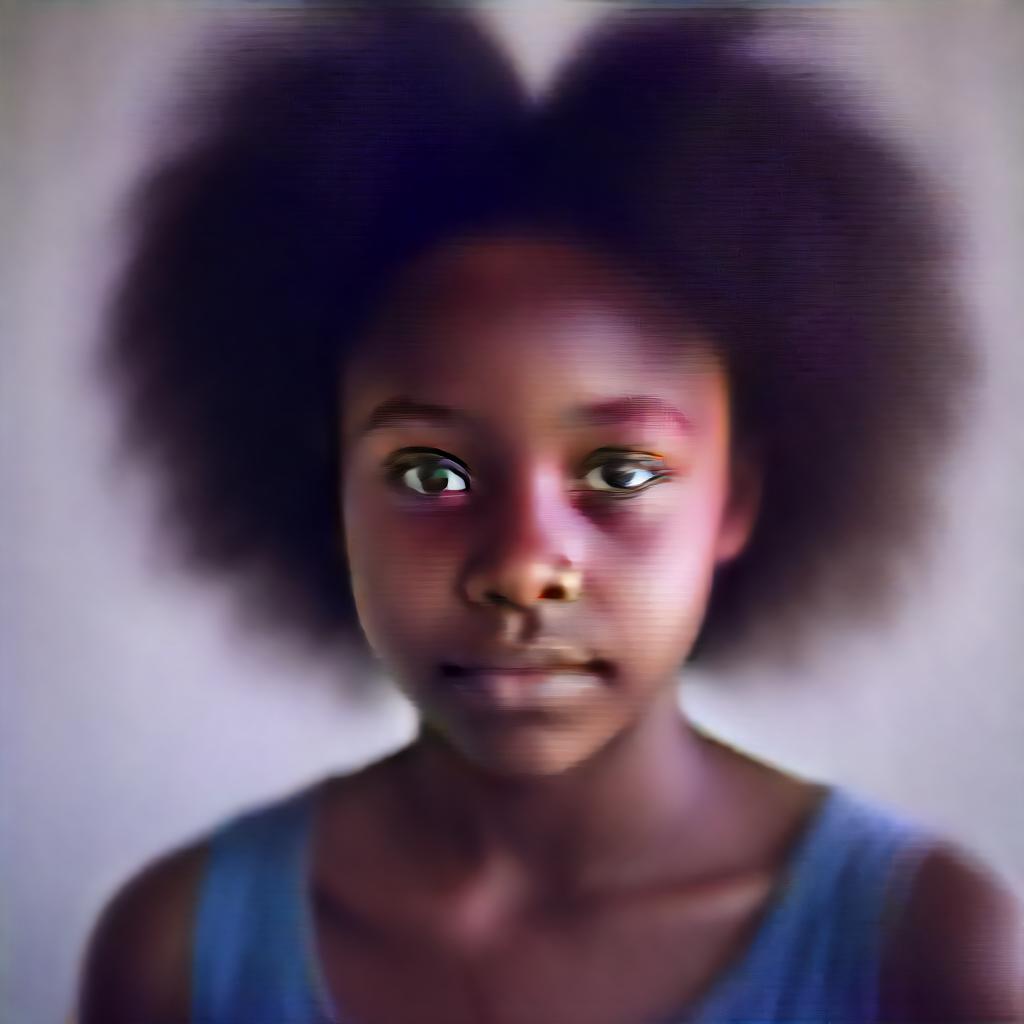} & \includegraphics[width=0.08\textwidth,height=0.08\textwidth]{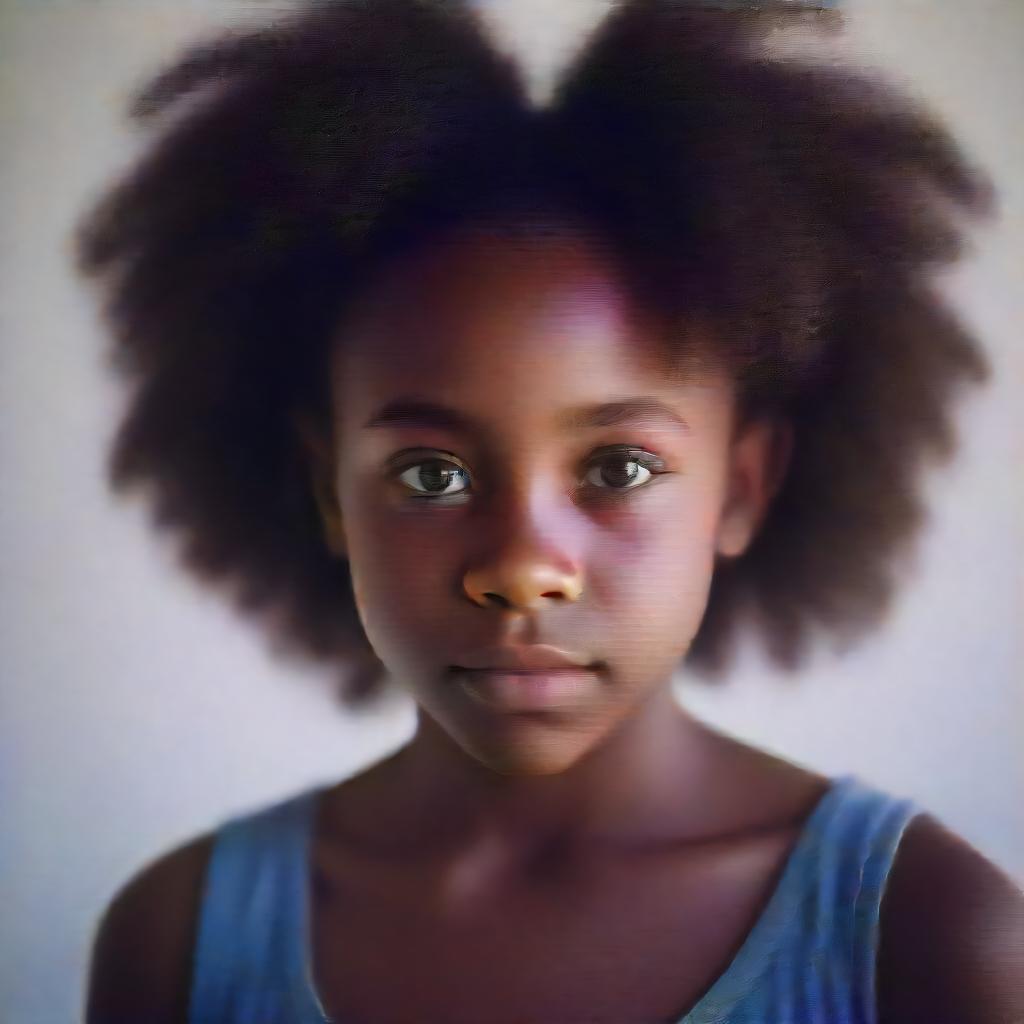} & \includegraphics[width=0.08\textwidth,height=0.08\textwidth]{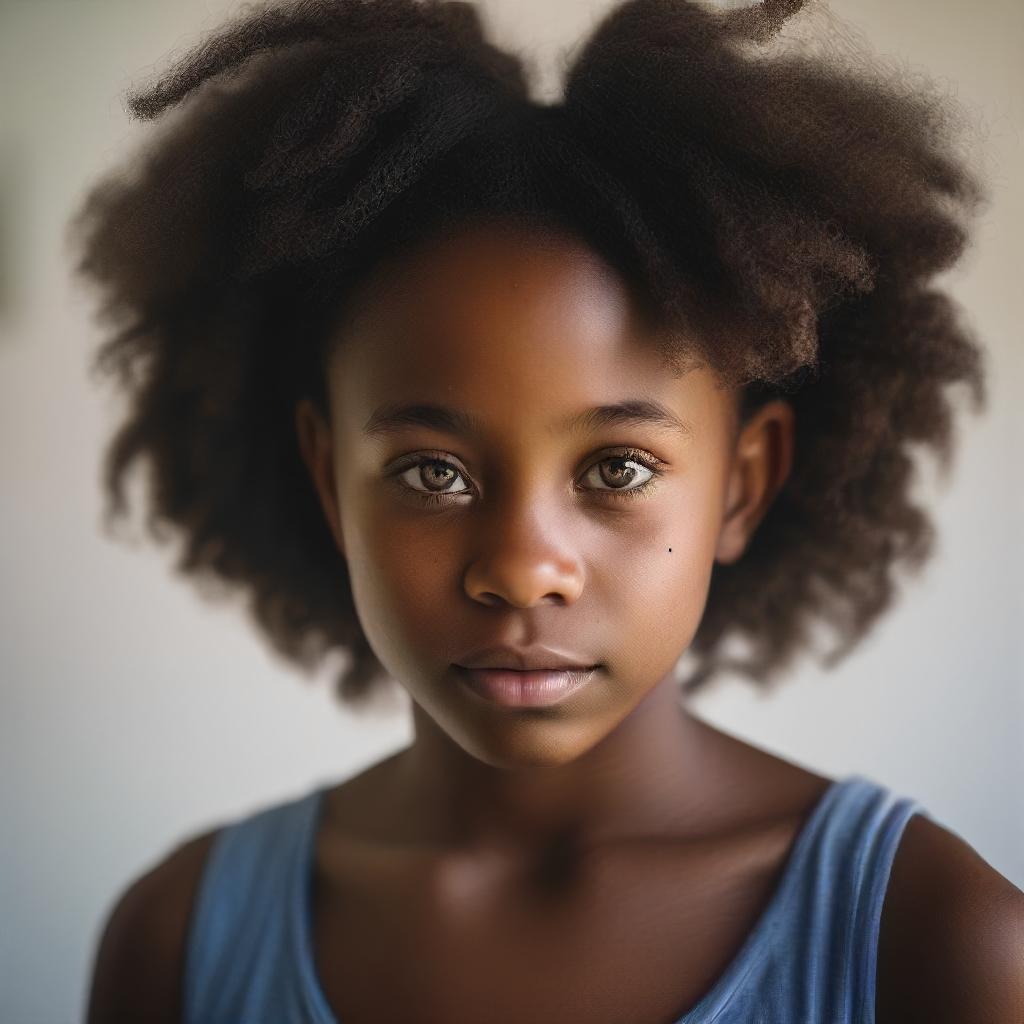}\\
        
        & 0\% & 10\% & 25\% & 50\% & 100\% \\
                 
    \end{tabular} \\ \\ 

    \begin{tabular}{cccccc}

        & \multicolumn{5}{c}{SDXL + LoRA DreamBooth} \\
        \raisebox{0.053\linewidth}{\rotatebox[origin=t]{90}{\fontsize{8pt}{8pt}\selectfont\begin{tabular}{c@{}c@{}c@{}c@{}} LCM \end{tabular}}} &
        \includegraphics[width=0.08\textwidth,height=0.08\textwidth]{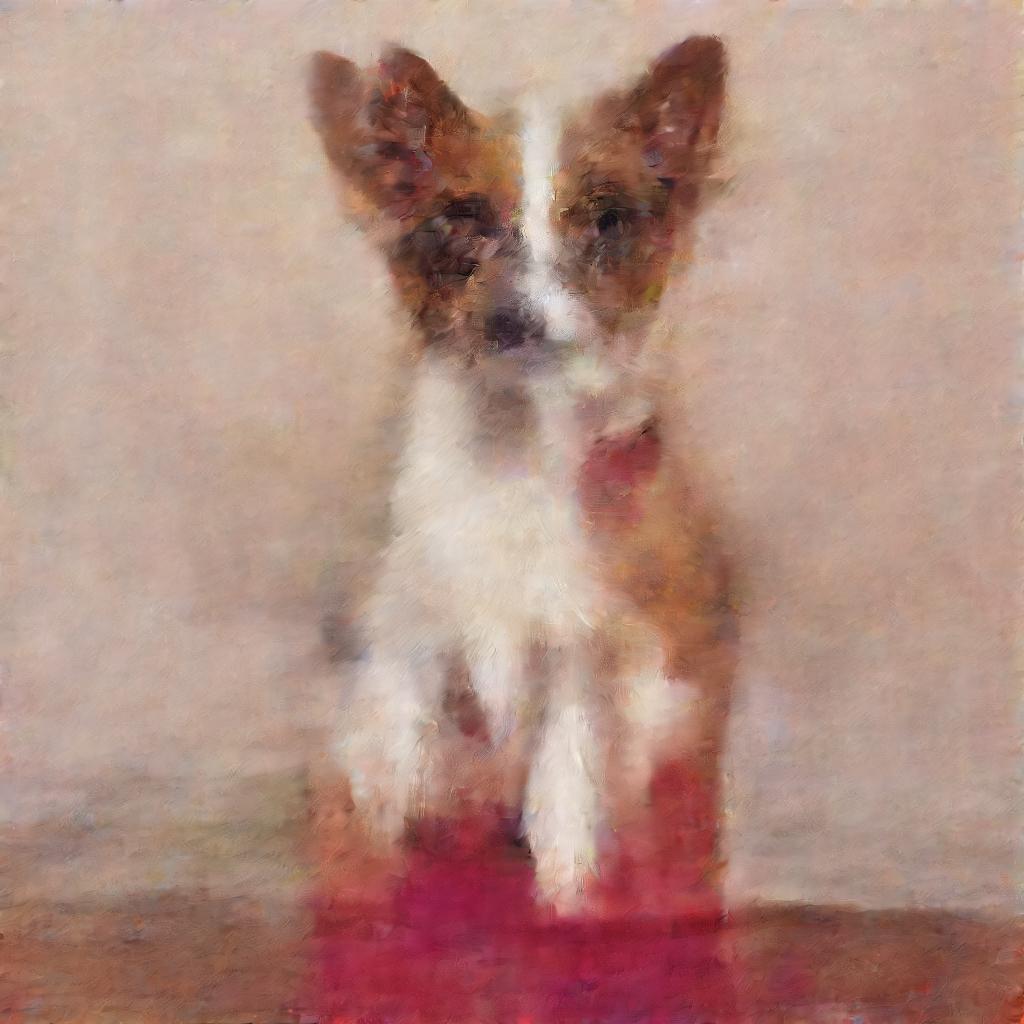} & \includegraphics[width=0.08\textwidth,height=0.08\textwidth]{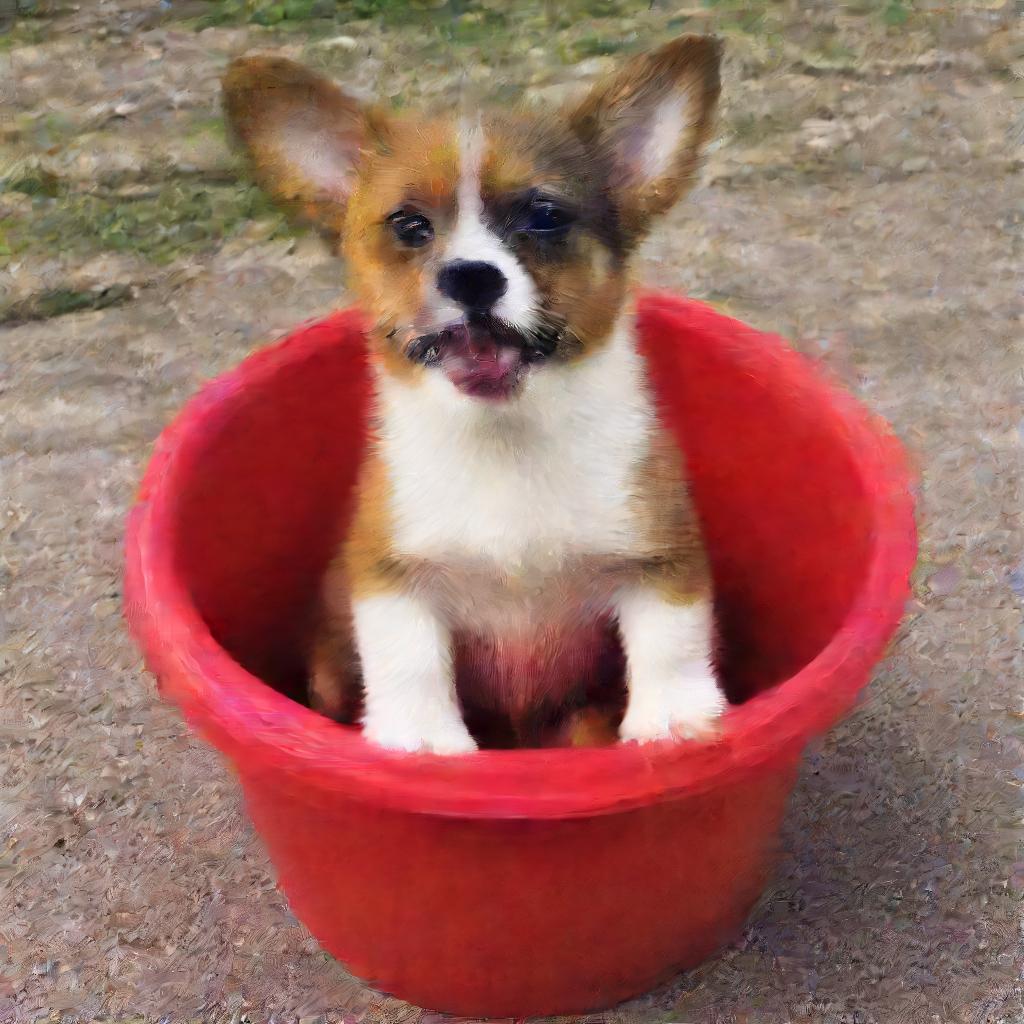} & \includegraphics[width=0.08\textwidth,height=0.08\textwidth]{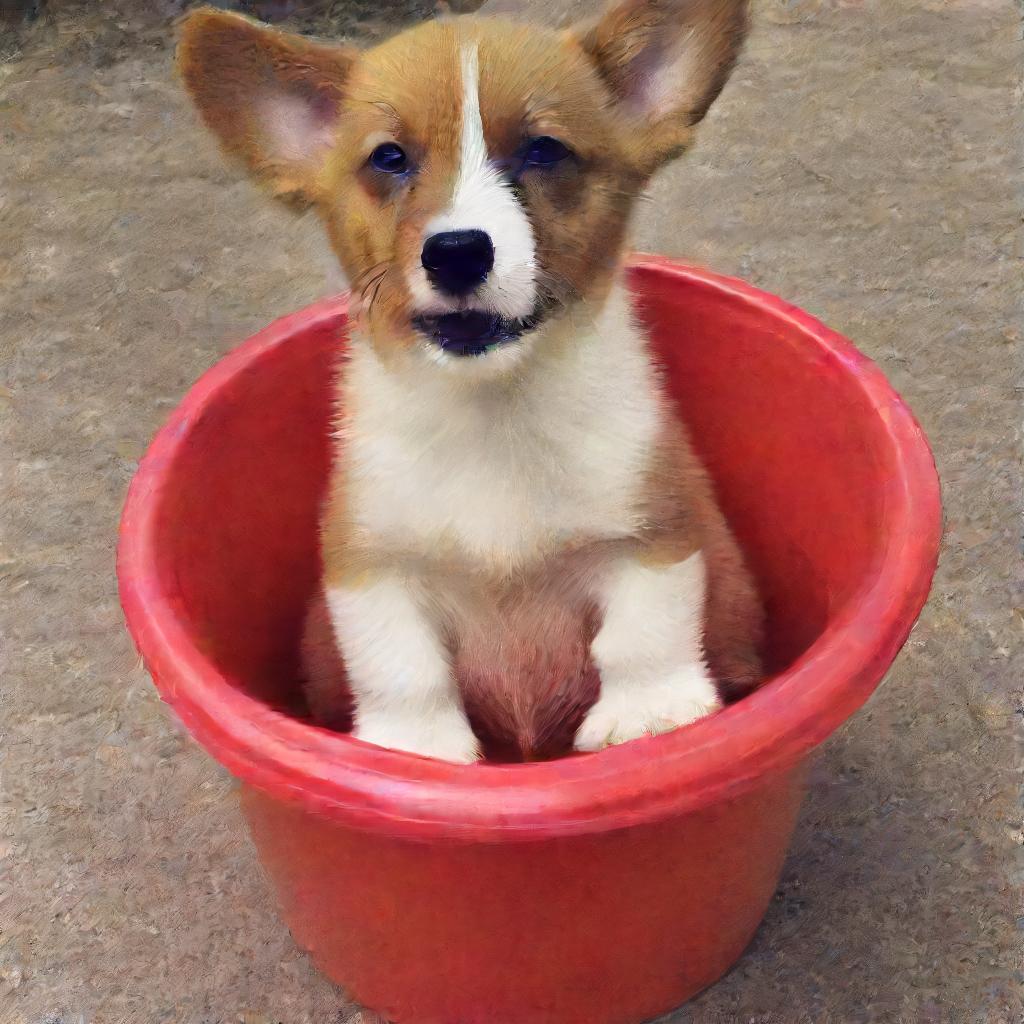} & \includegraphics[width=0.08\textwidth,height=0.08\textwidth]{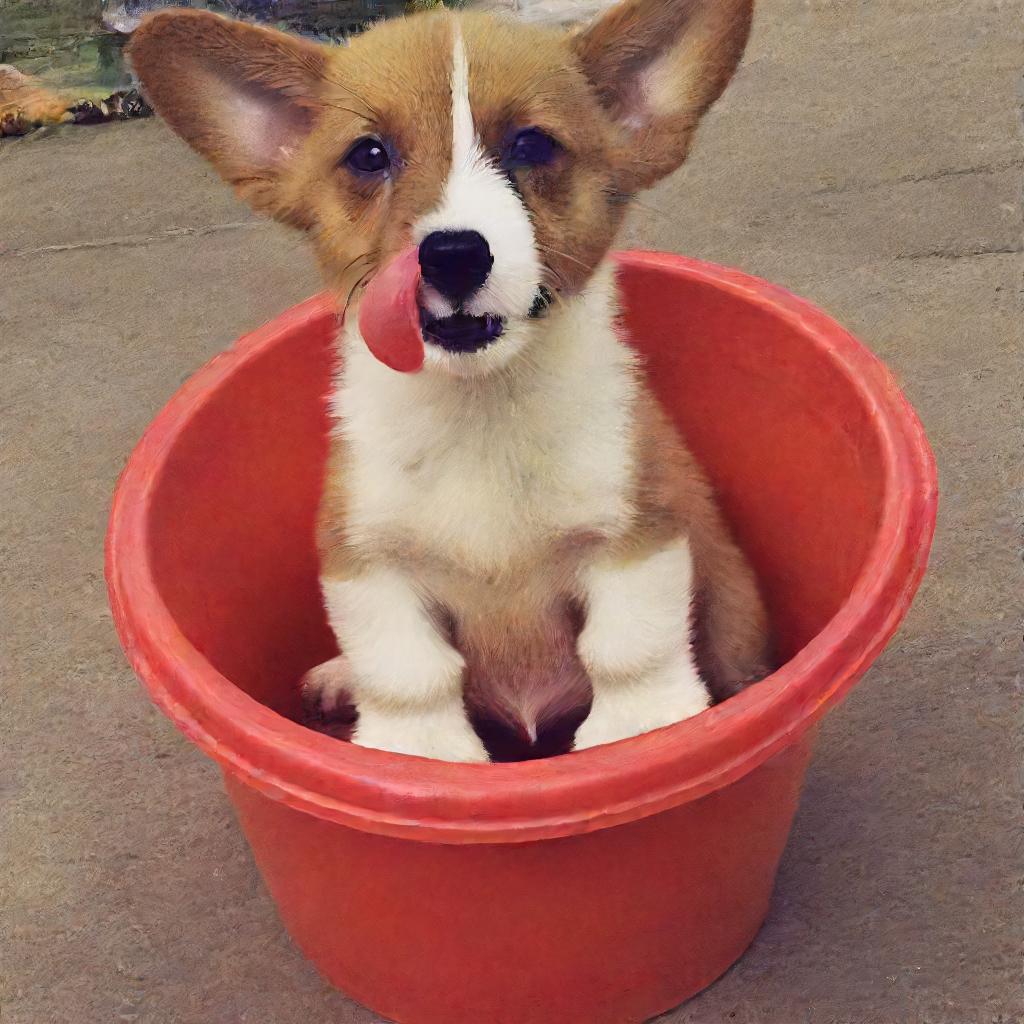} & \includegraphics[width=0.08\textwidth,height=0.08\textwidth]{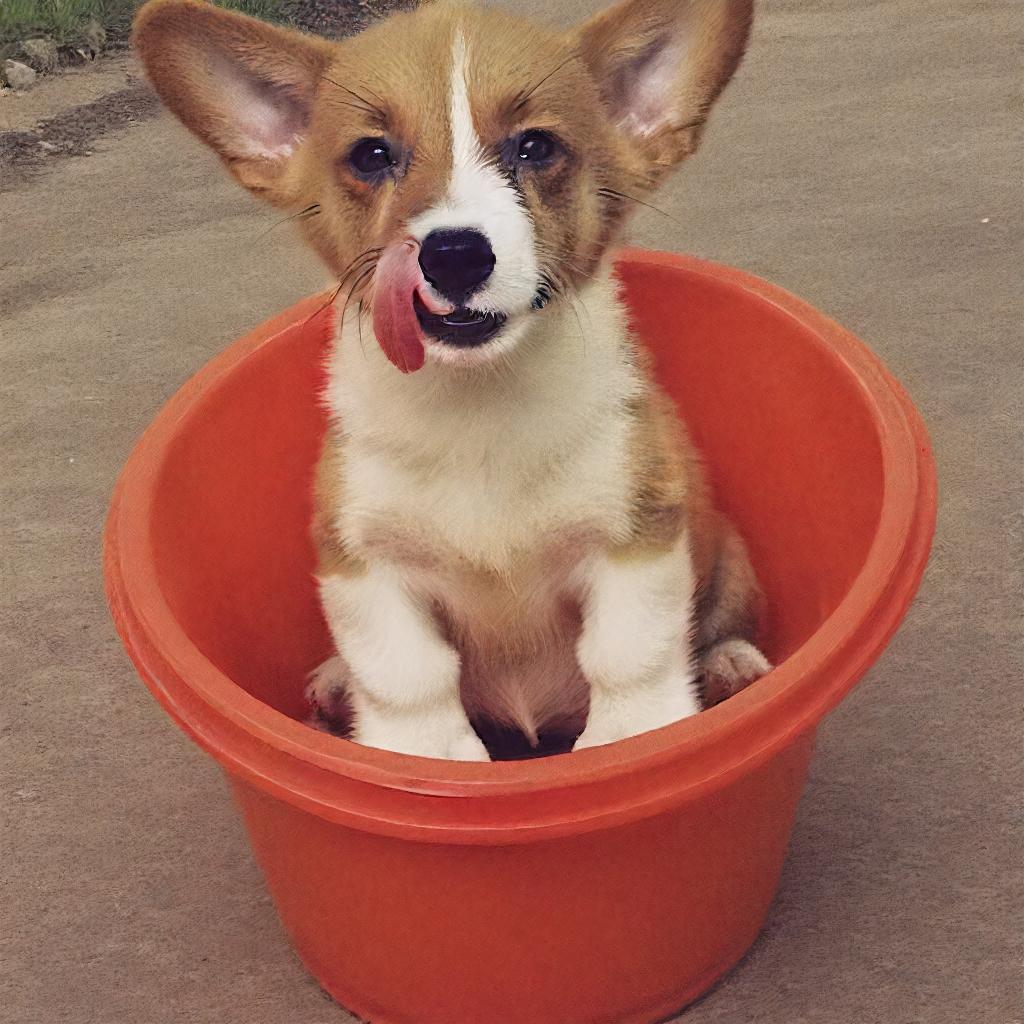}\\

        \raisebox{0.055\linewidth}{\rotatebox[origin=t]{90}{\fontsize{8pt}{8pt}\selectfont\begin{tabular}{c@{}c@{}c@{}c@{}} DDPM \end{tabular}}} & 
        \includegraphics[width=0.08\textwidth,height=0.08\textwidth]{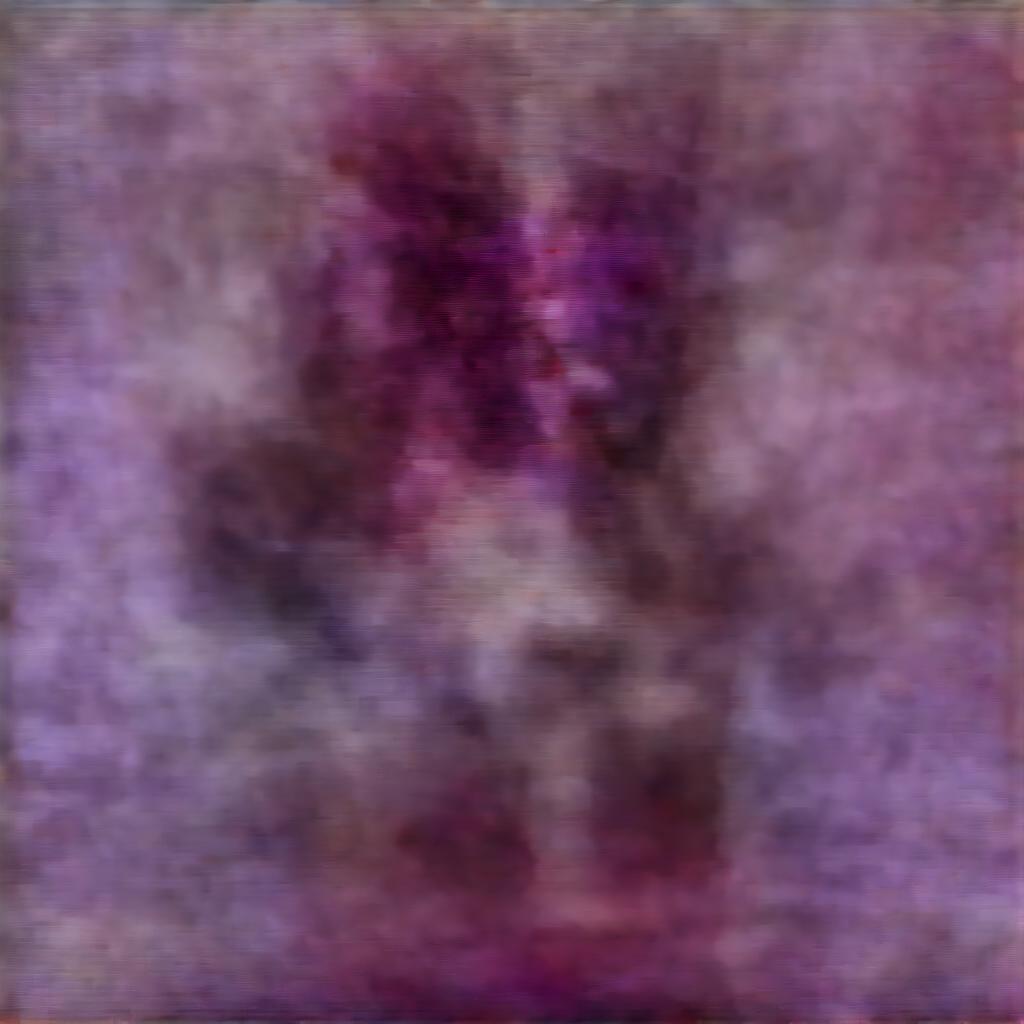} & \includegraphics[width=0.08\textwidth,height=0.08\textwidth]{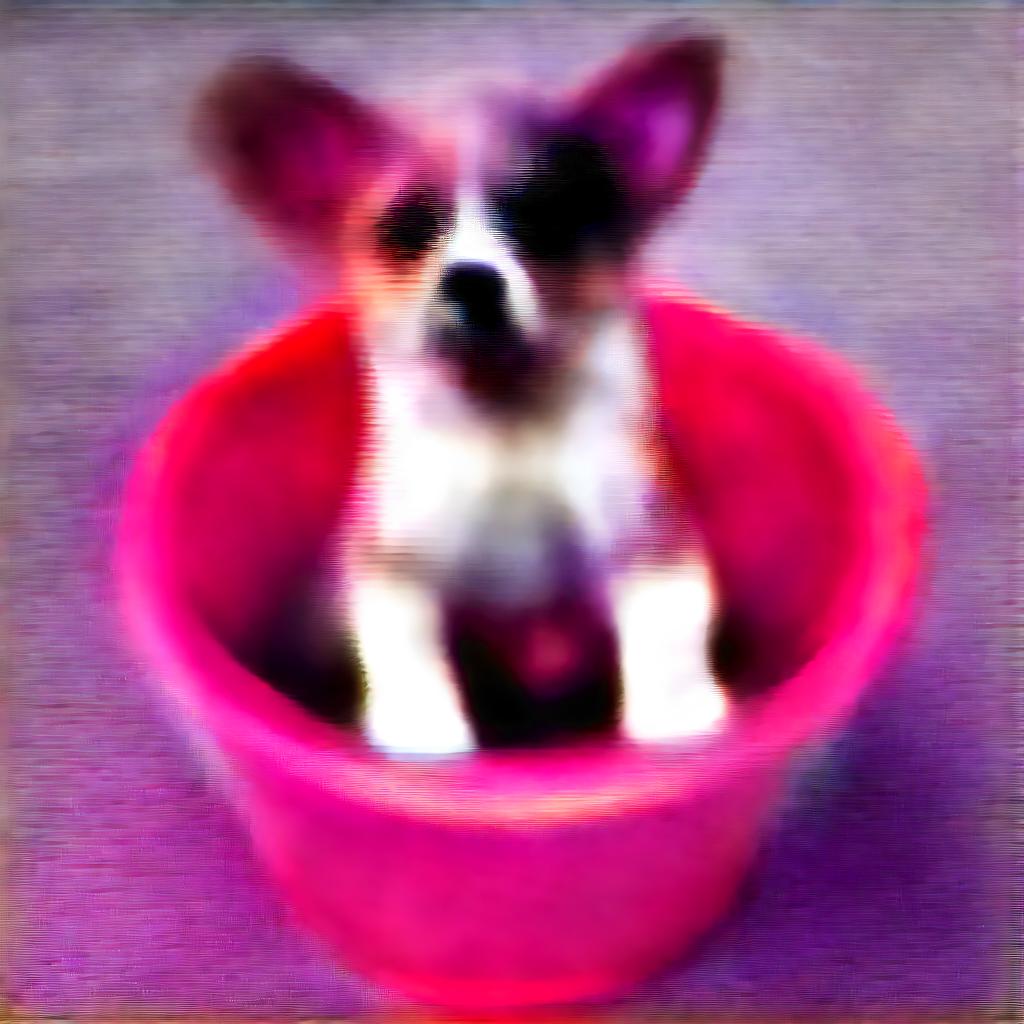} & \includegraphics[width=0.08\textwidth,height=0.08\textwidth]{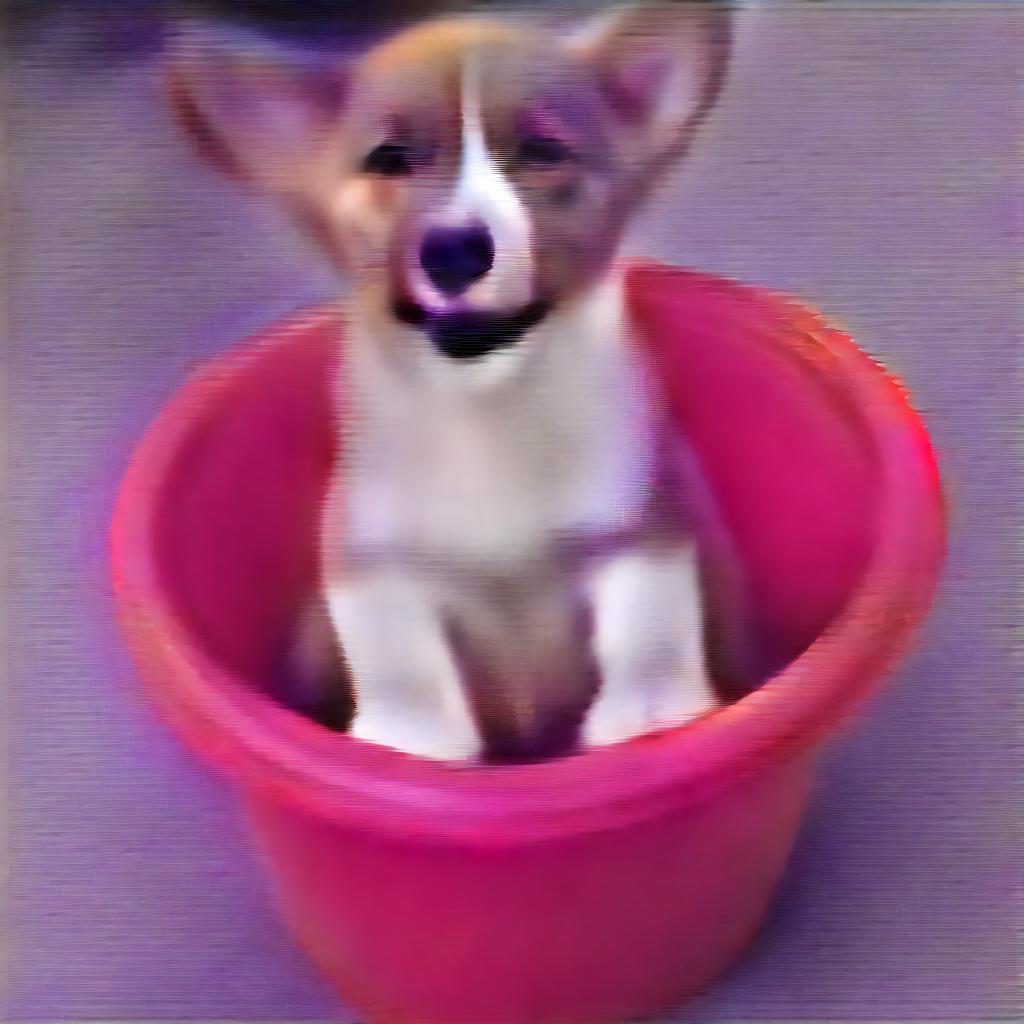} & \includegraphics[width=0.08\textwidth,height=0.08\textwidth]{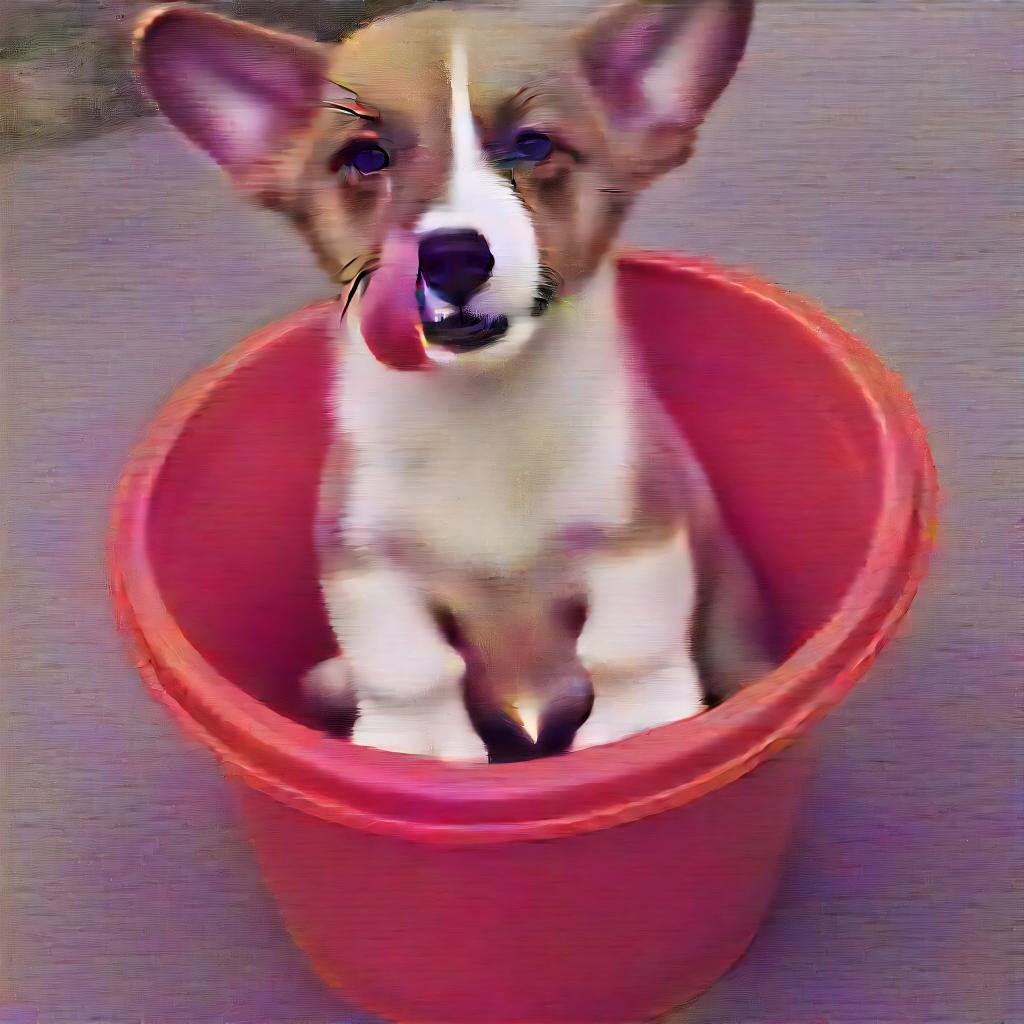} & \includegraphics[width=0.08\textwidth,height=0.08\textwidth]{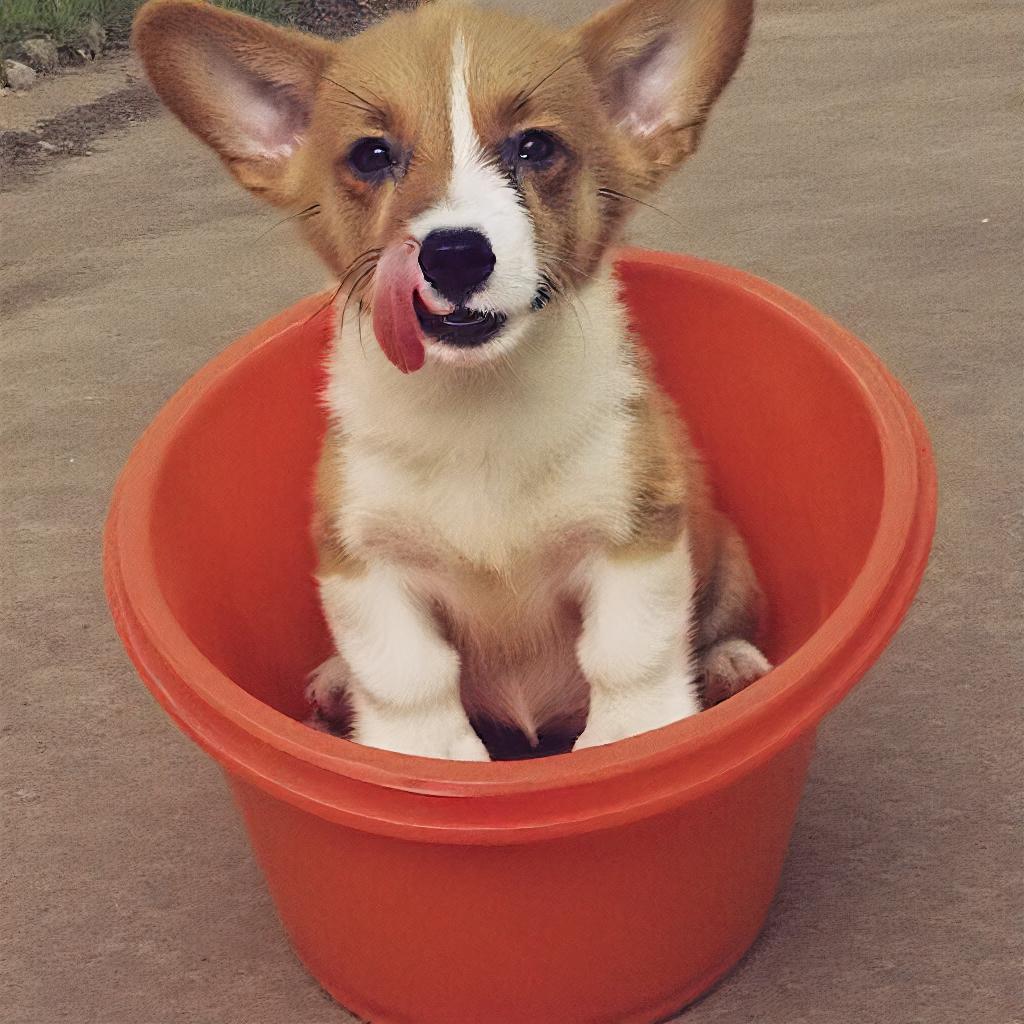}\\

        & 0\% & 10\% & 25\% & 50\% & 100\% \\
        
    \end{tabular}
    
    \end{tabular}
    
    \caption{LCM output alignment. We first denoise an image partway using DDPM~\cite{ho2020denoising} sampling with a baseline SDXL model~\cite{podell2024sdxl}. We then complete sampling in two manners: By performing a single LCM step, or by approximating the clean image using DDPM. Even at early timesteps, the LCM outputs provide a good approximation of the final DDPM prediction. This also holds for personalized models (\eg, a LoRA trained on the DreamBooth~\cite{ruiz2022dreambooth} dog, bottom). The numbers below each column indicate the fraction of standard DDPM denoising steps completed before applying the single-step prediction.}\label{fig:alignment}
\end{figure}

Here, we present a method for tackling this question by building on recent advancements in fast sampling methods~\cite{song2023consistency,sauer2023adversarial,gu2023boot}, and specifically latent consistency models (LCM,~\cite{luo2023latent,luo2023lcmlora}). To do so, we leverage an intriguing property of generative models: fine-tuning alignment, where a child model fine-tuned from a pretrained parent tends to preserve the semantics of the parent's latent space~\cite{wu2021stylealign,gal2021stylegan}. 
In \cref{fig:alignment}, we show a particular manifestation of this alignment in LCM models, where we compare a single-step LCM output to the DDPM single-step approximations at intermediate steps of the denoising process. The LCM results are not only sharper, but they maintain a high degree of similarity with the final DDPM prediction. This property also holds for personalized models (bottom, dog). We find this alignment holds particularly well in consistency-LoRA models~\cite{luo2023lcmlora}, likely by virtue of the consistency training loss itself~\cite{song2023consistency} and the minimal model changes.

We propose to leverage this alignment during the training of personalization encoders, where we can create a high-quality preview of the denoiser's final output by doing a single LCM step on the noisy latents, guided by the same personalization encoder. This preview can then be used to calculate image-space losses, such as those derived from an identity detection network. Under this approach, the LCM-model provides a ``shortcut'' through which gradients can back-propagate to earlier diffusion timesteps, without relying on low-quality approximations. In practice, we find that \naive applications of the shortcut loss can break the alignment between the baseline and LCM models. Hence, we investigate mechanisms for preserving alignment, and show that these can improve downstream performance.

In addition to the shortcut mechanism, we further explore an additional architectural modification inspired by recent video models and image editing works~\cite{wu2023tune,ceylan2023pix2video,Khachatryan_2023_ICCV,QI_2023_ICCV,cao_2023_masactrl,mou2023dragondiffusion,hertz2023StyleAligned}. There, it has been shown that extending the self-attention mechanism, such that a generated image can also observe self-attention keys and values of a real, source image, can allow for zero-shot appearance transfer from the source to the new image~\cite{alaluf2023crossimage,tewel2024trainingfree}. Hence, we propose to augment the encoder with an additional path where the input image is noised, passed through a copy of the diffusion U-Net, and its self-attention keys and values are extracted and appended to the forward pass of the newly generated image.

Finally, our encoder uses existing backbones~\cite{radford2021learning,ye2023ipadapter} which we tune on newly generated data. To create our dataset, we leverage SDXL Turbo~\cite{sauer2023adversarial}, a model distilled for single-step sampling from the vanilla SDXL using score distillation sampling~\cite{poole2023dreamfusion} and adversarial training~\cite{goodfellow2014generative}. Notably, we observe that this distillation process causes a significant collapse of diversity, such that a sufficiently detailed prompt will generate the same identity regardless of the input seed. We leverage this property to generate a dataset of images with repeated identities and differing styles. Importantly, the use of generated data allows us to avoid the need to collect sensitive personal data, and ensures that our dataset contains proper representation for minority classes that are typically under-represented in real datasets.

We validate our approach through comparisons to a range of recent baselines, and through a large set of ablation scenarios. These demonstrate that our approach can achieve higher identity fidelity and prompt alignment, and highlight the benefit of integrating image-space losses during model tuning.

Our code and models will be made public at \href{https://lcm-lookahead.github.io/}{\url{https://lcm-lookahead.github.io/}}.

\section{Related work}

\paragraph{\textbf{Text-to-image Personalization.}}
In the task of text-to-image personalization~\cite{gal2022textual,ruiz2022dreambooth}, the goal is to adapt a pretrained model to better represent a user-given concept which was unseen during the original training. Initial efforts show that this can be achieved by either optimizing a new text-embedding~\cite{gal2022textual} or by further tuning the denoising network itself~\cite{ruiz2022dreambooth}. This has spurred a variety of different approaches that try to extend the optimized embedding space~\cite{alaluf2023neural,voynov2023p+,dong2022dreamartist} or restrict the tuned part of the denoising network~\cite{simoLoRA2023,kumari2022customdiffusion,tewel2023key,avrahami2023bas,han2023svdiff}. A joint limitation of all these different approaches is that they require a per-subject optimization process which can be time-consuming.

\paragraph{\textbf{Encoder-Based Personalization.}} 
To overcome the limitations of optimization-based approaches, several works proposed that pretrained encoders can be used to initialize the optimization process, shortening tuning times by a factor of $x50$ or more~\cite{gal2023encoder,ruiz2023hyperdreambooth,arar2023domain,li2023blip}. Others tried to completely avoid any optimization, but this typically comes at the cost of accuracy~\cite{Wei_2023_ICCV,chen2023subjectdriven,shi2023instantbooth,ye2023ipadapter,jia2023taming}.

Recently, the problem of face personalization gained focus, with a range of works specifically tackling this domain. With facial data, a special focus is put on improving identity preservation across prompts~\cite{gal2023encoder,valevski2023face0,wang2024instantid,yuan2023celebbasis,xiao2023fastcomposer,ruiz2023hyperdreambooth}. In PhotoMaker~\cite{li2023photomaker} the features of a CLIP-model~\cite{radford2021learning} are modulated using a dedicated ID-oriented dataset. Face0~\cite{valevski2023face0} proposed to replace the CLIP encoder with an identity recognition network. Newer variants of IP-Adapter~\cite{ye2023ipadapter} introduced similar ideas, with identity network features being used either instead of, or on top of the CLIP image embedding. FaceStudio~\cite{yan2023facestudio} follows a similar mechanism but also uses a prior model to better adhere to the original prompt. The concurrent InsantID paper builds on IP-Adapter and extends it with a landmark-conditioned ControlNet~\cite{zhang2023adding} to preserve pose and simplify identity preservation. 

Common to all these methods is that their training signal is based solely on the standard diffusion training loss. This is in contrast with the common practice in GAN-Based encoder methods~\cite{richardson2020encoding,alaluf2021hyperstyle,alaluf2021restyle,tov2021designing,wang2021HFGI,dinh2022hyperinverter}, which demonstrated improved results by integrating perceptual losses, such as an identity loss. This gap can be attributed to the challenge of applying pixel-based losses during diffusion training. Such losses have been investigated by the concurrent PortraitBooth~\cite{peng2023portraitbooth}. They propose to apply an identity loss only to images sampled with low noise levels, where one can well approximate the clean image. However, this prevents the loss from influencing early diffusion steps, and hence limits its effectiveness.

Our work is a similar tuning-free face-identity encoder, but we propose to integrate an identity loss into to the model training by using an LCM-based shortcut mechanism, which allows us to create clean previews of the target image even at early sampling steps.

\paragraph{\textbf{Fast Diffusion Sampling.}}
Standard diffusion models~\cite{ho2020denoising,sohl2015deep} and their text-to-image variants~\cite{rombach2021highresolution,nichol2021glide,ramesh2022hierarchical} are trained to denoise an image over $1,000$ steps. At inference time, this span can be significantly shortened by leveraging ordinary differential equation solvers to navigate the diffusion flow in fewer steps~\cite{song2020denoising,lu2022dpmsolver,lu2023dpmsolverpp,liu2022pseudo,karras2022elucidating}.

More recently, several methods were proposed to distill diffusion models into versions that can be sampled from in fewer steps~\cite{salimans2022progressive}. The idea there is to teach the network to predict points farther along the diffusion flow, either by directly predicting a baseline model's output after multiple steps~\cite{salimans2022progressive}, through consistency modeling techniques~\cite{song2023consistency,luo2023lcmlora,luo2023latent,kim2024consistency}, by leveraging adversarial training~\cite{goodfellow2014generative} and score sampling~\cite{poole2023dreamfusion} methods~\cite{sauer2023adversarial} or by matching distributions between slow- and fast-sampling diffusion models~\cite{yin2024onestep}. LCM-LoRA~\cite{luo2023lcmlora} has further shown that the consistency modeling approach can be applied through a low-rank adaptation~\cite{Hu2021LoRALA} of the model weights.

Our work leverages such models, and LCM-LoRA in particular, in order to introduce image-space losses into personalization encoder training.

\label{sec:related}

\section{Preliminaries}\label{sec:preliminaries}

\subsection{Text-to-image Personalization}

Text-to-image personalization methods~\cite{gal2022textual,ruiz2022dreambooth,kumari2022customdiffusion,simoLoRA2023} introduce new, user-provided concepts to a pretrained T2I diffusion model, typically by tuning novel word-embeddings, parameters of the denoising network itself, or mixture of both. Such optimization-based methods use a small (typically $3$-$5$) image set depicting the concept, and tune the relevant parameters using the standard diffusion training loss~\cite{ho2020denoising}:
\begin{equation}
    \mathscr{L}_{Diffusion} := \mathbb{E}_{z, y, \epsilon \sim \mathcal{N}(0, 1), t }\Big[ \Vert \epsilon - \epsilon_\theta(z_{t},t,y) \Vert_{2}^{2}\Big] \, ,
    \label{eq:l_simple_orig}
\end{equation}
where $\epsilon$ is an unscaled noise sample, $\epsilon_\theta$ is the denoising network, $t$ is the time step, $z_t$ is an image or latent noised to time $t$, and $y$ is some conditioning prompt containing a placeholder token which is used to describe the new concept.

To speed up the personalization process, prior work suggested the use of encoders --- neural networks trained to take images of the subject, and map them to some conditioning code that can guide the pretrained diffusion network to generate an image of the subject. A common approach~\cite{Wei_2023_ICCV,jia2023taming} popularized by IP-Adapter~\cite{ye2023ipadapter} is to do so by extending the diffusion denoiser with additional cross-attention layers, which receive tokens derived from some external encoder module (e.g. a frozen CLIP-backbone~\cite{radford2021learning} with a small projection head). The novel cross-attention heads and the encoder module are then trained using the loss of \label{eq:l_simple_orig}, where images are drawn from large-scale datasets depicting an array of subjects, commonly from a single domain (e.g. human faces).

In this case, the diffusion loss can be rewritten as:
\begin{equation}
    \mathscr{L}_{Diffusion} := \mathbb{E}_{z_r, y, I_c, \epsilon \sim \mathcal{N}(0, 1), t }\Big[ \Vert \epsilon - \epsilon_\theta(z_{r,t},t,y,E(I_c)) \Vert_{2}^{2}\Big] \, ,
    \label{eq:l_simple}
\end{equation}
where $I_c$ is a conditioning image sampled from the training set, $z_r$ is the image latent of a reconstruction target showing the same subject as $I_c$, and $E(I_c)$ is a conditioning code derived from the encoder.

Our work builds on a pretrained IP-Adapter encoder, which we augment with additional self-attention-based features, and tune to improve identity and prompt alignment.

\section{Method}
\label{sec:method}

\begin{figure*}[t]

    \includegraphics[width=0.98\textwidth]{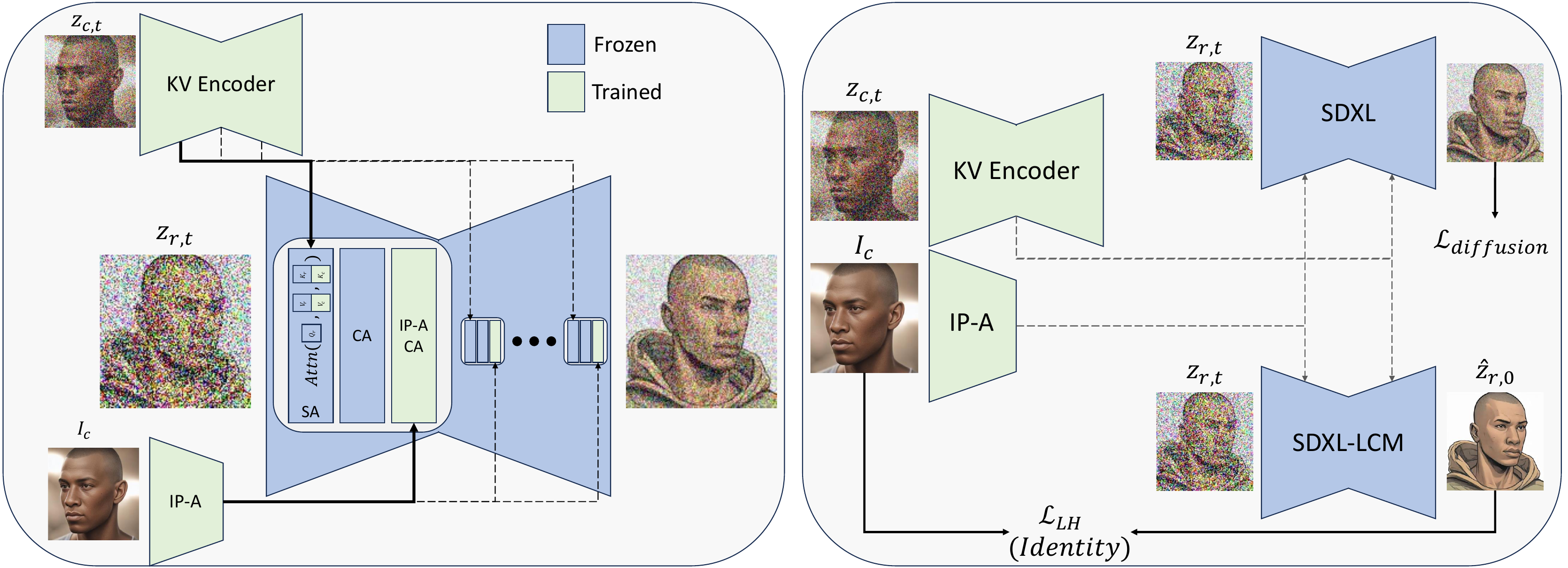}
    \caption{\textbf{(left)} \textbf{Encoder architecture:} Our encoder has two branches: One is the standard IP-Adapter~\cite{ye2023ipadapter} that provides conditions through a new cross-attention head. The second branch is a copy of the SDXL U-net from which we extract self-attention keys and values, which we concatenate with those of the main denoising branch. \textbf{(right)} \textbf{Training setup:} The two encoder paths are provided with a conditioning image (and its noisy latent), and their outputs are used to condition the denoising of a \textit{different} image of the same subject. We denoise the image with both the baseline SDXL~\cite{podell2024sdxl} model and an LCM-model~\cite{luo2023lcmlora}. The baseline model's output is used for calculating the standard diffusion loss (\cref{eq:l_simple}). The LCM output is used to calculate the lookhead identity loss (\cref{eq:lh_loss}). We portray latents as images for visual clarity.}\label{fig:method_outline}

\end{figure*}

Our work starts from an IP-Adapter model~\cite{ye2023ipadapter} pretrained to condition an SDXL model~\cite{podell2024sdxl} on facial identities. This adapter struggles to concurrently maintain both prompt-alignment and identity preservation (see \cref{fig:teaser}). Our goal is to improve it, so that newly created images will better align with both the textual prompts, and better preserve the subject's identity. For identity preservation, we introduce a novel LCM-based identity lookahead loss, and a self-attention-sharing module which can better draw visual features from the conditioning image. To improve prompt alignment, we utilize a synthetic dataset which contains consistent characters generated in an array of prompts and styles. \cref{fig:method_outline} provides a high-level outline of our encoder architecture (left) and training process (right). In the following section, we provide additional details on each of these components.

\subsection{LCM-Lookahead loss}

We begin with the goal of achieving high identity fidelity. Here, we would like to draw on core ideas from existing GAN inversion literature, and particularly the use of identity networks as a loss during encoder training~\cite{richardson2020encoding,nitzan2020face}. In the GAN-based literature, applying such a loss is trivial since the GAN can produce a clean image in a single forward pass. In the case of diffusion models, the standard training process involves sampling random intermediate diffusion time steps, and performing a single denoising step. These single-step predictions are typically noisy or blurry, and feeding them into a downstream image embedding network has previously been shown to be sub-optimal~\cite{dhariwal2021diffusion,wallace2023doodl,nichol2021glide}. These prior works showed that one can improve the quality of guidance by training the downstream embedding networks on noisy images~\cite{nichol2021glide}, or by performing multiple forward steps to create a clean image, and back-propagating through the entire diffusion chain~\cite{wallace2023doodl}. However, the first approach is expensive, and the second is impractical in training scenarios: both because such training is typically already memory constrained, and because end-to-end sampling increases iteration times by a factor of $\sim100$, making training infeasible. Instead, we propose to utilize a pretrained LCM-LoRA model~\cite{luo2023lcmlora}, tuned from the same baseline SDXL~\cite{podell2024sdxl} backbone, to create higher-quality previews of the fully-denoised images using a single diffusion step. This preview can be fed into the downstream feature extractor, in our case a face recognition network~\cite{deng2019arcface}, and gradients can be backpropagated to the encoder through this LCM-path. We focus on the LoRA variant of the LCM model as it best preserves the alignment with the baseline model. 

More concretely, let ($I_c$, $I_r$) be a conditioning image and reconstruction-target pair, let $z_{r,t}$ be the image latent $z_r$, noised to an intermediary time step $t$, and let $\epsilon_{LCM}$ be the LCM denoising network. The preview image is given by:
\begin{equation}
    \hat{z}_{r,0} = \frac{1}{\sqrt{\Tilde{\alpha}_t}} \left(z_{r,t} - \sqrt{\Tilde{\beta}_t} \cdot \epsilon_{LCM}(z_{r,t},y,t,E(I_c))\right),
\end{equation}
where $y$ and $E(I_c)$ are the prompt and encoder conditioning codes respectively, $\alpha_t$ and $\beta_t$ are parameters defined by the diffusion noising schedule. The lookahead loss is then:
\begin{equation}\label{eq:lh_loss}
    \mathscr{L}_{LH} = \mathcal{D}\left(D_{VAE}(\hat{z}_{r,0}), I_c\right),
\end{equation}
where $\mathcal{D}$ is some image-space distance metric (\eg an identity loss) and $D_{VAE}$ is the VAE~\cite{kingma2013auto} decoder which maps the latents back to image space. 

\subsection{Maintaining alignment}

In initial experiments, we found that applying the lookahead loss of \cref{eq:lh_loss} can improve identities over short-training runs. However, over time, this loss causes the LCM pathway to focus solely on the loss-metric at the expense of its prior output, breaking the alignment with the baseline model. We investigated a series of options for improving alignment preservation, including the use of existing distribution matching options like Score Distillation Sampling~\cite{poole2023dreamfusion}, or appending the standard Consistency Model loss~\cite{song2020denoising} to the LCM-path. Full details and results from this investigation are presented in the supplementary. In practice, the best downstream performance was achieved using a model interpolation approach. There, for half of our training iterations, we randomly re-scale the LoRA component of the LCM-LoRA using $\alpha_{LoRA} \in [0.1, 1.0]$. We hypothesize that applying the loss through the continuously interpolated model makes it more difficult for the encoder to converge to a solution which works differently for the LCM- and non-LCM paths. Moreover, this serves as a form of augmentation which makes adversarial solutions less likely. Lower values of $\alpha_{LoRA}$ can still act as a preview for intermediate (but not overly noisy) outputs, which can still provide a signal through the image-space model.

Finally, similarly to \citet{wallace2023doodl}, we find it useful to focus on early (noisy) diffusion time steps. Here, we do so by applying annealing to the training time step sampling. We use the importance-weighting function of \citet{huang2023reversion}:
$f(t) = \frac{1}{T}(1 - \alpha \cos{\frac{\pi t}{T}})$, where $f(t)$ is the un-normalized probability to sample time step $t$, $T$ is the total number of diffusion time steps (\ie $1,000$) and $\alpha$ is a hyperparameter which we empirically set to $0.2$.

\subsection{Extended self-attention features}

As a second component for improving identity fidelity, we propose to leverage recent ideas in video-based modeling~\cite{wu2023tune,ceylan2023pix2video} and appearance transfer~\cite{alaluf2023crossimage,hertz2023StyleAligned,cao_2023_masactrl}. In these works, it was shown that expanding the self-attention mechanism such that a generated image can attend to the keys and values derived from a source image, can lead to a significant increase in visual similarity between the generated image and the source. Here, we use a similar idea to transfer identity features from the conditioning image to the generated output. These are applied on top of the baseline encoder from which we start. 

Specifically, we create a copy of the denoising U-Net which we call a ``KV encoder". We pass a noisy version of our conditioning image through this U-Net, and cache the self-attention keys and values derived from this pass. Then, when performing the diffusion denoising pass, we append these keys and values to those derived from the denoised image at each self-attention layer: $K^{l} := K_{z_{r,t}}^{l} \odot K_{z_{c,t}}^{l}$, $V_{z_{r,t}}^{'l} := V^{l} \odot V_{z_{c,t}}^{l}$, where $l$ is the layer index and $z_{r,t}, z_{c,t}$ subscripts denote attention features coming from the reconstruction target latent and the conditioning image latent respectively. 
This mechanism is illustrated in ~\cref{fig:method_outline} (left).

Directly applying this approach with a pretrained U-net as our encoder can lead to excessive appearance transfer and loss of editing. Hence, we do not keep the encoder frozen, but rather tune it using LoRA~\cite{Hu2021LoRALA}. As our training set (detailed below) contains target images which differ in style from the source, this draws the network towards discarding appearance properties that are related to the style and not to the content, greatly improving prompt alignment. We note that a similar attention-expansion idea was used for personalization in the concurrent work of \citet{purushwalkam2024bootpig}. However, their target images are all photo-realistic, and their results do indeed suffer from reduced editability.

Finally, using this pathway requires us to slightly modify the standard classifier-free guidance~\cite{ho2021classifier} equation to account for the new conditioning path. See the supplementary for more details.

\subsection{Consistent data generation}

Having introduced components to improve identity preservation, we now turn to improving prompt alignment. Here, we hypothesize that the limited editability in current encoder-based methods is partially grounded in their training set, which typically focuses on reconstructing real images. 
Moreover, existing large-scale sets are either closed and proprietary or use the withdrawn LAION dataset~\cite{schuhmann2021laion}. They also commonly contain biases, which may result in models whose performance deteriorates on minority classes.

Instead of relying on such data, we propose to generate a novel consistent dataset in which we generated the same \textit{synthetic} subjects across a wide range of prompts. These can then serve as training data for our encoder, where the encoder itself is provided with one image of a given identity, and the denoising goal considers another image portraying the same subject. Such cross-image training has already been shown to be beneficial with real data~\cite{jia2023taming,li2023photomaker}. Here, the use of the generated data allows us to take the idea a step further, and ensure our reconstruction targets also contain stylized images. These can in turn prevent the encoder from focusing on photo-realism.

To create our data, we investigate multiple approaches for consistent generation, including: (1) generating celebrities, which the model is already familiar with and can consistently generate across various prompts, (2) ConsiStory~\cite{tewel2024trainingfree}, a recently introduced approach that aligns identities through attention feature sharing, and (3) using SDXL-Turbo~\cite{sauer2023adversarial}.

\begin{figure}

    \centering
    \setlength{\tabcolsep}{1.5pt}
    {\small
    
    \begin{tabular}{c c c}

    \includegraphics[width=0.15\textwidth]{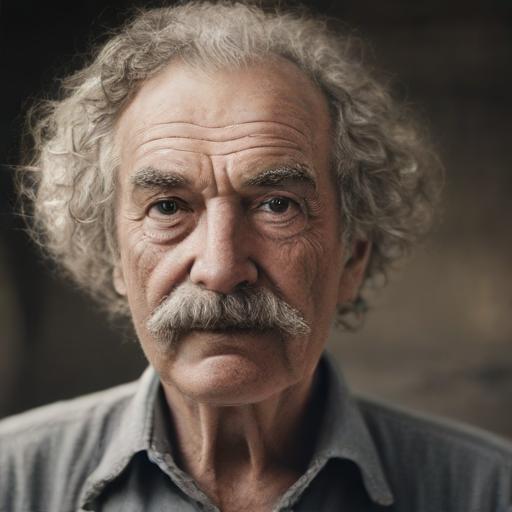} &
    \includegraphics[width=0.15\textwidth]{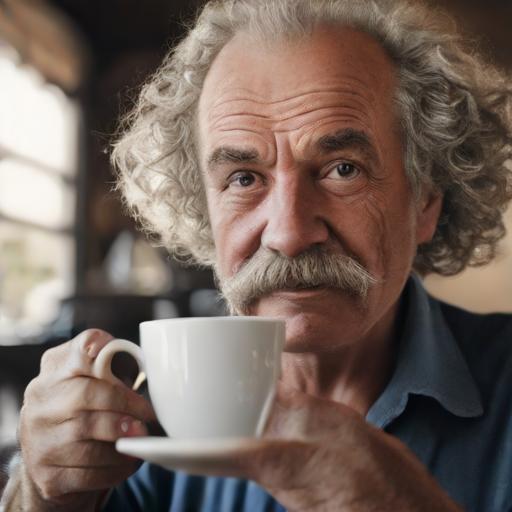} &
    \includegraphics[width=0.15\textwidth]{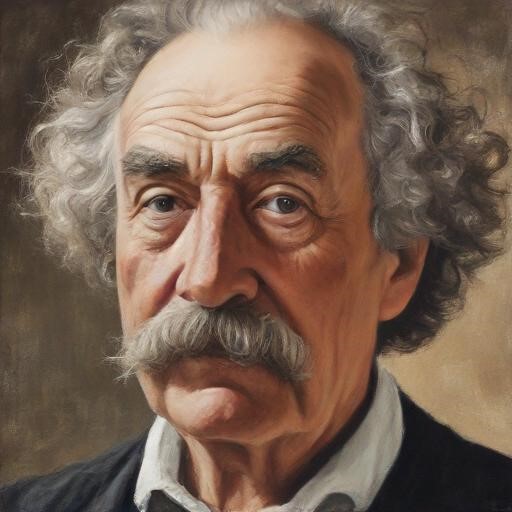} \\

    {\small{portrait photo}} &   
    {\small{drinking coffee}} &   
    {\small{oil painting}} \\

    \includegraphics[width=0.15\textwidth]{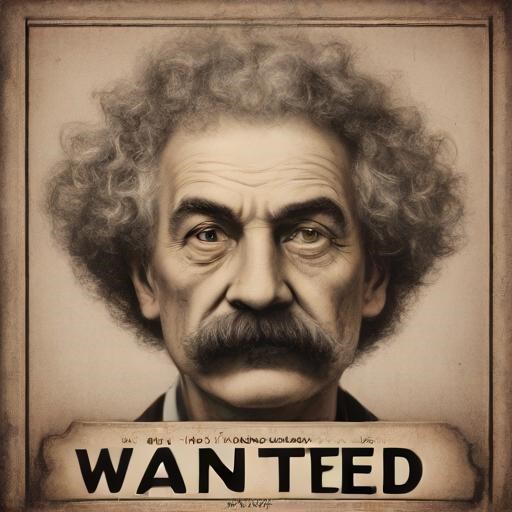} &
    \includegraphics[width=0.15\textwidth]{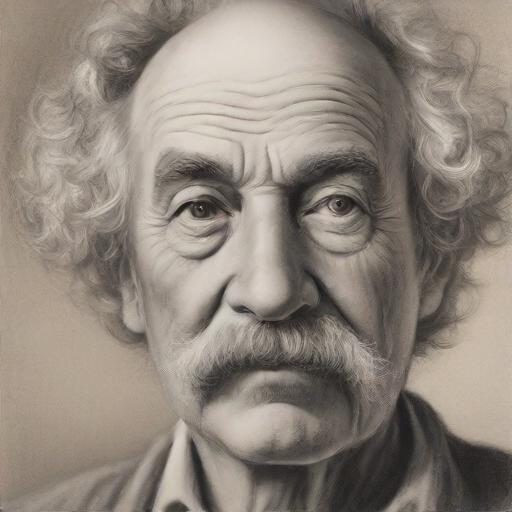} &
    \includegraphics[width=0.15\textwidth]{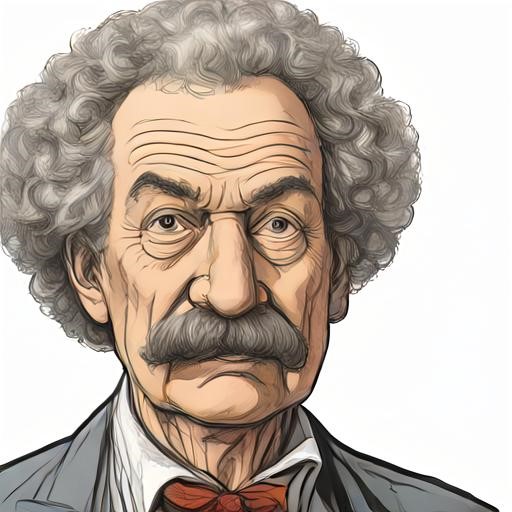} 
    \\ 
    {\small{wanted poster}} &   
    {\small{pencil drawing}} &   
    {\small{comic book}}

    \end{tabular}
    
    }
    \caption{\textbf{Consistent Data.} Consistent data generated using SDXL-Turbo with the description ``old man with curly hair and a moustache'' incorporated into different prompts (e.g. ``as an oil painting'', ``as a wanted poster'')
    }
    \label{fig:turbo_results}
\end{figure}

While SDXL-Turbo is designed for fast sampling and not for consistent image generation, its adversarial training leads to mode-collapse. We find that, as a consequence, conditioning it on sufficiently detailed subject prompts will lead to a fixed identity across seeds and styles, as shown in ~\cref{fig:turbo_results}. We find this approach to achieve the best trade-off between generation time, identity consistency, and ability to change styles across the generated images. Hence, we use it to generate $500k$ images spanning roughly $100k$ identities. In practice, we trained for less than an epoch ($40k$ identities in total). Further experiments and comparisons on consistent data generation can be found in the supplementary.

\subsection{Implementation Details}

We initialize our IP-Adapter backbone using the CLIP-based~\cite{radford2021learning} Face-ID model, and tune the same parameters as the original IP Adapter~\cite{ye2023ipadapter} using a learning rate of $1e-5$. The encoder and denoiser U-Nets~\cite{ronneberger2015u} are initialized from a pretrained SDXL model. We tune the encoder U-Net using LoRA with rank $4$ and a learning rate of $5e-6$. The decoder is kept frozen. 
For our LCM-Lookahead we use TinyVAE~\cite{bohan2023tinyvae} to decode the latents. 
This lighter model reduces our memory consumption and also improves gradient flow during backpropagation.
We tune the models over $5,000$ iterations with a total batch size of $8$ split across $2$ NVIDIA A100 GPUs. 

During training, both the KV-encoder and the denoising U-Net are conditioned on the prompt that generated their respective input images ($I_c$ and $I_r$). We modify these prompts to drop the detailed subject description, so that a prompt of the form ``a photo of an old Slavic female with tan skin and bags under eyes and brown hair in a forest" will be compressed to ``a photo of a face in a forest". At inference, we condition the KV-encoder on the prompt "a photo of a face", and the denoising U-Net on the user-provided prompt.

\begin{figure*}[t]

    \centering
    \setlength{\tabcolsep}{1.5pt}
    {\small

    \begin{tabular}{c c c c @{\hspace{0.2cm}} c c c}

    &
    {{\begin{tabular}{c}\small Guiding image \end{tabular}}} &   
    {{\begin{tabular}{c}\small $\hat{x}_0$ guidance \end{tabular}}} &   
    {{\begin{tabular}{c}\small LCM guidance\end{tabular}}} &   
    {{\begin{tabular}{c}\small Guiding  image \end{tabular}}} &   
    {{\begin{tabular}{c}\small $\hat{x}_0$ guidance \end{tabular}}} &   
    {{\begin{tabular}{c}\small LCM guidance \end{tabular}}} \\

    \raisebox{0.053\linewidth}{\rotatebox[origin=t]{90}{\fontsize{8pt}{8pt}\selectfont\begin{tabular}{c@{}c@{}c@{}c@{}} LPIPS Guidance \end{tabular}}} &

    \includegraphics[width=0.13\textwidth]{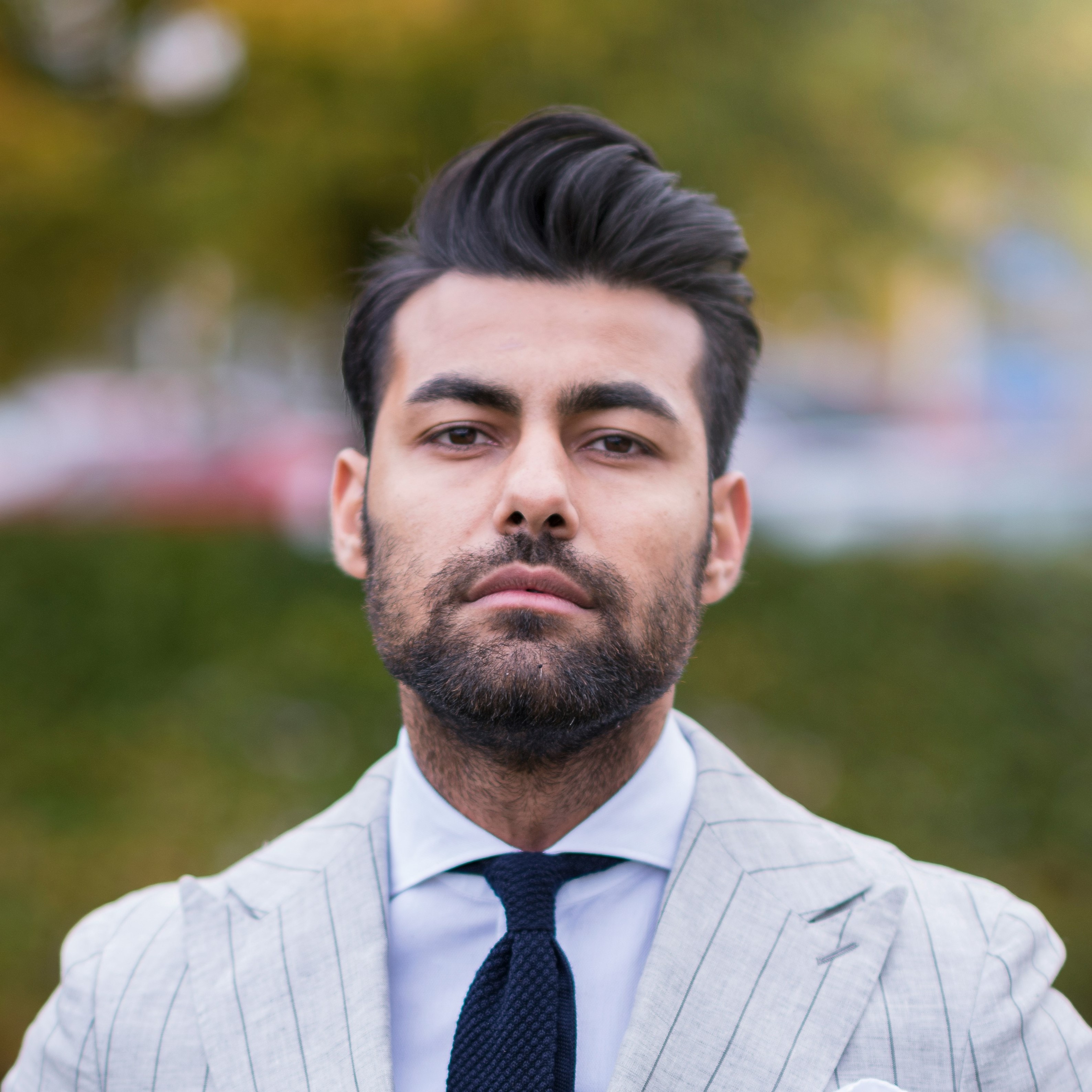} &
    \includegraphics[width=0.13\textwidth]{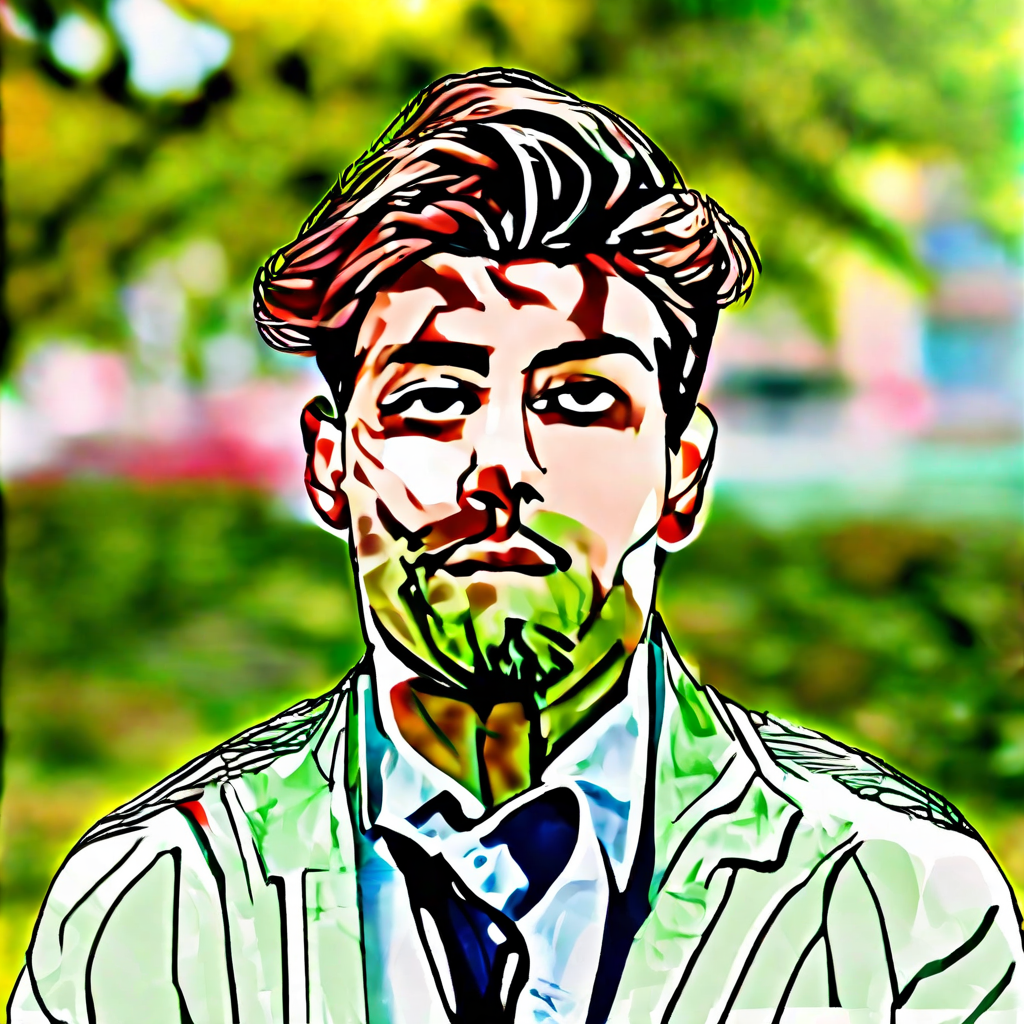} &
    \includegraphics[width=0.13\textwidth]{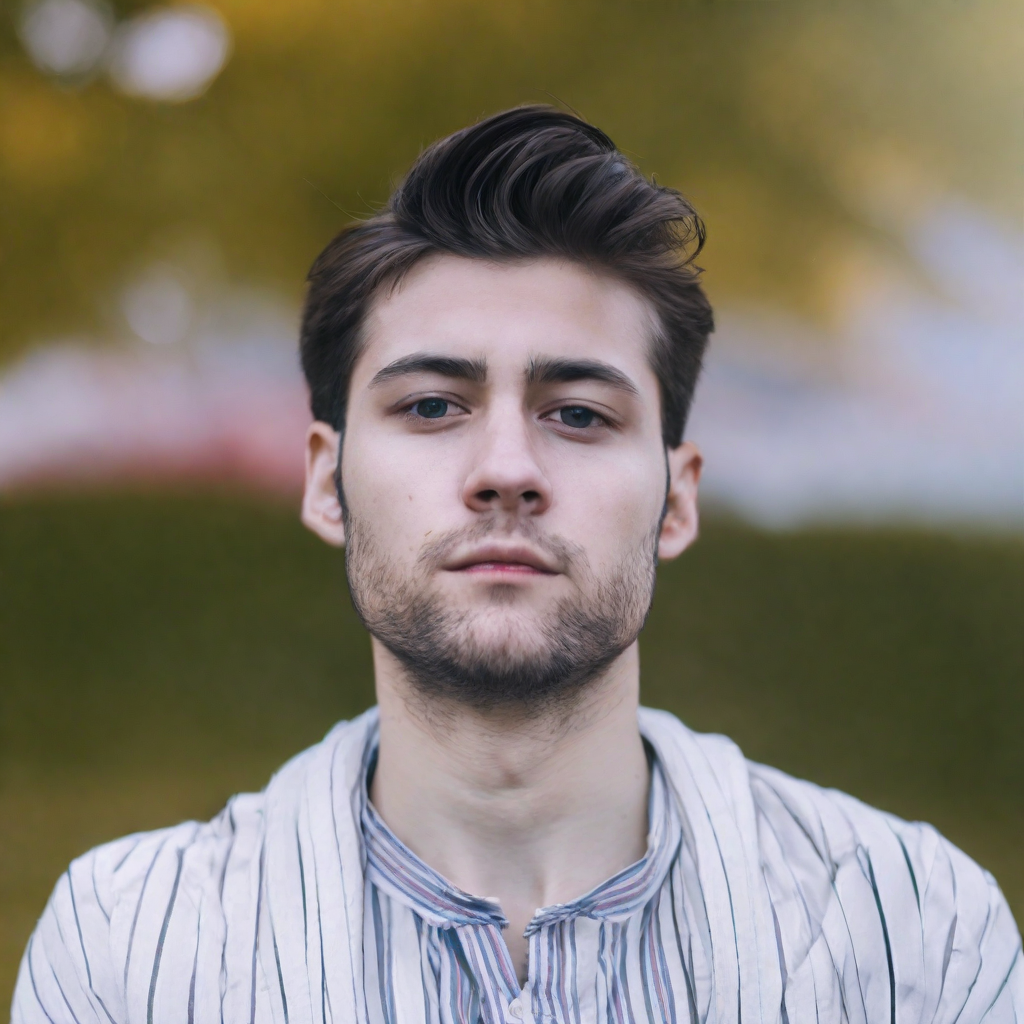} &

    \includegraphics[width=0.13\textwidth]{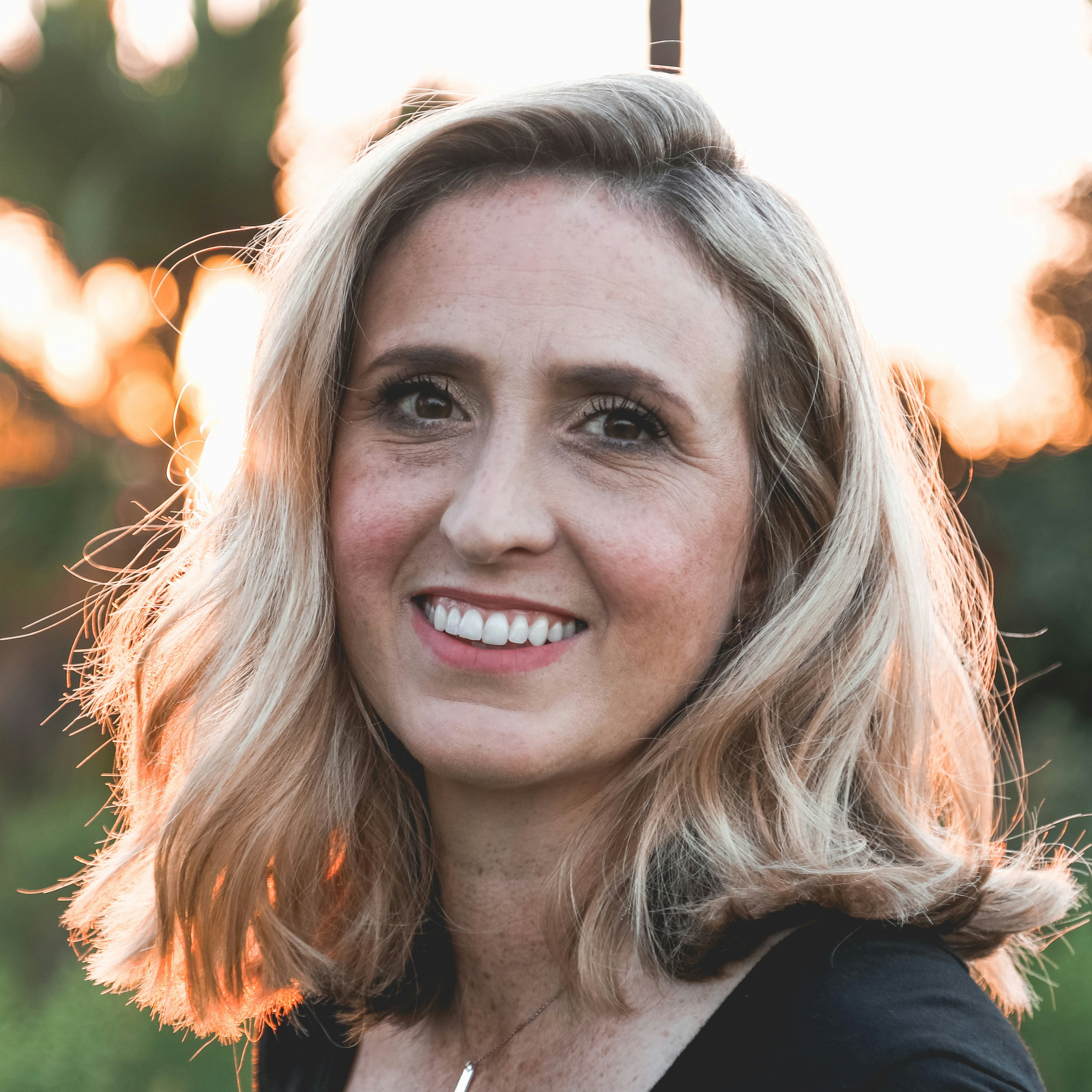} &
    \includegraphics[width=0.13\textwidth]{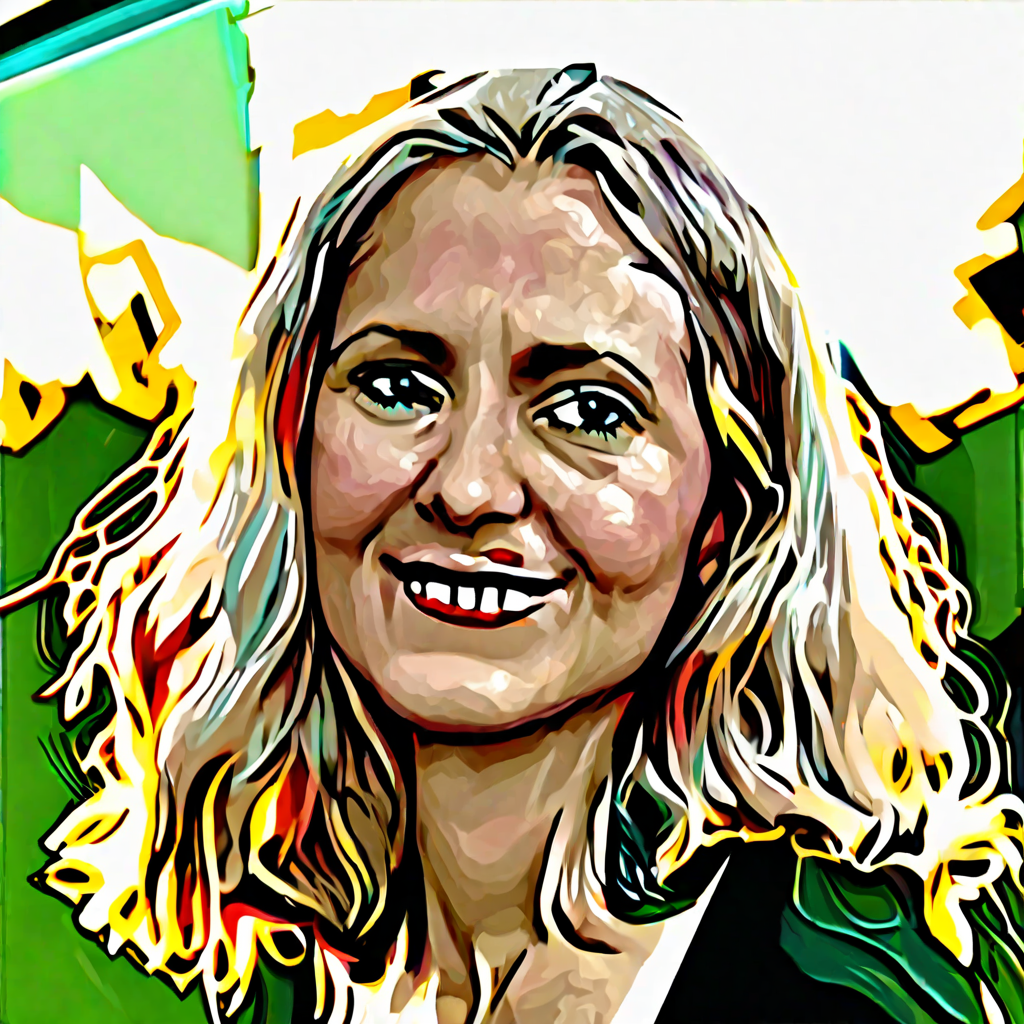} &
    \includegraphics[width=0.13\textwidth]{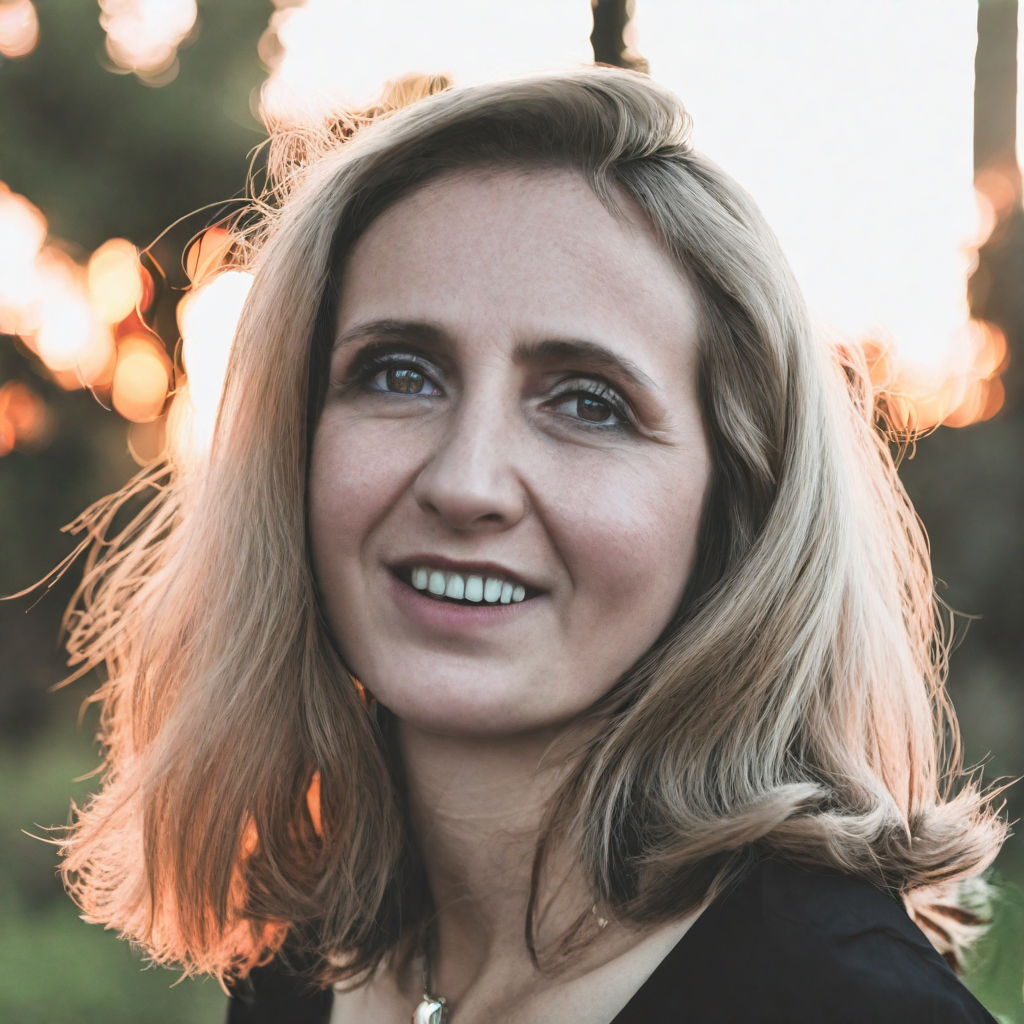}
    
    \\

    \raisebox{0.053\linewidth}{\rotatebox[origin=t]{90}{\fontsize{8pt}{8pt}\selectfont\begin{tabular}{c@{}c@{}c@{}c@{}} CLIP Guidance \end{tabular}}} &

    \includegraphics[width=0.13\textwidth]{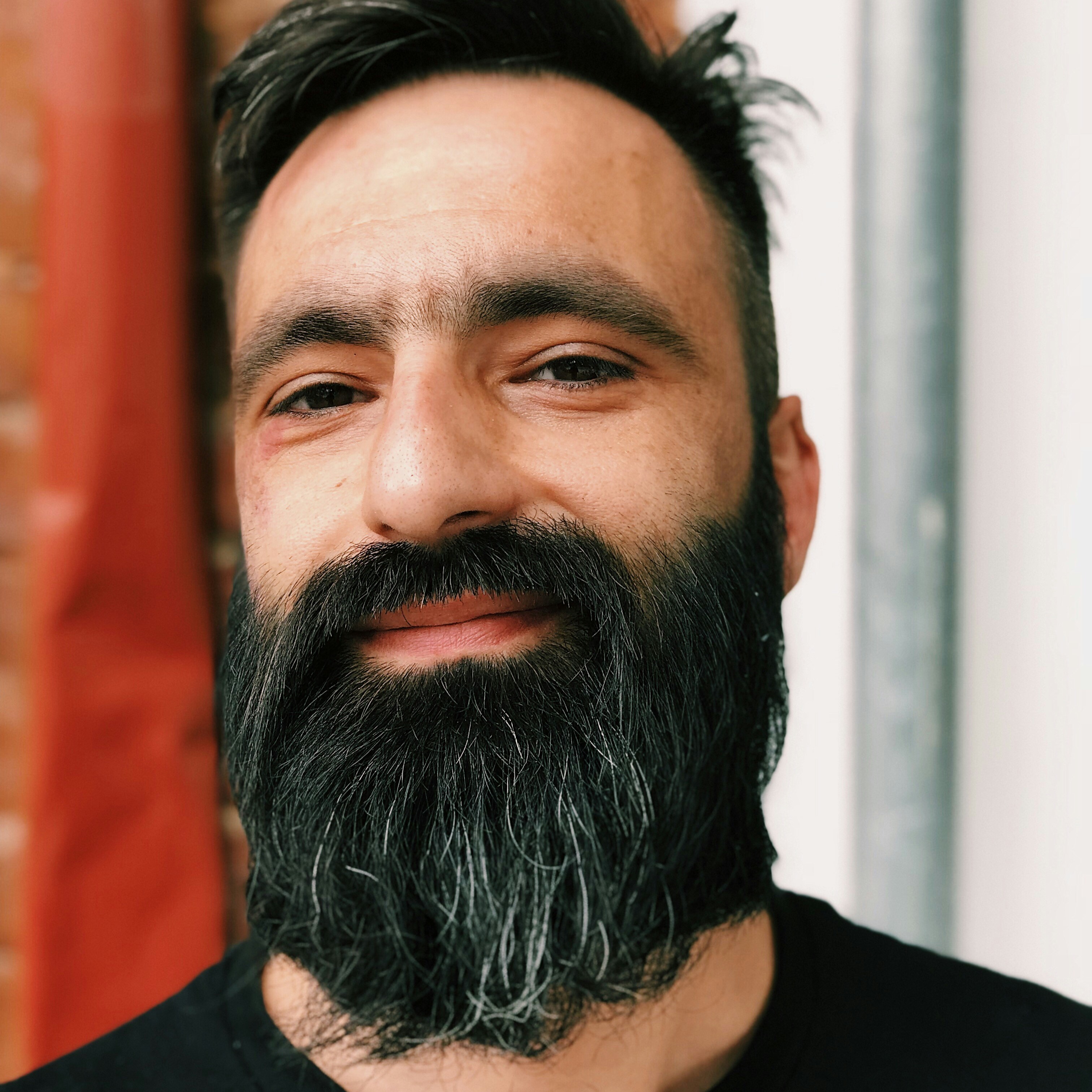} &
    \includegraphics[width=0.13\textwidth]{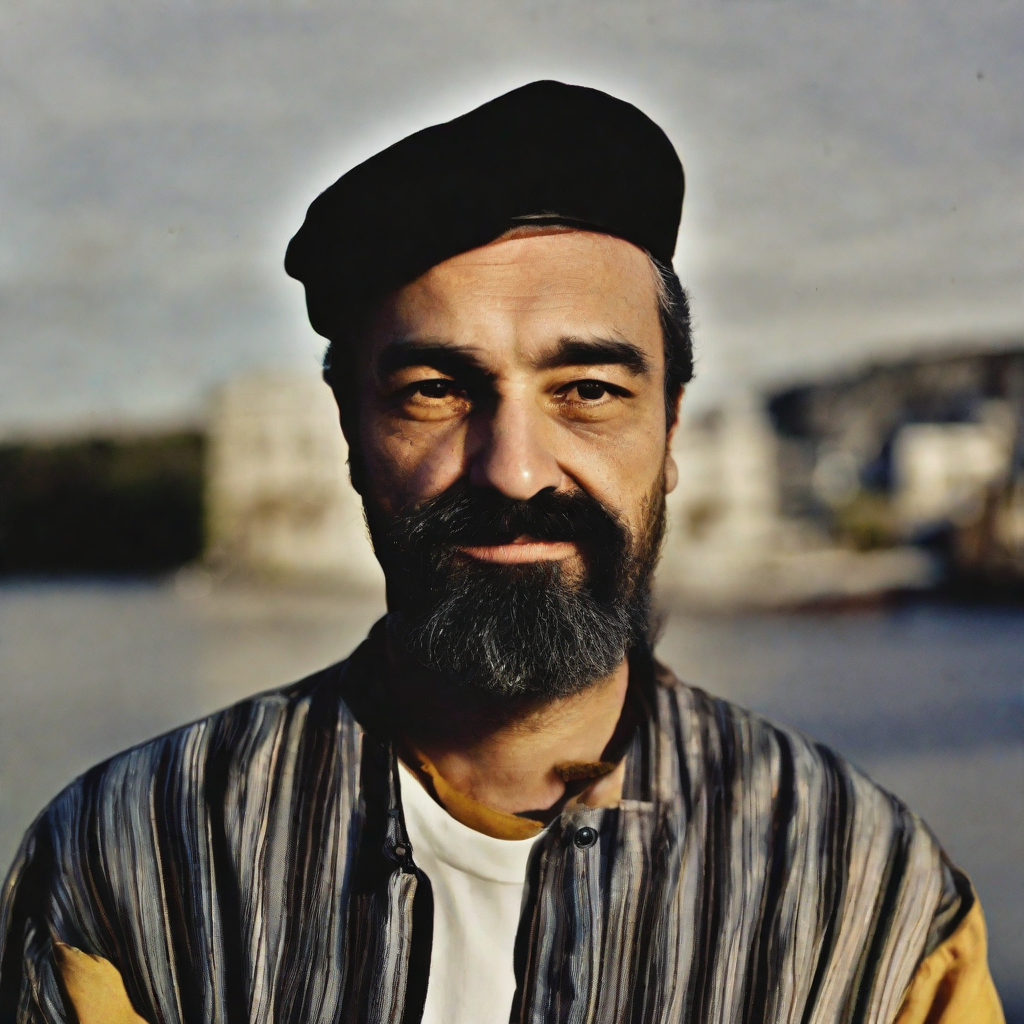} &
    \includegraphics[width=0.13\textwidth]{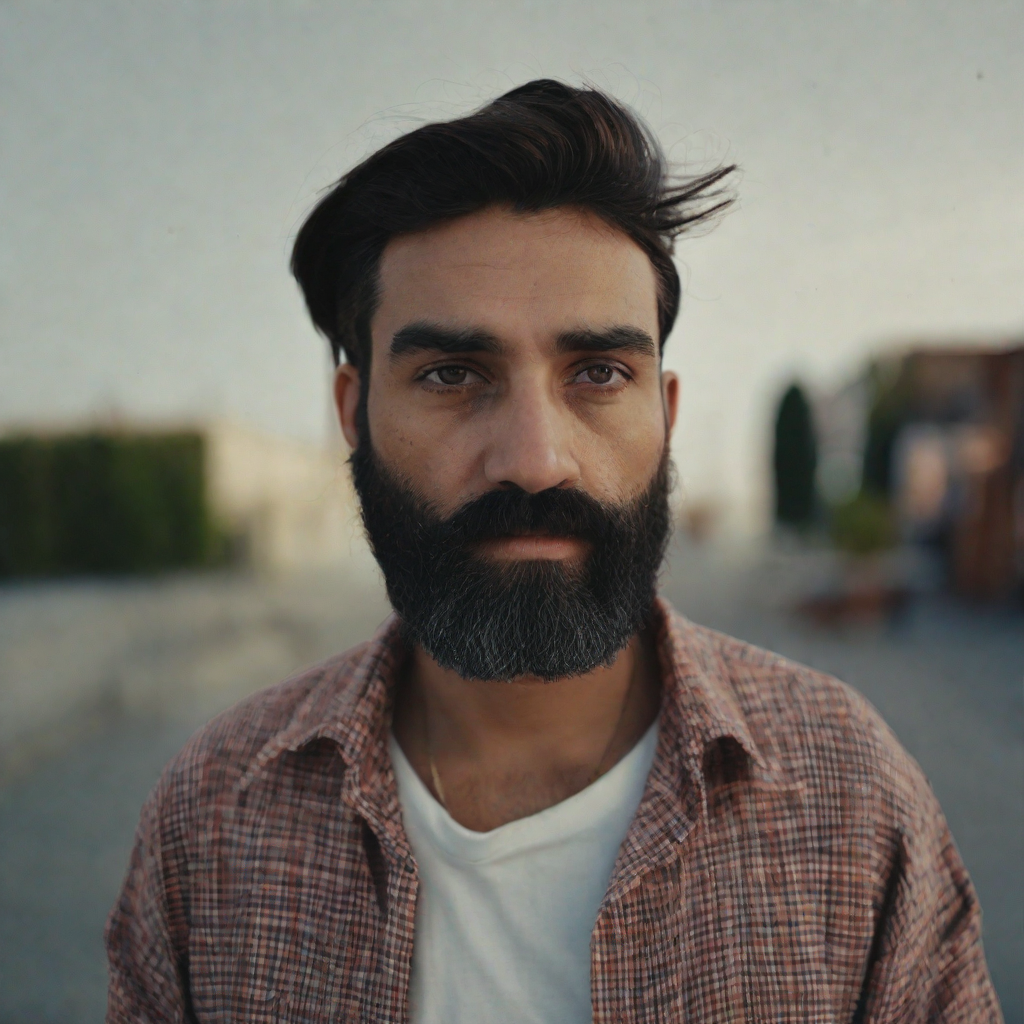} &

    \includegraphics[width=0.13\textwidth]{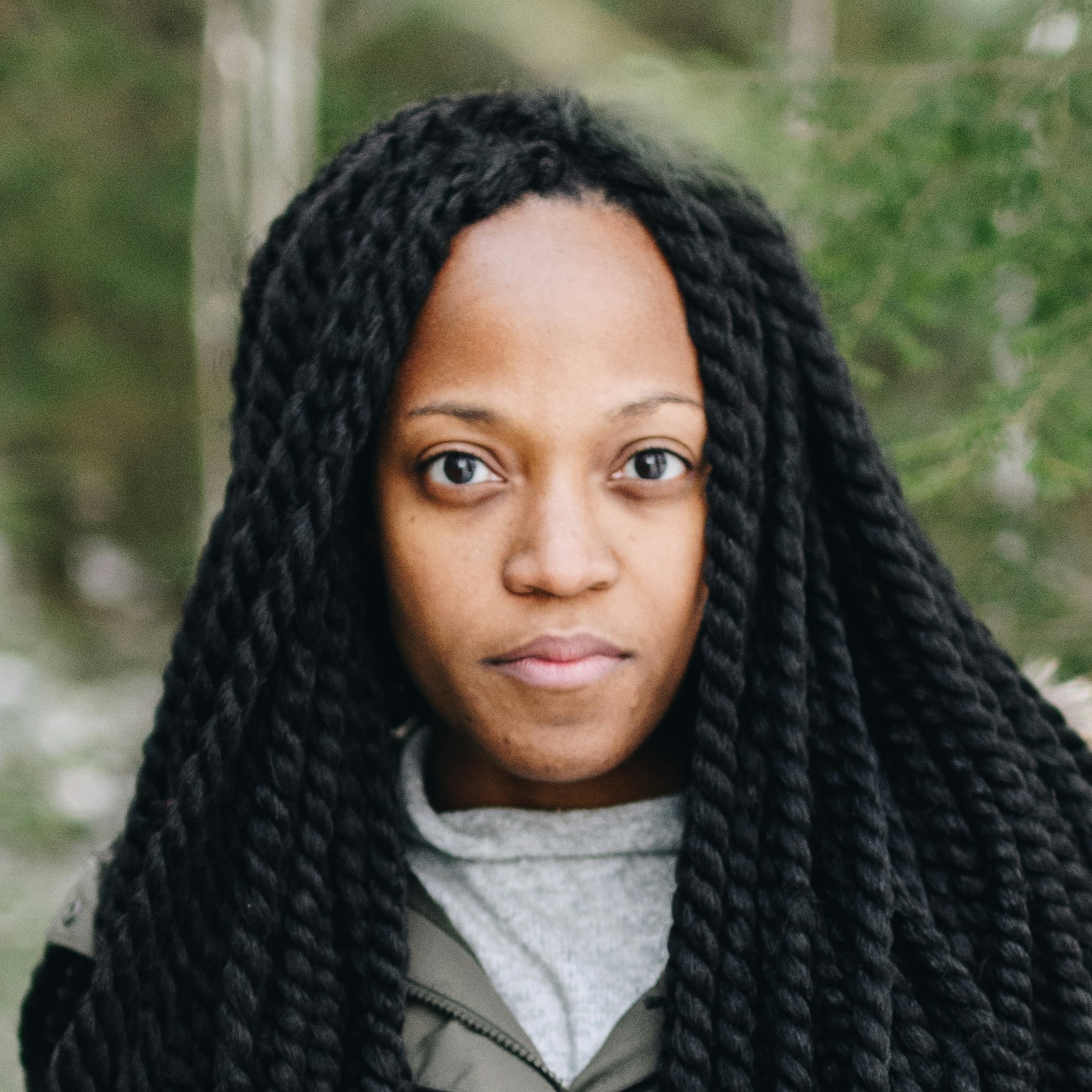} &
    \includegraphics[width=0.13\textwidth]{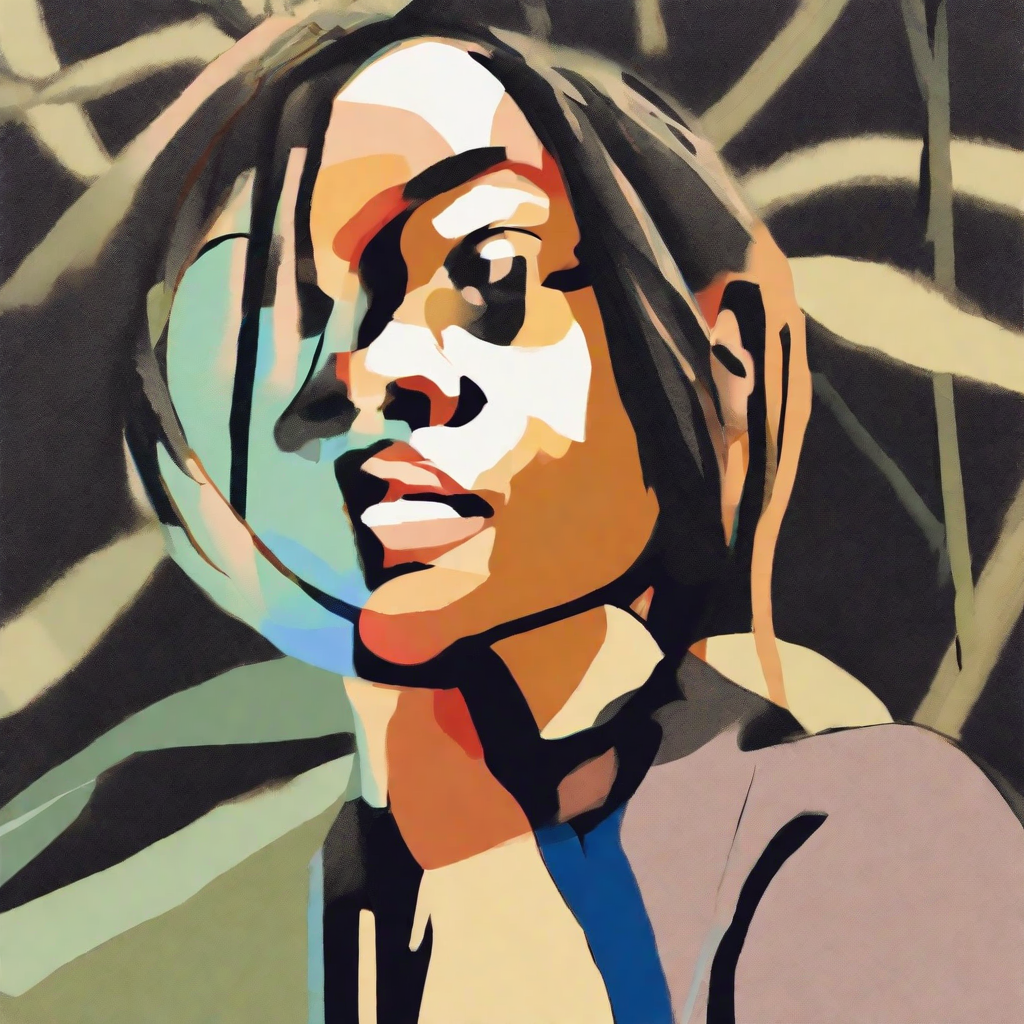} &
    \includegraphics[width=0.13\textwidth]{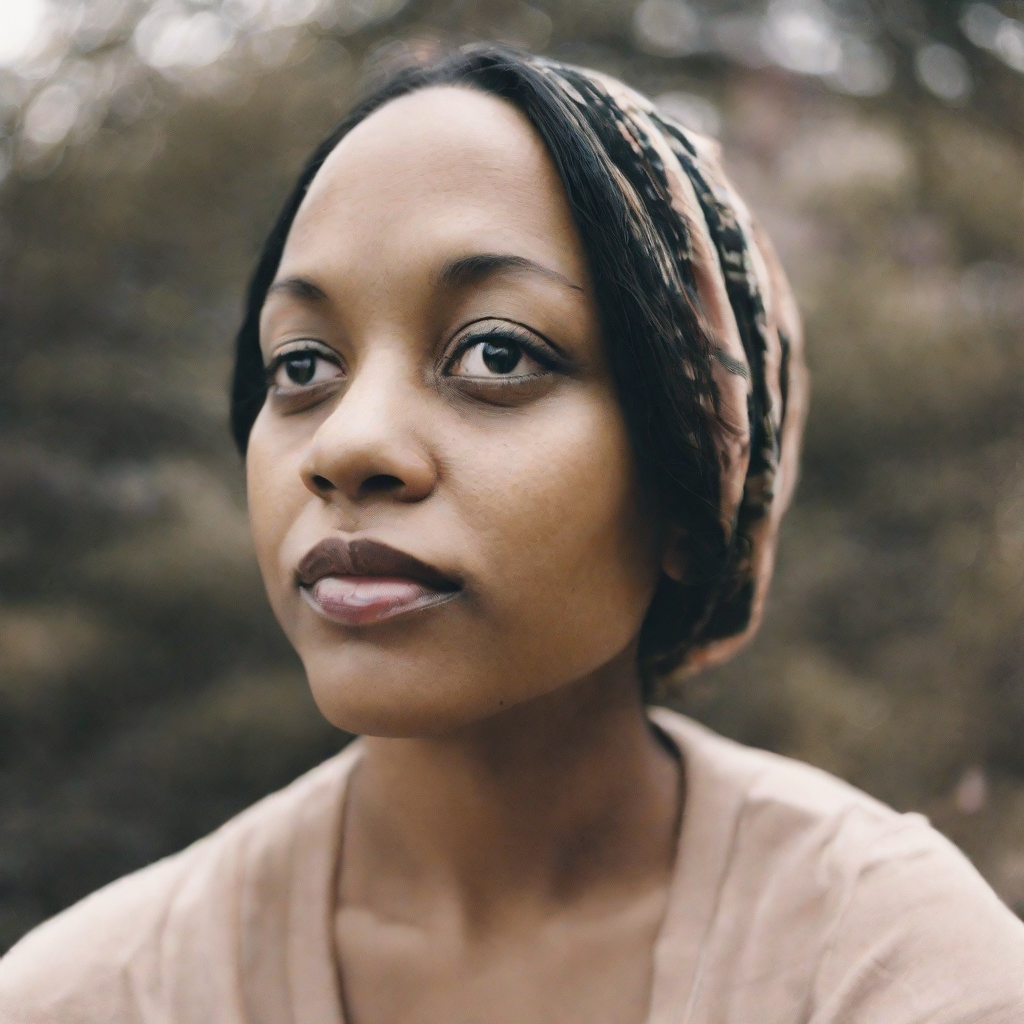}
    
    \\

    \raisebox{0.053\linewidth}{\rotatebox[origin=t]{90}{\fontsize{8pt}{8pt}\selectfont\begin{tabular}{c@{}c@{}c@{}c@{}} ID Guidance \end{tabular}}} &

    \includegraphics[width=0.13\textwidth]{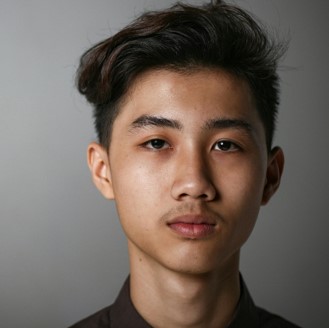} &
    \includegraphics[width=0.13\textwidth]{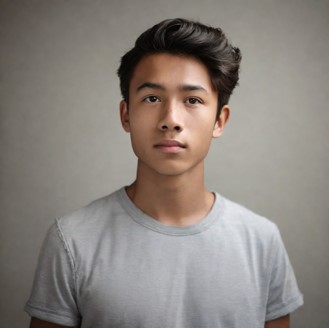} &
    \includegraphics[width=0.13\textwidth]{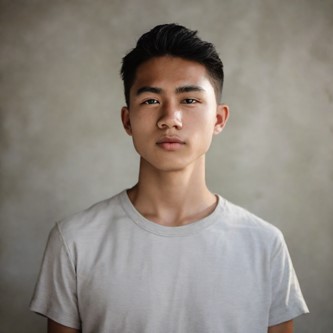} &

    \includegraphics[width=0.13\textwidth]{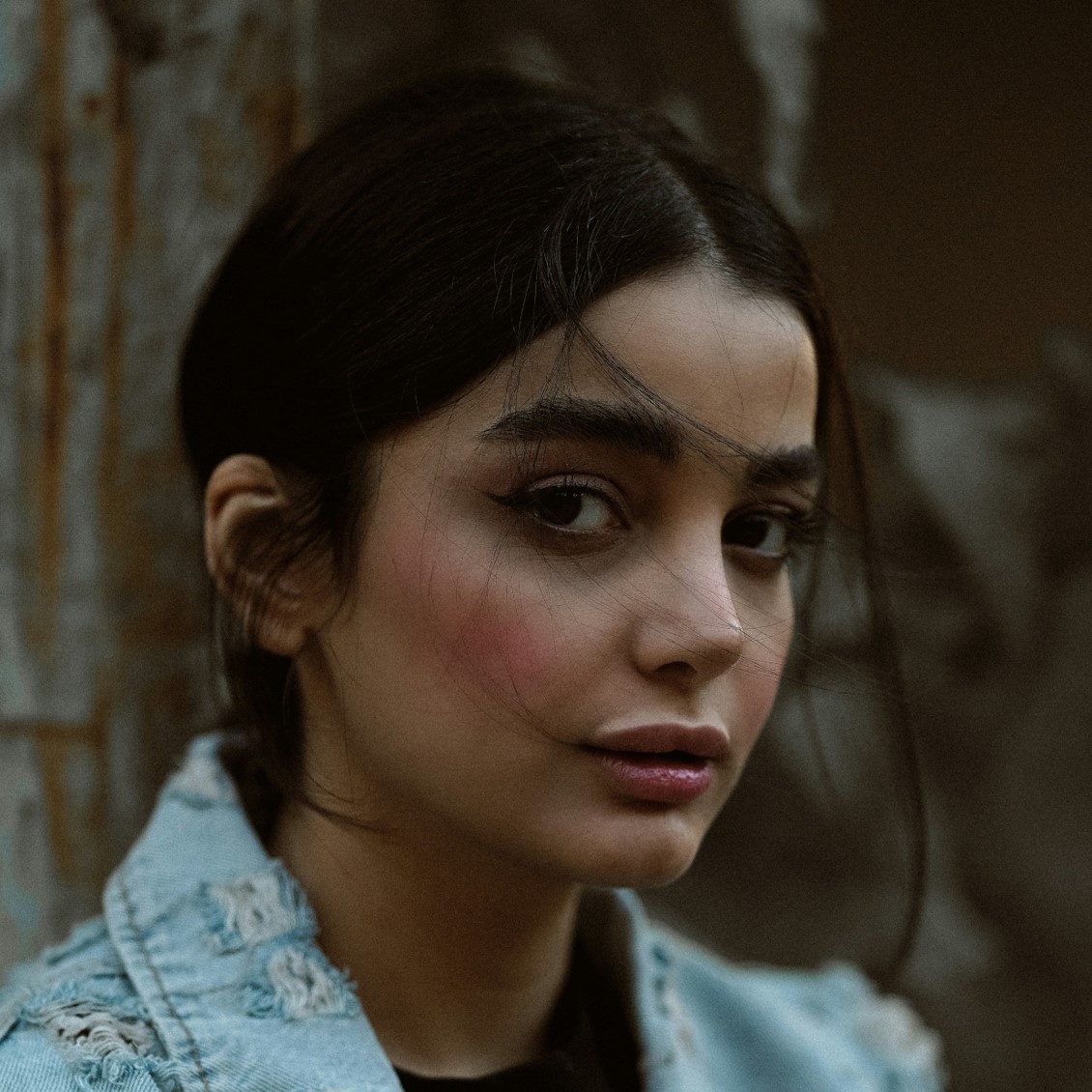} &
    \includegraphics[width=0.13\textwidth]{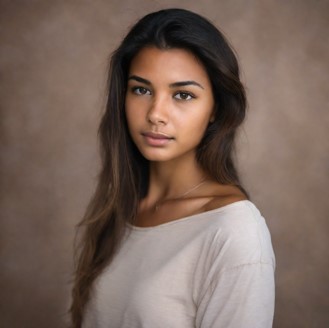} &
    \includegraphics[width=0.13\textwidth]{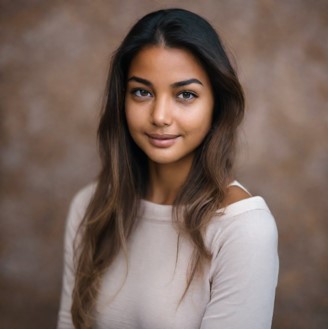}
    
    \\
    &
     &   
    {{\begin{tabular}{c}\small  $ID_s=0.18$\end{tabular}}} &   
    {{\begin{tabular}{c}\small  $ID_s=0.38$\end{tabular}}} &   
     &   
    {{\begin{tabular}{c}\small  $ID_s=0.13$\end{tabular}}} &   
    {{\begin{tabular}{c}\small $ID_s=0.42$\end{tabular}}} \\

    \end{tabular}
    
    }
    \caption{\textbf{LCM-Lookahead Guidance.} Results of classifier guidance when using different classifiers on top of our LCM-Lookahead, or standard $\hat{x}_0$ approximation. Each classifier preserves different attributes of the guiding image. $\hat{x}_0$ guidance may result in reduced quality or visible artifacts. Identity similarity values ($\uparrow$, measured using \cite{huang2020curricularface}) are shown at the bottom.
    }
    \label{fig:toy_guide}
\end{figure*}

\section{Experiments}
\label{sec:experiments}

\subsection{Classifer guidance with LCM-Lookahead}
We begin with an exploration of the LCM-Lookahead mechanism using a toy experiment, where we investigate its application to classifier guidance. Although this training-free method is less effective compared to encoder-based approaches, it serves as a simple use-case in which we can analyze the lookahead mechanism and discern its potential.
Specifically, we follow~\cite{wallace2023doodl} and apply repeated guidance iterations on an early diffusion time step ($t=44$ out of $50$ DDIM~\cite{song2020denoising} steps). At each iteration, we denoise the current latents using LCM-LoRA~\cite{luo2023lcmlora}, decode them to an image, apply our pixel-space loss, and backpropagate to modify the latents. Final results are generated by continuing unguided-DDIM sampling  with SDXL~\cite{podell2024sdxl} for the remaining diffusion steps. \cref{fig:toy_guide} shows several outputs of such guidance, using a perceptual LPIPS loss~\cite{zhang2018perceptual}, CLIP loss~\cite{radford2021learning} and an identity loss~\cite{deng2019arcface}. Each loss preserves different attributes of the guiding image. LPIPS preserves the semantic layout of the image, CLIP preserves semantic attributes such as facial hair, and the identity loss explicitly improves facial similarity. When applying the same guidance to the single-step DDIM-approximations ($\hat{x}_0$ in \cref{fig:toy_guide}) we observe artifacts or reduced performance, attributed to the fact that $\hat{x}_0$ is typically blurry and discolored in early timesteps. The identity loss is particularly robust to blur, but its performance still improves by using a lookhead loss.

Next, we investigate the use of the lookahead identity loss to improve encoder-based personalization methods.

\subsection{Encoder evaluation}

\begin{figure*}[t]

    \centering
    \setlength{\tabcolsep}{1.5pt}
    {\footnotesize
    \begin{tabular}{c c c c c c}

        \includegraphics[width=0.15\textwidth,height=0.15\textwidth]{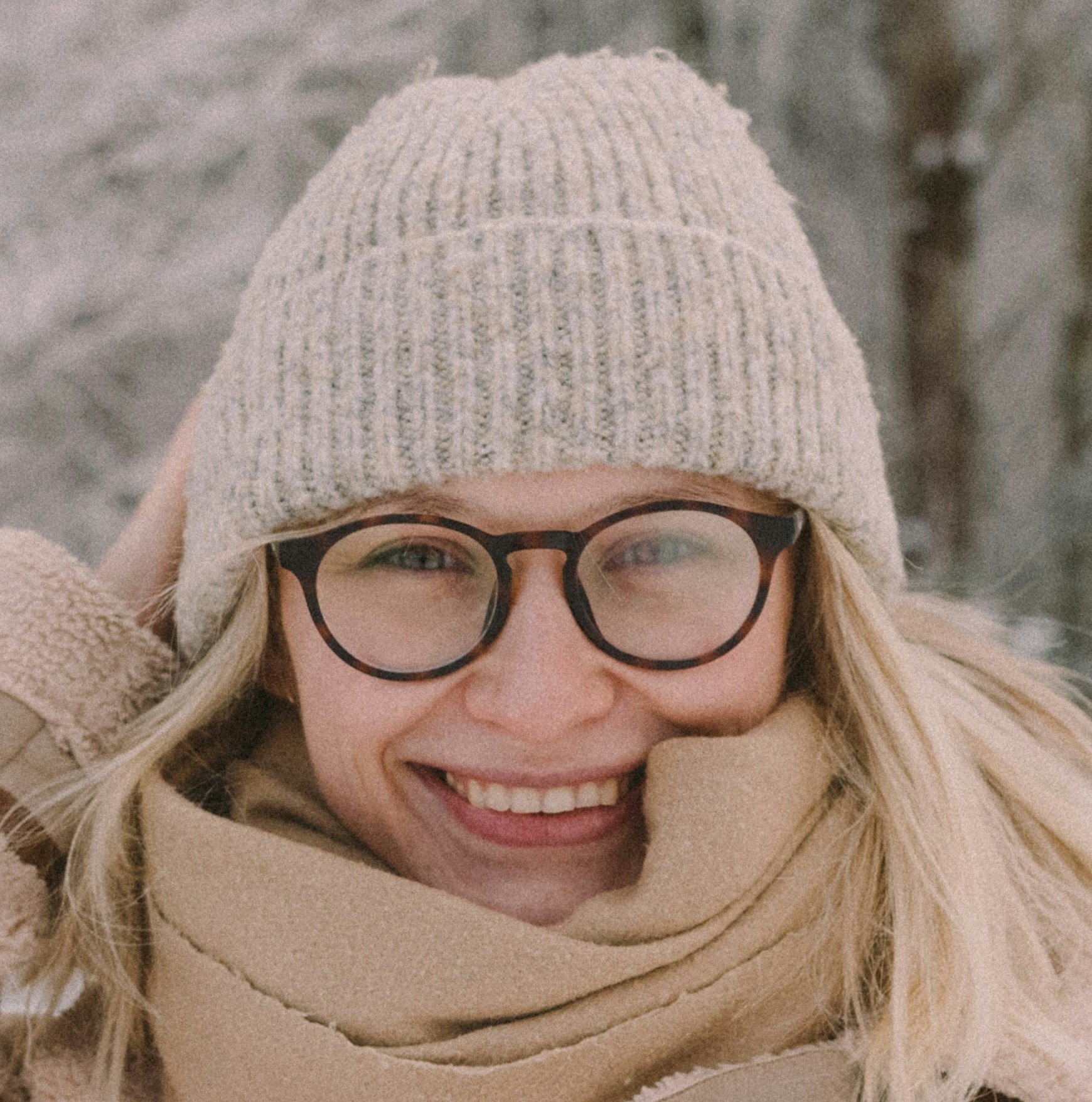} &
        \includegraphics[width=0.15\textwidth,height=0.15\textwidth]{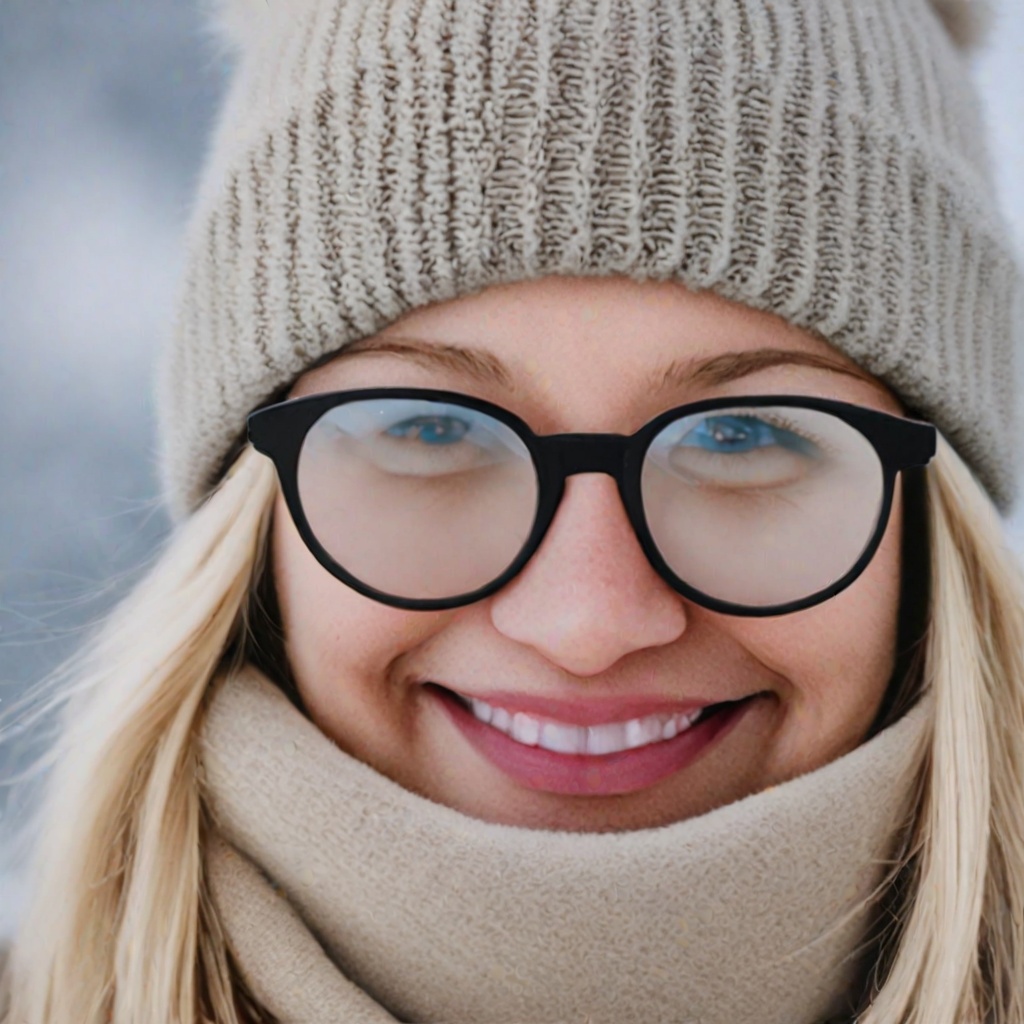} &
        \includegraphics[width=0.15\textwidth,height=0.15\textwidth]{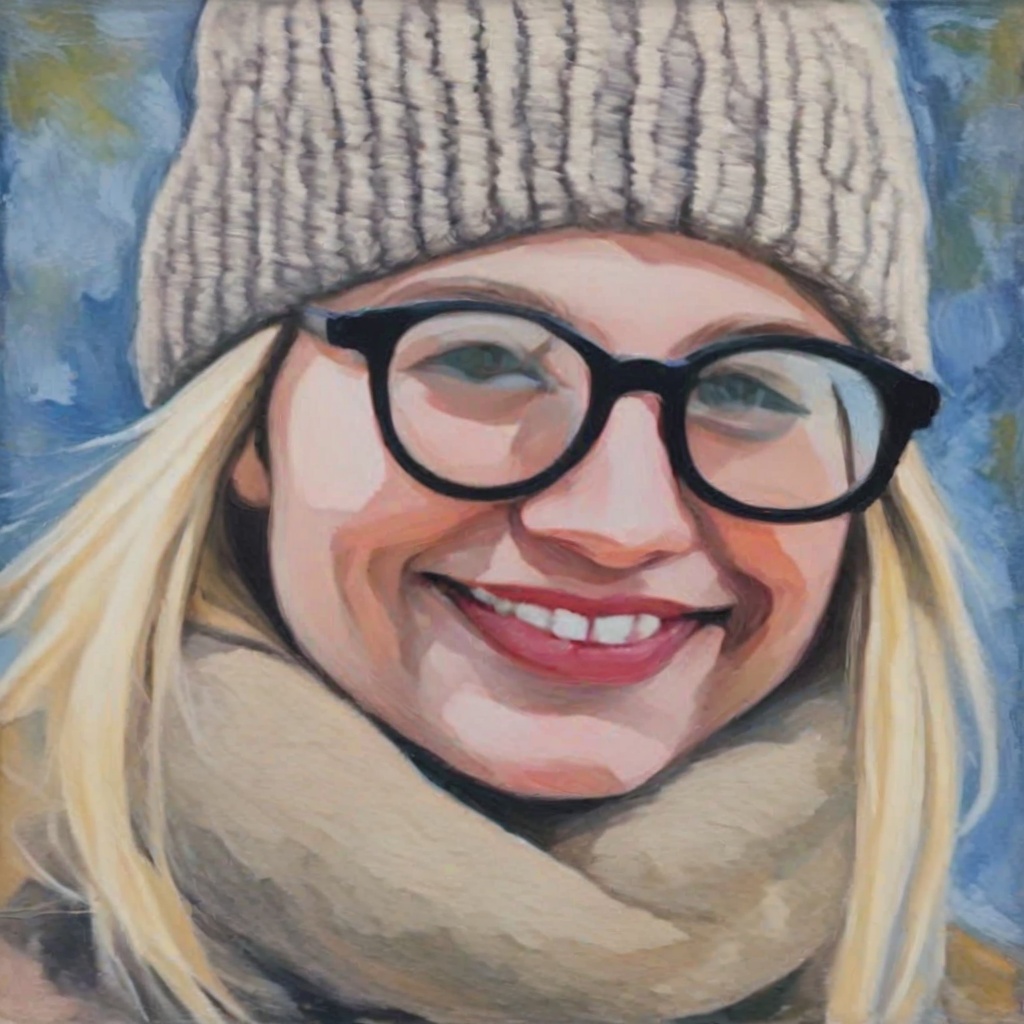} &
        \includegraphics[width=0.15\textwidth,height=0.15\textwidth]{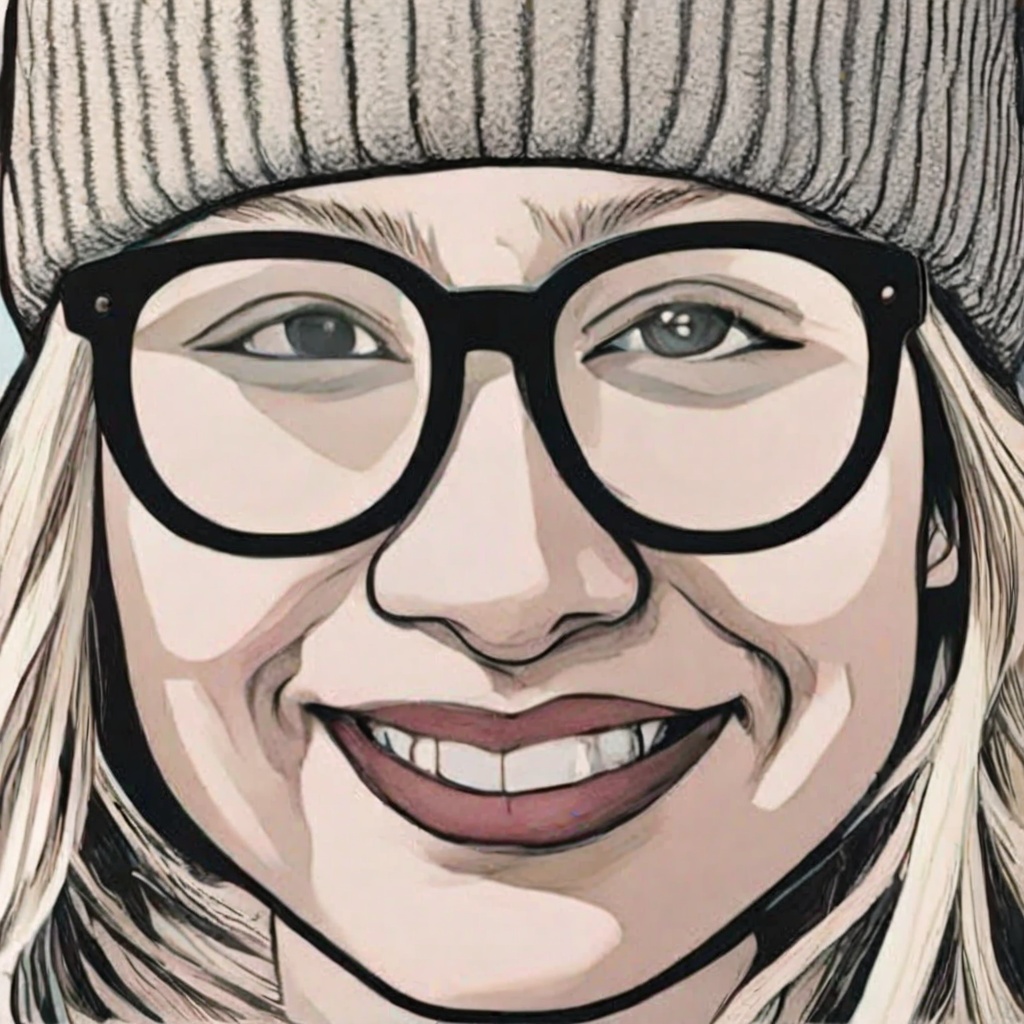} &
        \includegraphics[width=0.15\textwidth,height=0.15\textwidth]{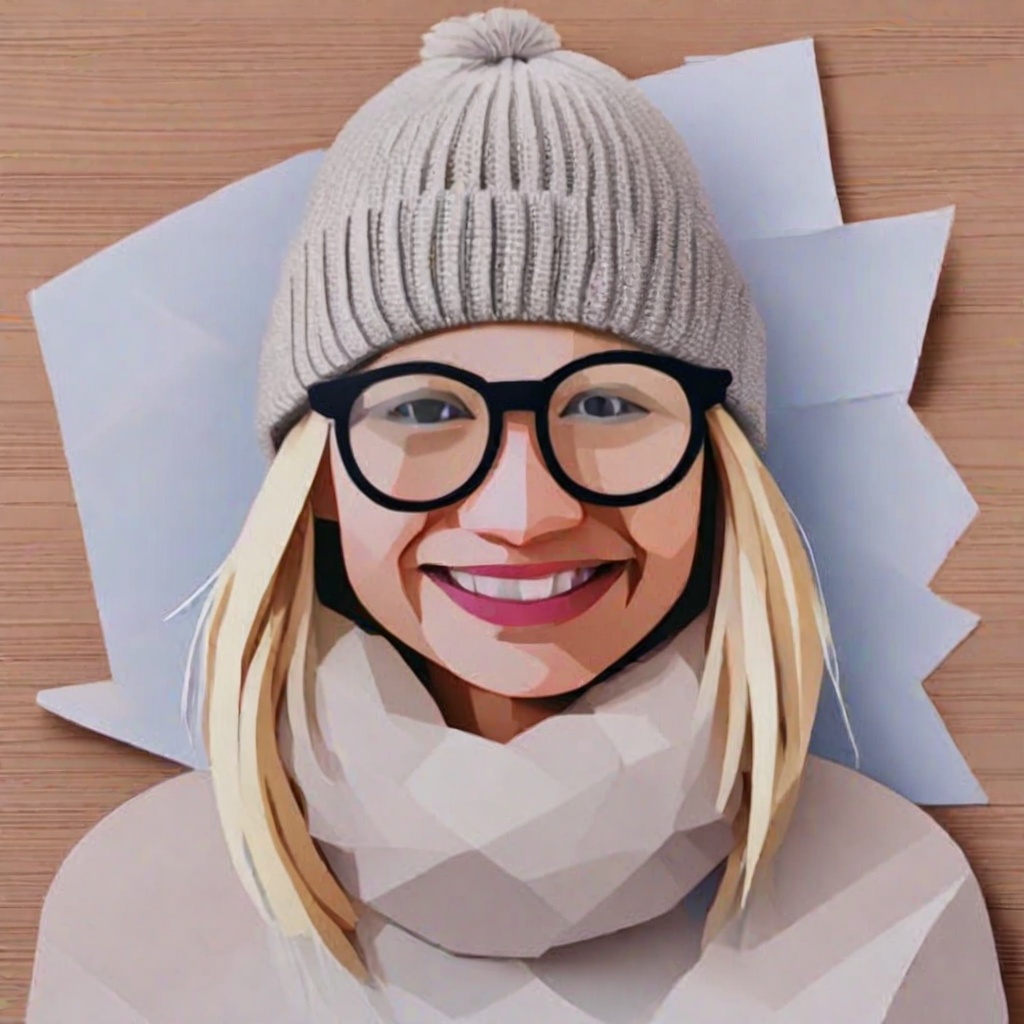} &
        \includegraphics[width=0.15\textwidth,height=0.15\textwidth]{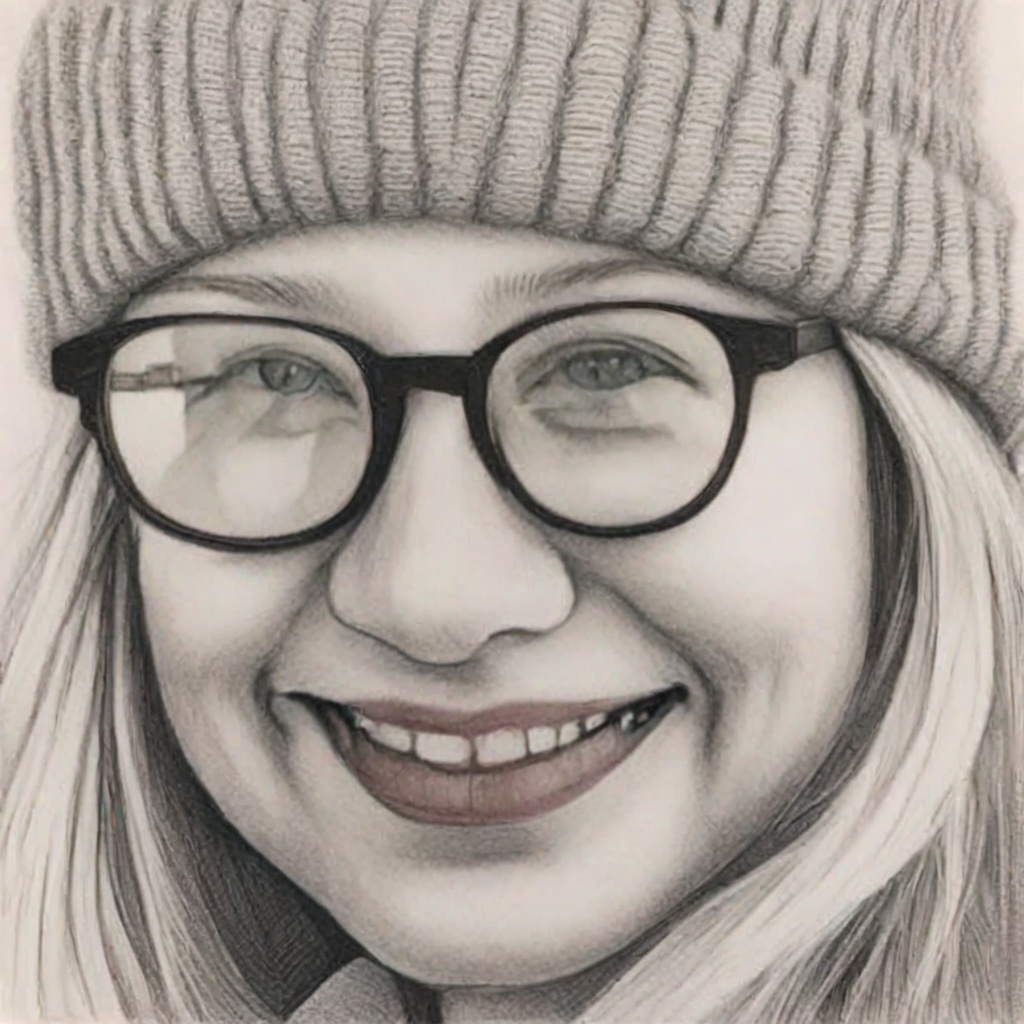} 
        \\

        \includegraphics[width=0.15\textwidth,height=0.15\textwidth]{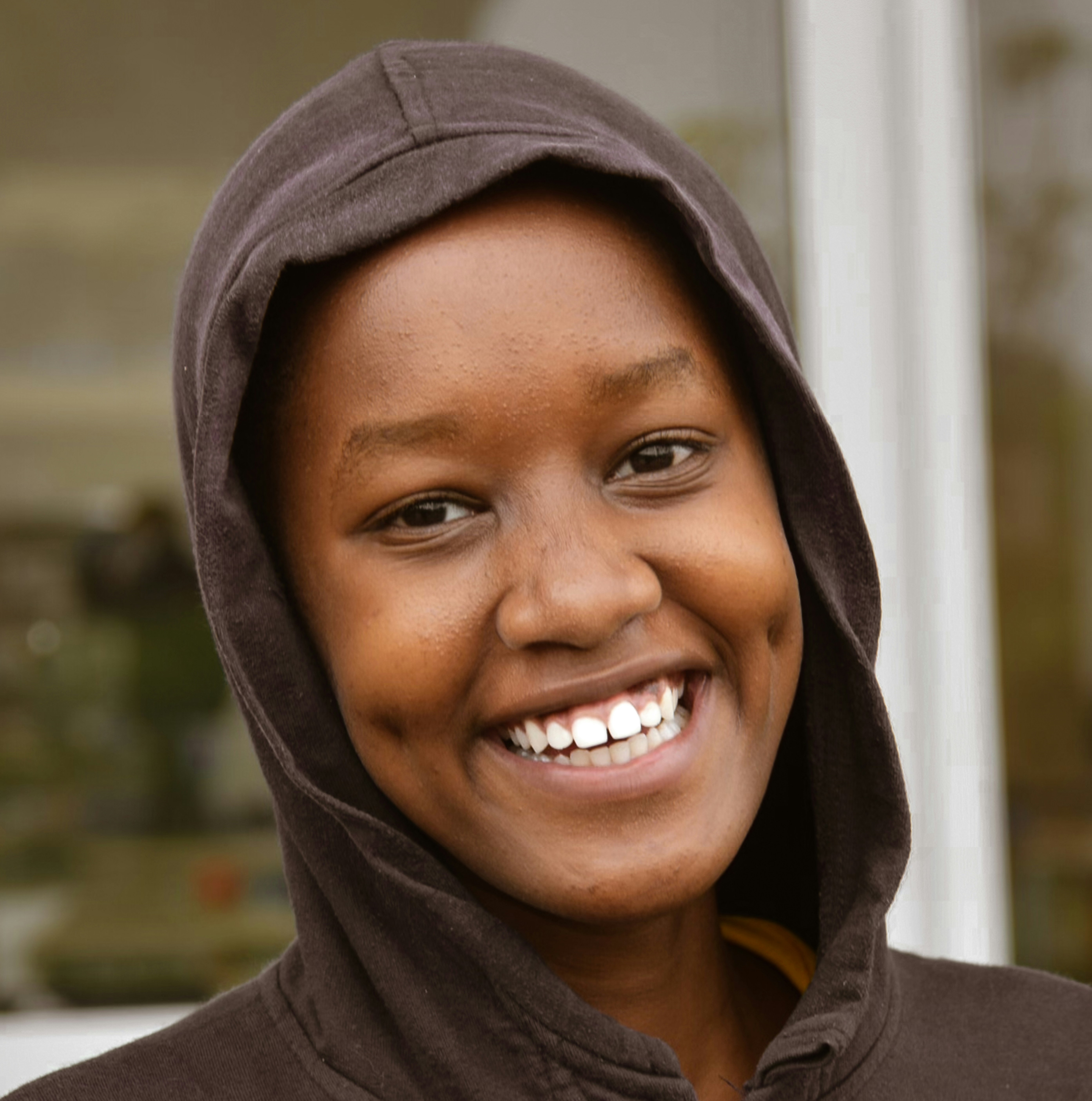} &
        \includegraphics[width=0.15\textwidth,height=0.15\textwidth]{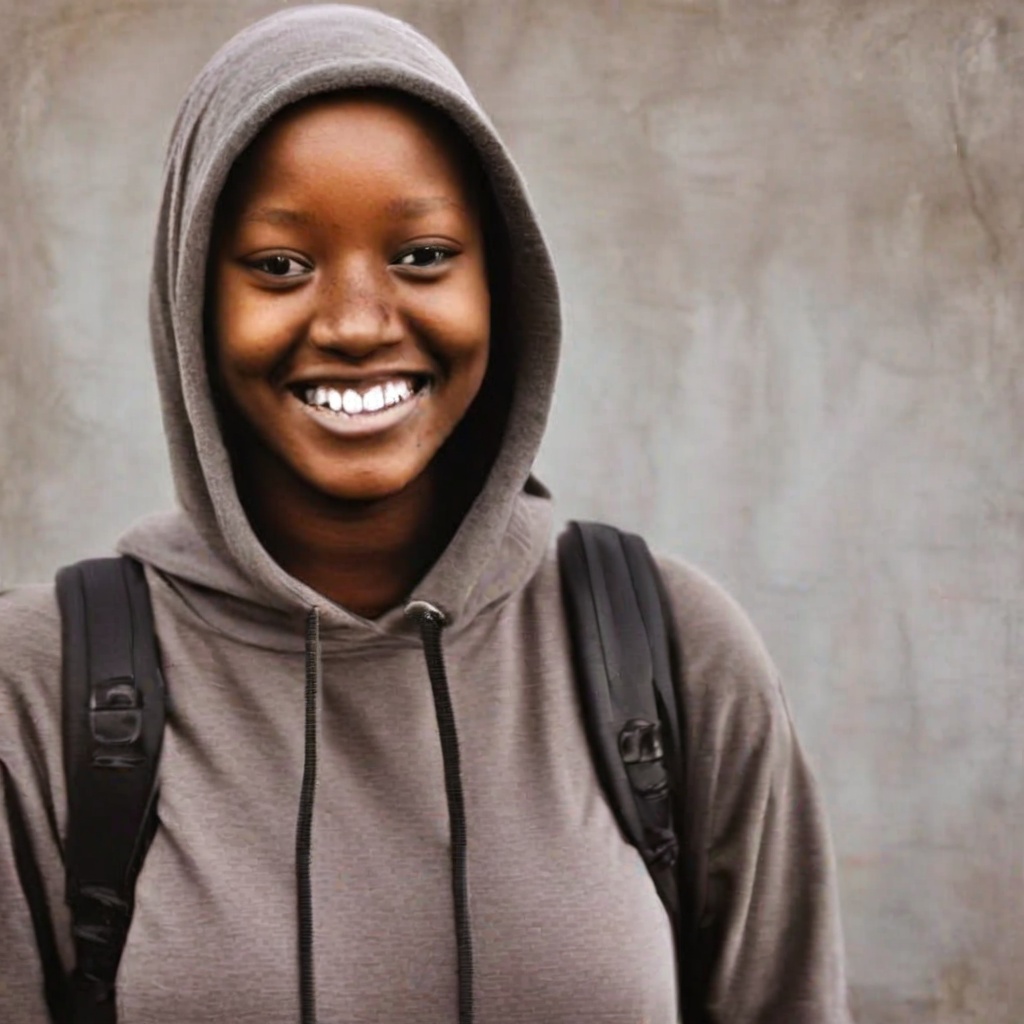} &
        \includegraphics[width=0.15\textwidth,height=0.15\textwidth]{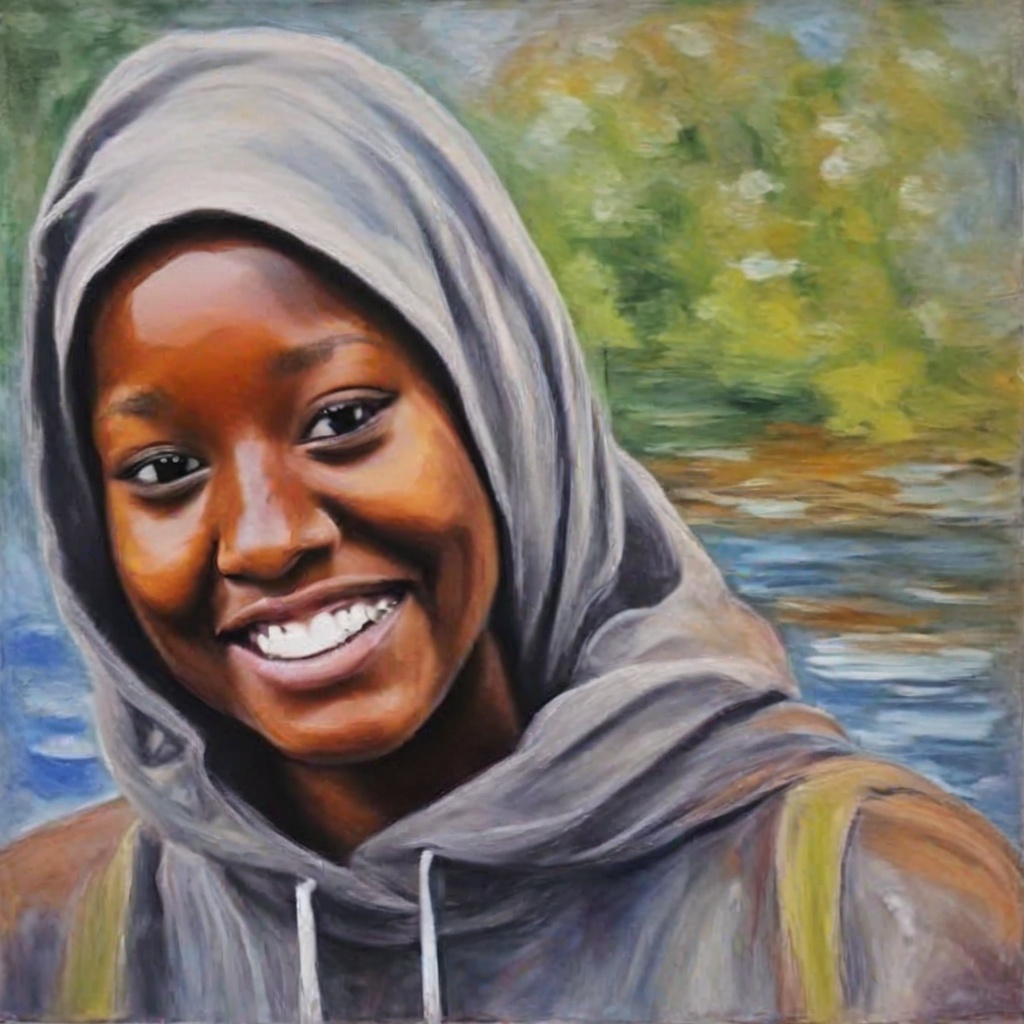} &
        \includegraphics[width=0.15\textwidth,height=0.15\textwidth]{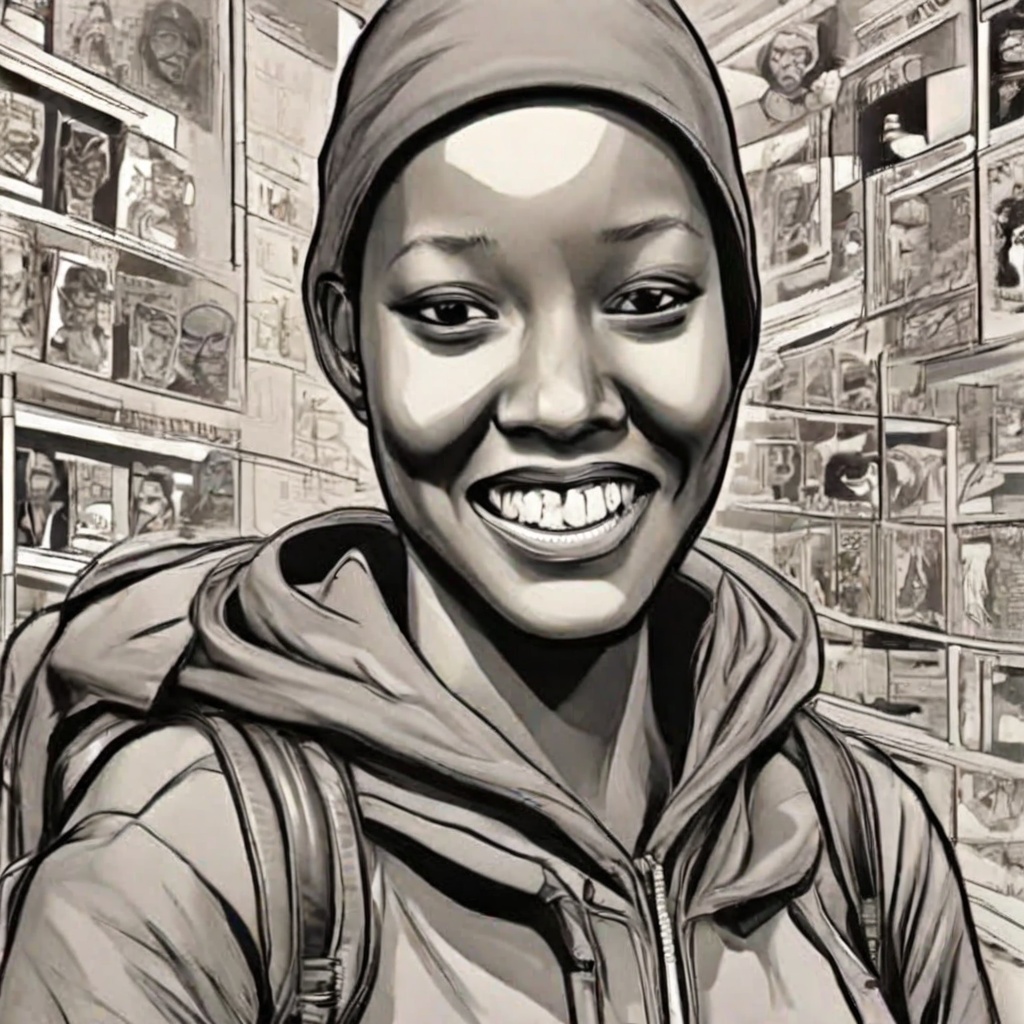} &
        \includegraphics[width=0.15\textwidth,height=0.15\textwidth]{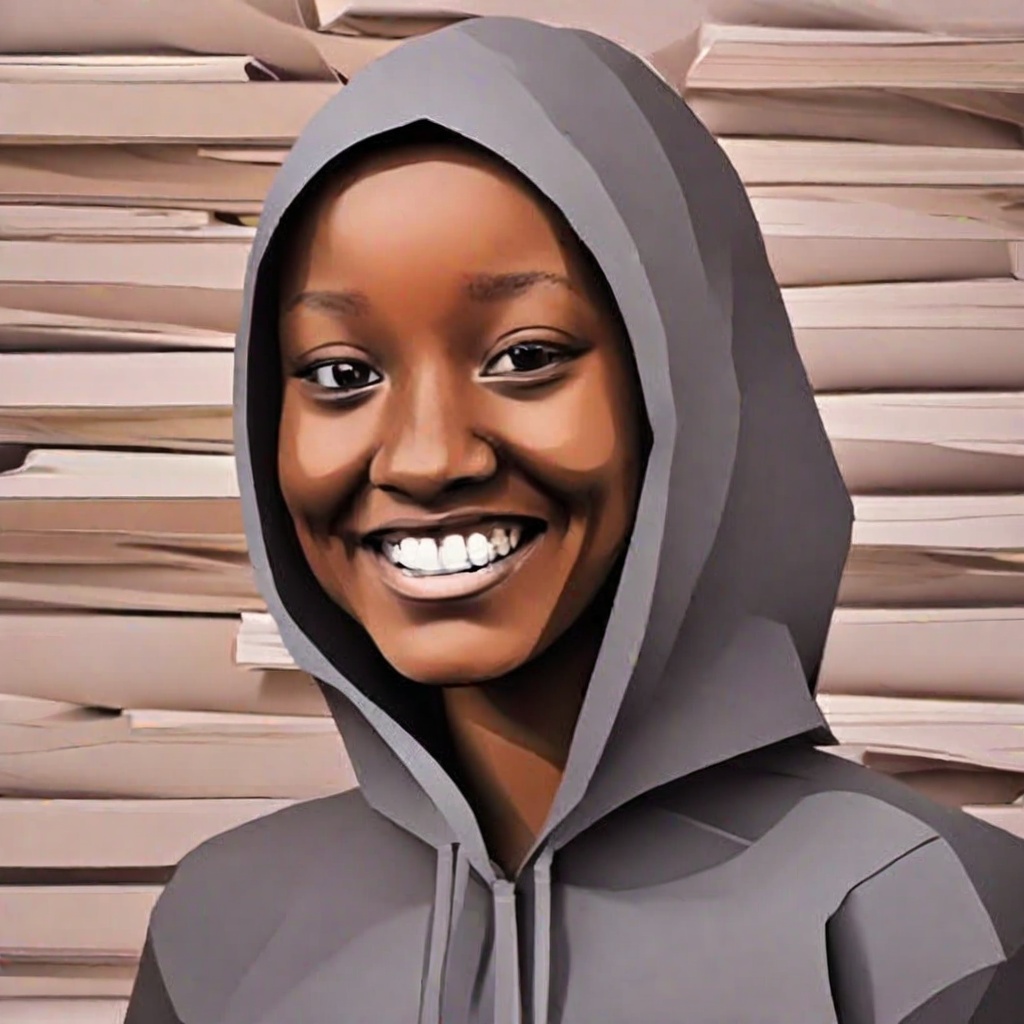} &
        \includegraphics[width=0.15\textwidth,height=0.15\textwidth]{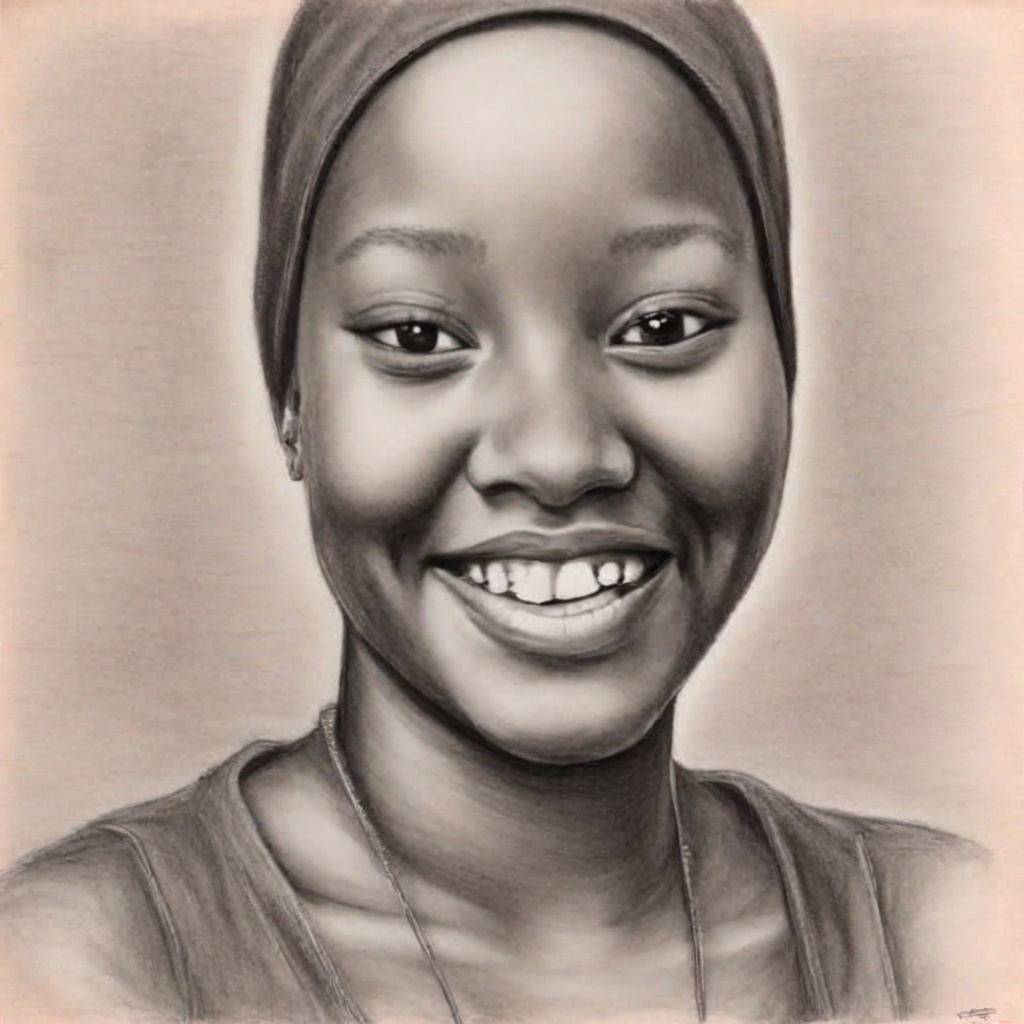} 
        \\

        \includegraphics[width=0.15\textwidth,height=0.15\textwidth]{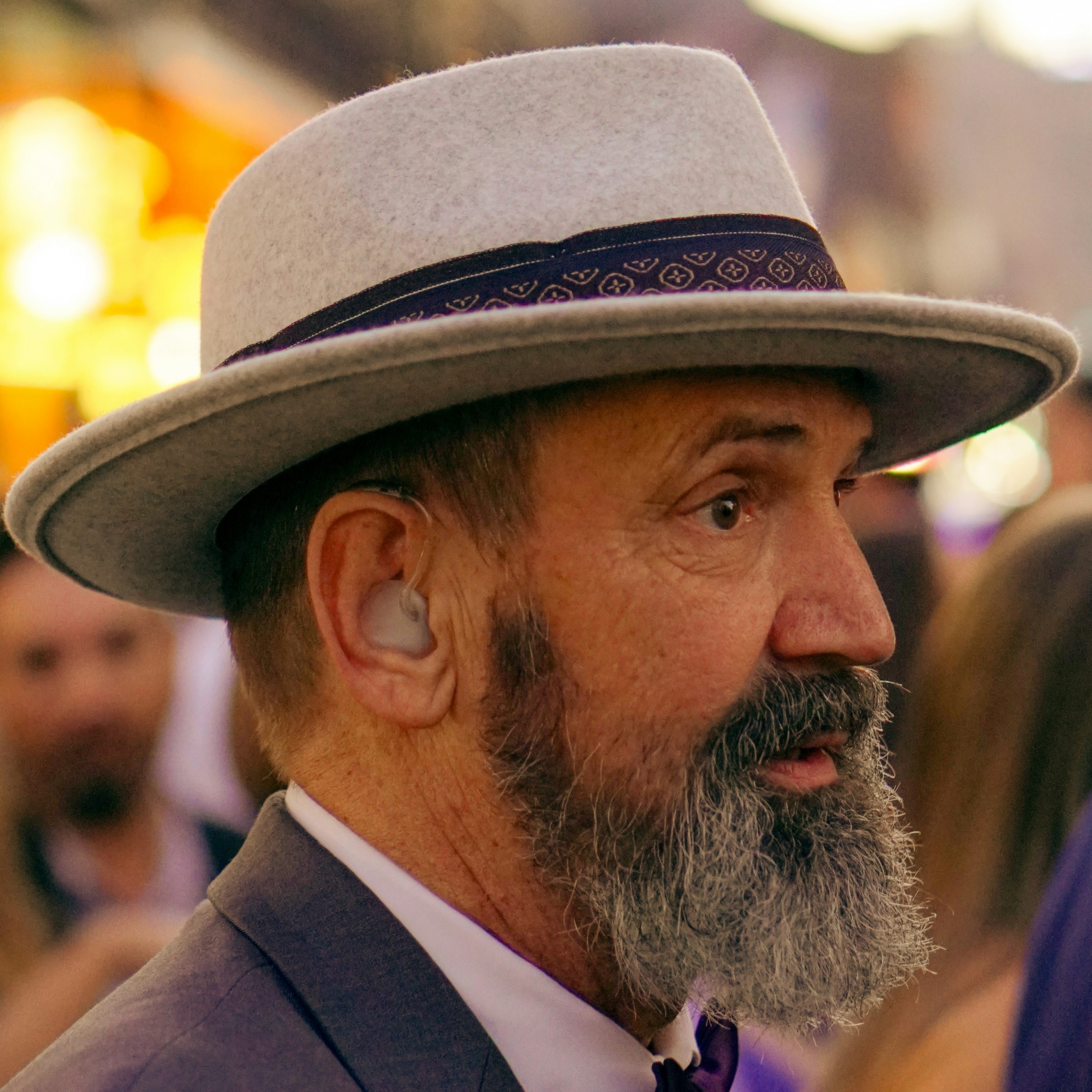} &
        \includegraphics[width=0.15\textwidth,height=0.15\textwidth]{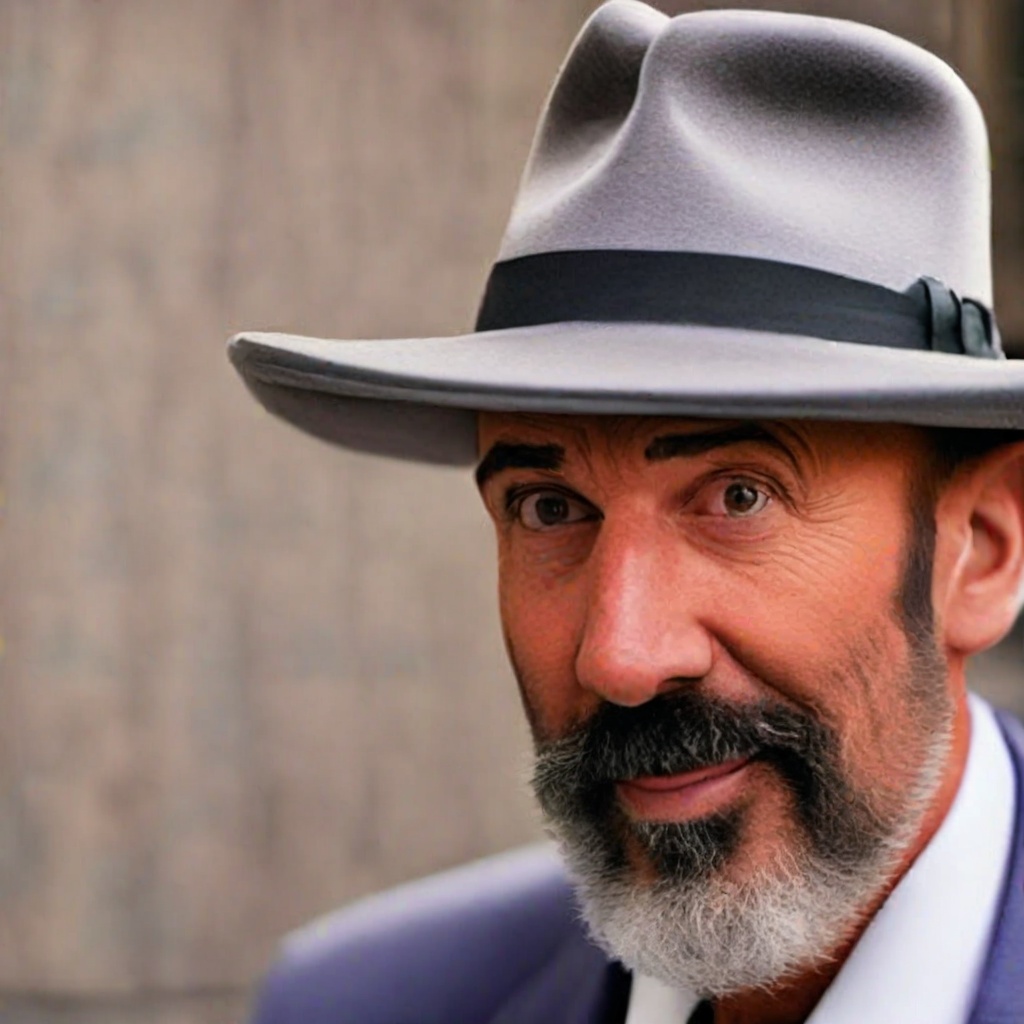} &
        \includegraphics[width=0.15\textwidth,height=0.15\textwidth]{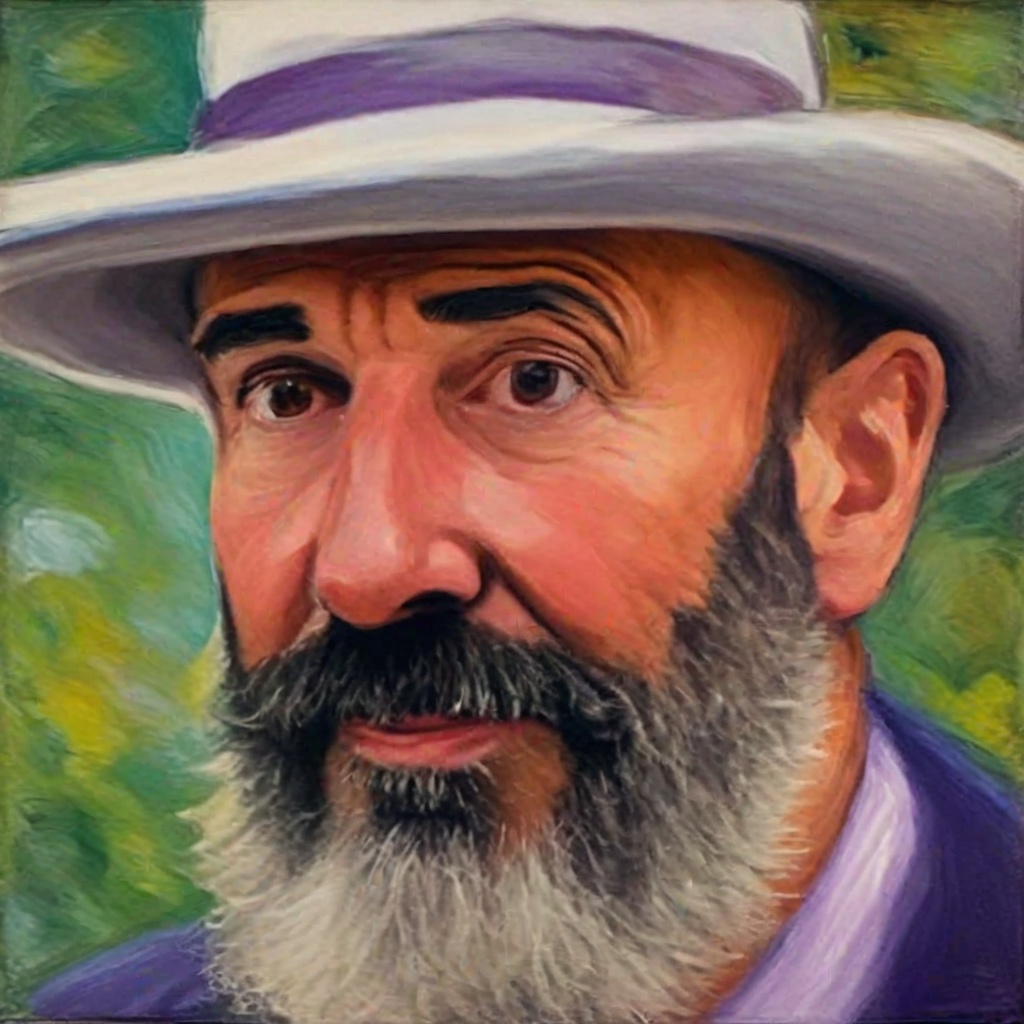} &
        \includegraphics[width=0.15\textwidth,height=0.15\textwidth]{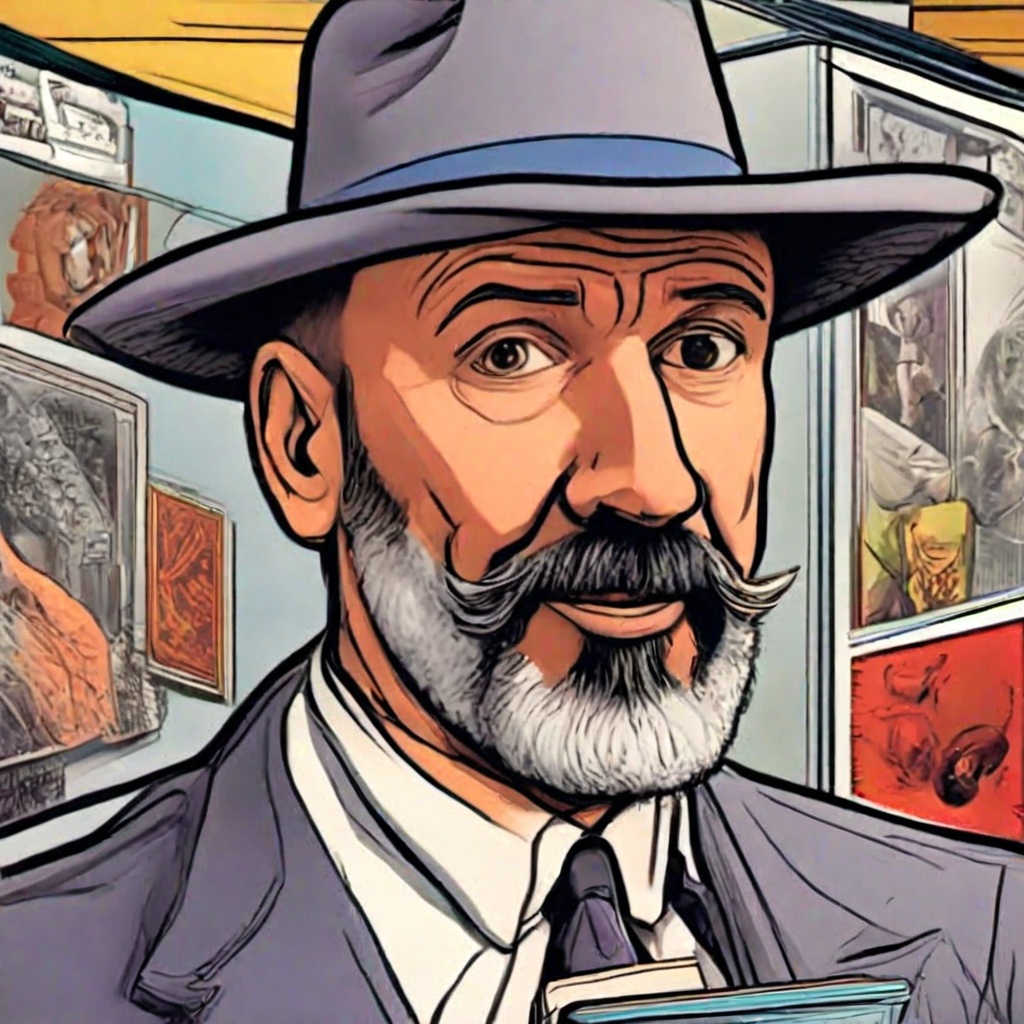} &
        \includegraphics[width=0.15\textwidth,height=0.15\textwidth]{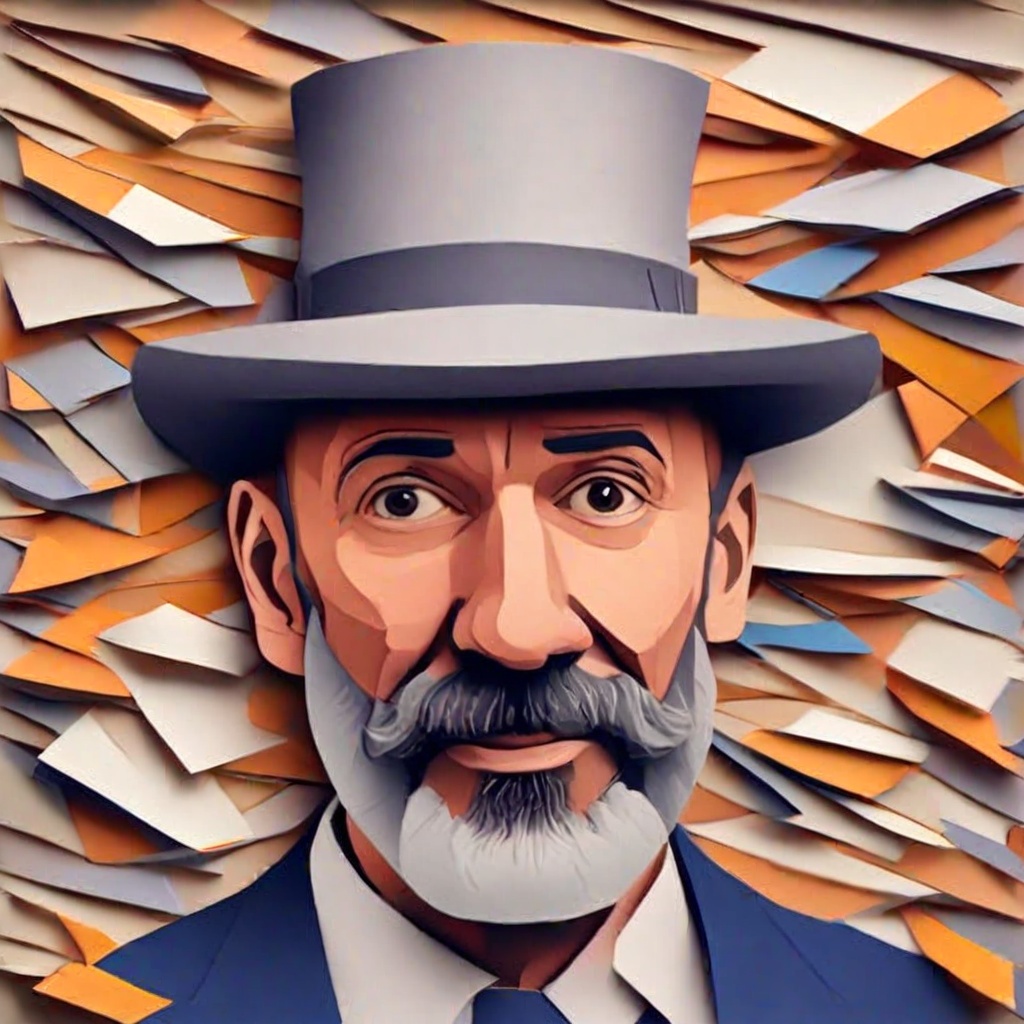} &
        \includegraphics[width=0.15\textwidth,height=0.15\textwidth]{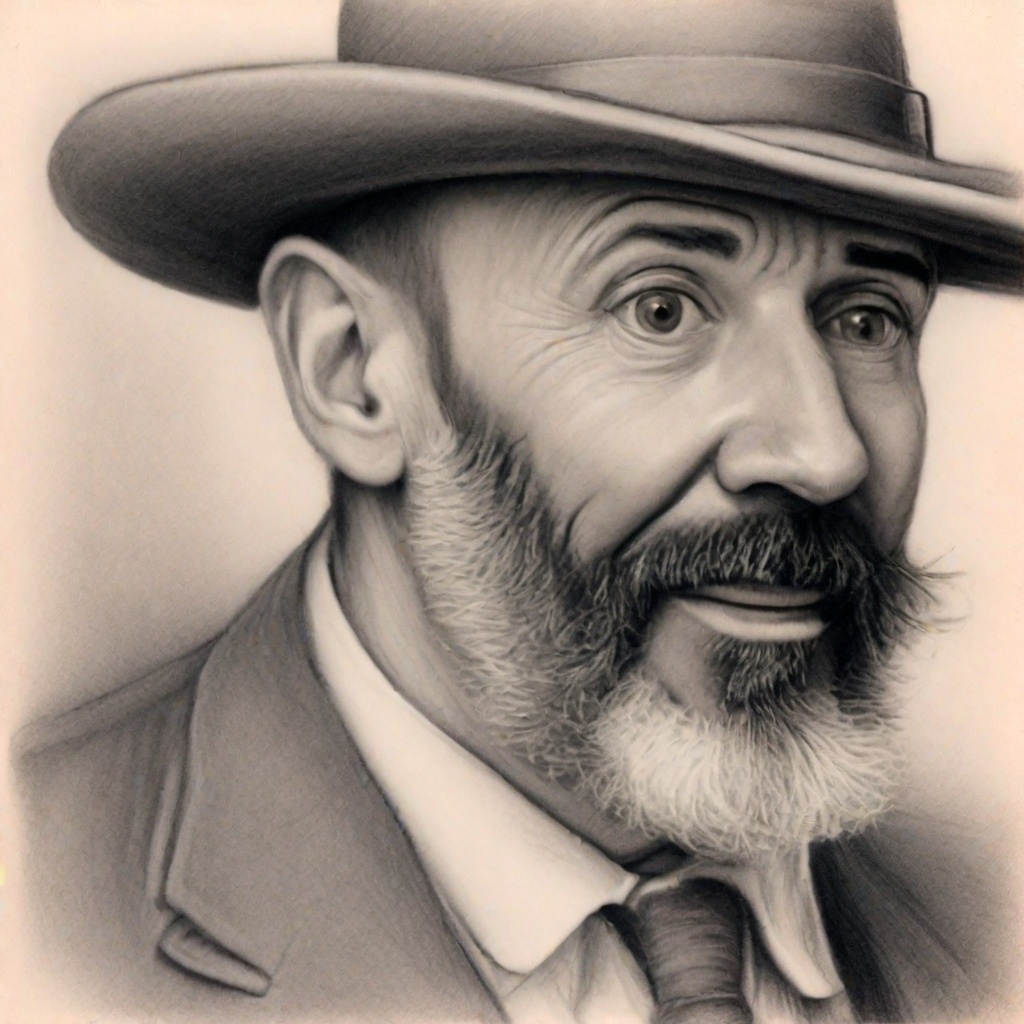} 
        \\

        \includegraphics[width=0.15\textwidth,height=0.15\textwidth]{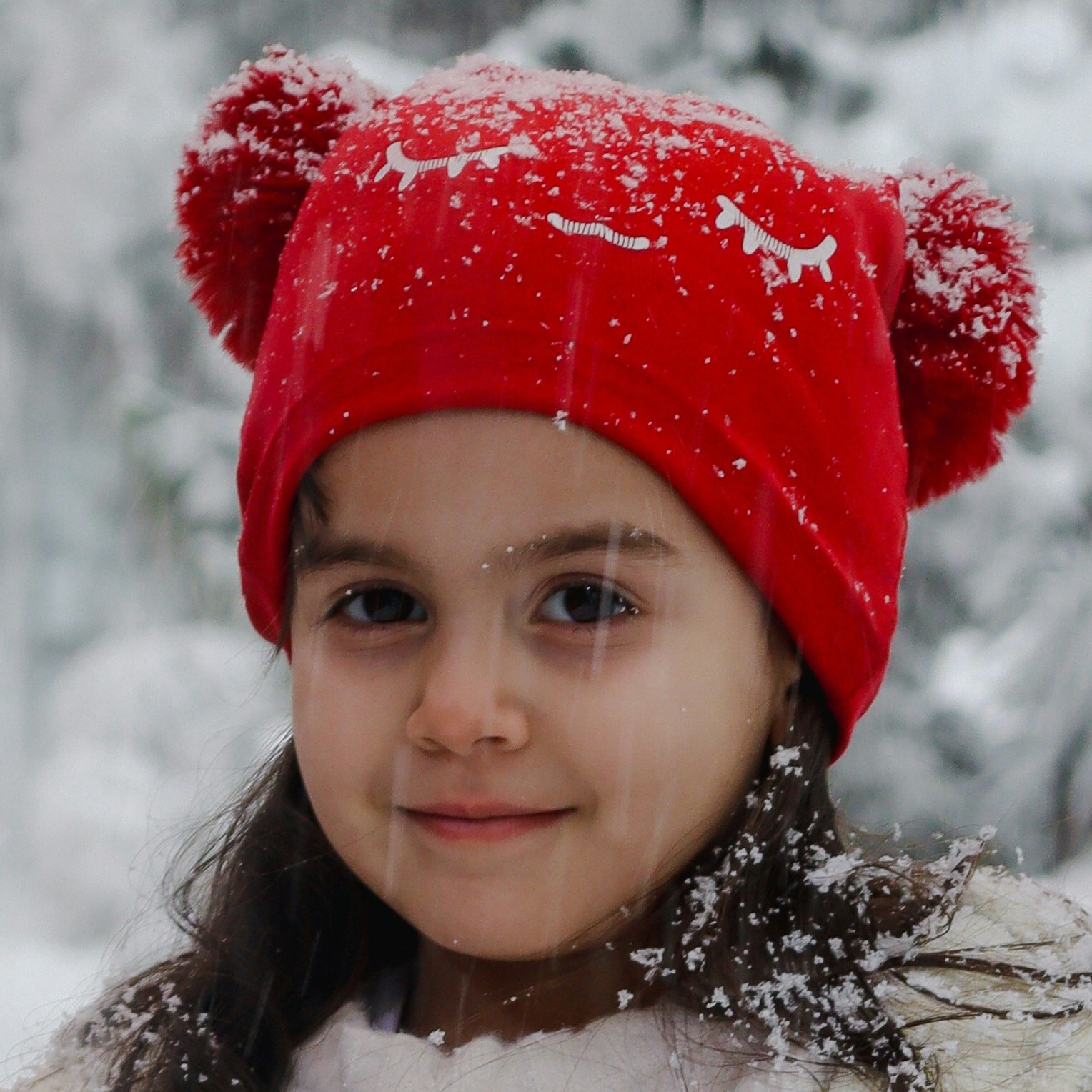} &
        \includegraphics[width=0.15\textwidth,height=0.15\textwidth]{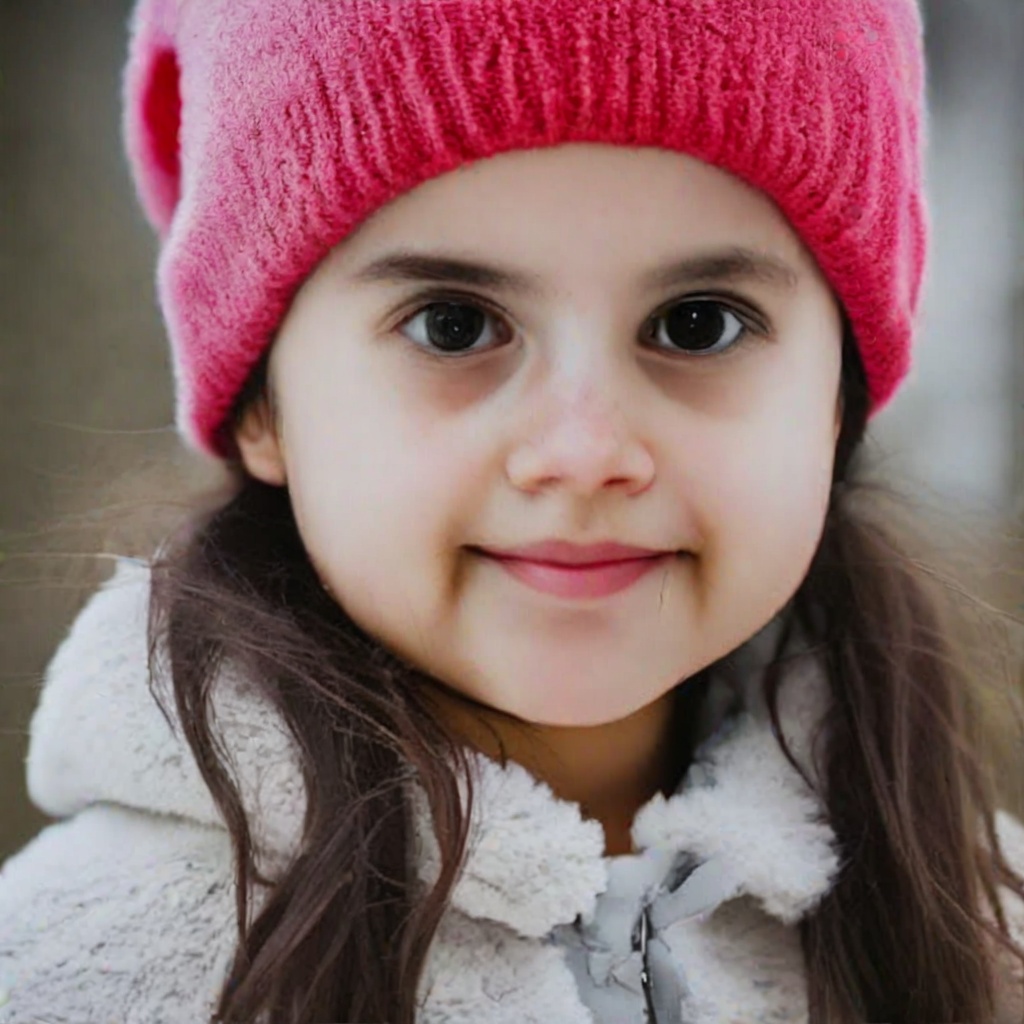} &
        \includegraphics[width=0.15\textwidth,height=0.15\textwidth]{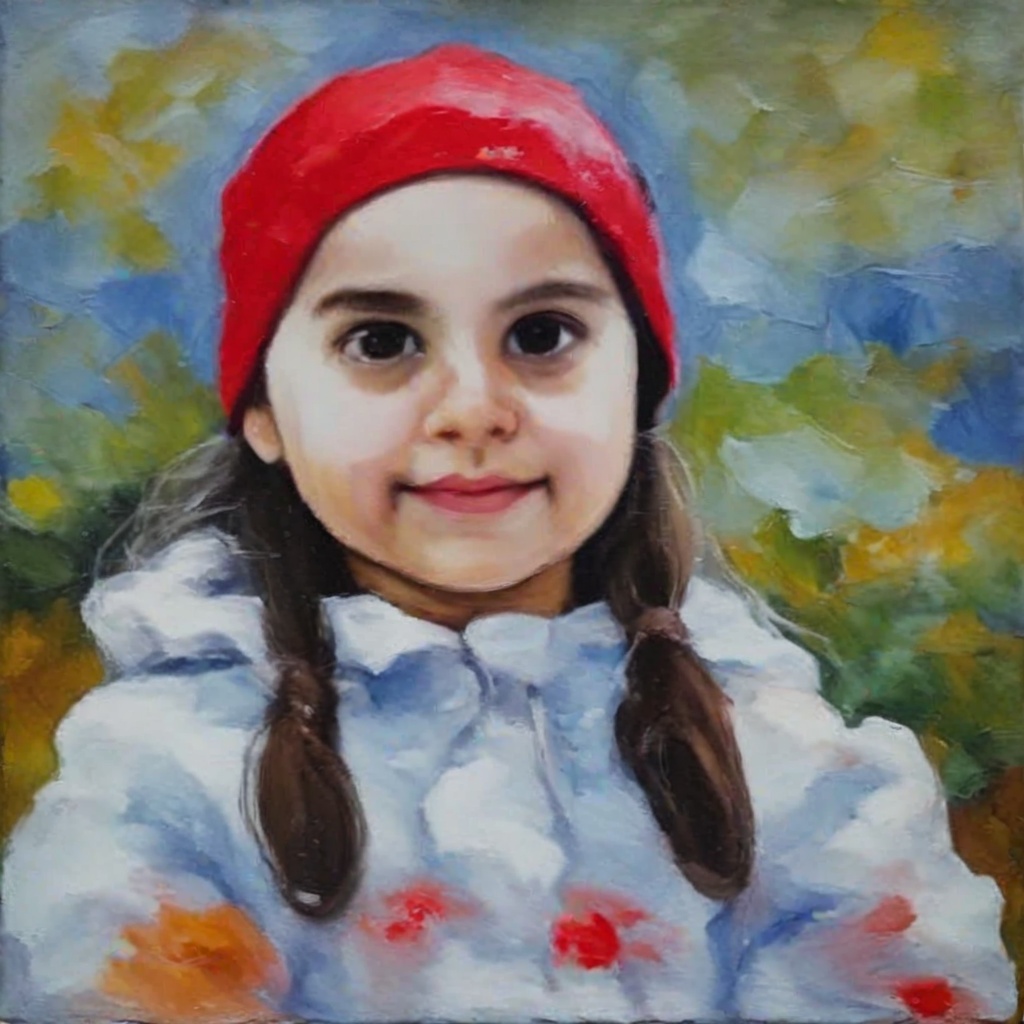} &
        \includegraphics[width=0.15\textwidth,height=0.15\textwidth]{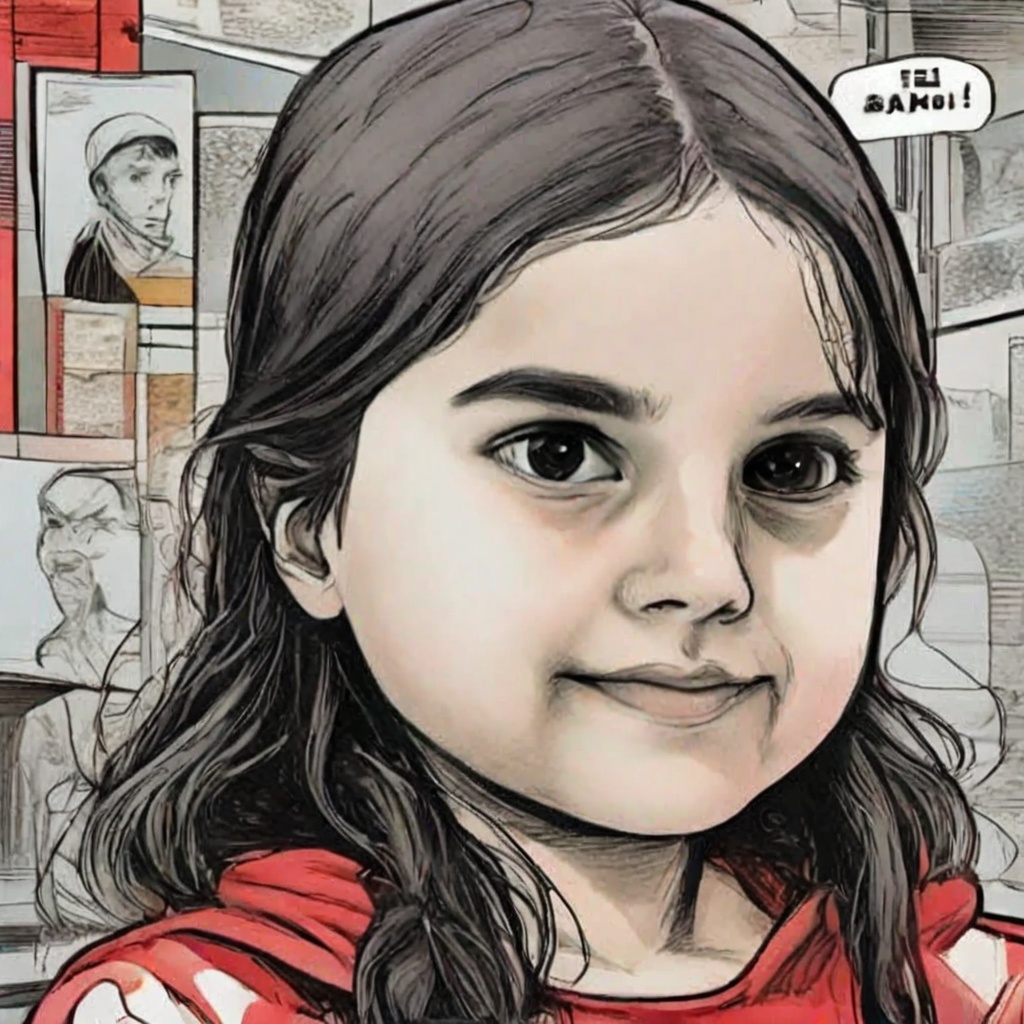} &
        \includegraphics[width=0.15\textwidth,height=0.15\textwidth]{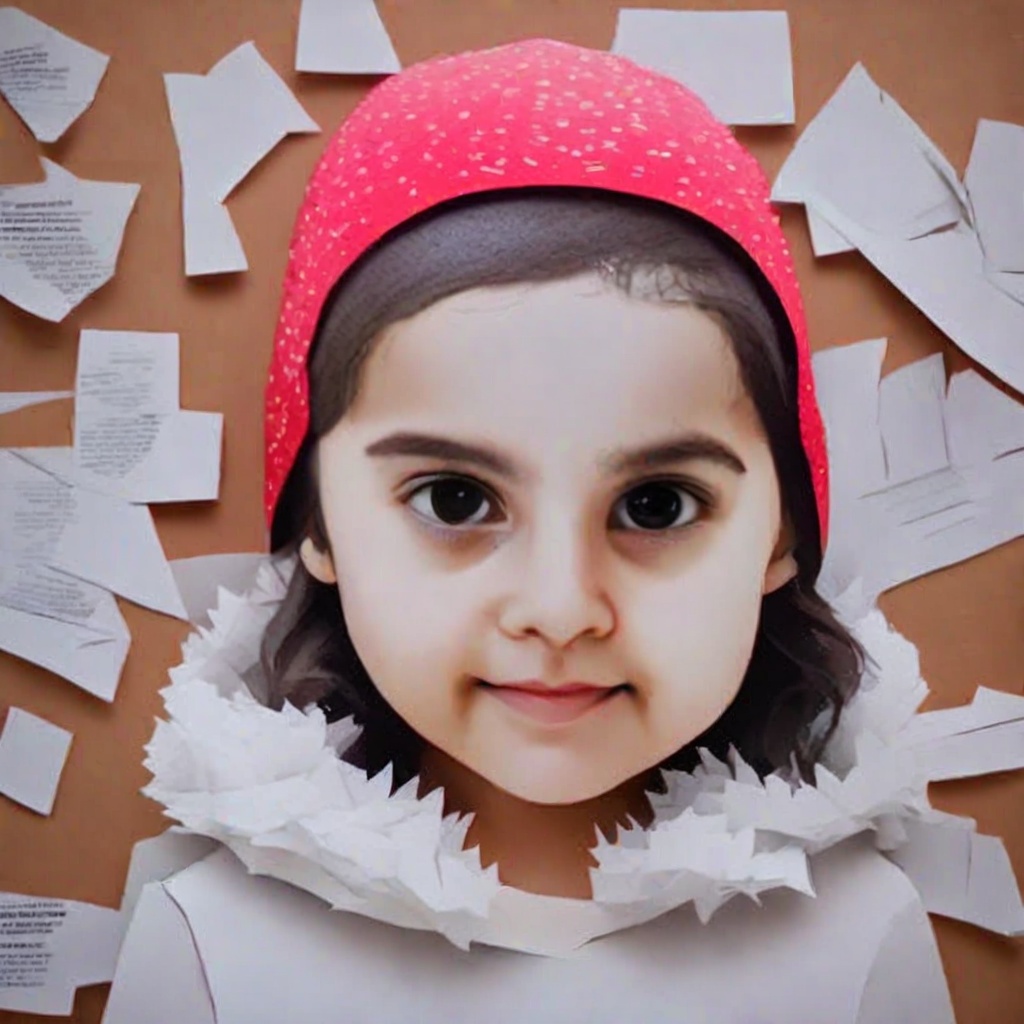} &
        \includegraphics[width=0.15\textwidth,height=0.15\textwidth]{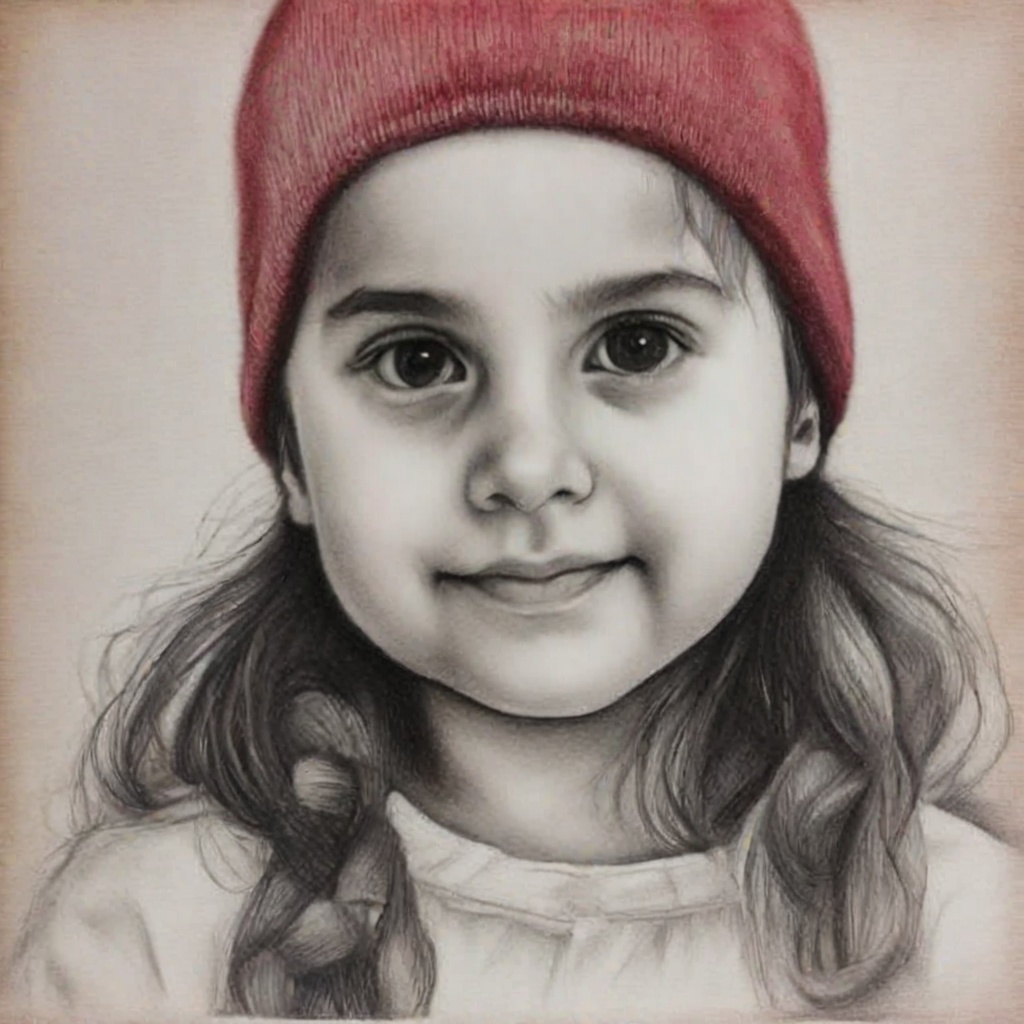} 
        \\

        \includegraphics[width=0.15\textwidth,height=0.15\textwidth]{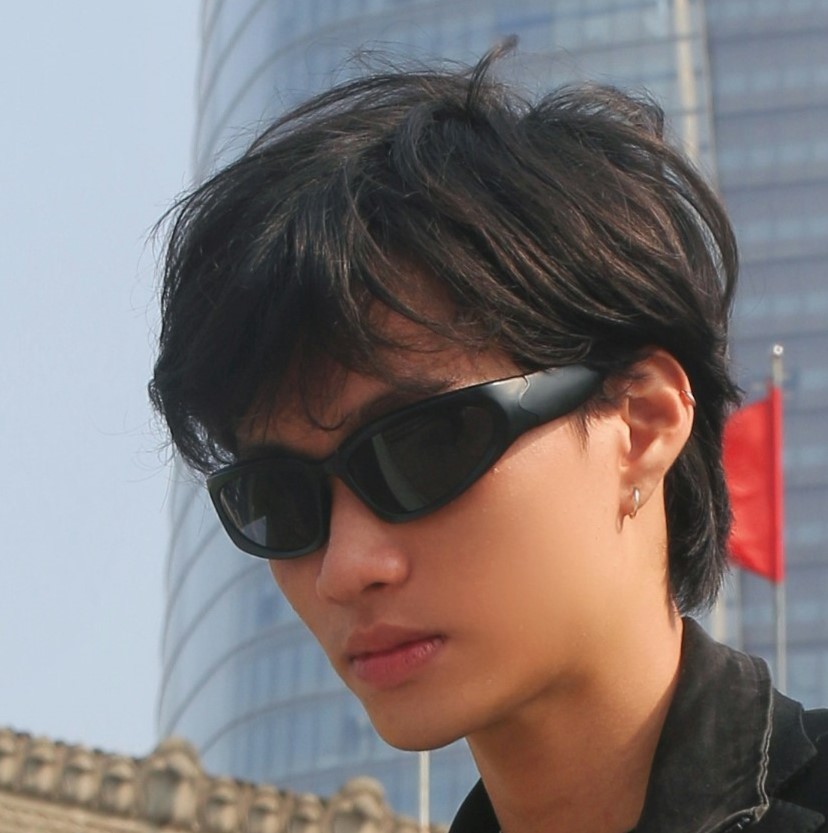} &
        \includegraphics[width=0.15\textwidth,height=0.15\textwidth]{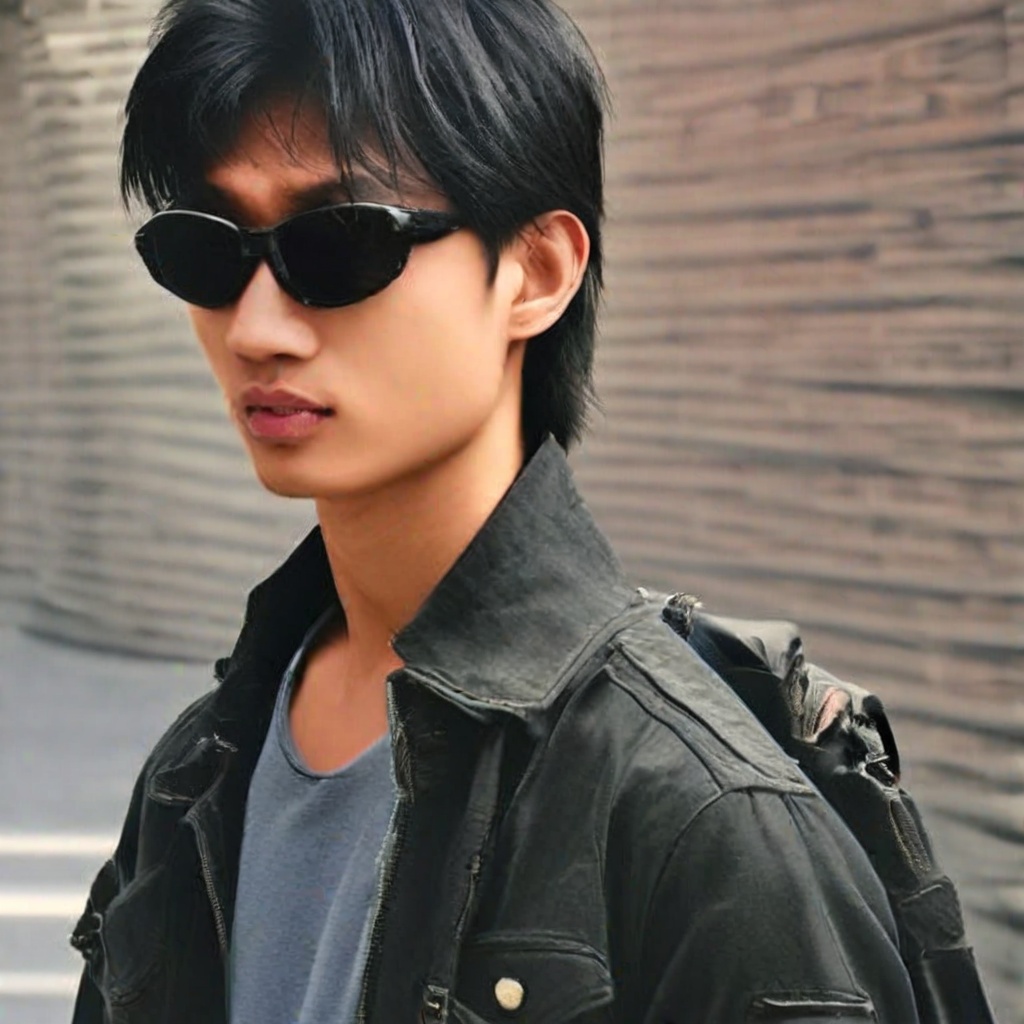} &
        \includegraphics[width=0.15\textwidth,height=0.15\textwidth]{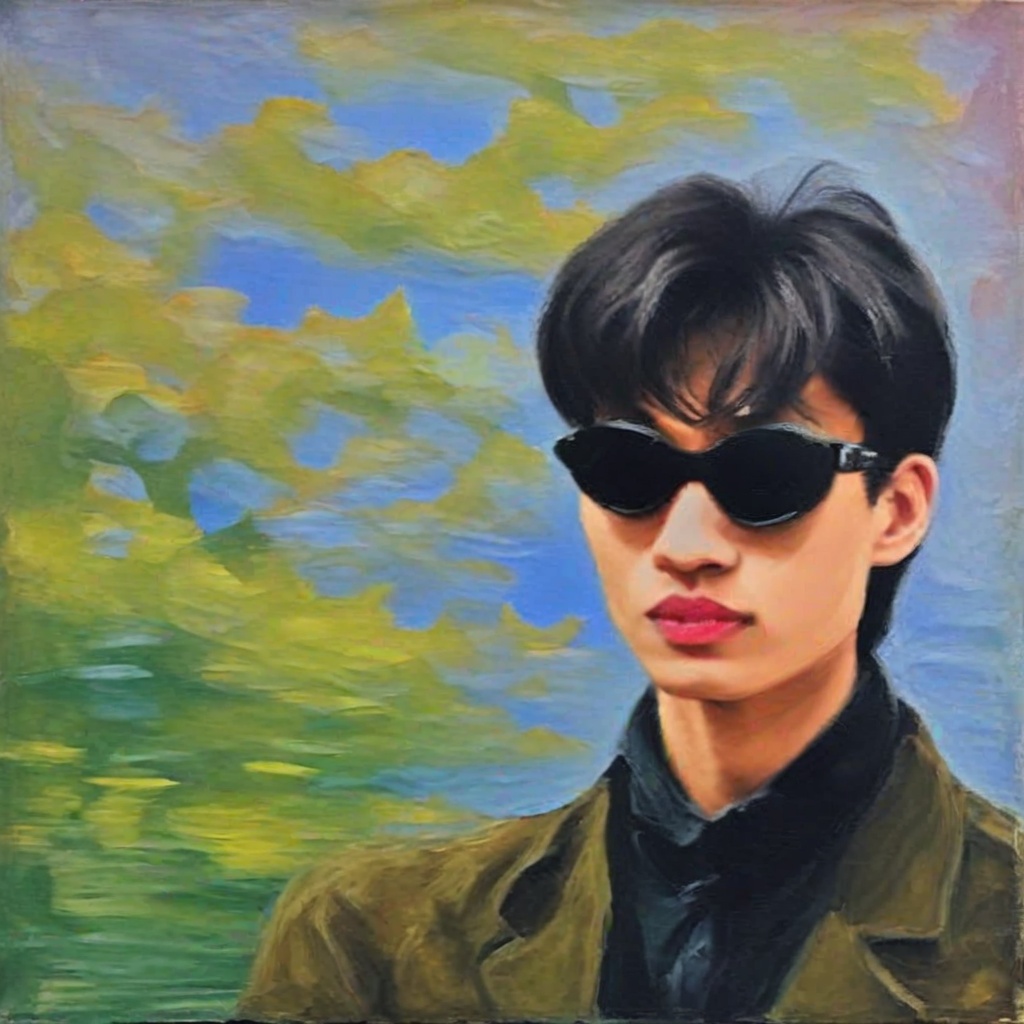} &
        \includegraphics[width=0.15\textwidth,height=0.15\textwidth]{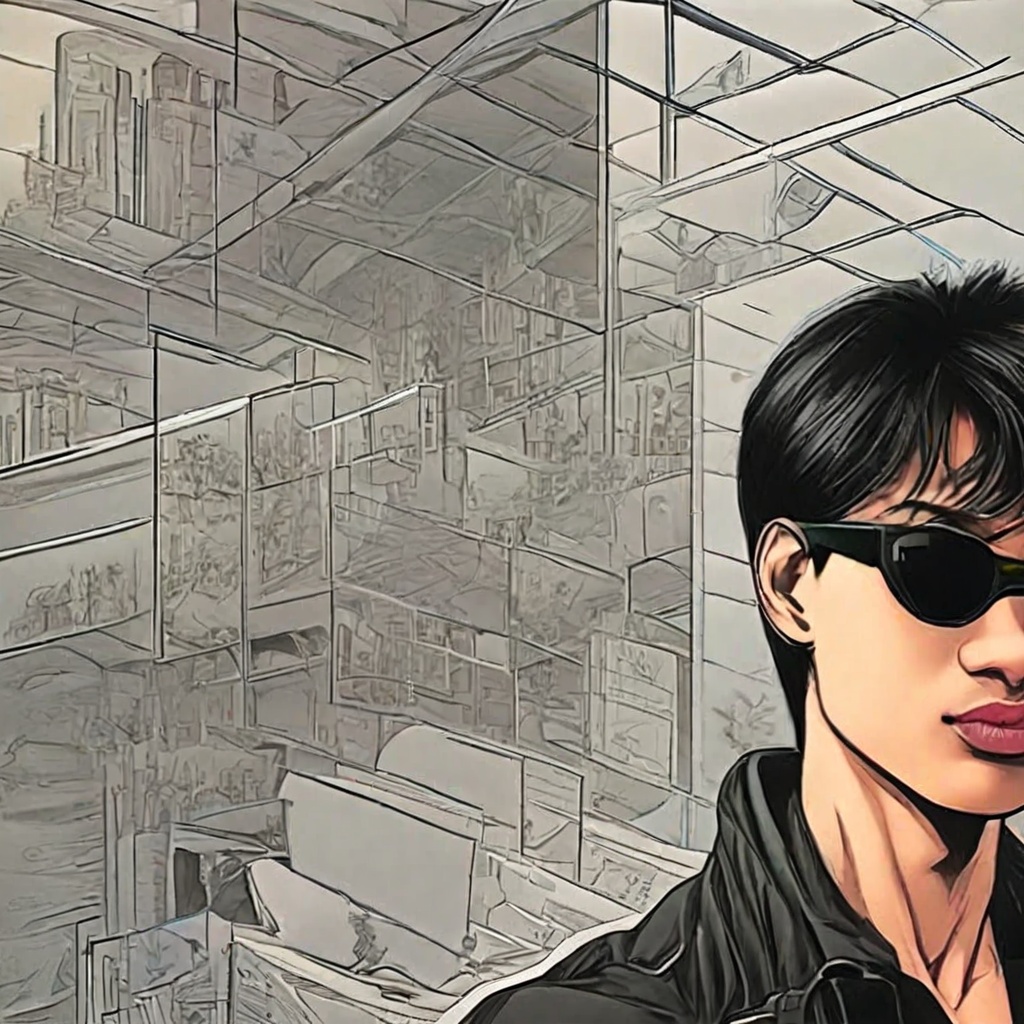} &
        \includegraphics[width=0.15\textwidth,height=0.15\textwidth]{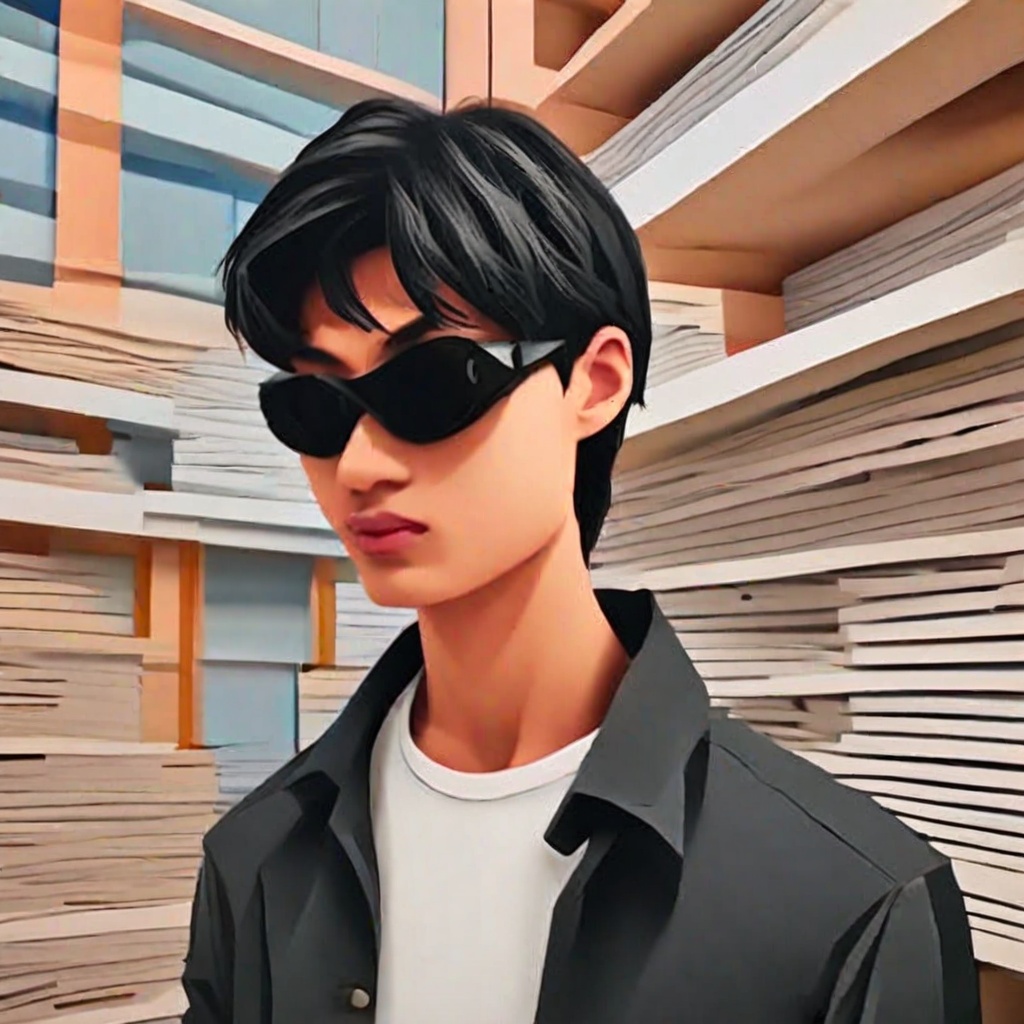} &
        \includegraphics[width=0.15\textwidth,height=0.15\textwidth]{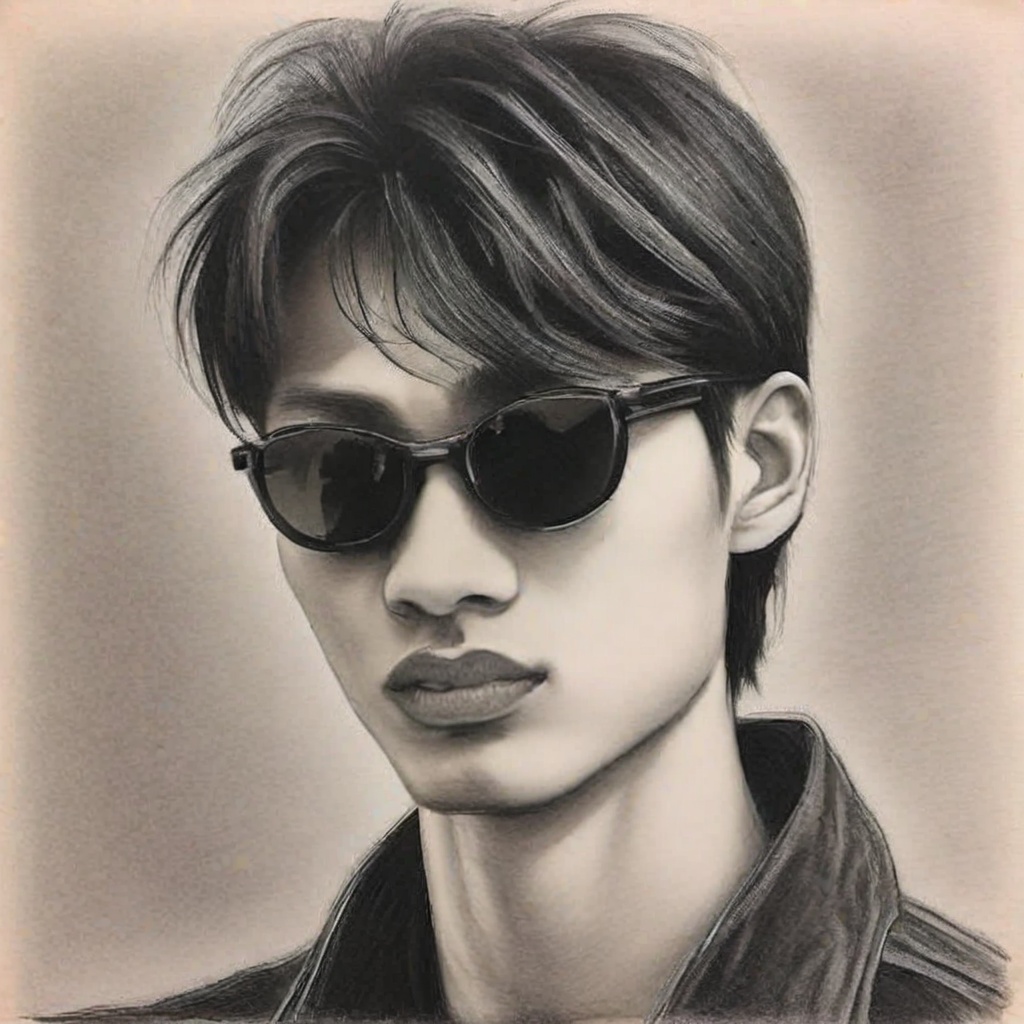} 
        \\

        {\normalsize Input} & {\normalsize ``A photo of.."} & {\normalsize``..Monet"} & {\normalsize``..Comic Book"} & {\normalsize``..Papercraft"} & {\normalsize``Pencil.."} \\

    \end{tabular}
    
    }
    \caption{\textbf{Qualitative results.} Our method can personalize a model to specific face identities at inference time, and align with both photo realistic and stylized prompts.}
    \vspace{-0.1cm}
    \label{fig:qual_ours}
\end{figure*}

\cref{fig:qual_ours} shows a set of images synthesized using our encoder, across a range of identities and prompts. Our method can produce both photo-realistic results and stylized outputs, while largely preserving the subject's identity.

To better gauge the quality of our results, we evaluate our method against a set of prior and concurrent works on personalized face generation. Specifically, we consider the three leading tuning-free approaches which have public SDXL implementations: IP-Adapter~\cite{ye2023ipadapter}, InstantID~\cite{wang2024instantid}, and PhotoMaker~\cite{li2023photomaker}.
For IP-Adapter, we use the face model that serves as our backbone, and evaluate it using two adapter-scale settings: The '`official'' value ($1.0$), and the one commonly used by the community ($0.5$) which significantly improves alignment with the textual prompts but harms identity.

\begin{figure*}[t]

    \centering
    \setlength{\tabcolsep}{1.5pt}
    {\normalsize
    \begin{tabular}{c c c c c c c}

        {\normalsize Input} & 
        {\normalsize InstantID} &{\normalsize PhotoMaker} & {\normalsize IP-A (0.5)} & {\normalsize IP-A (1.0)} & {\normalsize Ours} & \\

        \includegraphics[width=0.14\textwidth,height=0.14\textwidth]{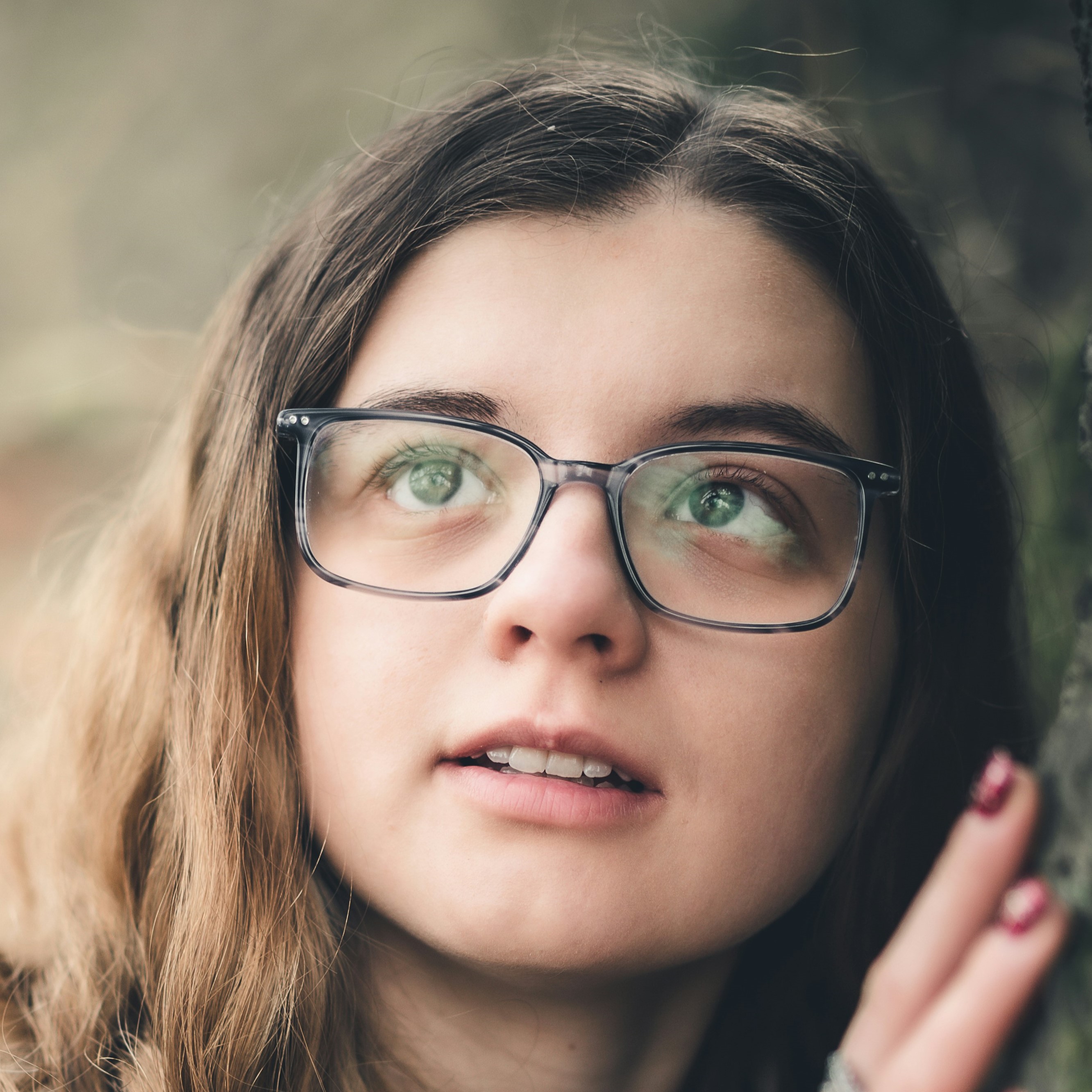} &

        \includegraphics[width=0.14\textwidth,height=0.14\textwidth]{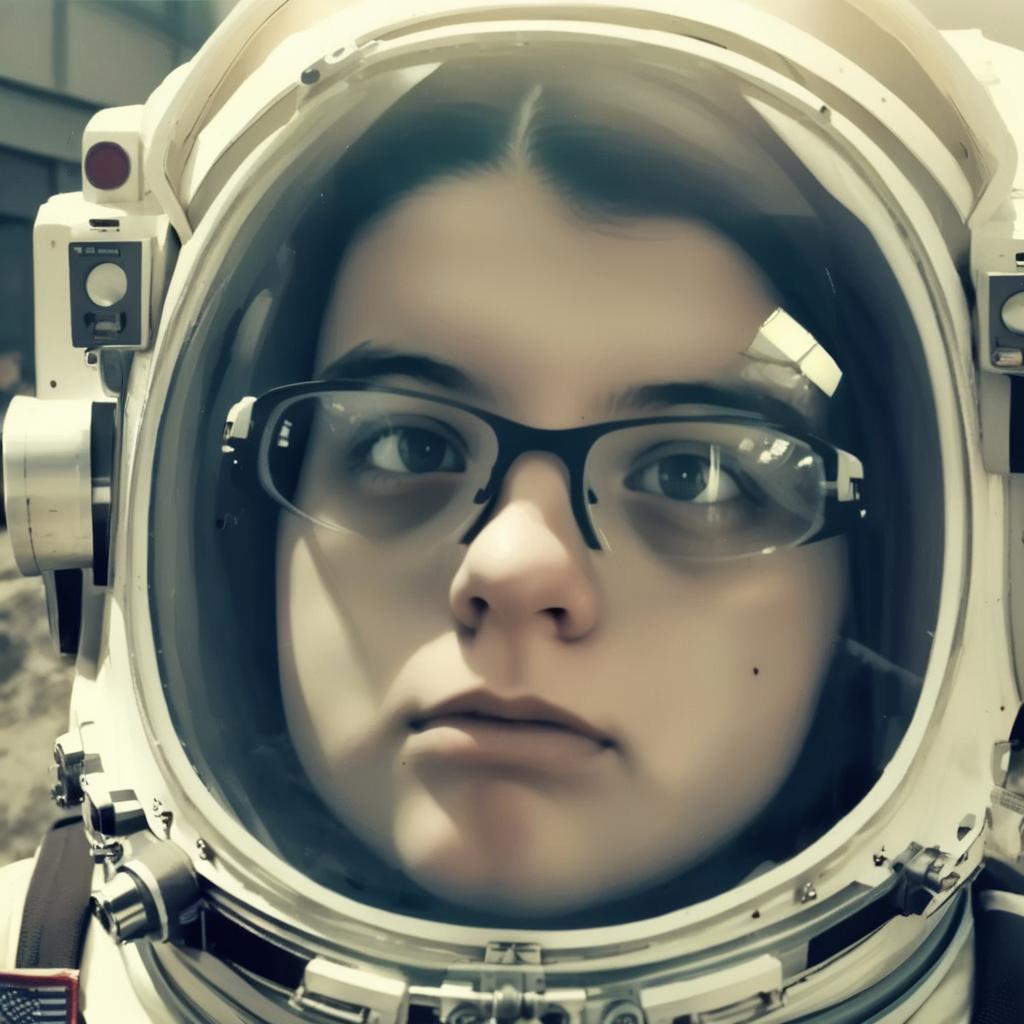} &
        \includegraphics[width=0.14\textwidth,height=0.14\textwidth]{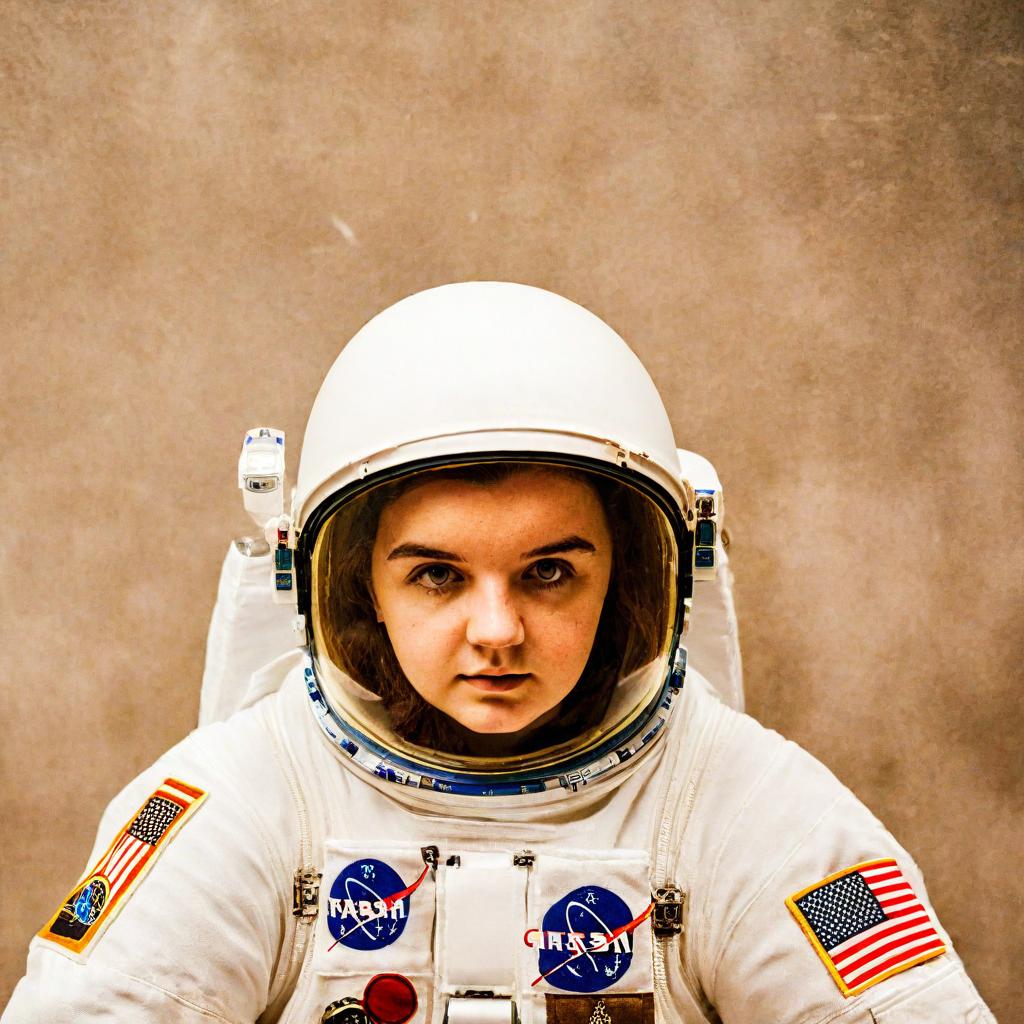} &
        \includegraphics[width=0.14\textwidth,height=0.14\textwidth]{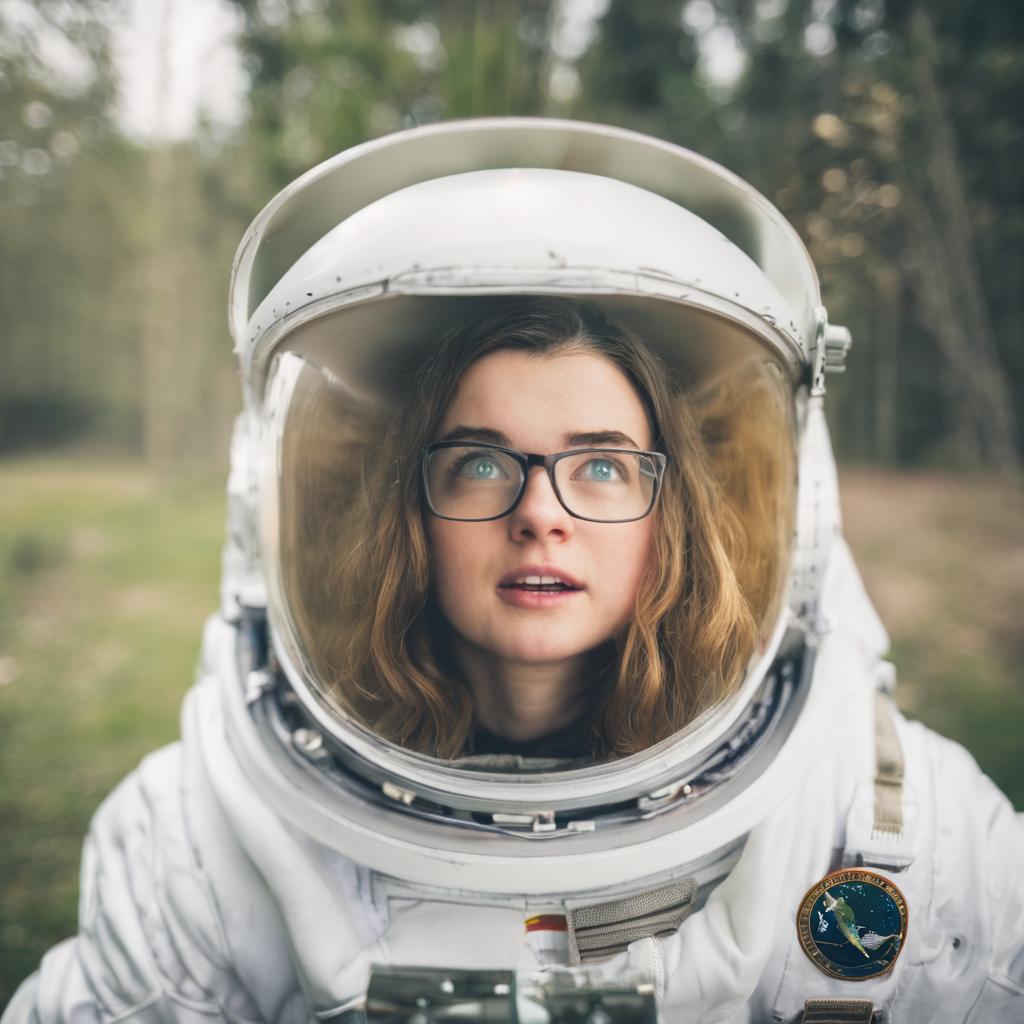} &
        \includegraphics[width=0.14\textwidth,height=0.14\textwidth]{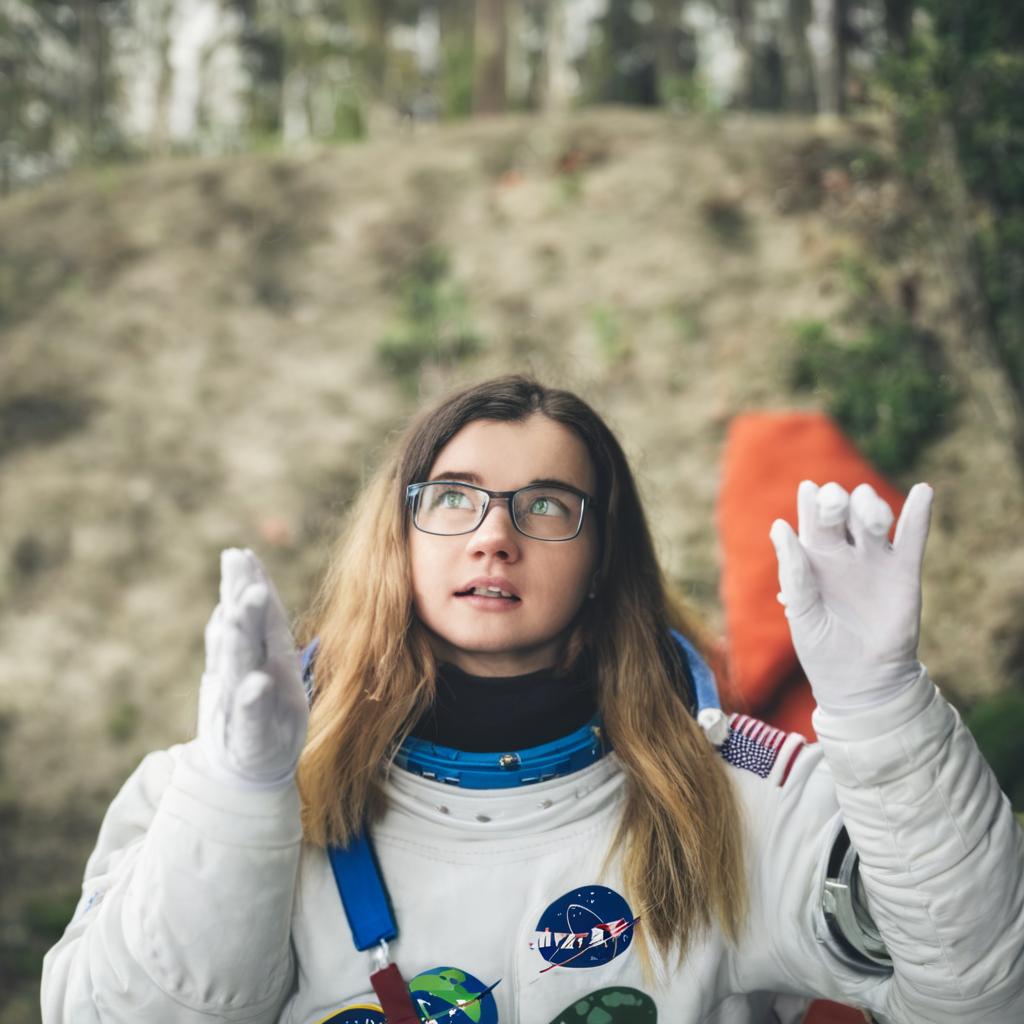} &
        \includegraphics[width=0.14\textwidth,height=0.14\textwidth]{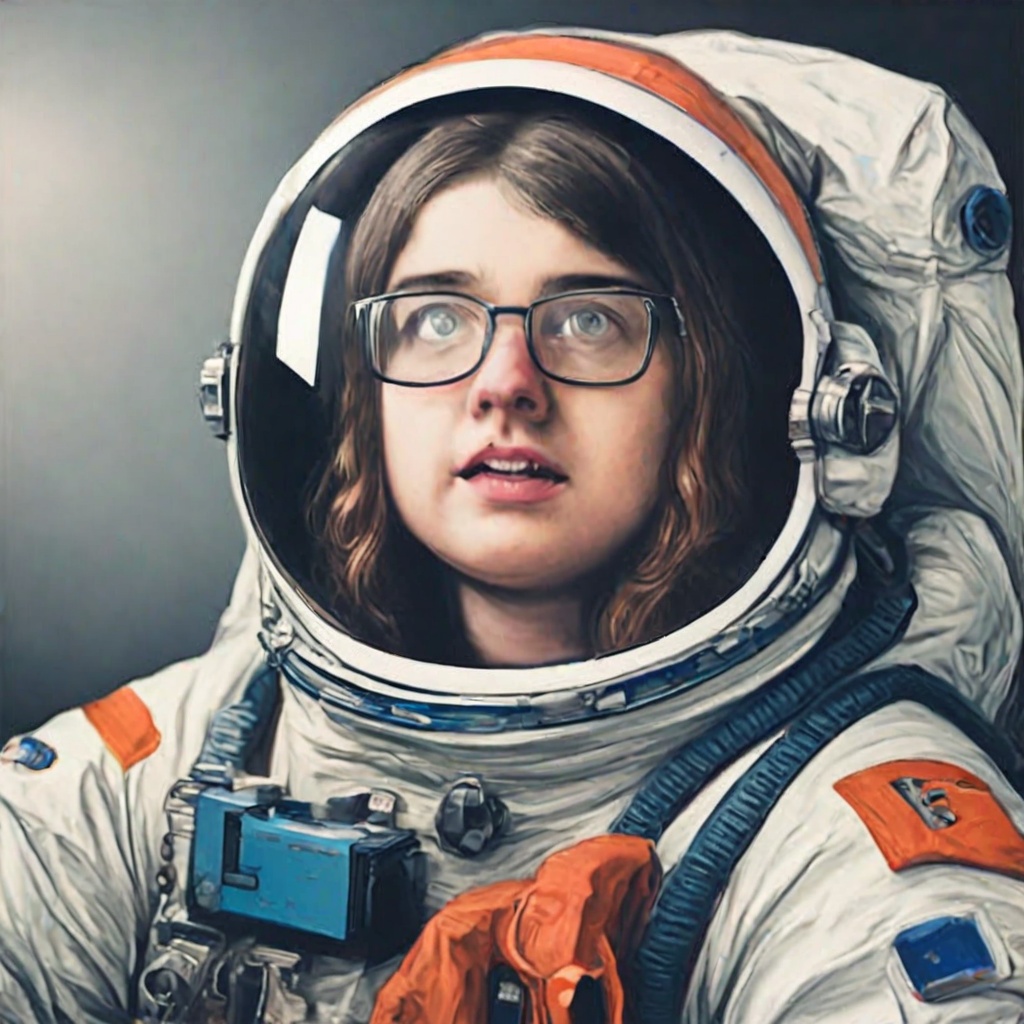} &
        
        \raisebox{0.054\textwidth}{\rotatebox[origin=t]{-90}{\scalebox{0.9}{\begin{tabular}{c@{}c@{}c@{}} Astronaut\end{tabular}}}}
        \\

        \includegraphics[width=0.14\textwidth,height=0.14\textwidth]{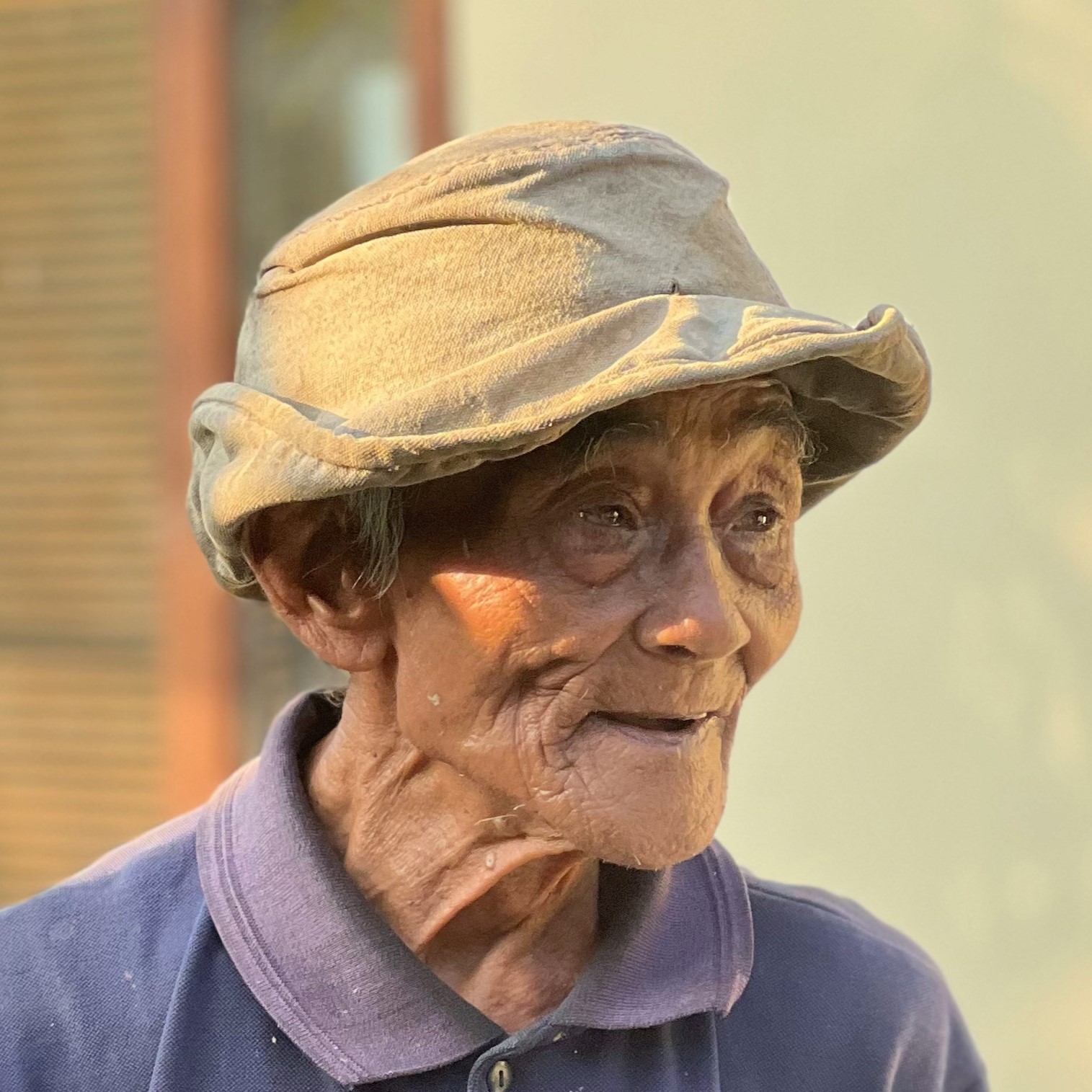} &
        
        \includegraphics[width=0.14\textwidth,height=0.14\textwidth]{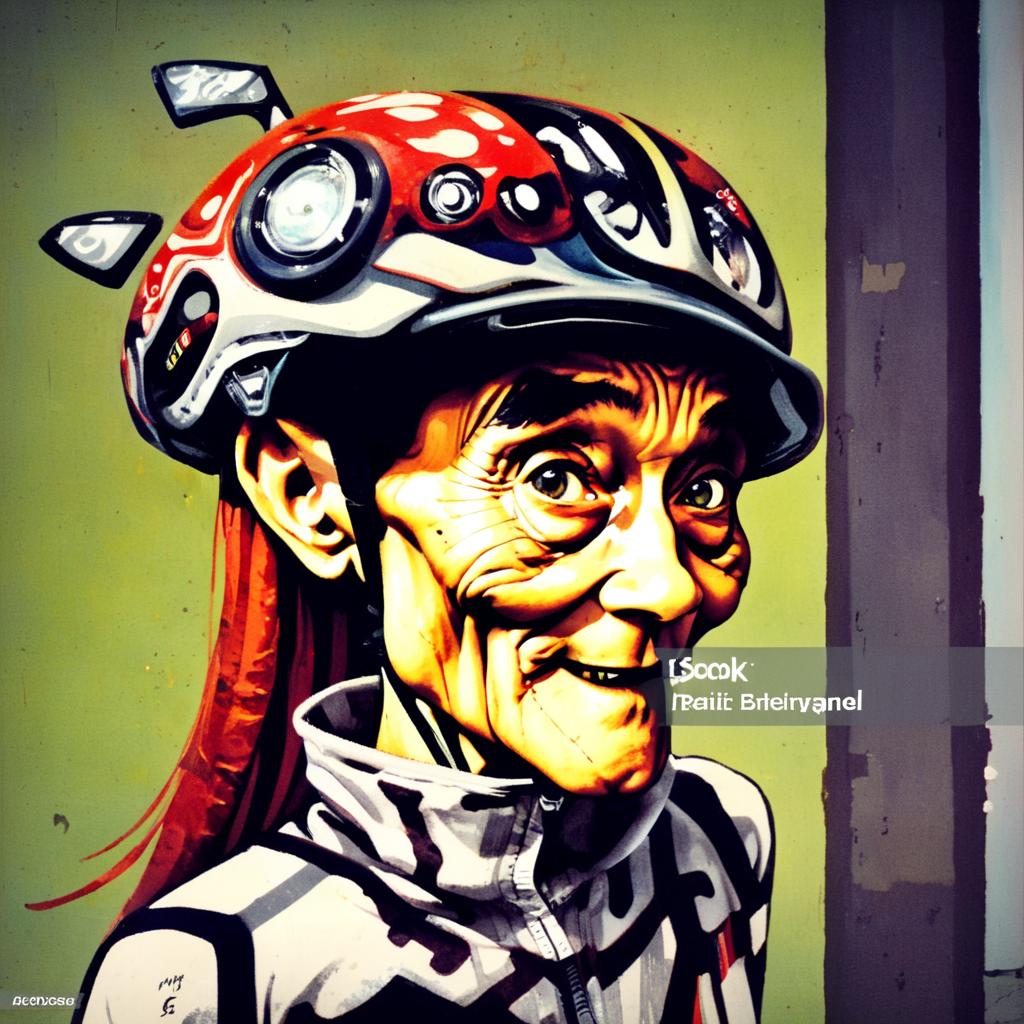} &
        \includegraphics[width=0.14\textwidth,height=0.14\textwidth]{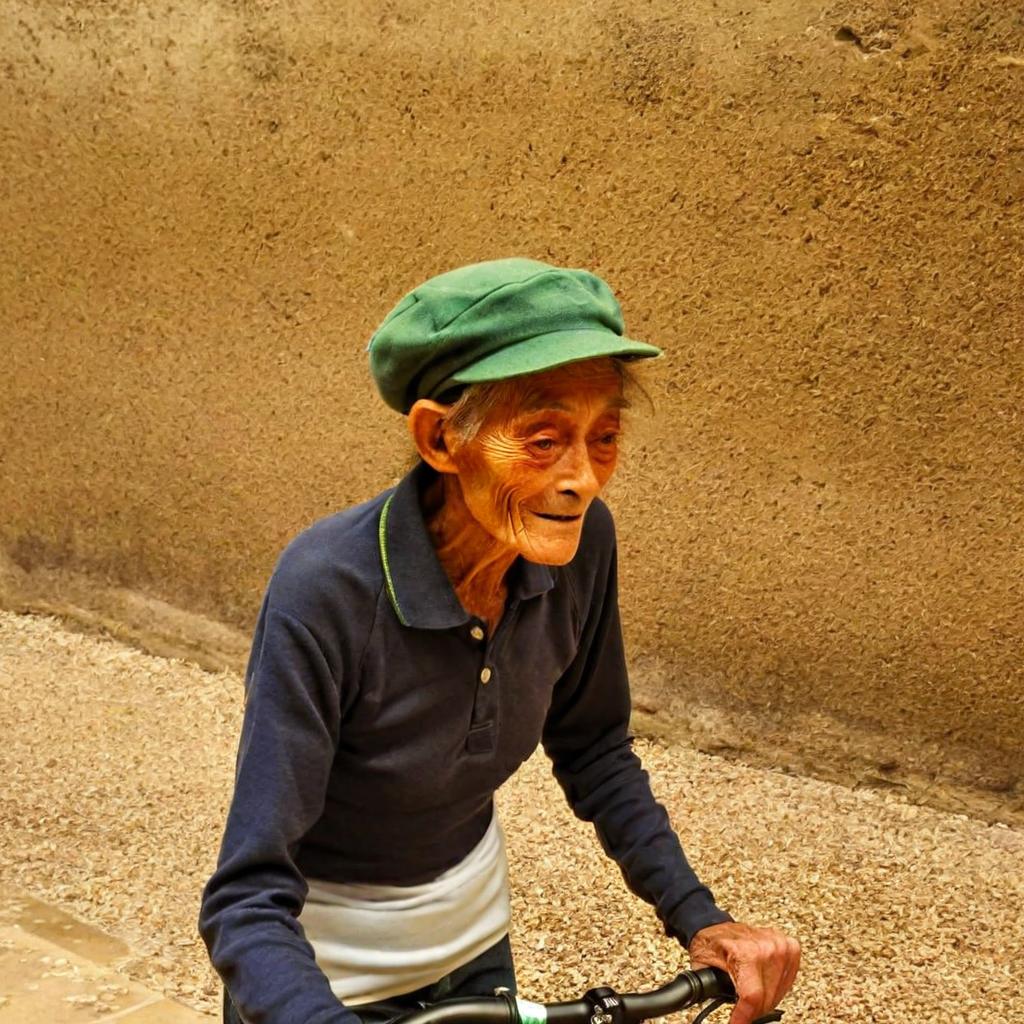} &
        \includegraphics[width=0.14\textwidth,height=0.14\textwidth]{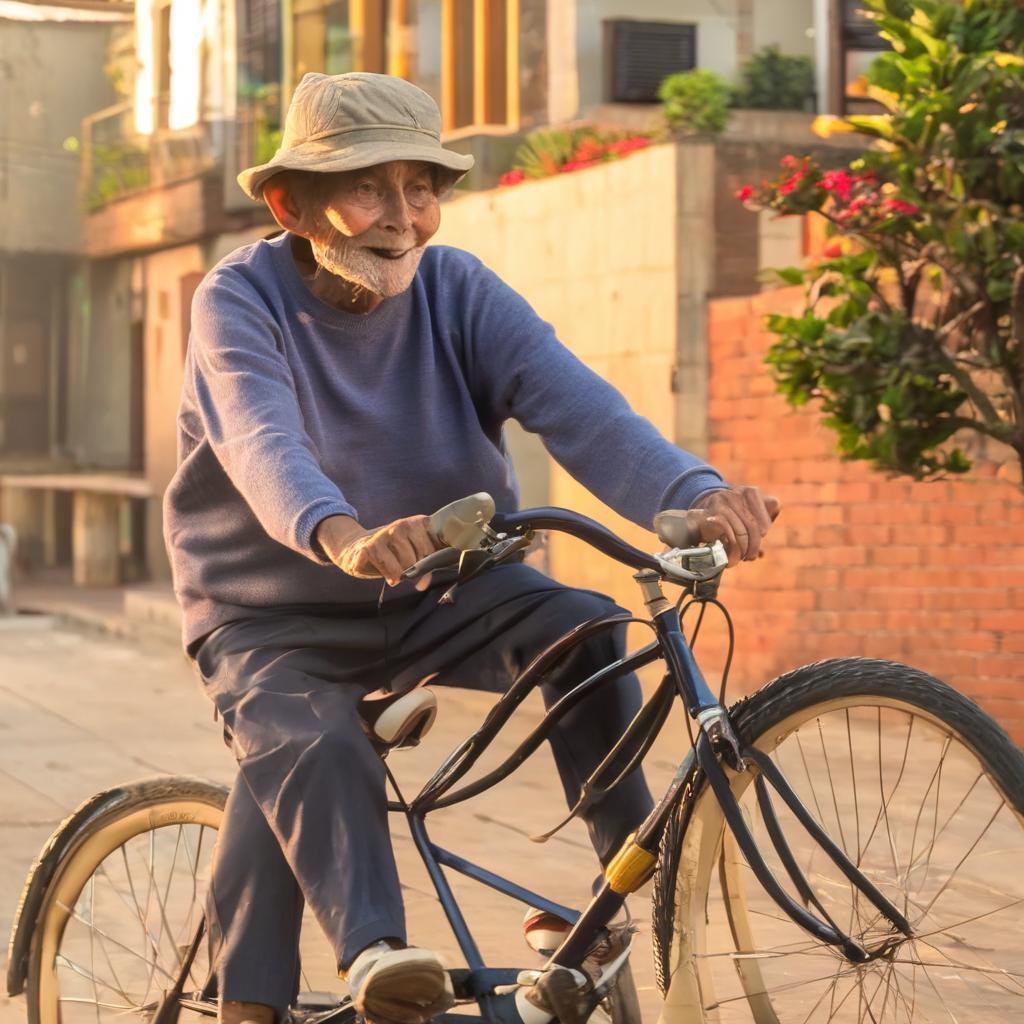} &
        \includegraphics[width=0.14\textwidth,height=0.14\textwidth]{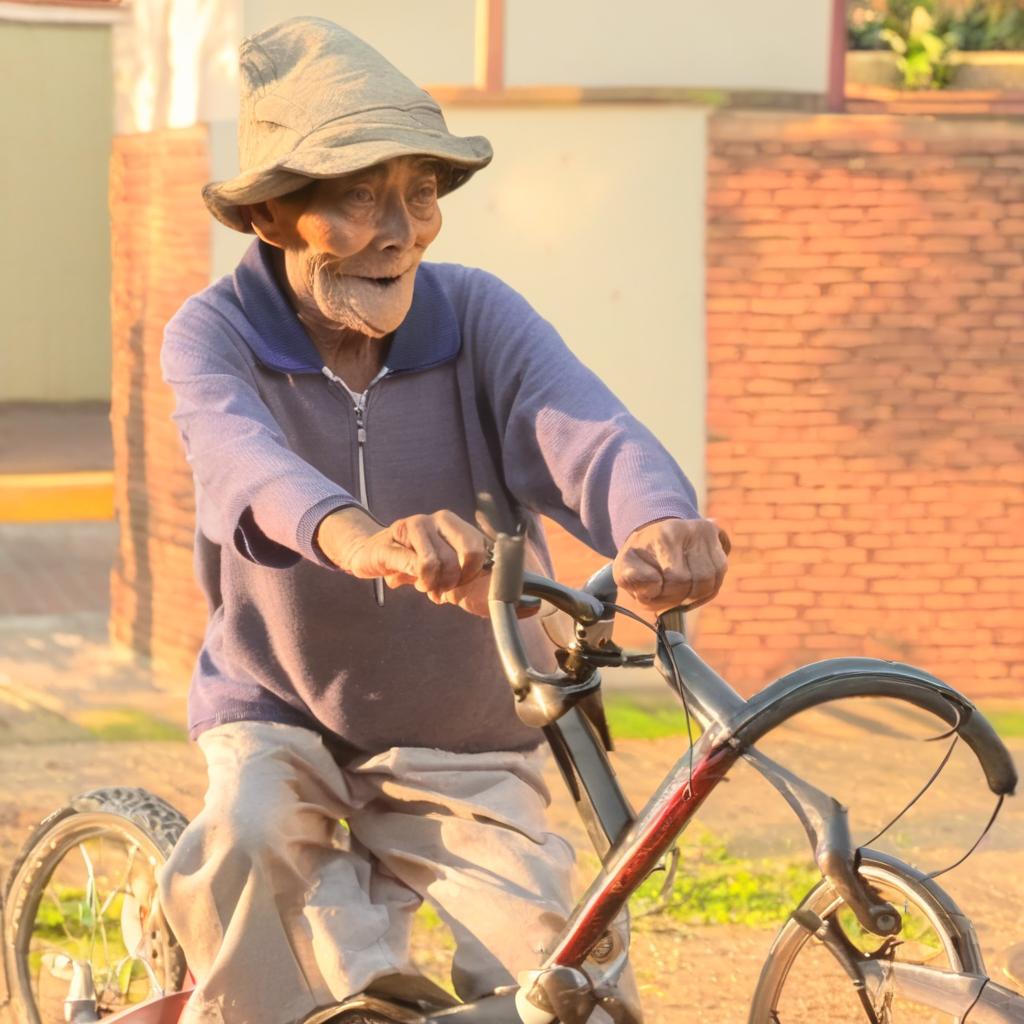} &
        \includegraphics[width=0.14\textwidth,height=0.14\textwidth]{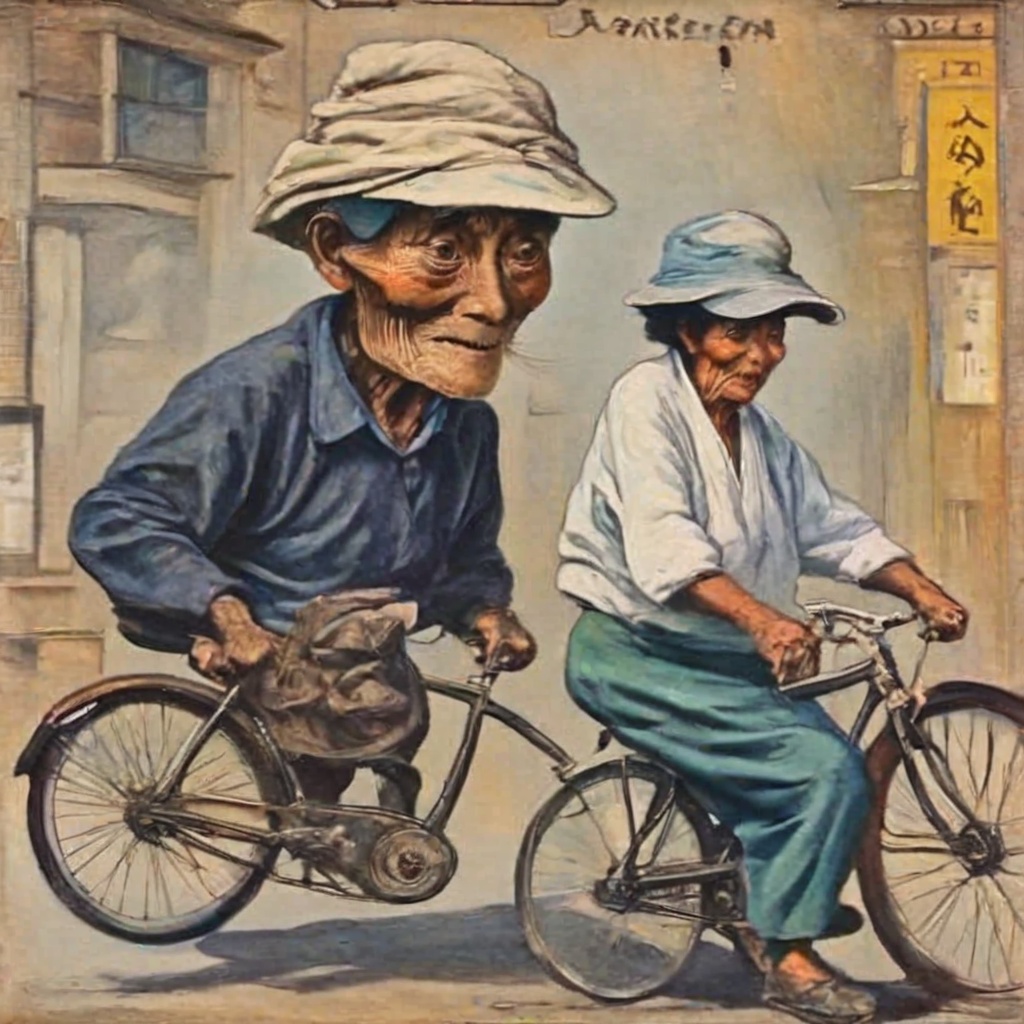} &
        
        \raisebox{0.06\textwidth}{\rotatebox[origin=t]{-90}{\scalebox{0.9}{\begin{tabular}{c@{}c@{}c@{}} On a bike\end{tabular}}}} \\

        \includegraphics[width=0.14\textwidth,height=0.14\textwidth]{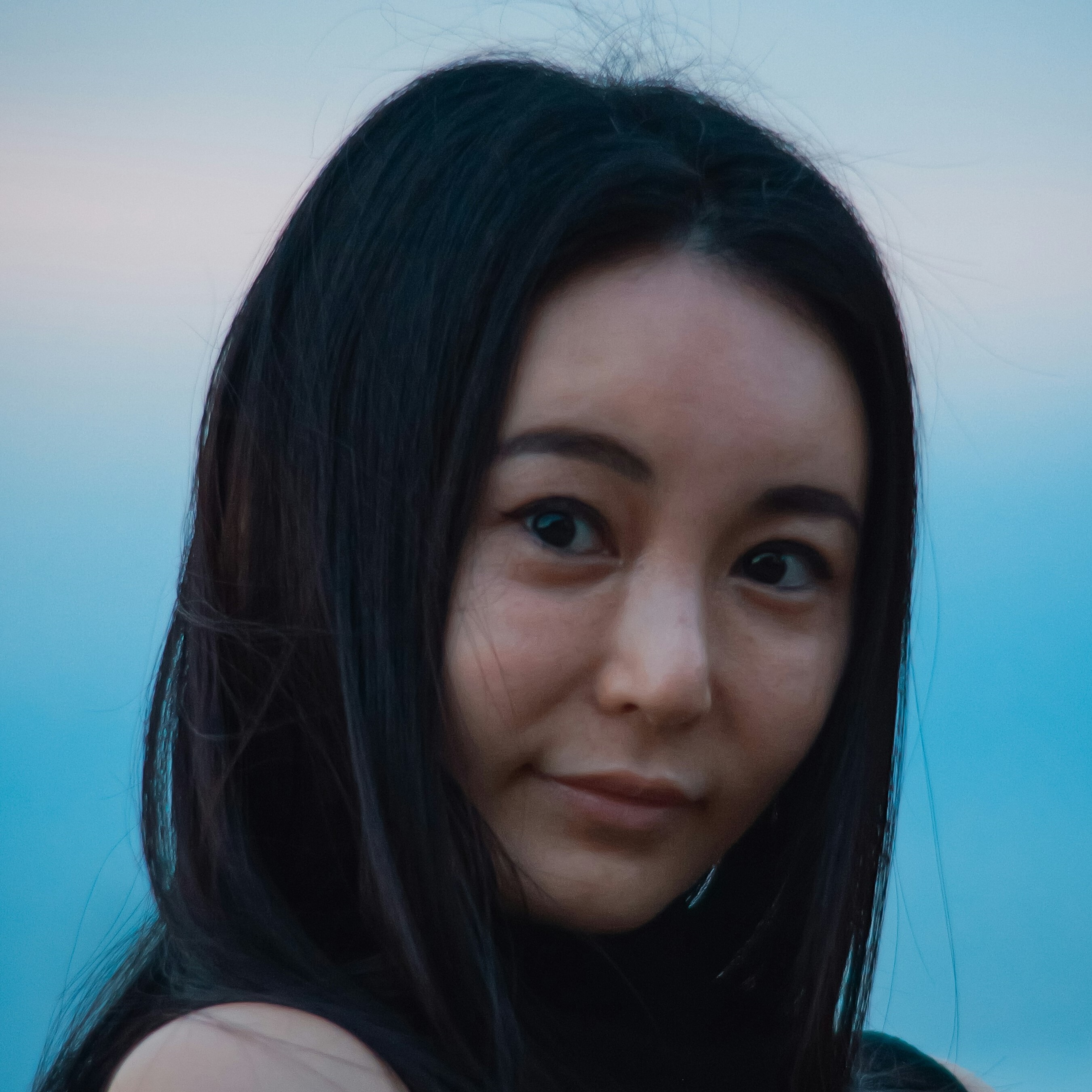} &

        \includegraphics[width=0.14\textwidth,height=0.14\textwidth]{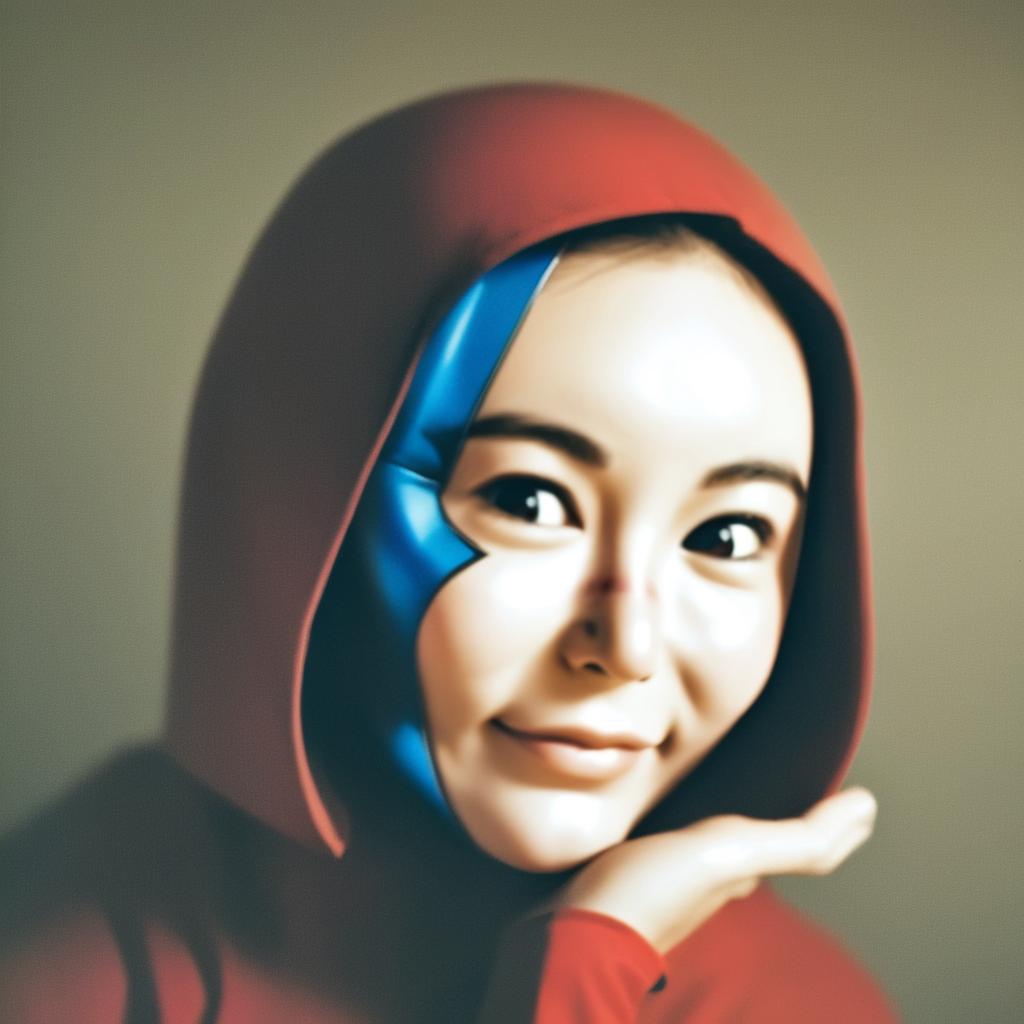} &
        \includegraphics[width=0.14\textwidth,height=0.14\textwidth]{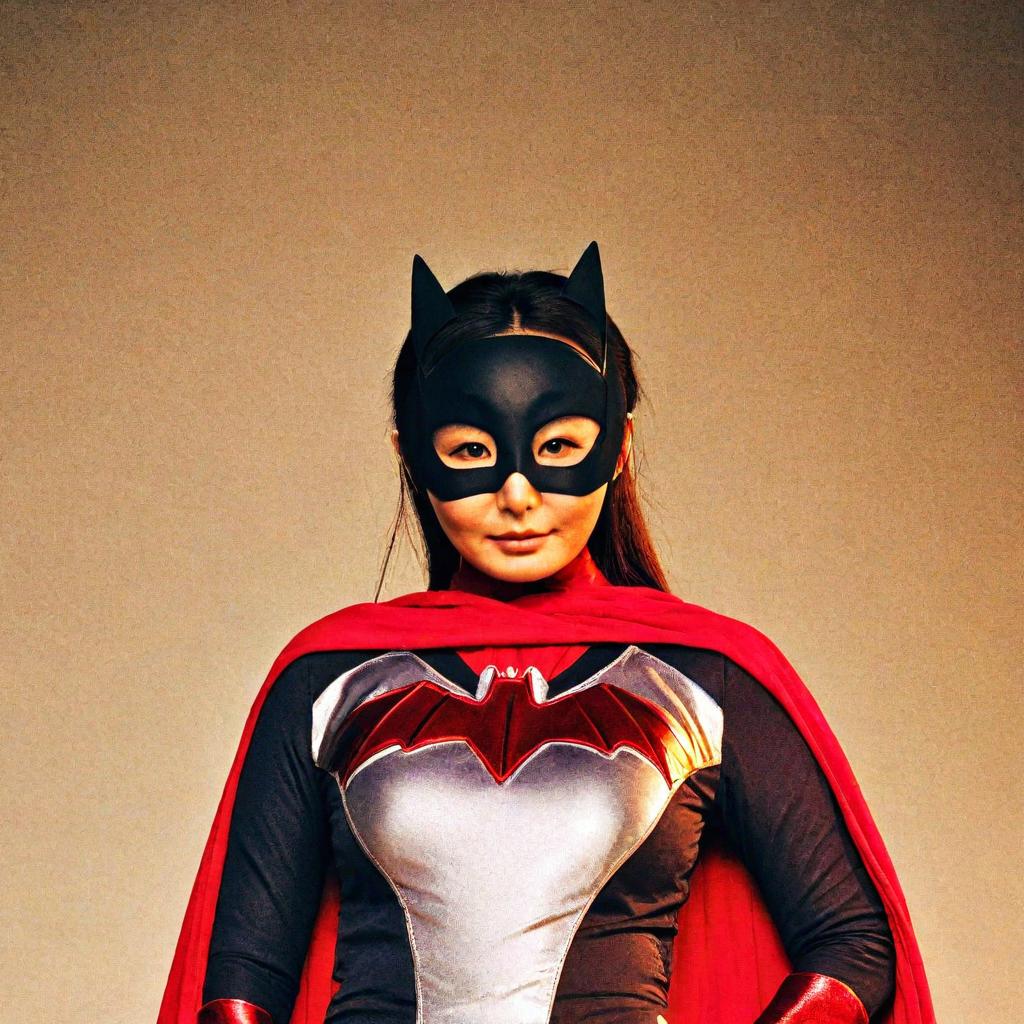} &
        \includegraphics[width=0.14\textwidth,height=0.14\textwidth]{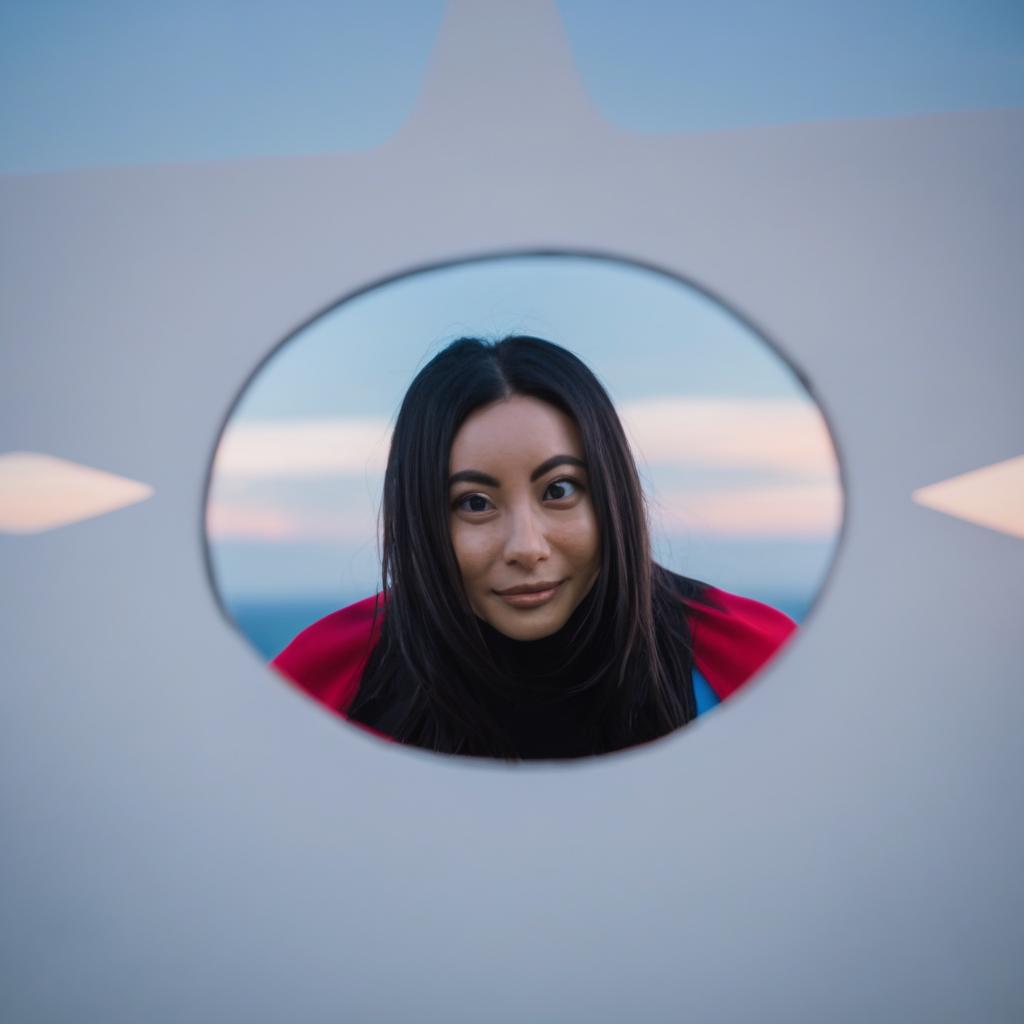} &
        \includegraphics[width=0.14\textwidth,height=0.14\textwidth]{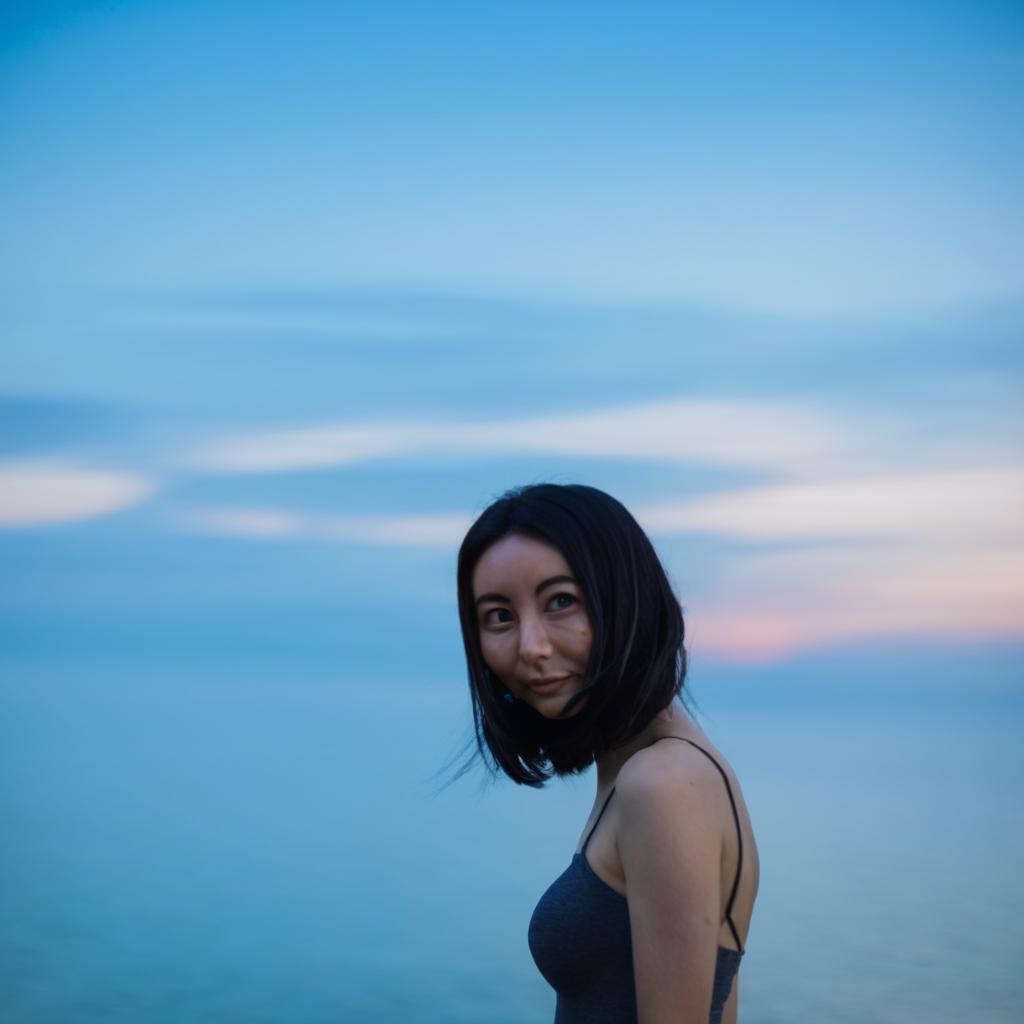} &
        \includegraphics[width=0.14\textwidth,height=0.14\textwidth]{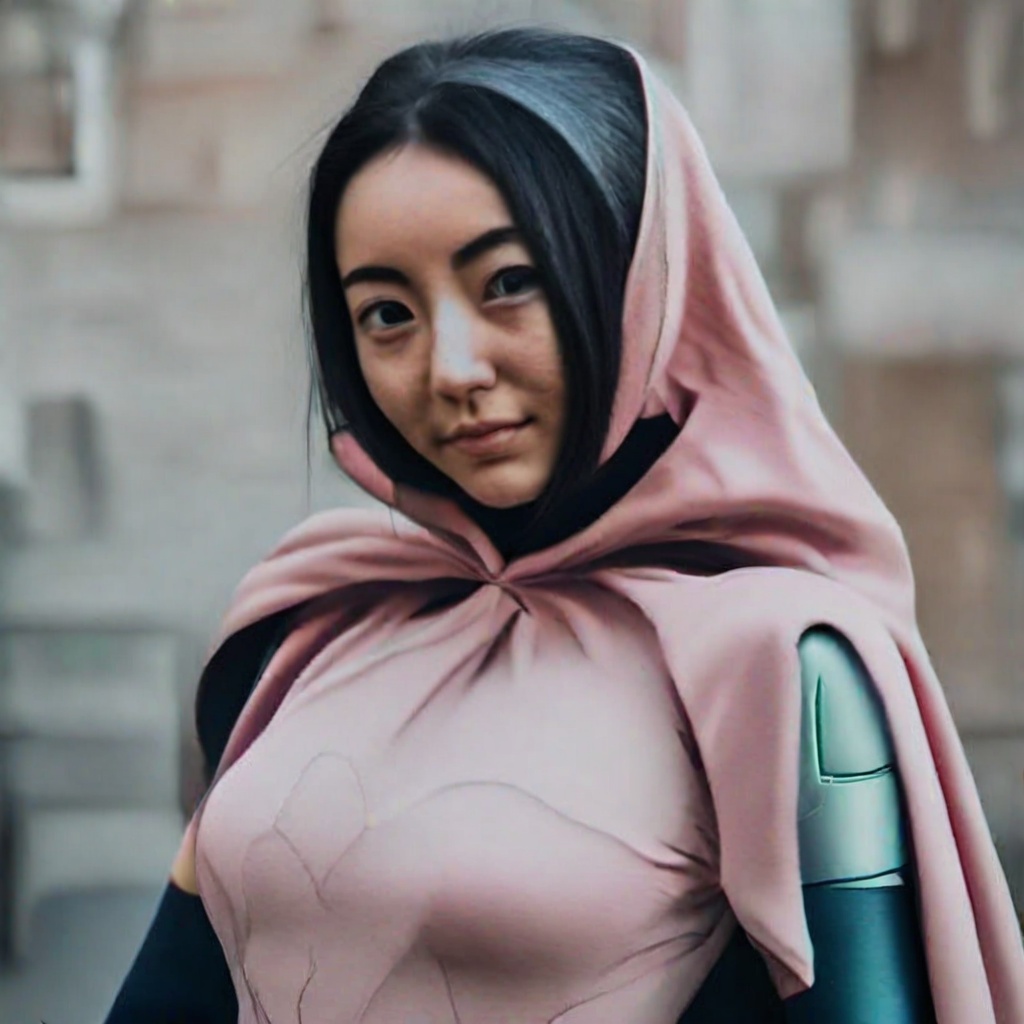} &
        
        \raisebox{0.06\textwidth}{\rotatebox[origin=t]{-90}{\scalebox{0.9}{\begin{tabular}{c@{}c@{}c@{}} Superhero \end{tabular}}}} \\

        \includegraphics[width=0.14\textwidth,height=0.14\textwidth]{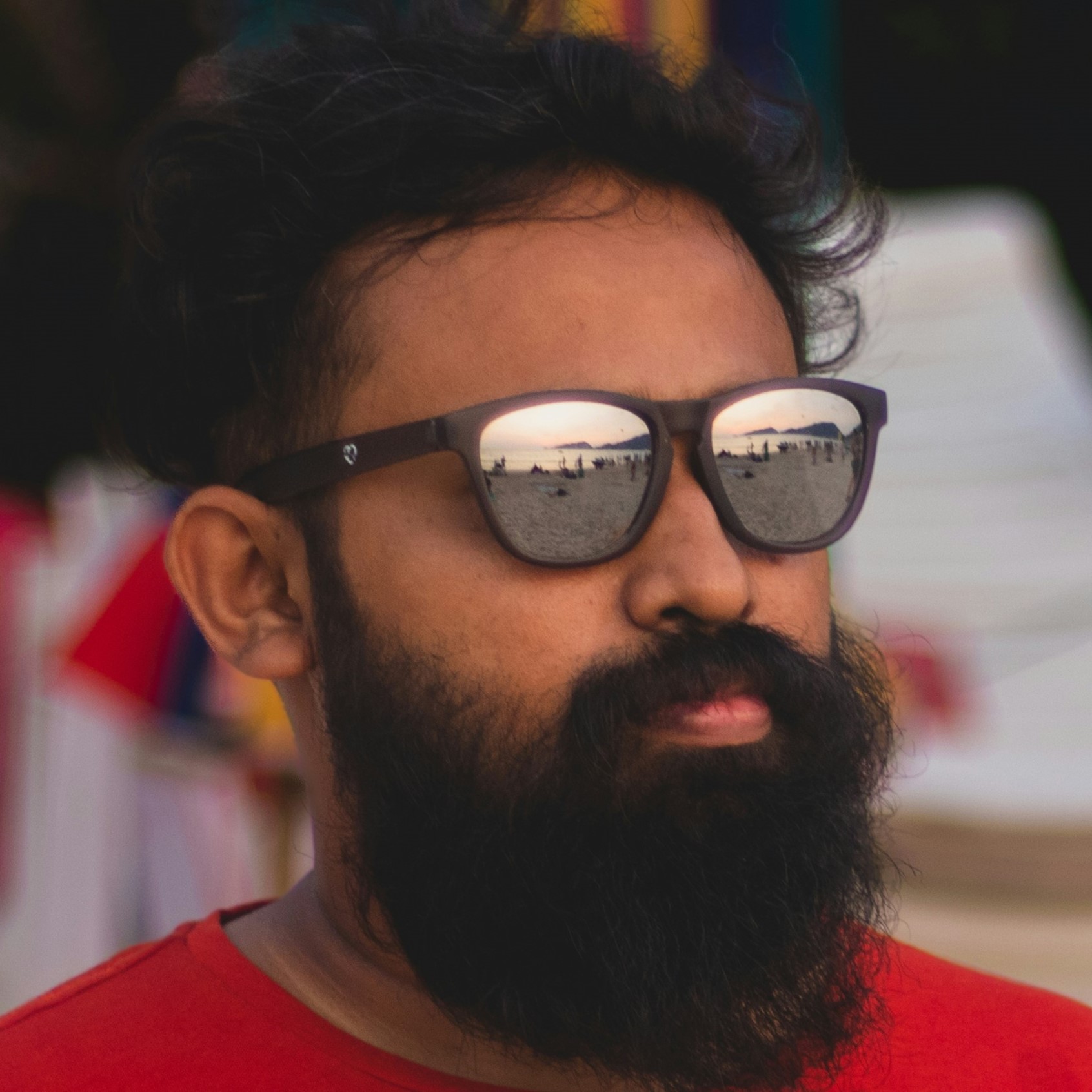} &

        \includegraphics[width=0.14\textwidth,height=0.14\textwidth]{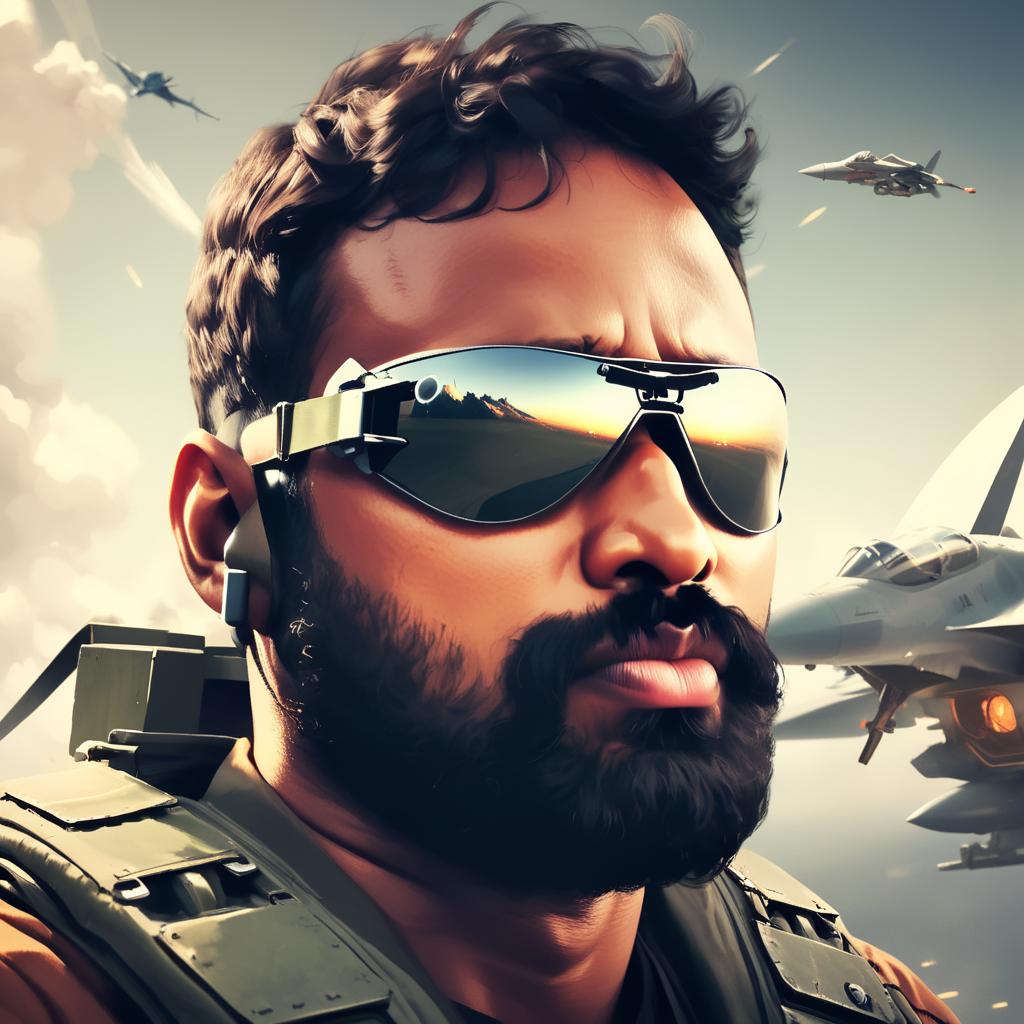} &
        \includegraphics[width=0.14\textwidth,height=0.14\textwidth]{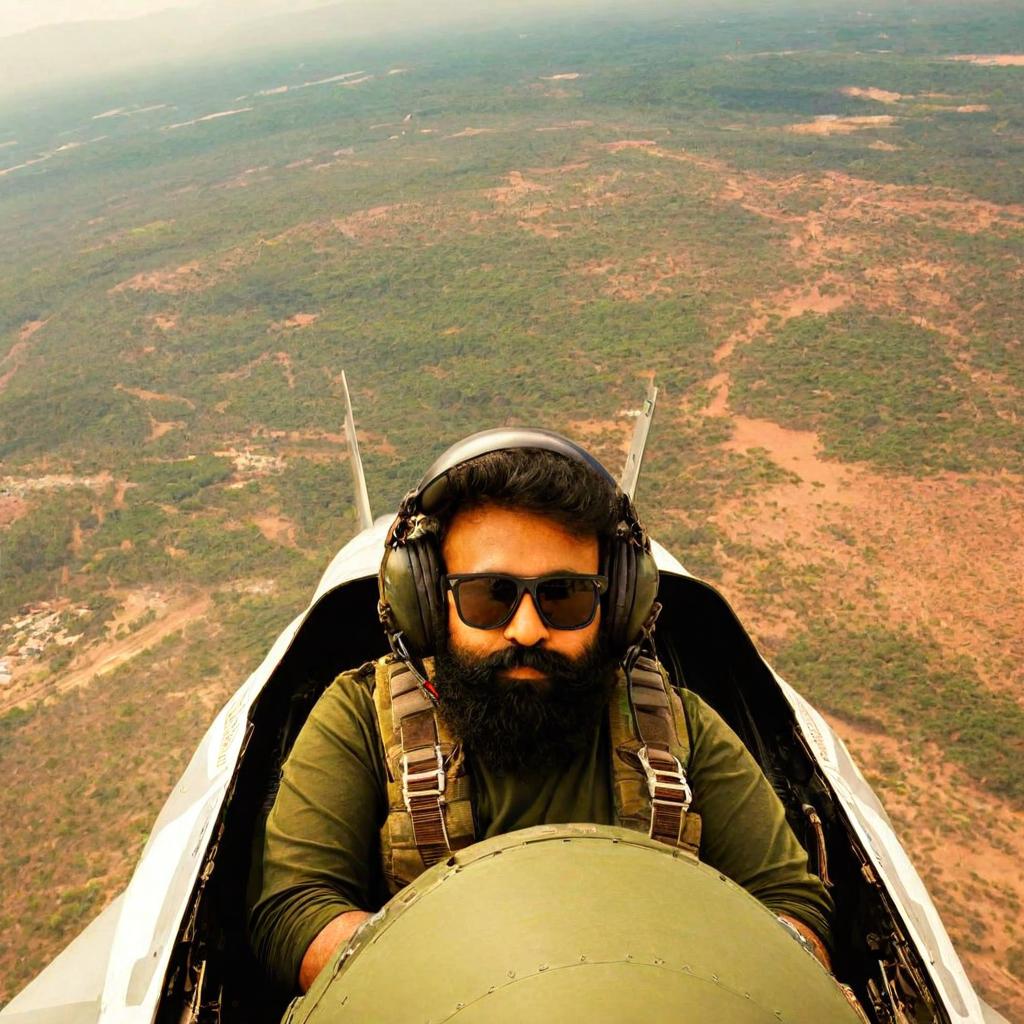} &
        \includegraphics[width=0.14\textwidth,height=0.14\textwidth]{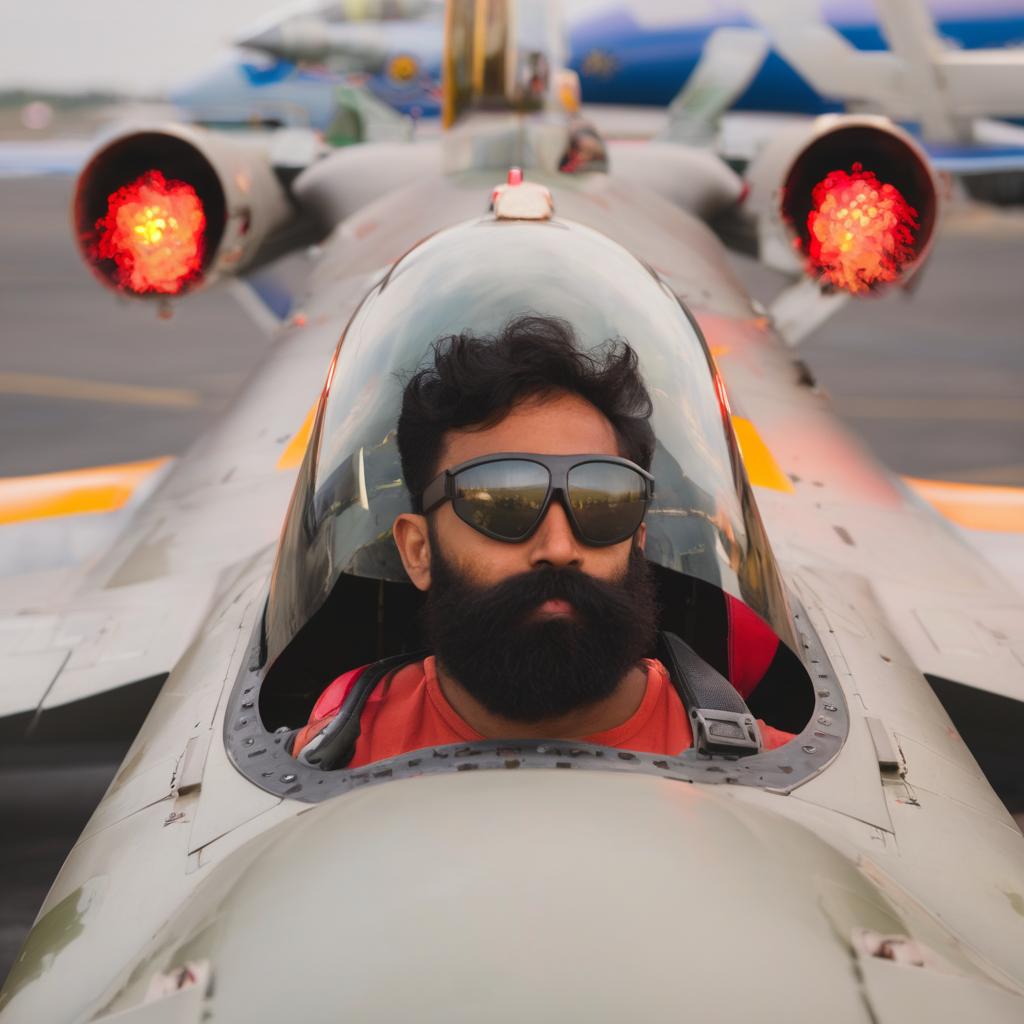} &
        \includegraphics[width=0.14\textwidth,height=0.14\textwidth]{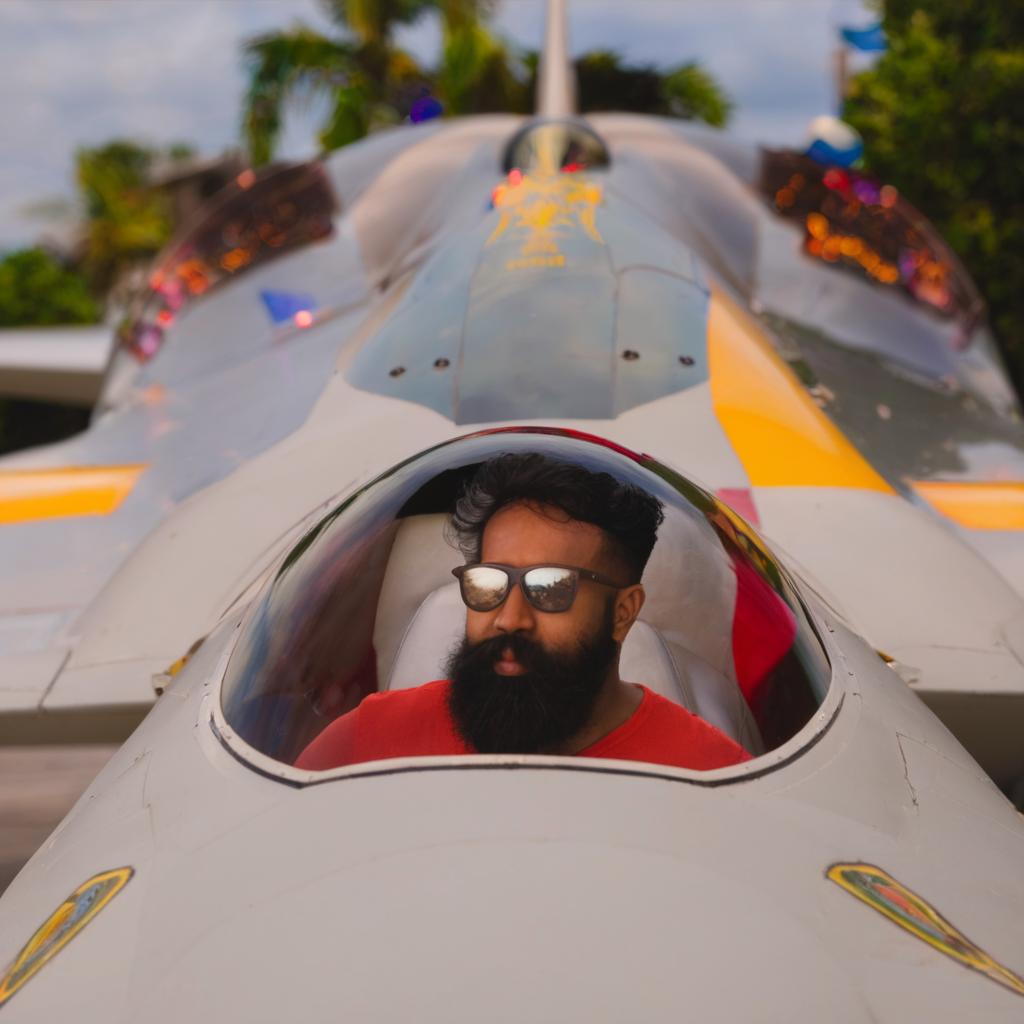} &
        \includegraphics[width=0.14\textwidth,height=0.14\textwidth]{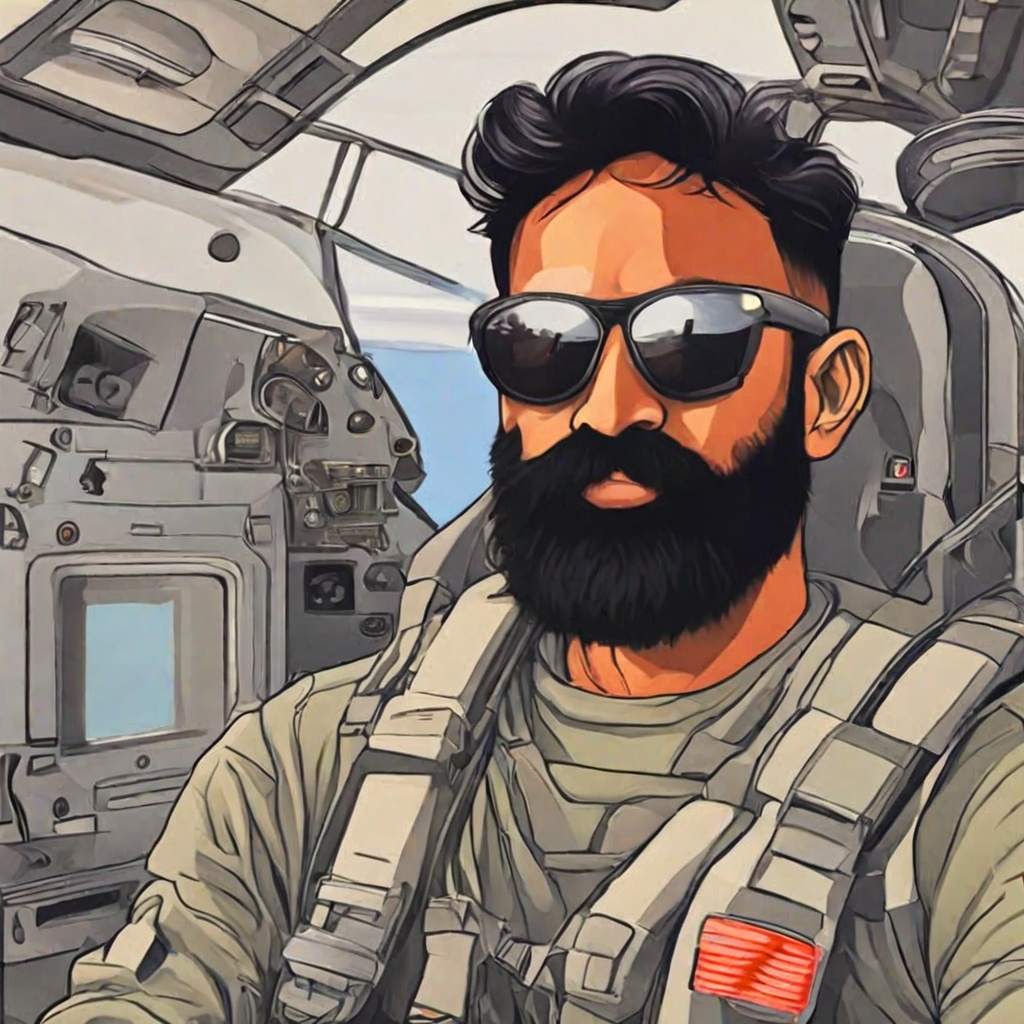} &
        
        \raisebox{0.06\textwidth}{\rotatebox[origin=t]{-90}{\scalebox{0.9}{\begin{tabular}{c@{}c@{}c@{}} Jet pilot\end{tabular}}}} \\

        \includegraphics[width=0.14\textwidth,height=0.14\textwidth]{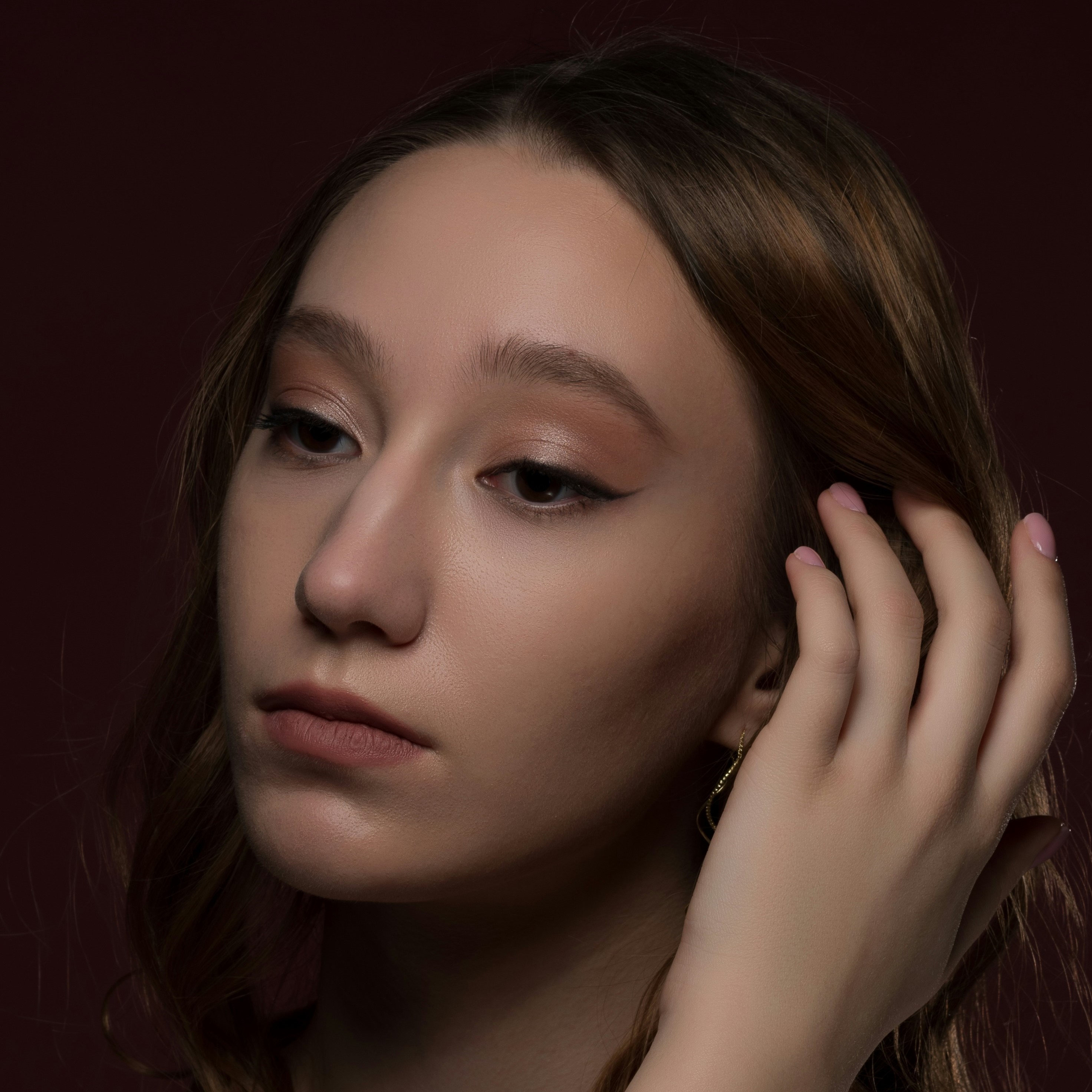} &

        \includegraphics[width=0.14\textwidth,height=0.14\textwidth]{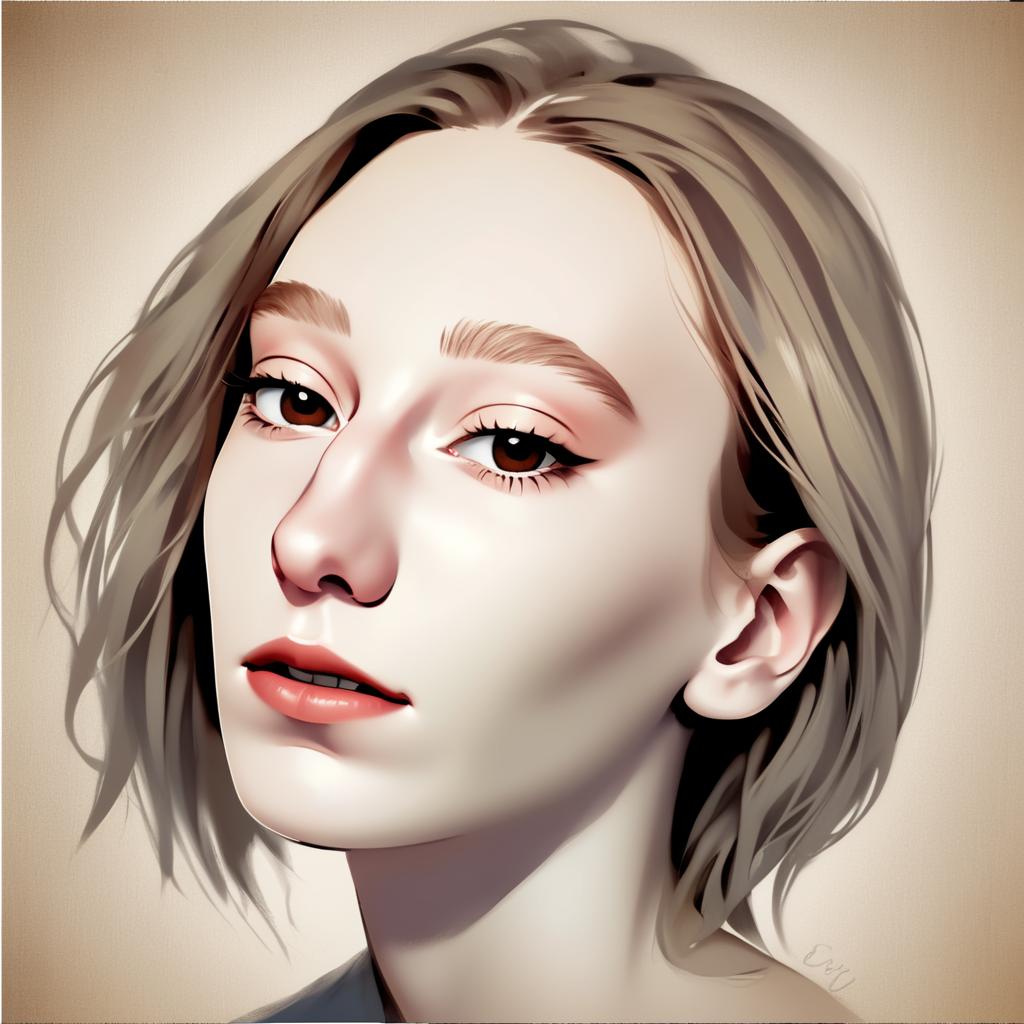} &
        \includegraphics[width=0.14\textwidth,height=0.14\textwidth]{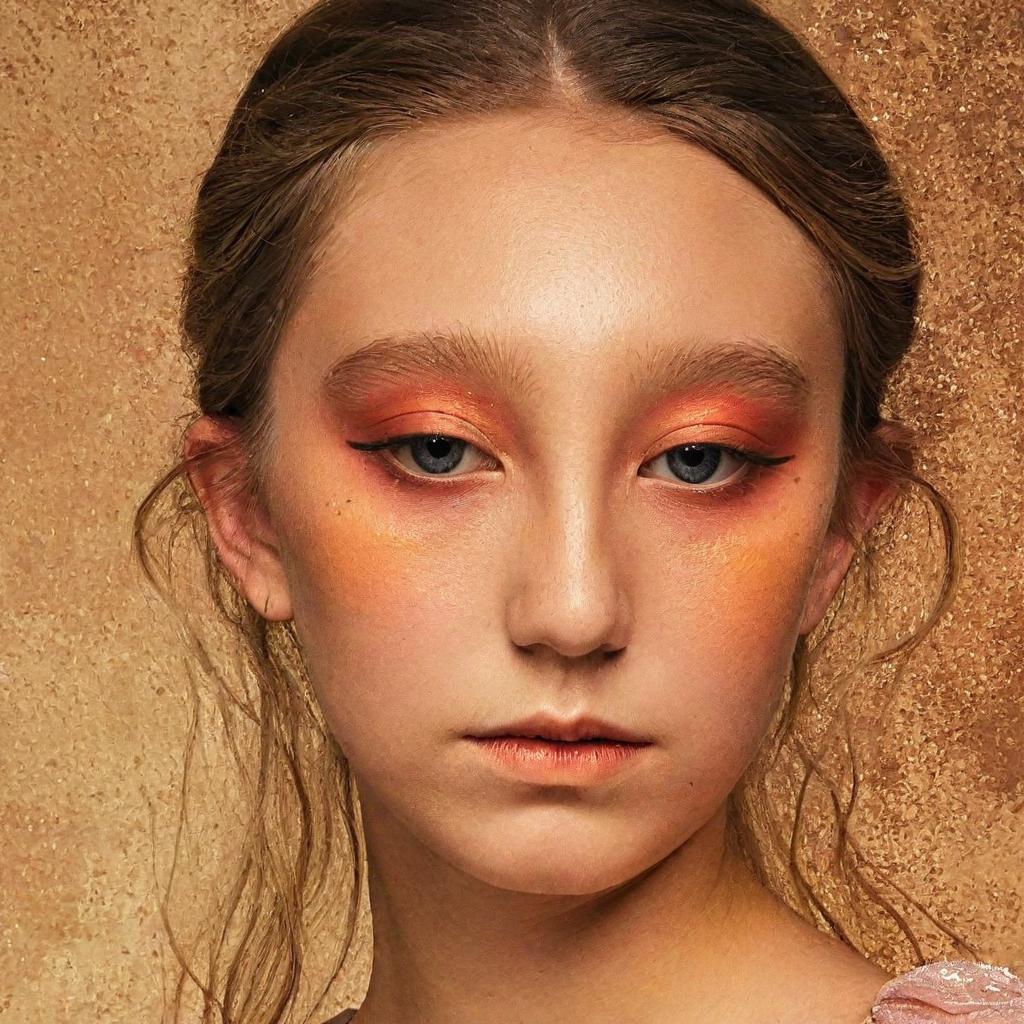} &
        \includegraphics[width=0.14\textwidth,height=0.14\textwidth]{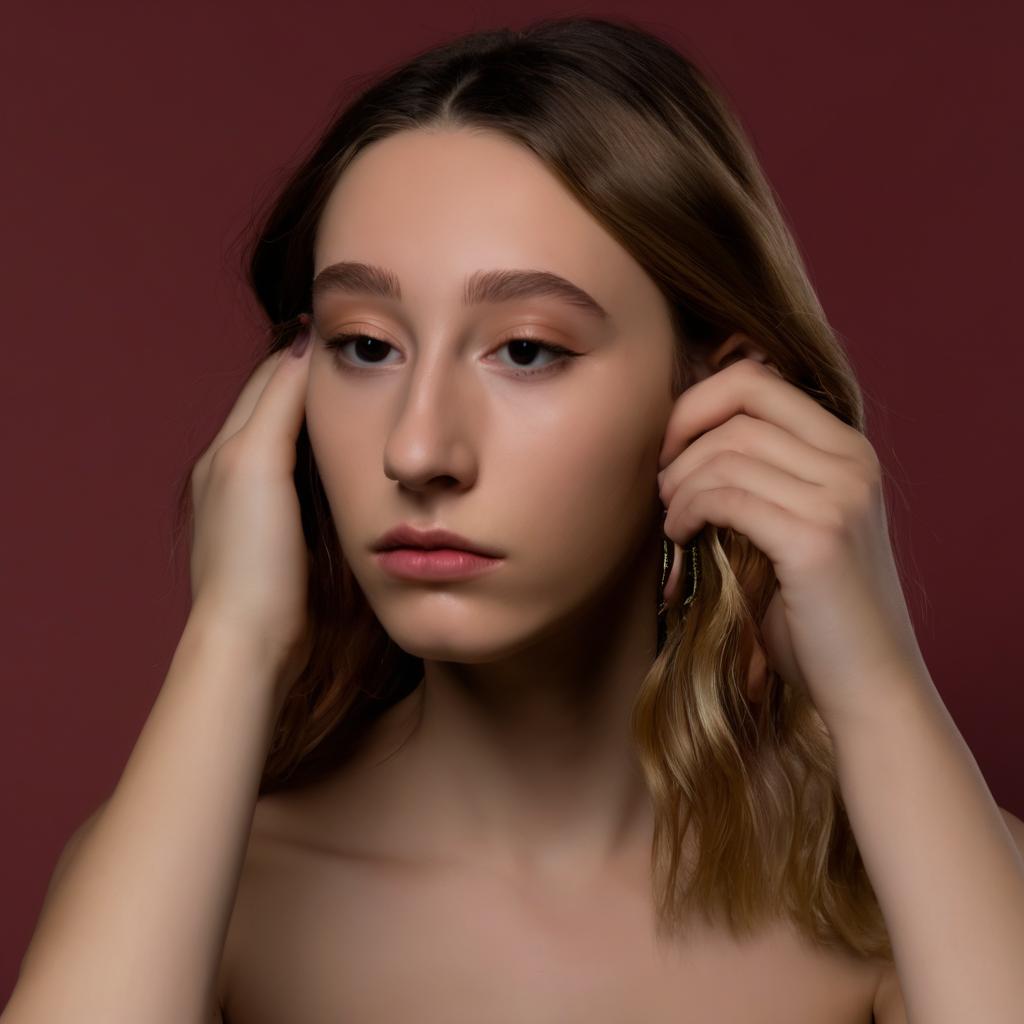} &
        \includegraphics[width=0.14\textwidth,height=0.14\textwidth]{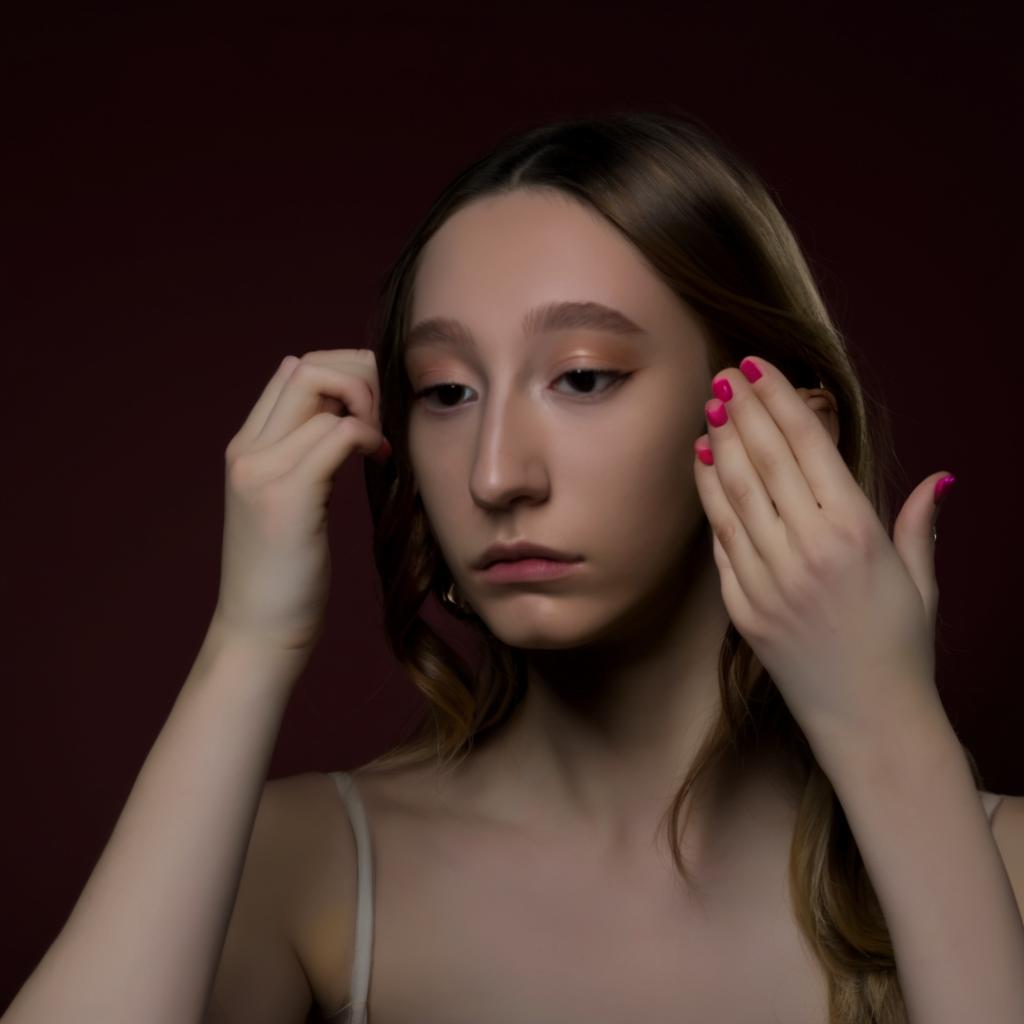} &
        \includegraphics[width=0.14\textwidth,height=0.14\textwidth]{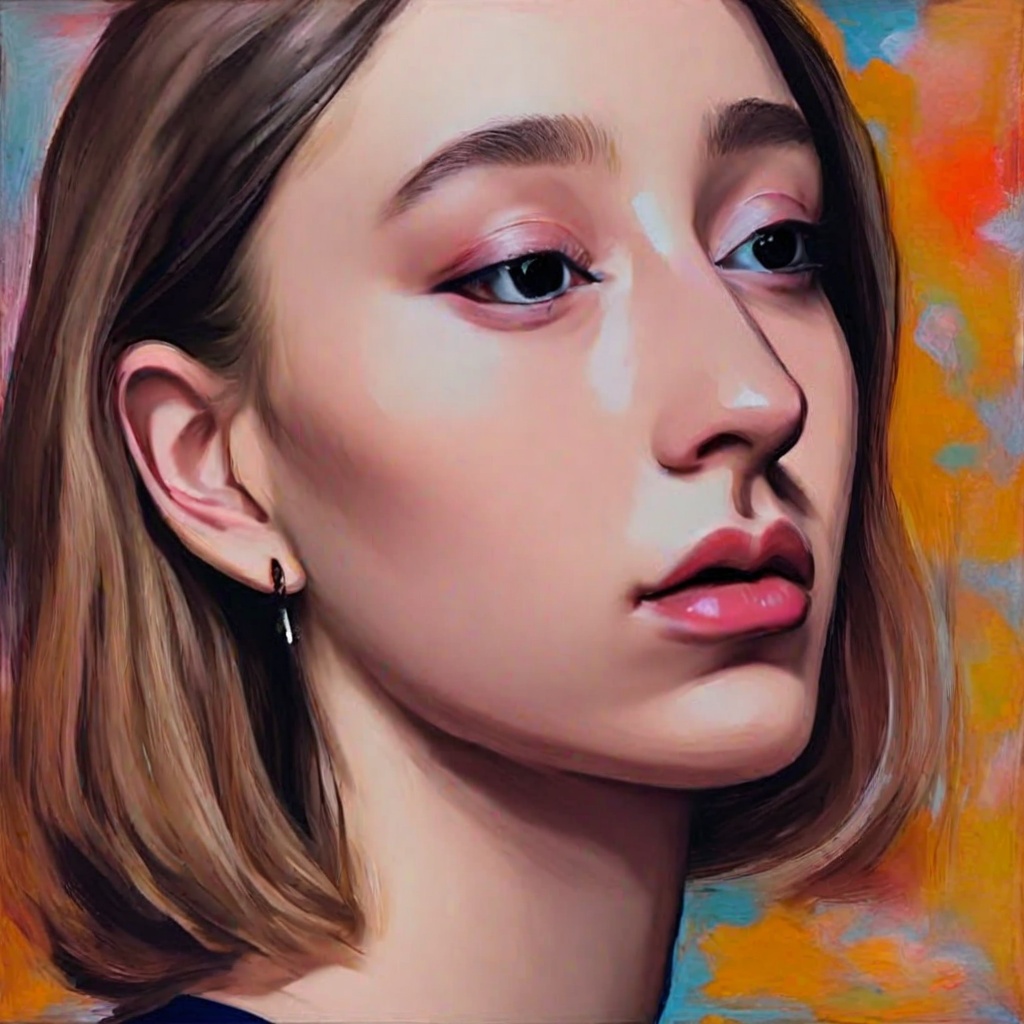} &
        
        \raisebox{0.056\textwidth}{\rotatebox[origin=t]{-90}{\scalebox{0.9}{\begin{tabular}{c@{}c@{}c@{}} Digital art\end{tabular}}}} \\

    \end{tabular}
    
    }
    \caption{Comparisons against prior and concurrent face-personalization encoders. IP-A ($1.0$) and ($0.5$) represent the IP-Adapter results with a scale of $1.0$ and $0.5$, respectively. Notably, IP-A ($1.0$) serves as the backbone which we fine-tune.}
    \label{fig:qual_comp}
\end{figure*}

\paragraph{\textbf{Qualitative Comparison.}}
We first conduct a qualitative comparison. Prior art often shows results on well-known celebrities. However, it is excessively easy to overfit on such identities (indeed, SDXL already contains tokens describing them, relegating an encoder's job to simply finding these tokens). Moreover, some recent papers test on identities contained in their training set (\eg the LAION-Faces dataset contains dozens of images of famous researchers such as Yann Lecun). To paint a full picture, we provide such comparisons in the supplementary. Here, we instead collect a small set of $50$ images with permissive licenses which were uploaded to \url{https://unsplash.com/} over the period of Feb 19th - March 4th (2024). Our assumption is that these portray novel individuals which are less likely to exist in any prior training set, and thus offer the 'cleanest' benchmark. Comparison results are shown in \cref{fig:qual_comp}. 

Notably, our method outperforms the IP-Adapter model that serves as our backbone in prompt-alignment, while matching or exceeding it on identity preservation. InstantID achieves comparable results. However, it heavily restricts the target's pose to that of the conditioning image. Moreover, InstantID was trained on significantly more data and compute ($60$ million images and $48$ GPUs over an undisclosed time-frame). We are hopeful that our approach could also be applied on top of an InstantID backbone and lead to improved results, but were unable to approach their scale of training here.

\paragraph{\textbf{Quantitative Comparison.}}
Next, we move to quantitative evaluations. Here, we follow prior work~\cite{gal2023encoder} and compare the baselines across two metrics - identity similarity and prompt alignment. Our prompts (detailed in the supplementary) include photo-realistic reconstructions, but also stylization and material change which prior art often struggles with. To measure identity similarity we use CurricularFace~\cite{huang2020curricularface}, which differs from both our loss network and from the backbones used to extract features for the baselines. Text alignment is measured using CLIP similarity (using the ViT-B/16 version which differs from the IP-Adapter backbone). We report both metrics using two sets: (1) $5,000$ identities randomly sampled from the FFHQ dataset~\cite{karras2019style} and (2) The $50$ unsplash identities. The results are shown in \cref{tab:quant}. Our method is situated on the pareto front, providing good identity preservation and high prompt alignment. Note in particular that we outperform the backbone IP-Adapter in its typically used setup ($\alpha=0.5$) on both metrics. 

Since identity metrics are highly sensitive to both the success of editing (\ie stylized images will have lower identity scores than failed edits), and to poses (which InstantID copies), we also verify our results using a user-study. There, for each question we showed users a reference image and a prompt pair, and the outputs of two models conditioned on this pair (ours, and a random baseline). We asked users to select the image that better preserves the reference identity and better aligns with the prompt. In total, we collected 460 responses from 43 different users. In \cref{tab:ustudy} we report the percentage of users that preferred our method over each baseline. The results largely align with the automatic metrics, showing that our approach is preferred to the backbone on which we build. InstantID still outperforms all methods, primarily on account of its much improved editability, but the margin is less severe.

\begin{table}
\caption{Quantitative comparisons against prior and concurrent SDXL face-encoders}\label{tab:quant}
\begin{tabular}{l : c c : c c}
& \multicolumn{2}{c}{FFHQ-5000} & \multicolumn{2}{c}{Unsplash-50} \\
& ID $\uparrow$ & CLIP-T $\uparrow$ & ID $\uparrow$ & CLIP-T $\uparrow$ \\
\toprule
Ours & 0.345 & 26.33 & 0.308 & 26.79 \\
IP-A (0.5) & 0.268 & 25.82 & 0.250 & 26.36 \\
IP-A (1.0) & 0.368 & 21.39 & 0.387 & 22.06 \\
PhotoMaker & 0.344 & 26.69 & 0.218 & 27.19 \\
\midrule
InstantID & 0.631 & 28.58 & 0.612 & 29.06 \\
\end{tabular}
\end{table}

\begin{table}
\caption{Ablation study results}\label{tab:ablation_quant}
\begin{tabular}{l : c c : c c}
& \multicolumn{2}{c}{FFHQ-5000} & \multicolumn{2}{c}{Unsplash-50} \\
& ID $\uparrow$ & CLIP-T $\uparrow$ & ID $\uparrow$ & CLIP-T $\uparrow$ \\
\toprule
Baseline IPA   & 0.368 & 21.39 & 0.387 & 22.06 \\
\midrule
$+$ Our Data   & 0.220 & 28.14 & 0.205 & 28.25 \\
w/ ID-Loss x0  & 0.282 & 27.50 & 0.272 & 27.74 \\
w/ ID-Loss LCM & 0.301 & 27.31 & 0.281 & 27.74 \\
$+$ KV Injection & 0.345 & 26.33 & 0.308 & 26.79 \\
\end{tabular}
\end{table}

\begin{table}
\caption{User study results. For each matchup, we report the fraction of users who prefered our method, and the fraction that preferred the baseline.}\label{tab:ustudy}
\renewcommand{\tabcolsep}{5pt}
\begin{tabular}{c c c c c}
& PhotoMaker & IP-A ($1.0$) & IP-A ($0.5$) & InstantID \\
\toprule
Ours     & $71.18\%$ & $82.25\%$ & $57.32\%$ & $44.06\%$ \\
Baseline & $28.82\%$ & $17.75\%$ & $42.68\%$ & $55.94\%$ 
\end{tabular}
\end{table}

\subsection{Ablation}

We conduct an ablation study to evaluate the contribution of each of our suggested components. Here, we evaluate the following setups: 
(1) The baseline IP-Adapter tuned on our data, without extended attention or identity losses. (2) The setup of (1) + an identity loss derived through standard $x0$ approximations. (3) The setup of (1) + an identity loss derived through an LCM shortcut. (4) Our full model (the setup of (3) + attention injection branch). For comparison, we again provide the backbone IP Adapter results. Quantitative results are provided in \cref{tab:ablation_quant}. The use of our novel dataset greatly improves prompt alignment, at the cost of some identity fidelity. We attribute this in part to our smaller training scale, and to the fact that the identity score is impacted by stylization. Adding an identity loss significantly increases identity preservation, while passing this loss through the LCM shortcut provides a noticeably larger increase. Finally, injecting attention features leads to even better identity alignment, but at the cost of editability.

Our results demonstrate that appending losses through an LCM-shortcut mechanism can provide improvements over the direct approximation approach. All-in-all, the combination of our components leads to a high level of both identity preservation and prompt alignment.

\section{Limitations and ethic concerns}
While our model can improve existing encoders, it is not free of limitations. First, like prior tuning-free encoders, it still falls short of the quality provided by optimization-based methods. This is particularly noticeable when working with inputs that are noticeably out-of-domain compared to standard face imagery. It also tends to copy accessories, and may default to artistic styles if photo-realism is not requested (see supplementary for examples).

Secondly, our model may still suffer from biases inherent in both the backbone that we built on, as well as the diffusion model itself. Hence, it may serve to amplify social biases.
Moreover, facial editing and generation software can be used to spread disinformation or defame individuals. Existing detection tools can help mitigate such risks~\cite{wang2023detection,corvi2023detection}, and we hope that these continue to improve.

\section{Conclusions}
\label{sec:conclusions}

In this work, we presented LCM-Lookahead, a novel mechanism for applying image-space losses to diffusion training using a fast-sampling-based shortcut mechanism. We then build on top of this mechanism to provide better identity signal to a personalization encoder, leading to improved identity preservation. 
Our work further explores the shortcomings of current personalization encoders and proposes two additional techniques to further improve their results. First, we show that consistent data generation methods can greatly impact prompt-alignment quality, and that SDXL-Turbo can serve to create such data. Indeed, fine-tuning on such data can even restore editability to encoders which have overfit on the photo-realistic domain. Finally, we show that the common self-attention key-value injection mechanism can also be applied to encoder-based personalization, improving the faithfulness of the generated results. We hope that both our LCM-Lookahead and our improved training scheme will serve to further push the boundaries of text-to-image personalization.

\begin{acks}
This work was partially supported by BSF (grant 2020280) and ISF (grants 2492/20 and 3441/21). 
\end{acks}

\bibliographystyle{ACM-Reference-Format}
\bibliography{main}

\end{document}


\title{LCM-Lookahead for Encoder-based Text-to-Image Personalization - Supplementary Materials} 





 
 

\maketitle

\section{Additional comparisons}

\subsection{Celebrity Comparisons}

\begin{figure*}[t]

    \centering
    \setlength{\tabcolsep}{1.5pt}
    \hspace{-0.045\textwidth}
    {\large 
    \begin{tabular}{c c c c c c c}

        {\large Input} & 
        {\large InstantID} &{\large PhotoMaker} & {\large IP-A (0.5)} & {\large IP-A (1.0)} & {\large Ours} & \\

        \includegraphics[width=0.16\textwidth,height=0.16\textwidth]{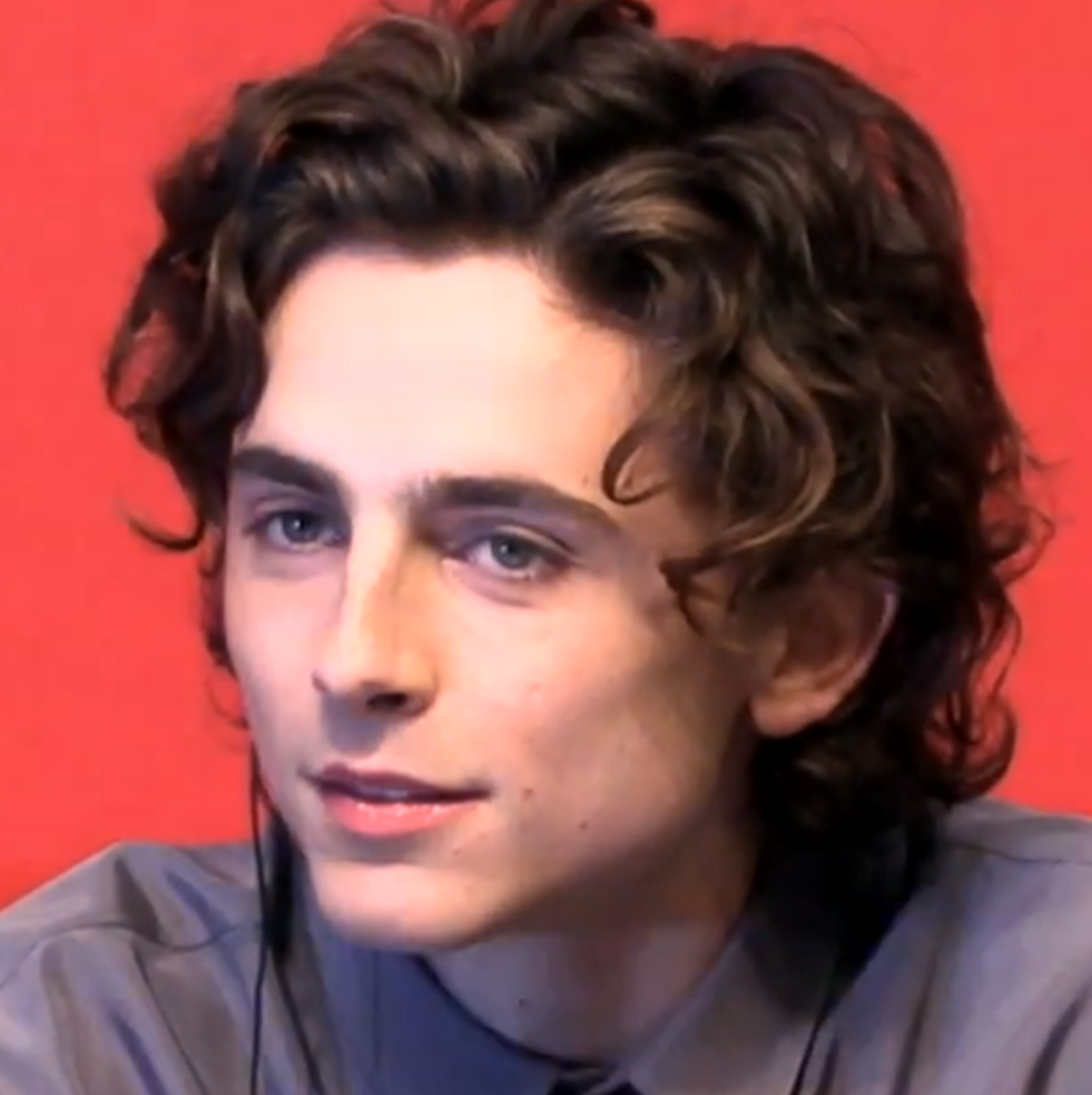} &

        \includegraphics[width=0.16\textwidth,height=0.16\textwidth]{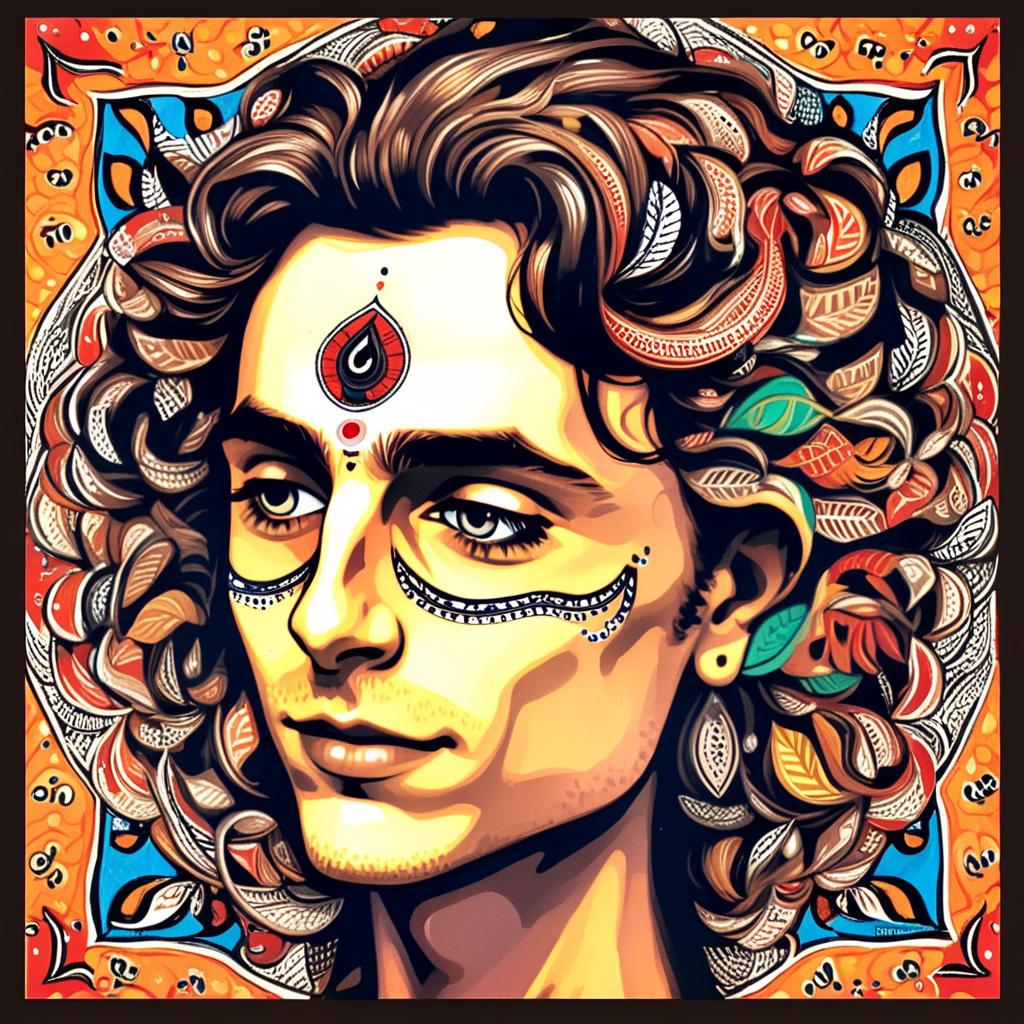} &
        \includegraphics[width=0.16\textwidth,height=0.16\textwidth]{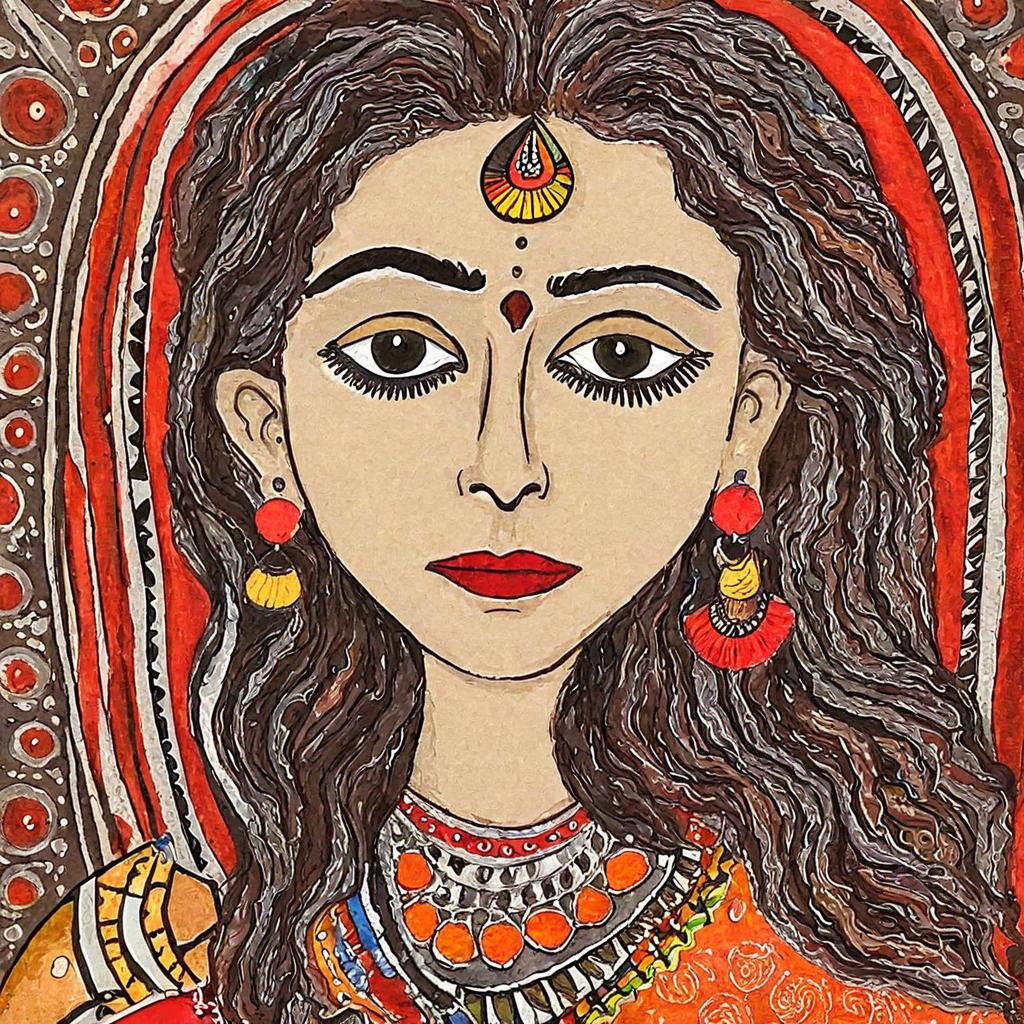} &
        \includegraphics[width=0.16\textwidth,height=0.16\textwidth]{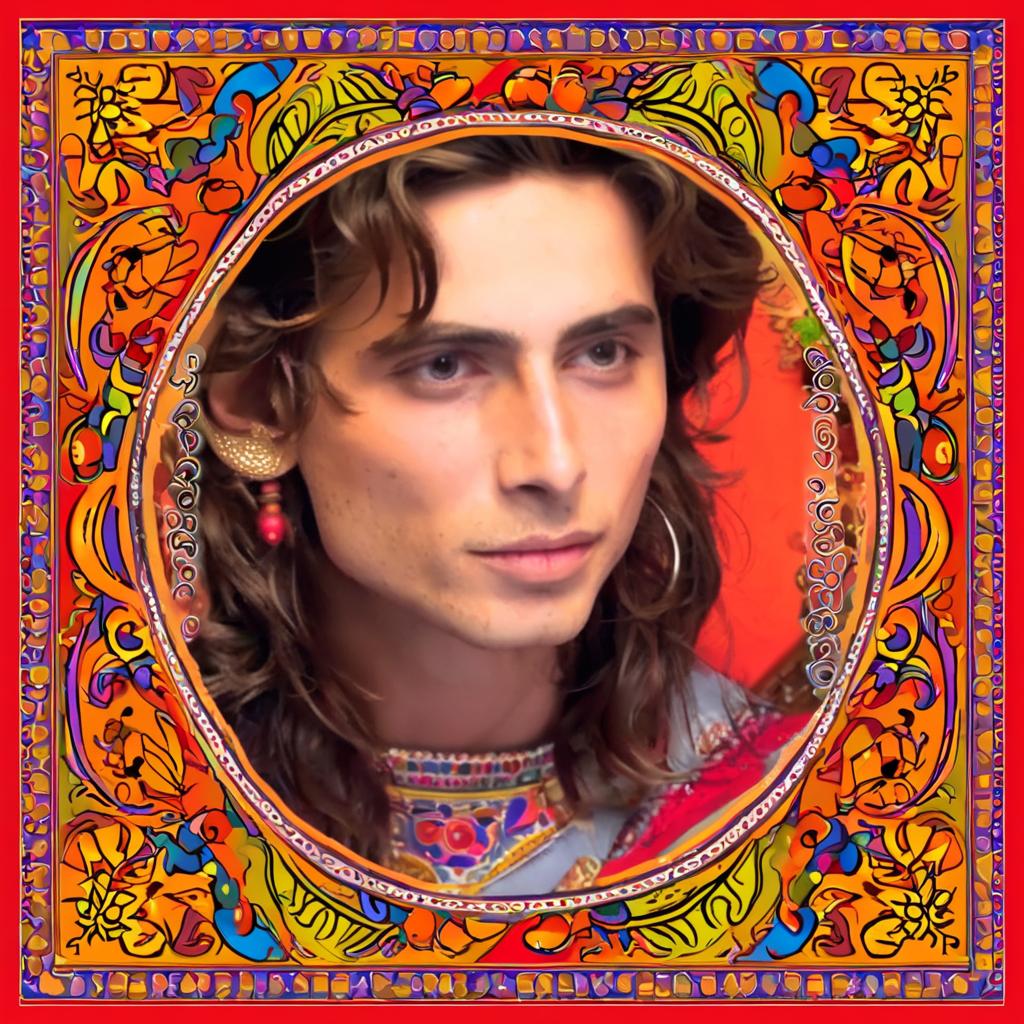} &
        \includegraphics[width=0.16\textwidth,height=0.16\textwidth]{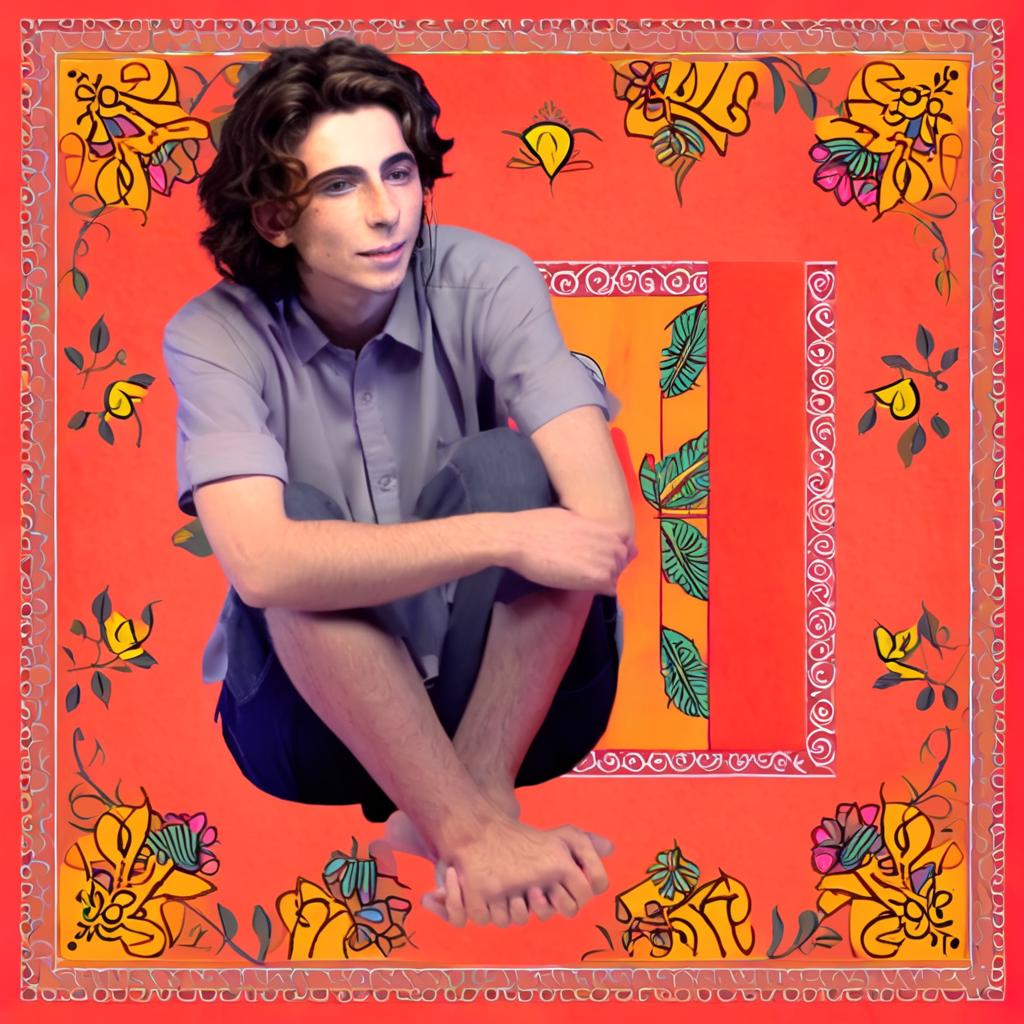} &
        \includegraphics[width=0.16\textwidth,height=0.16\textwidth]{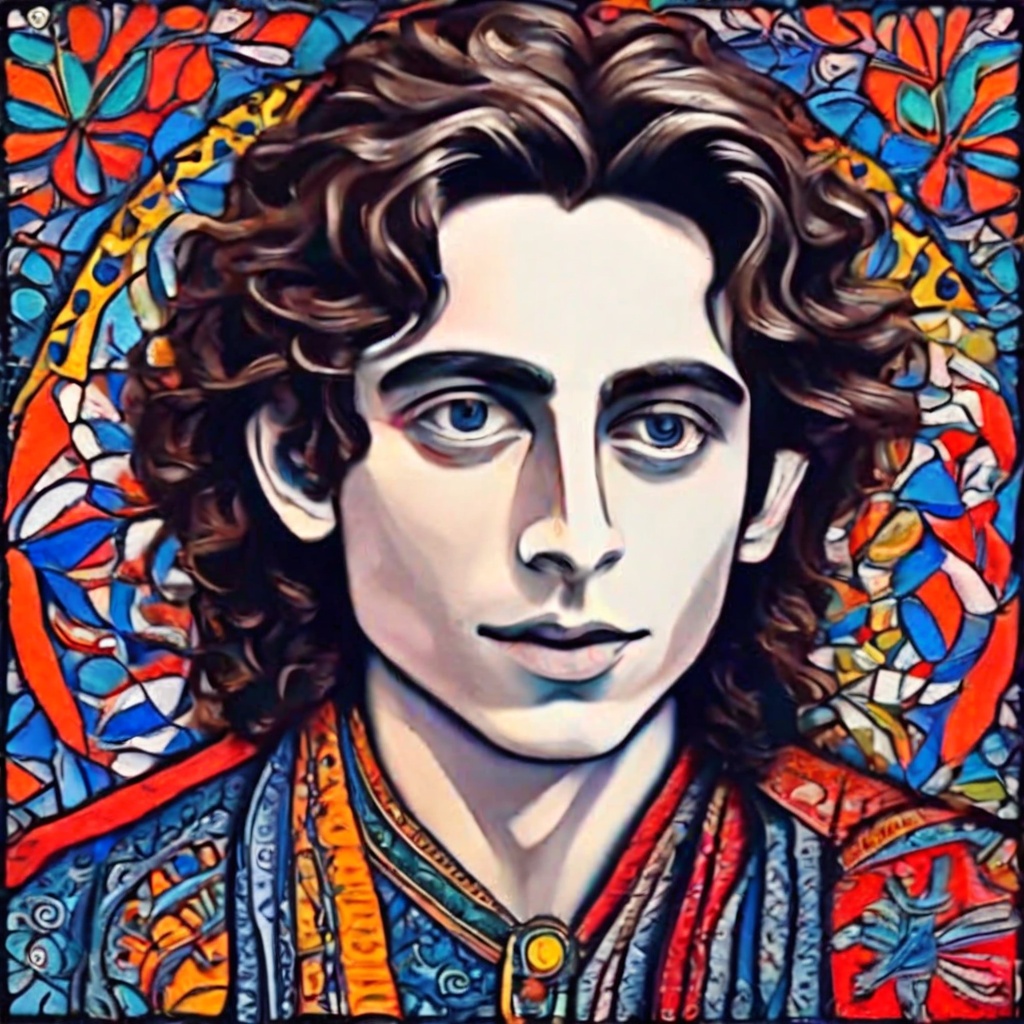} &
        
        \raisebox{0.068\textwidth}{\rotatebox[origin=t]{-90}{\scalebox{0.9}{\begin{tabular}{c@{}c@{}c@{}} Madhubani\end{tabular}}}}
        \\

        \includegraphics[width=0.16\textwidth,height=0.16\textwidth]{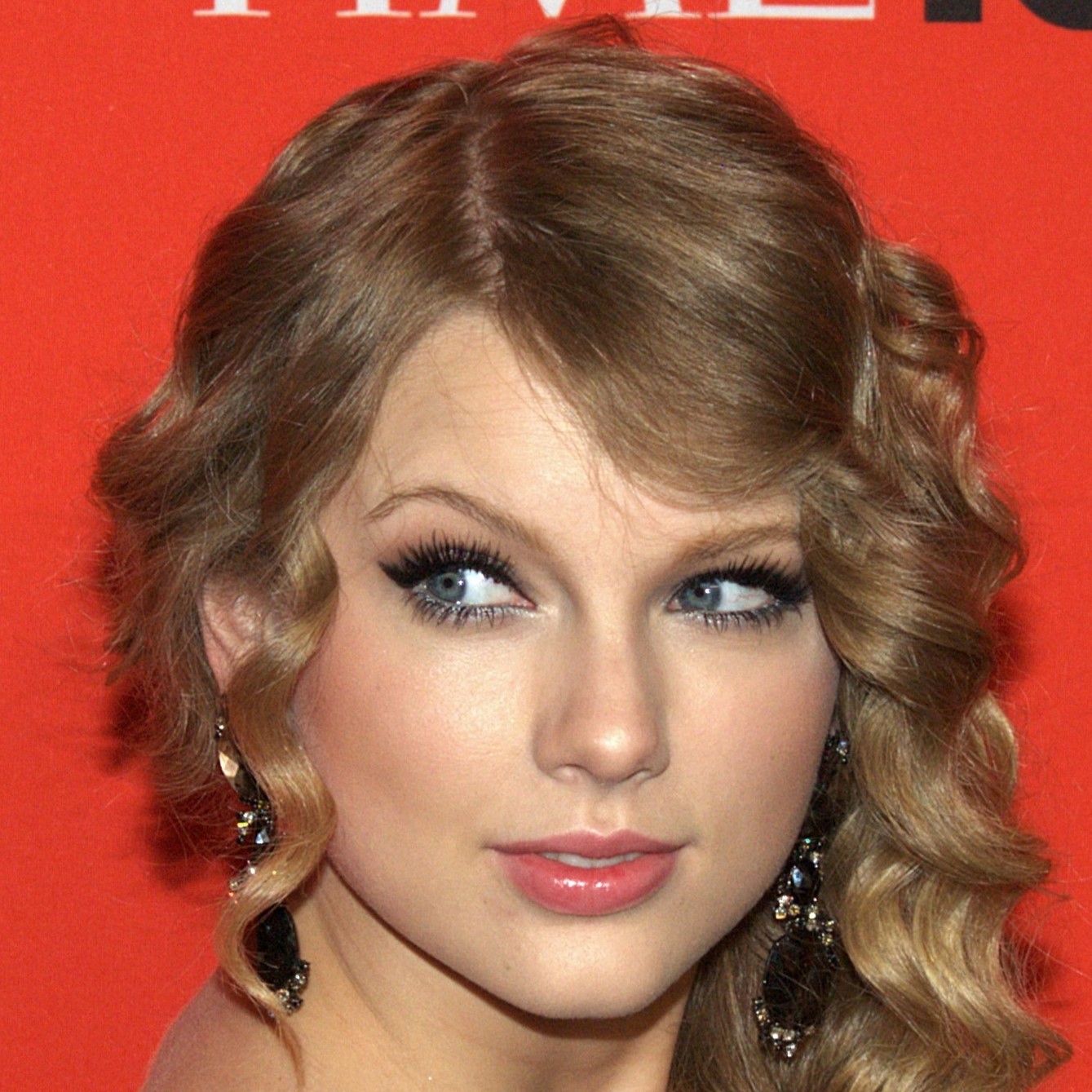} &
        
        \includegraphics[width=0.16\textwidth,height=0.16\textwidth]{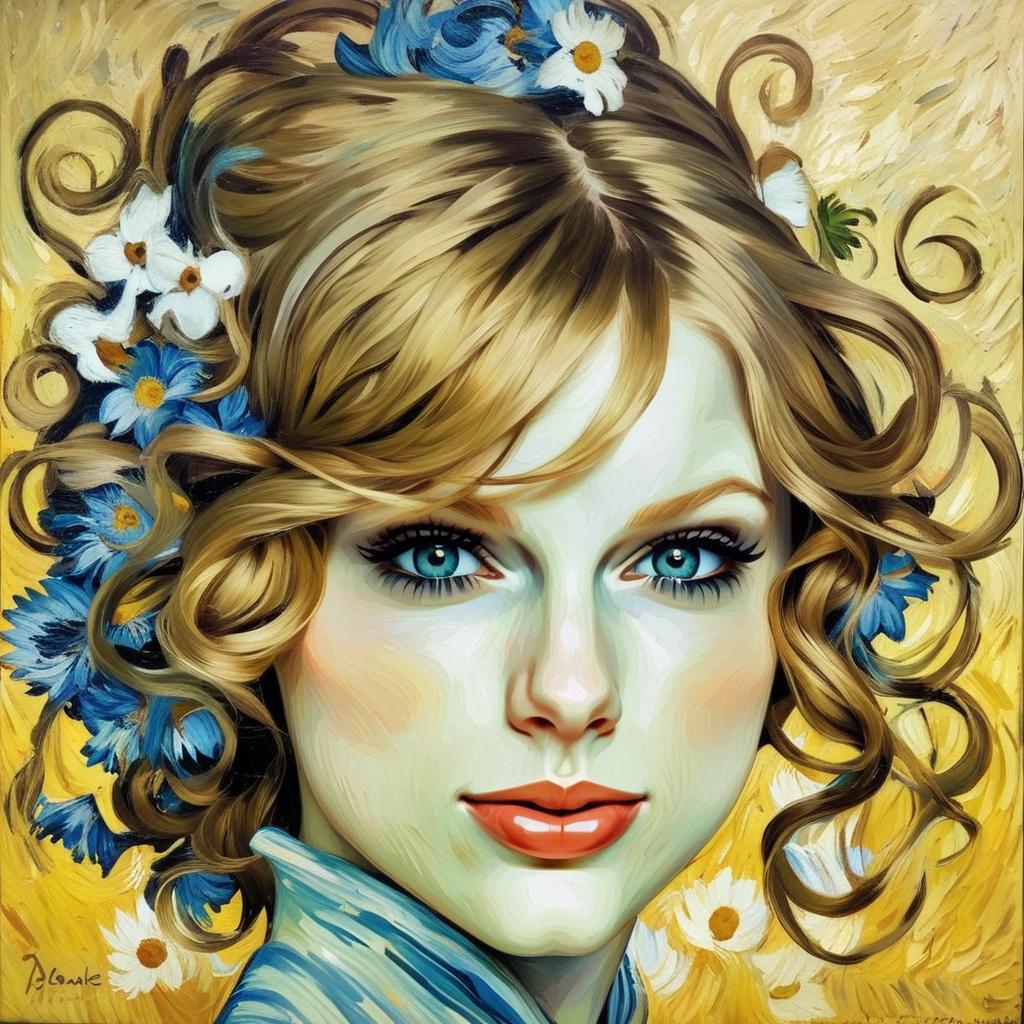} &
        \includegraphics[width=0.16\textwidth,height=0.16\textwidth]{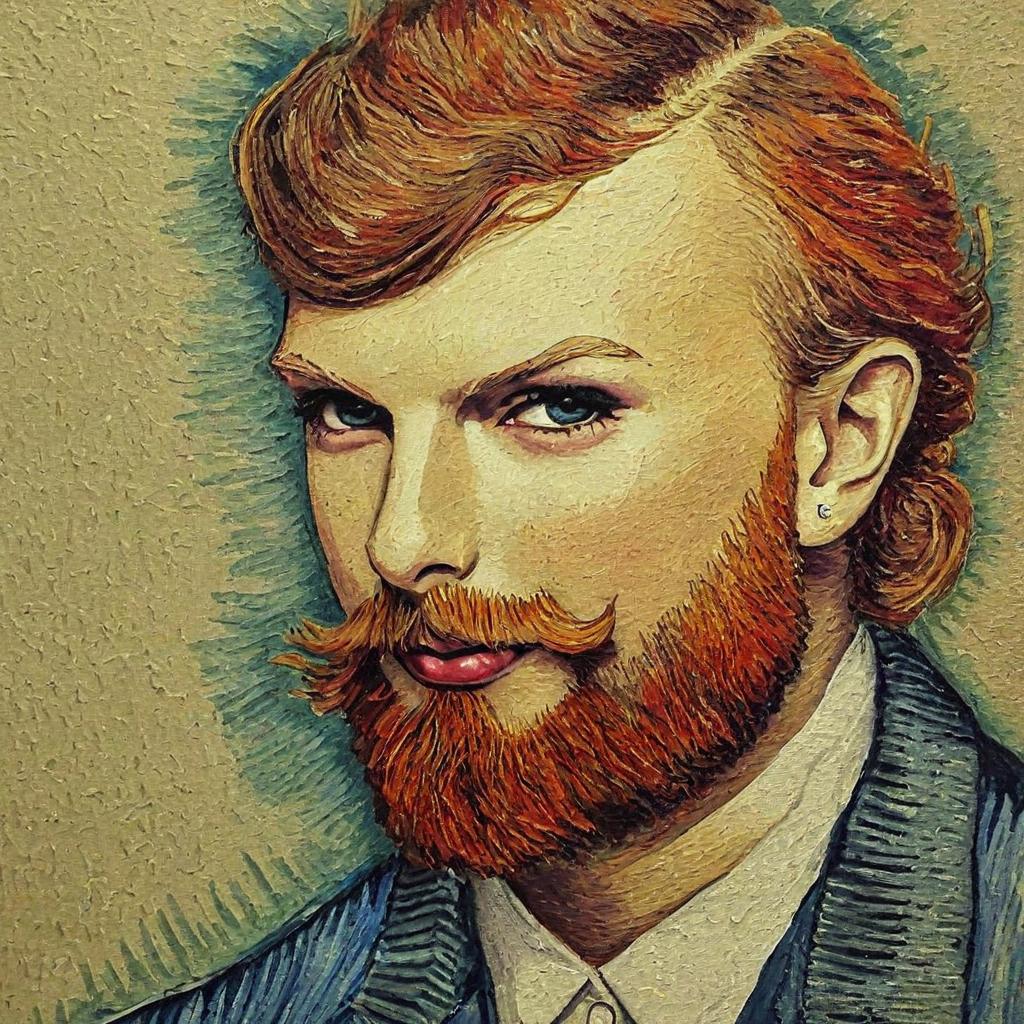} &
        \includegraphics[width=0.16\textwidth,height=0.16\textwidth]{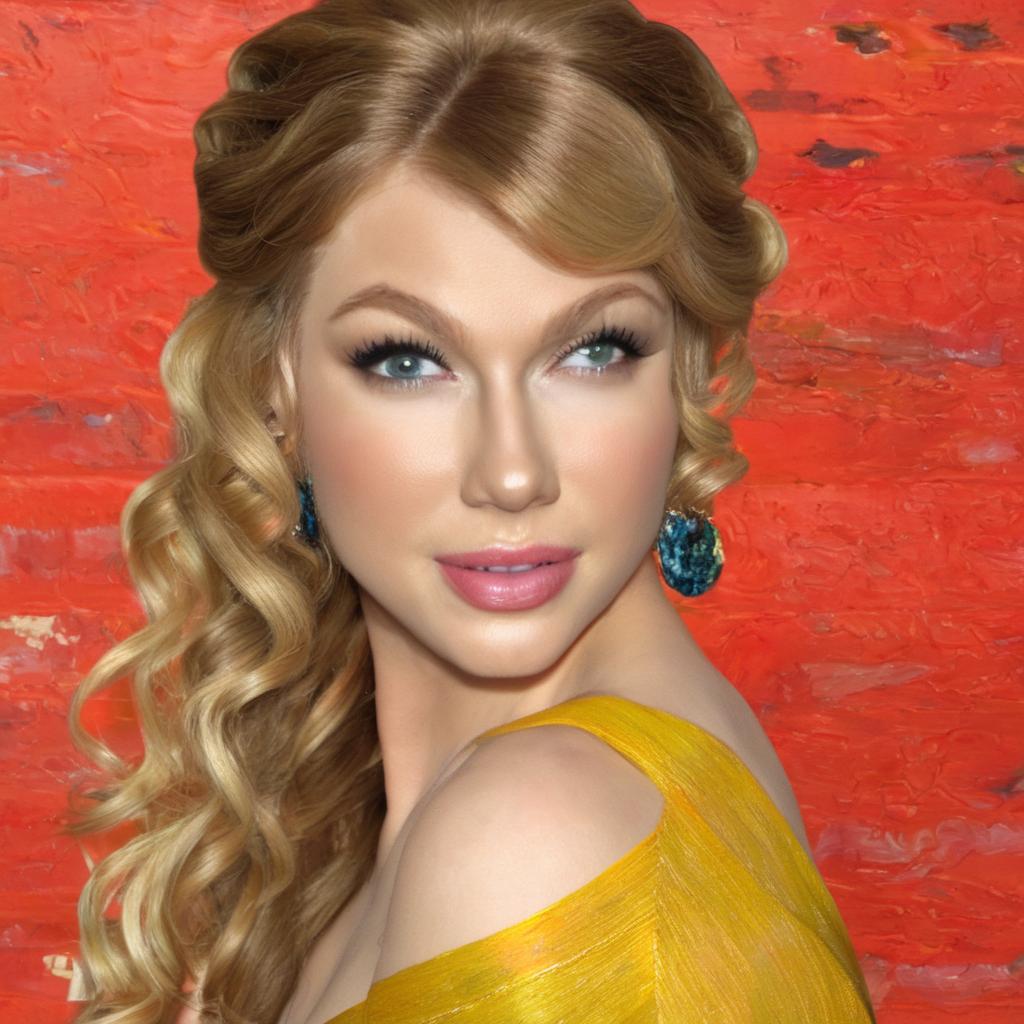} &
        \includegraphics[width=0.16\textwidth,height=0.16\textwidth]{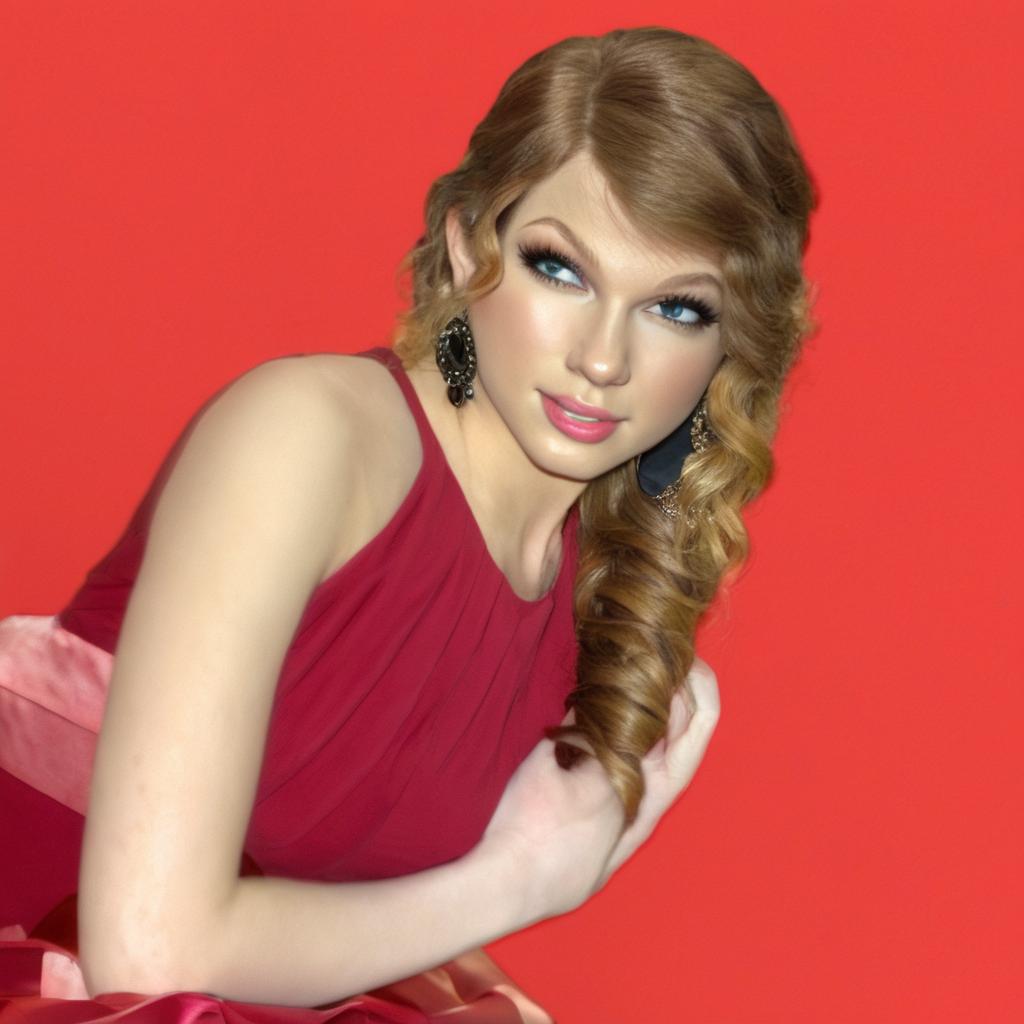} &
        \includegraphics[width=0.16\textwidth,height=0.16\textwidth]{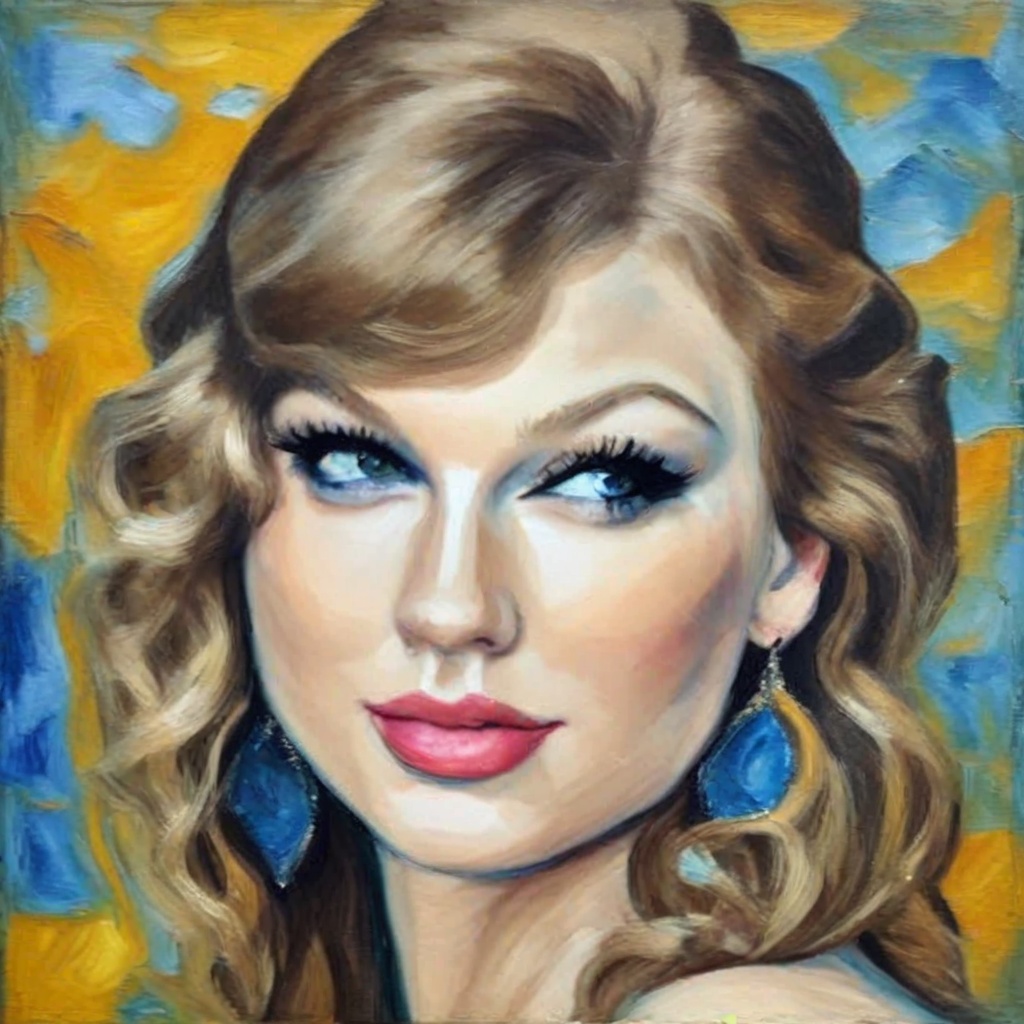} &
        
        \raisebox{0.068\textwidth}{\rotatebox[origin=t]{-90}{\scalebox{0.9}{\begin{tabular}{c@{}c@{}c@{}} Van Gogh\end{tabular}}}} \\

        \includegraphics[width=0.16\textwidth,height=0.16\textwidth]{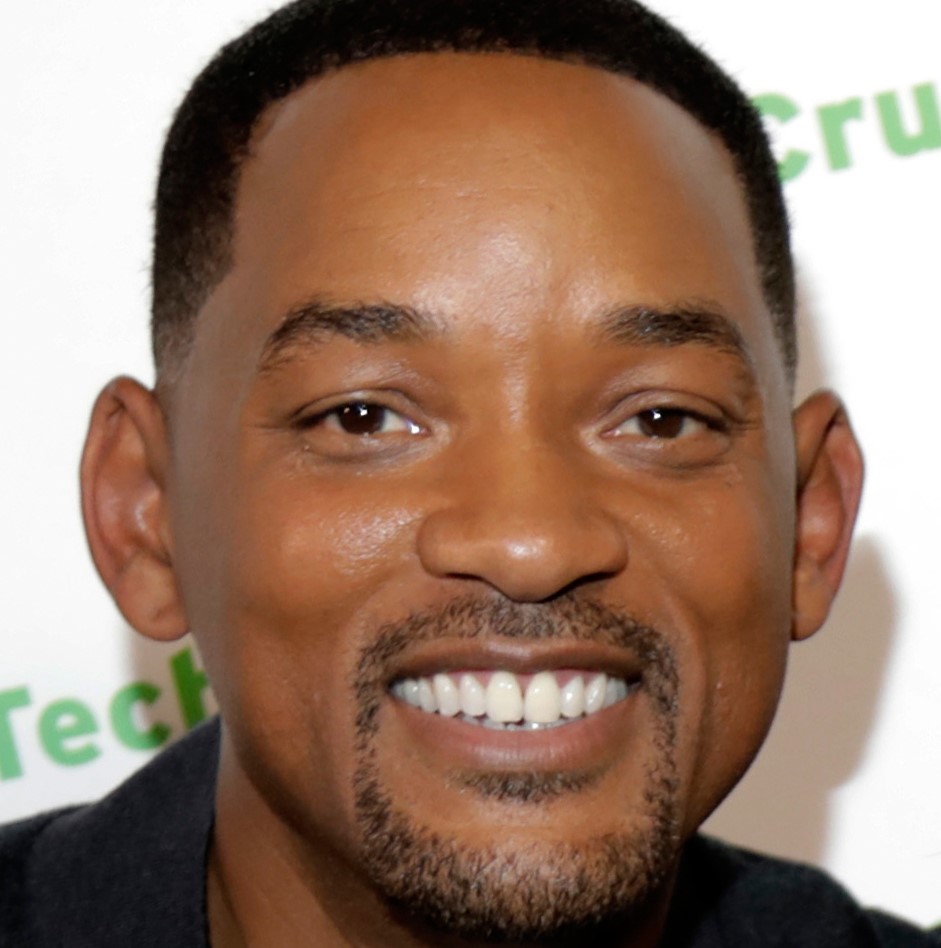} &

        \includegraphics[width=0.16\textwidth,height=0.16\textwidth]{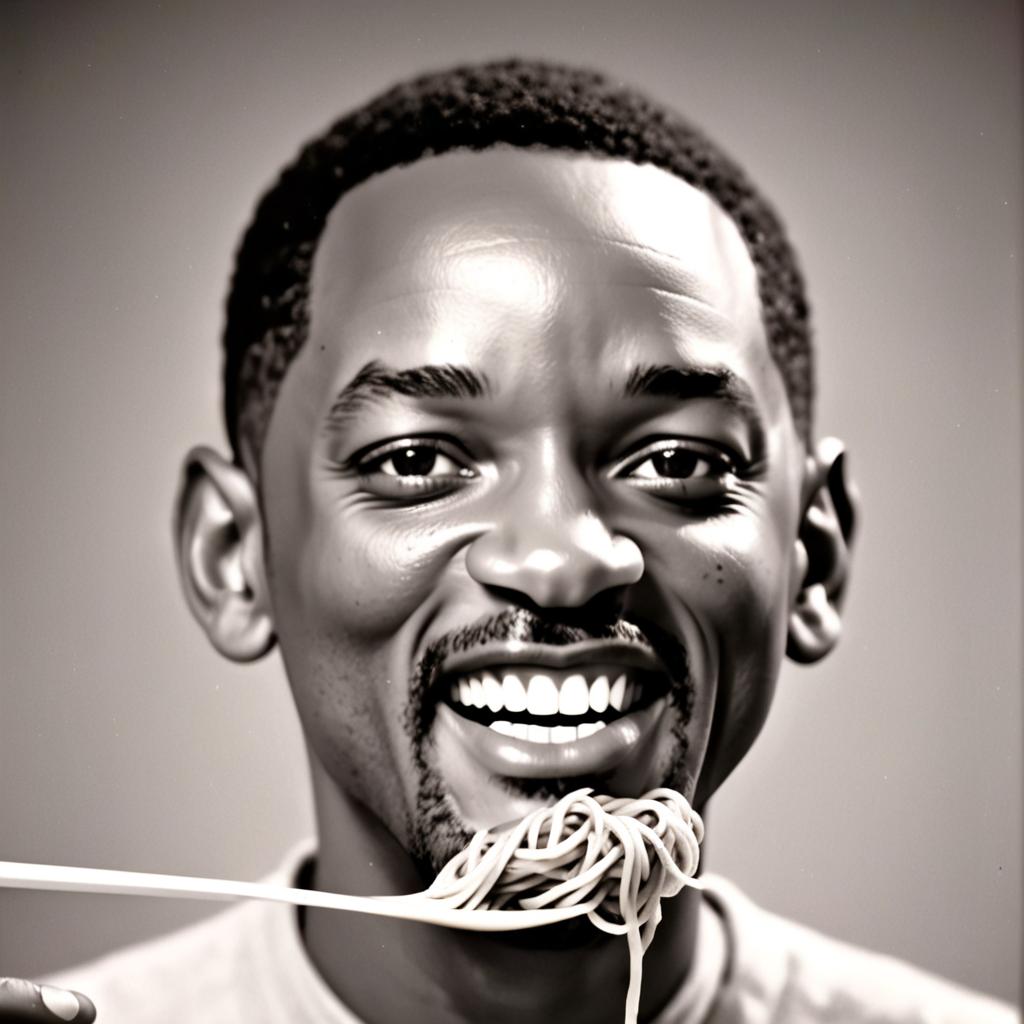} &
        \includegraphics[width=0.16\textwidth,height=0.16\textwidth]{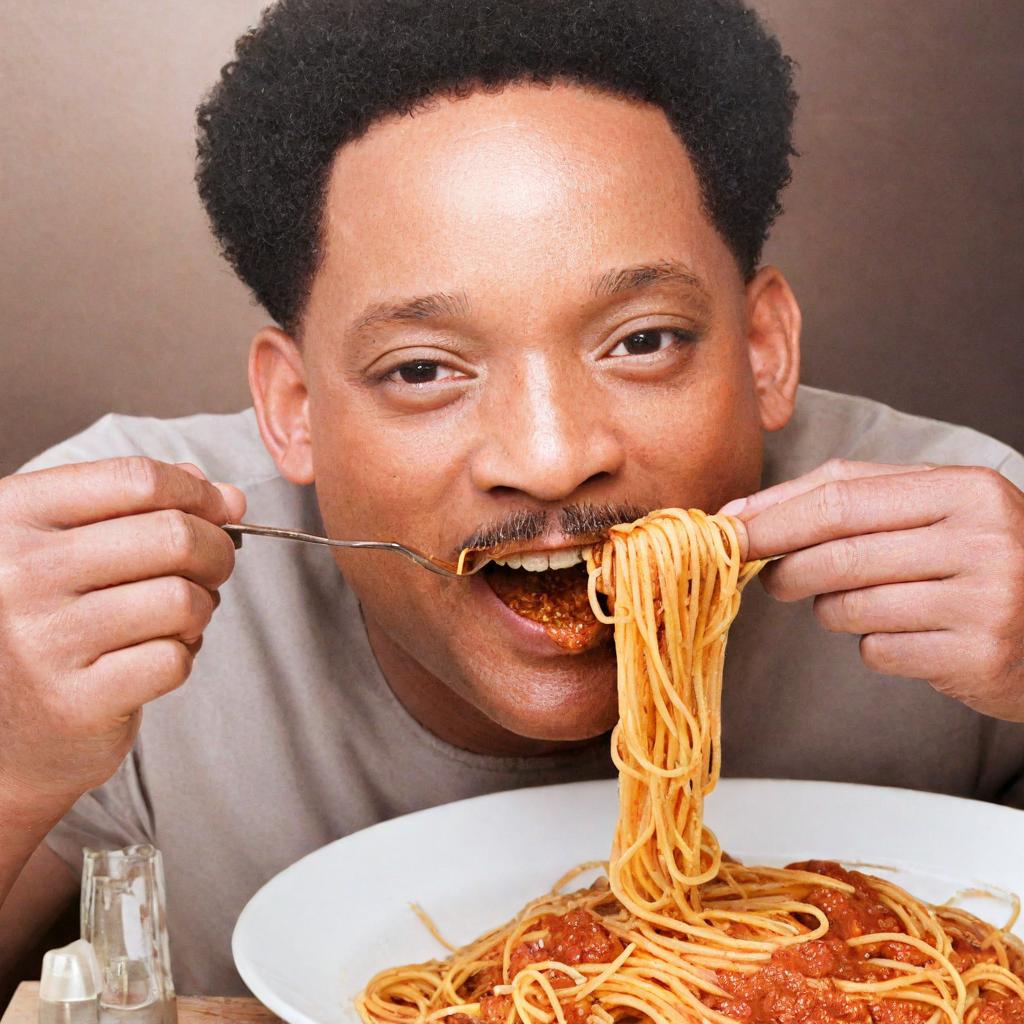} &
        \includegraphics[width=0.16\textwidth,height=0.16\textwidth]{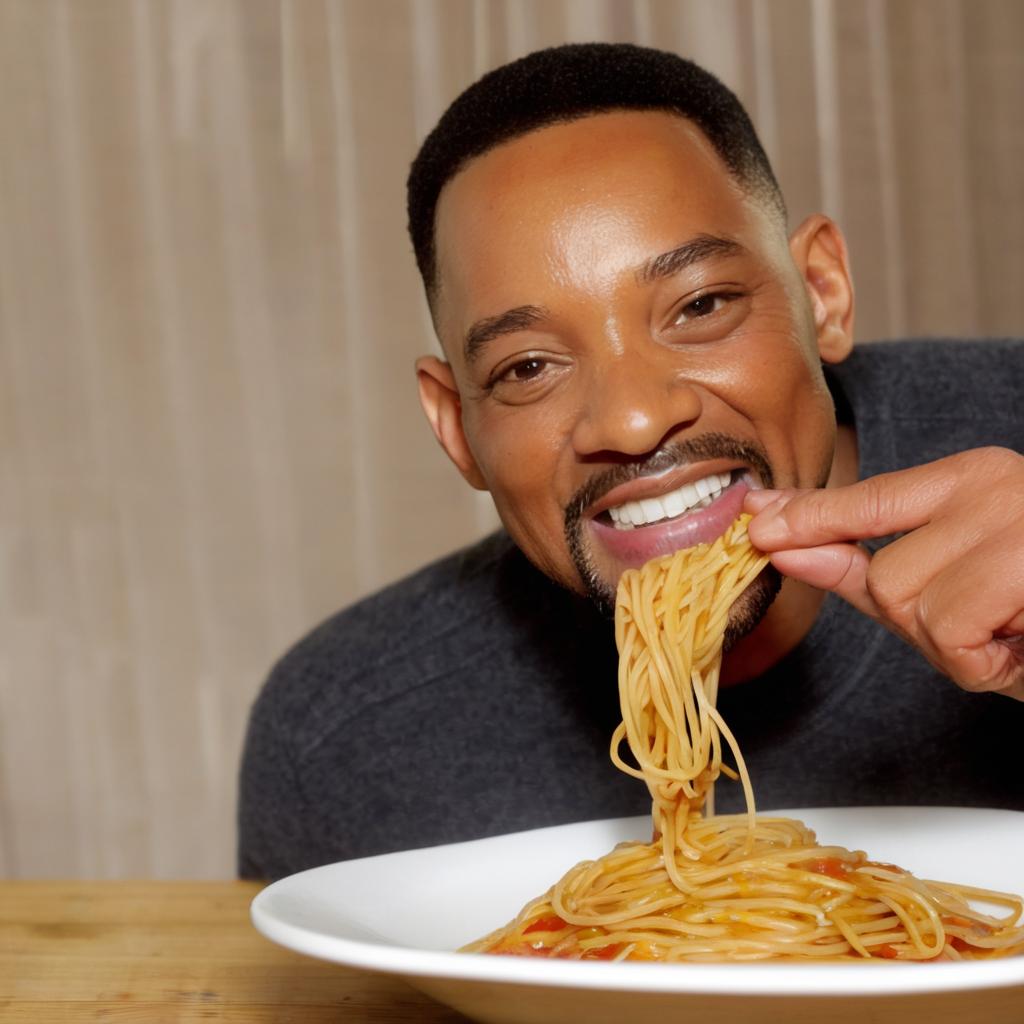} &
        \includegraphics[width=0.16\textwidth,height=0.16\textwidth]{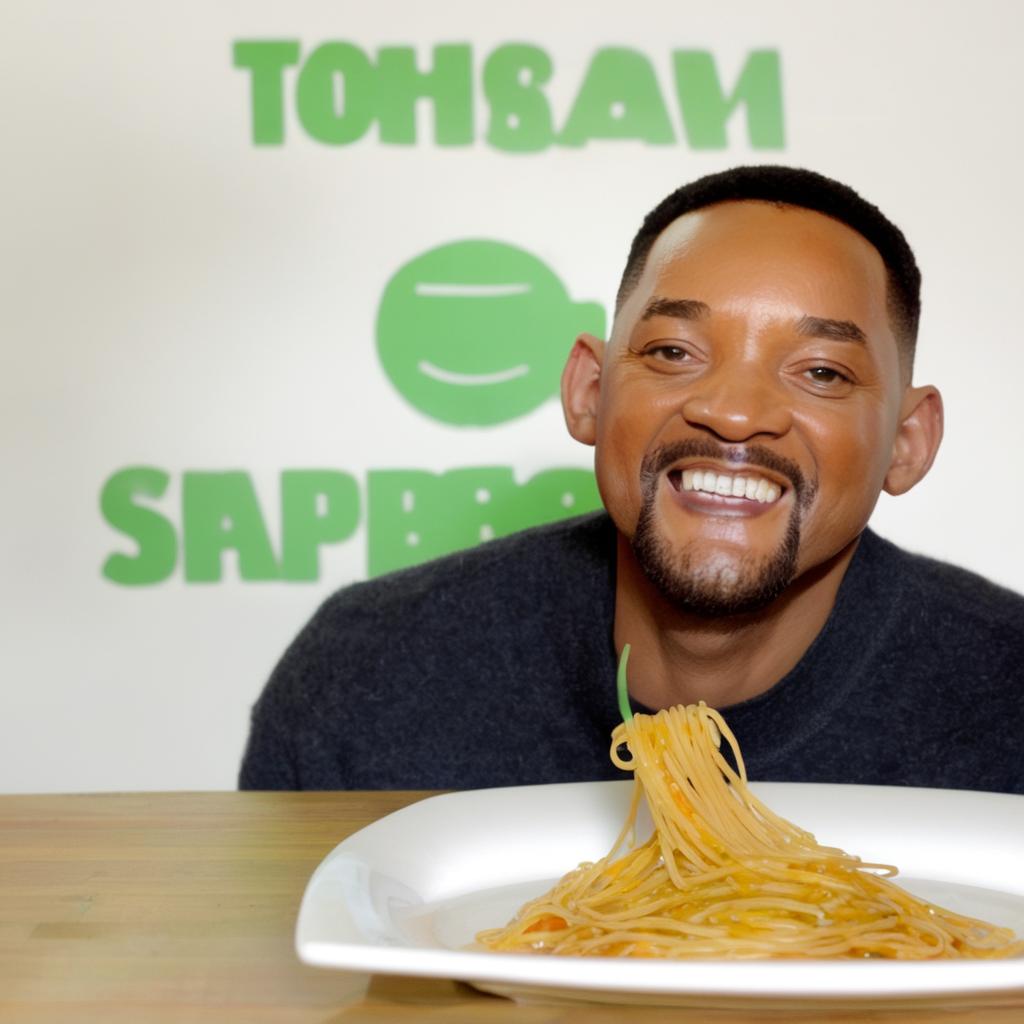} &
        \includegraphics[width=0.16\textwidth,height=0.16\textwidth]{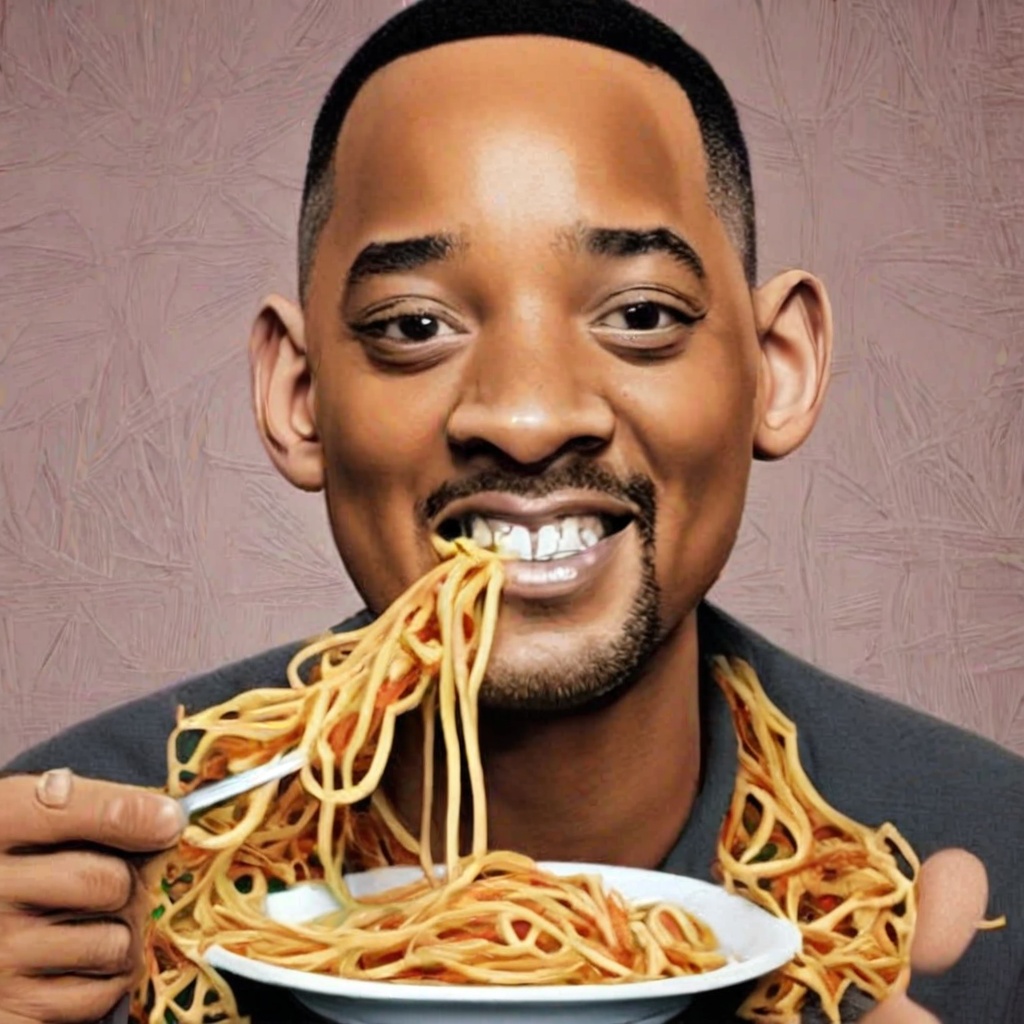} &
        
        \raisebox{0.065\textwidth}{\rotatebox[origin=t]{-90}{\scalebox{0.9}{\begin{tabular}{c@{}c@{}c@{}} Eating spaghetti \end{tabular}}}} \\

        \includegraphics[width=0.16\textwidth,height=0.16\textwidth]{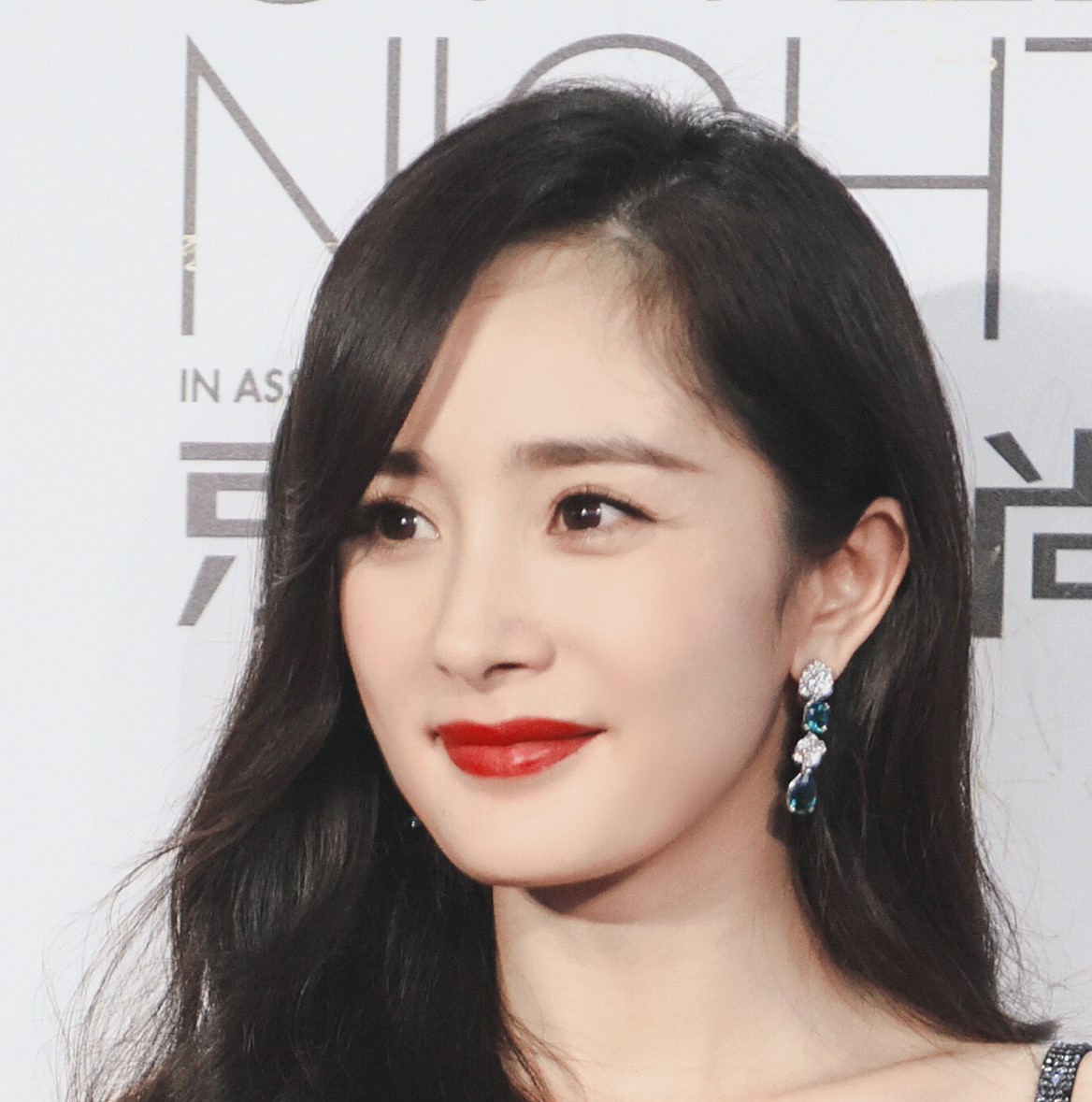} &

        \includegraphics[width=0.16\textwidth,height=0.16\textwidth]{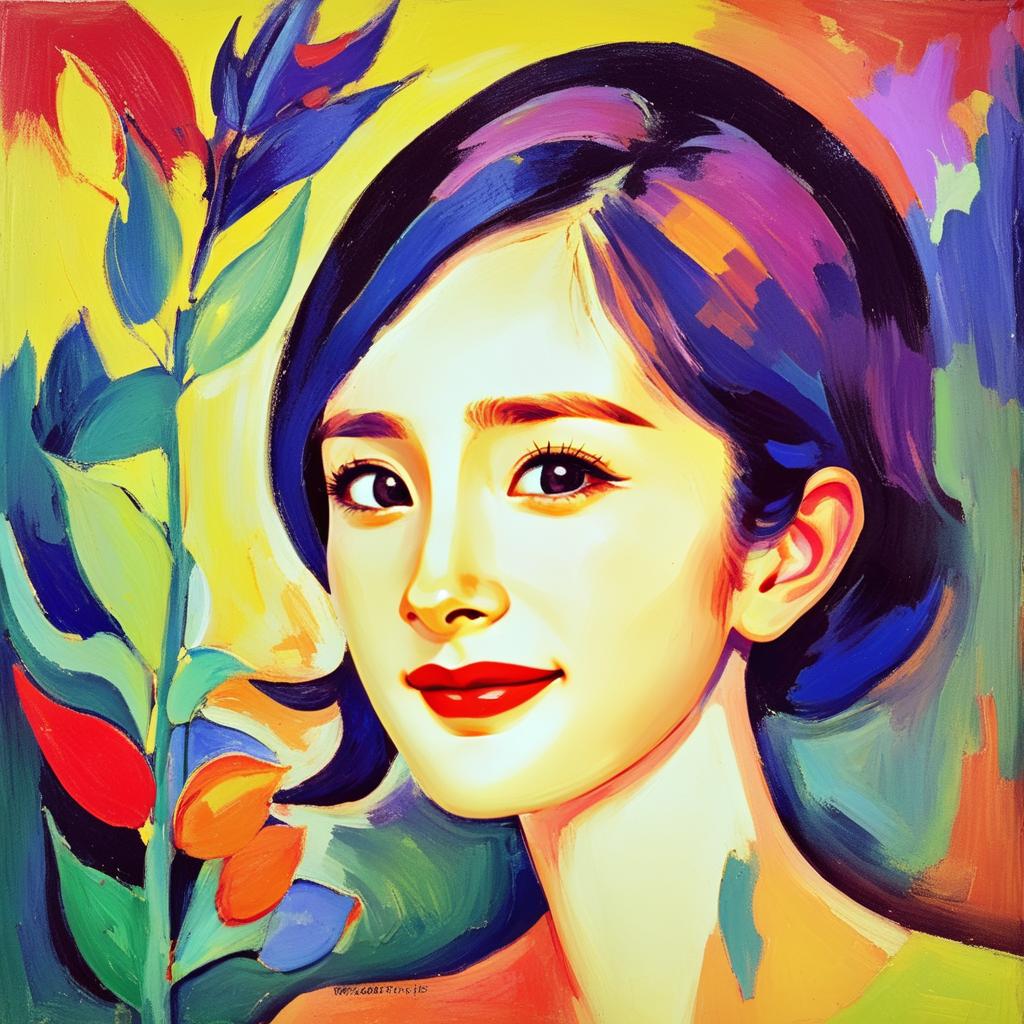} &
        \includegraphics[width=0.16\textwidth,height=0.16\textwidth]{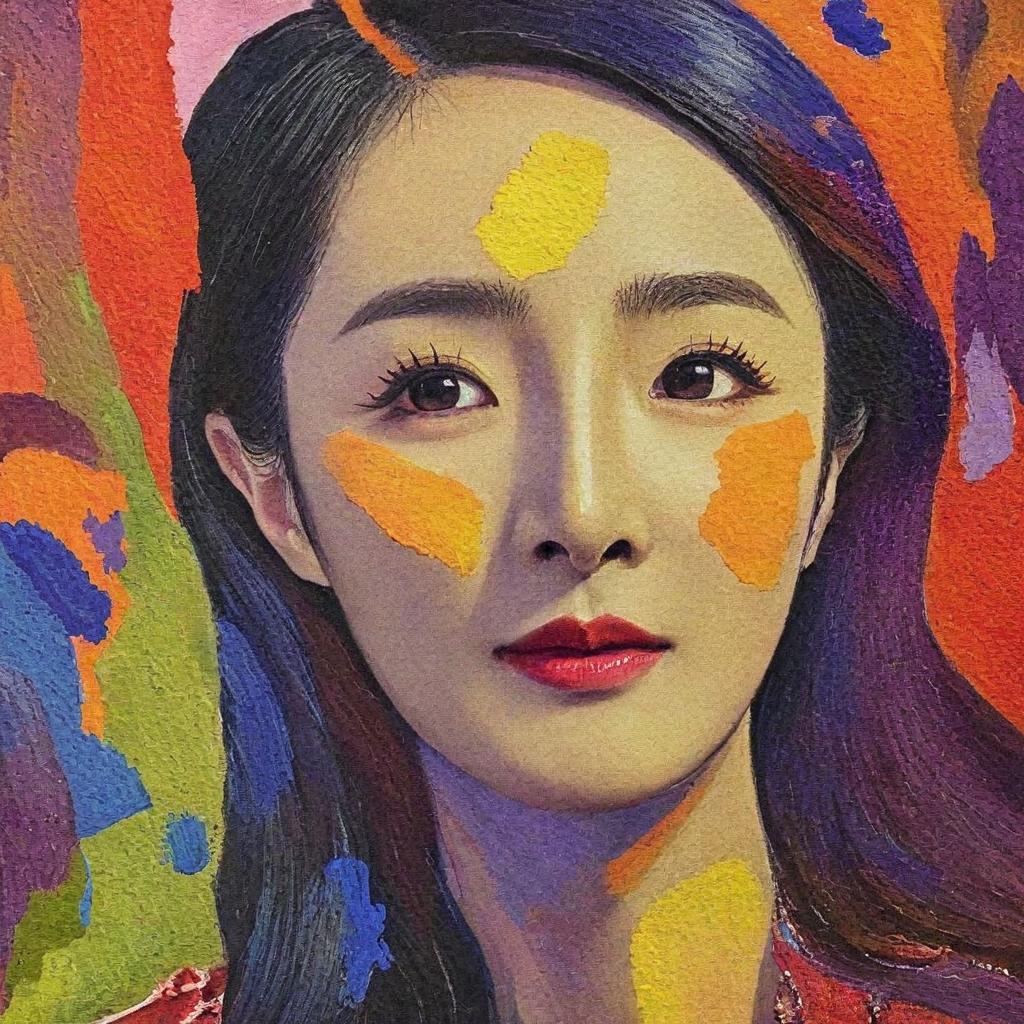} &
        \includegraphics[width=0.16\textwidth,height=0.16\textwidth]{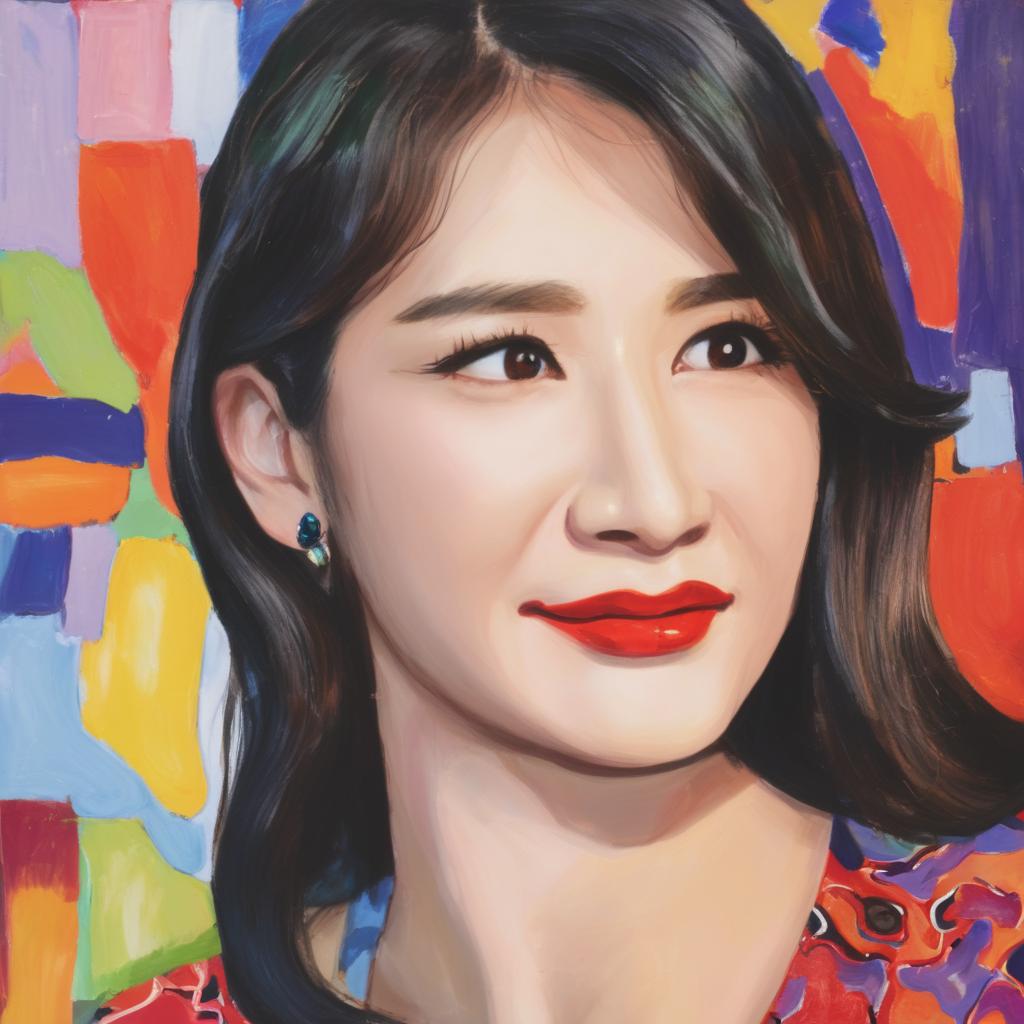} &
        \includegraphics[width=0.16\textwidth,height=0.16\textwidth]{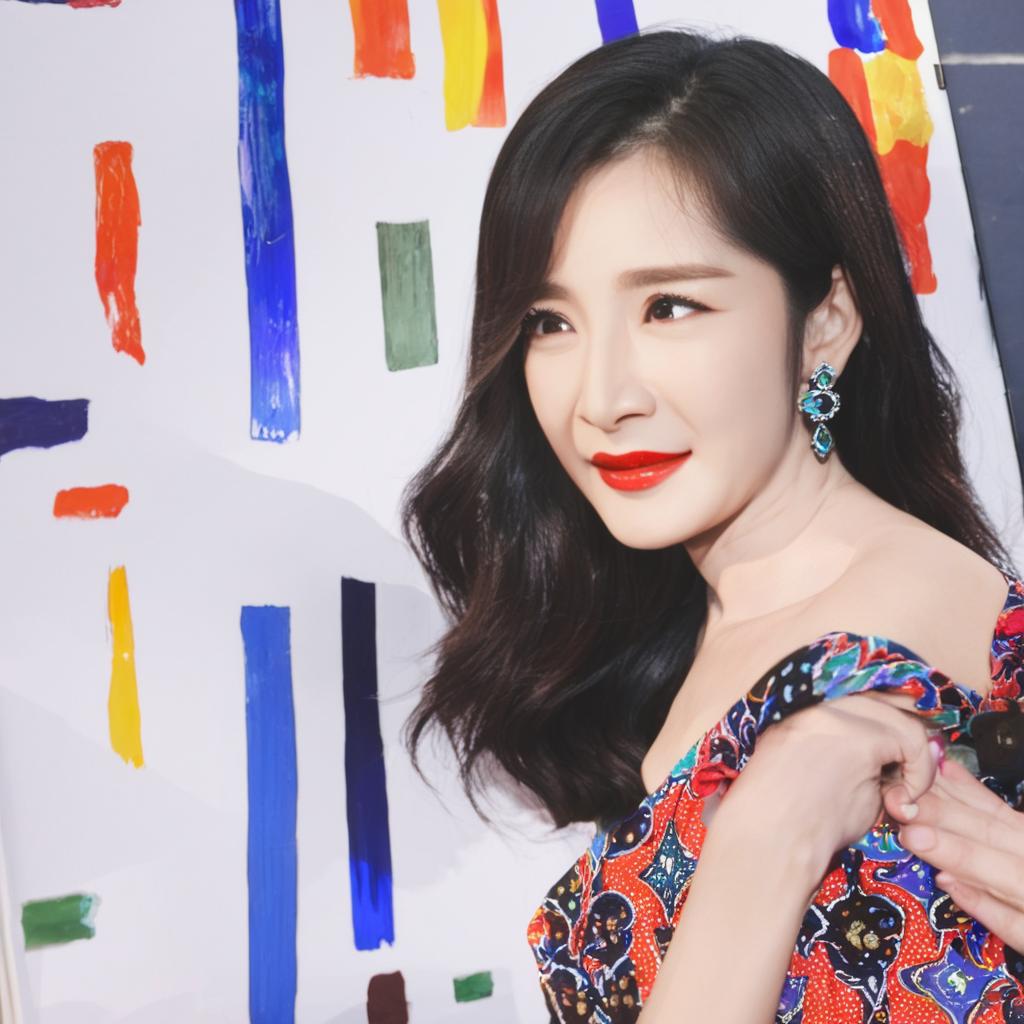} &
        \includegraphics[width=0.16\textwidth,height=0.16\textwidth]{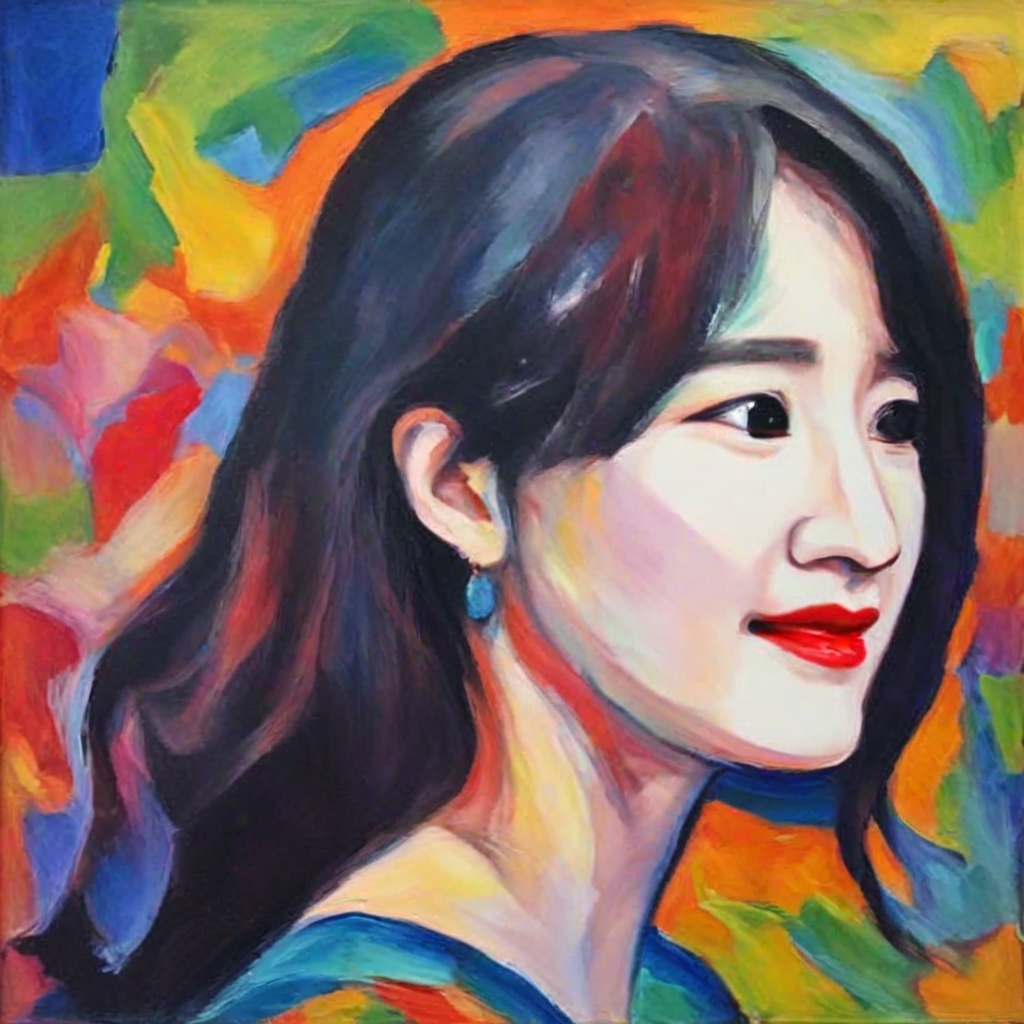} &
        
        \raisebox{0.068\textwidth}{\rotatebox[origin=t]{-90}{\scalebox{0.9}{\begin{tabular}{c@{}c@{}c@{}} Fauvism\end{tabular}}}} \\

        \includegraphics[width=0.16\textwidth,height=0.16\textwidth]{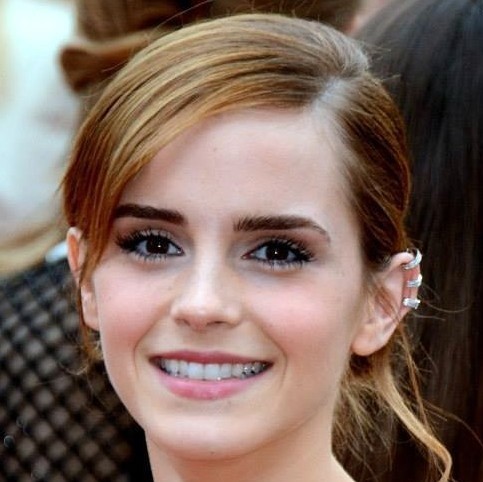} &

        \includegraphics[width=0.16\textwidth,height=0.16\textwidth]{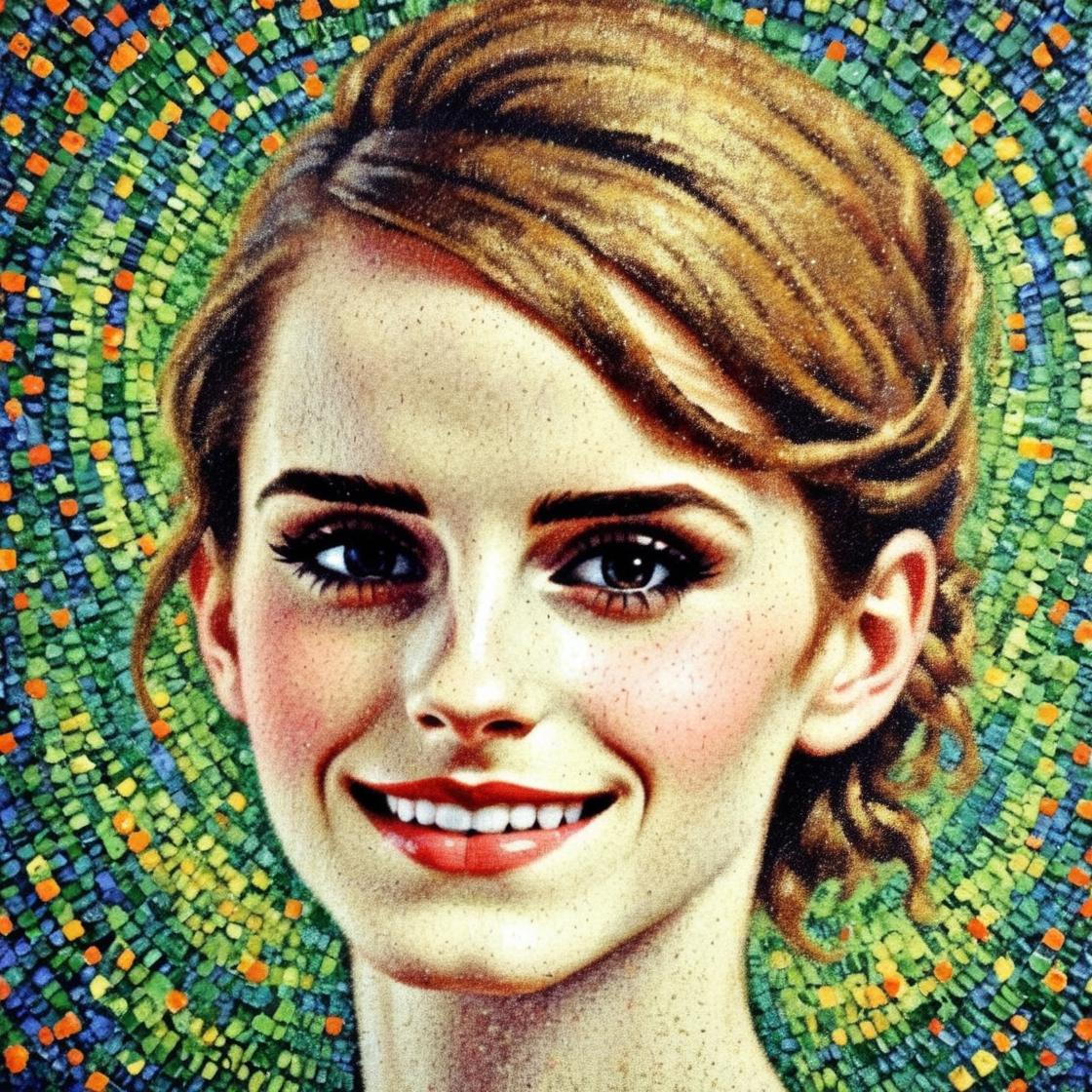} &
        \includegraphics[width=0.16\textwidth,height=0.16\textwidth]{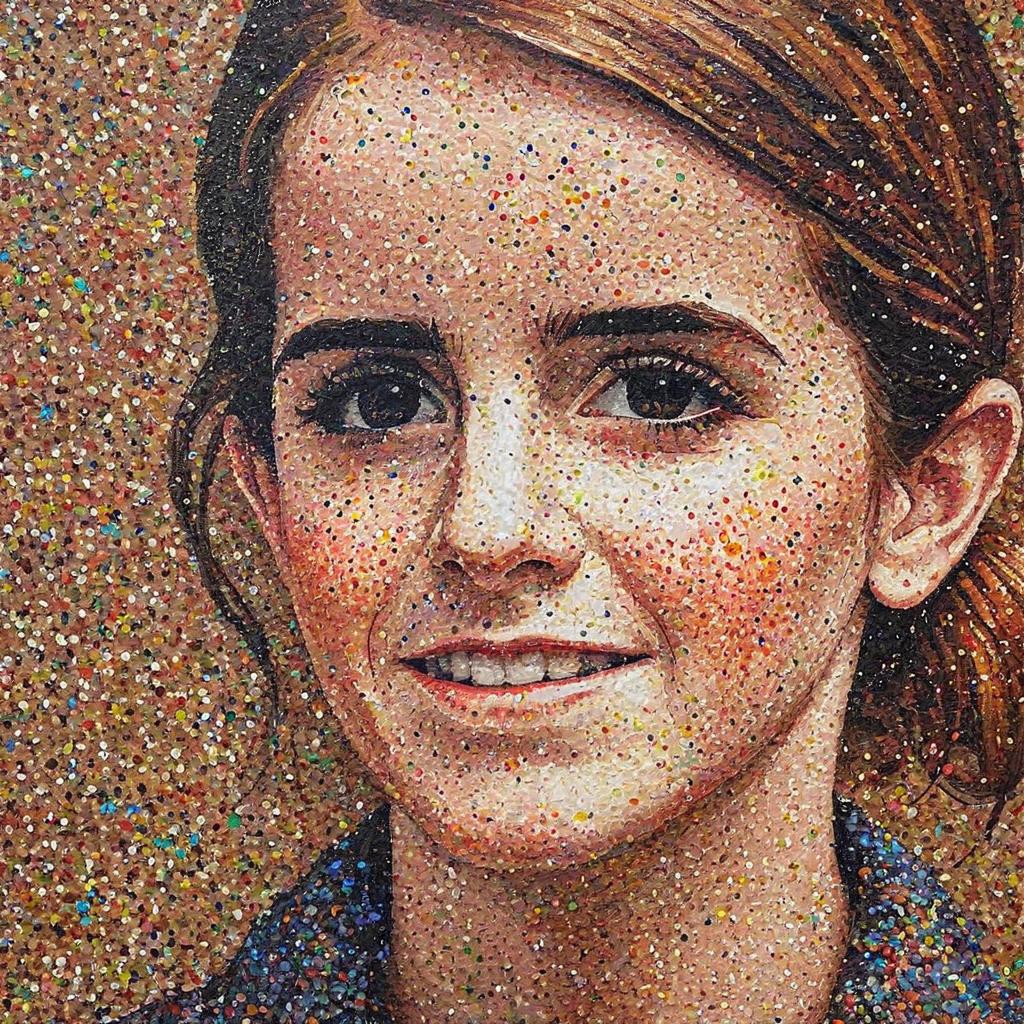} &
        \includegraphics[width=0.16\textwidth,height=0.16\textwidth]{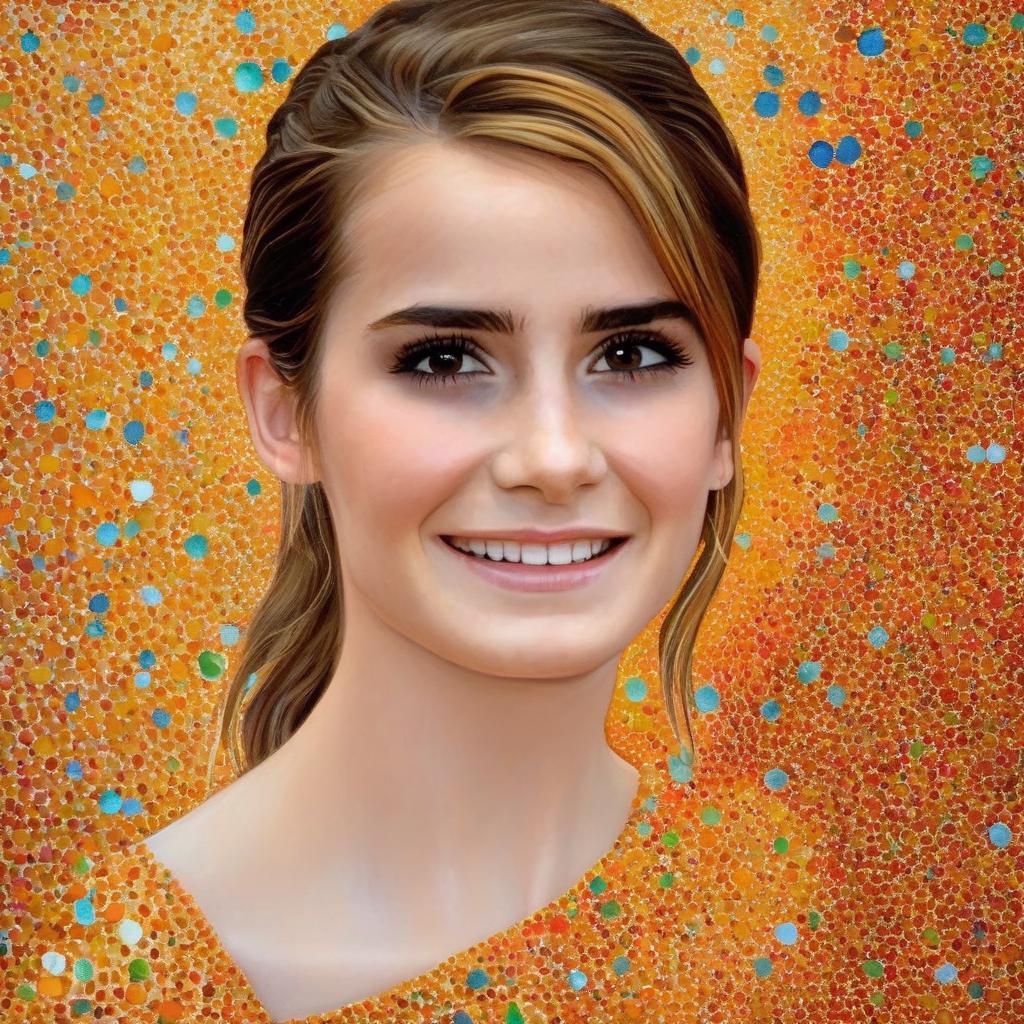} &
        \includegraphics[width=0.16\textwidth,height=0.16\textwidth]{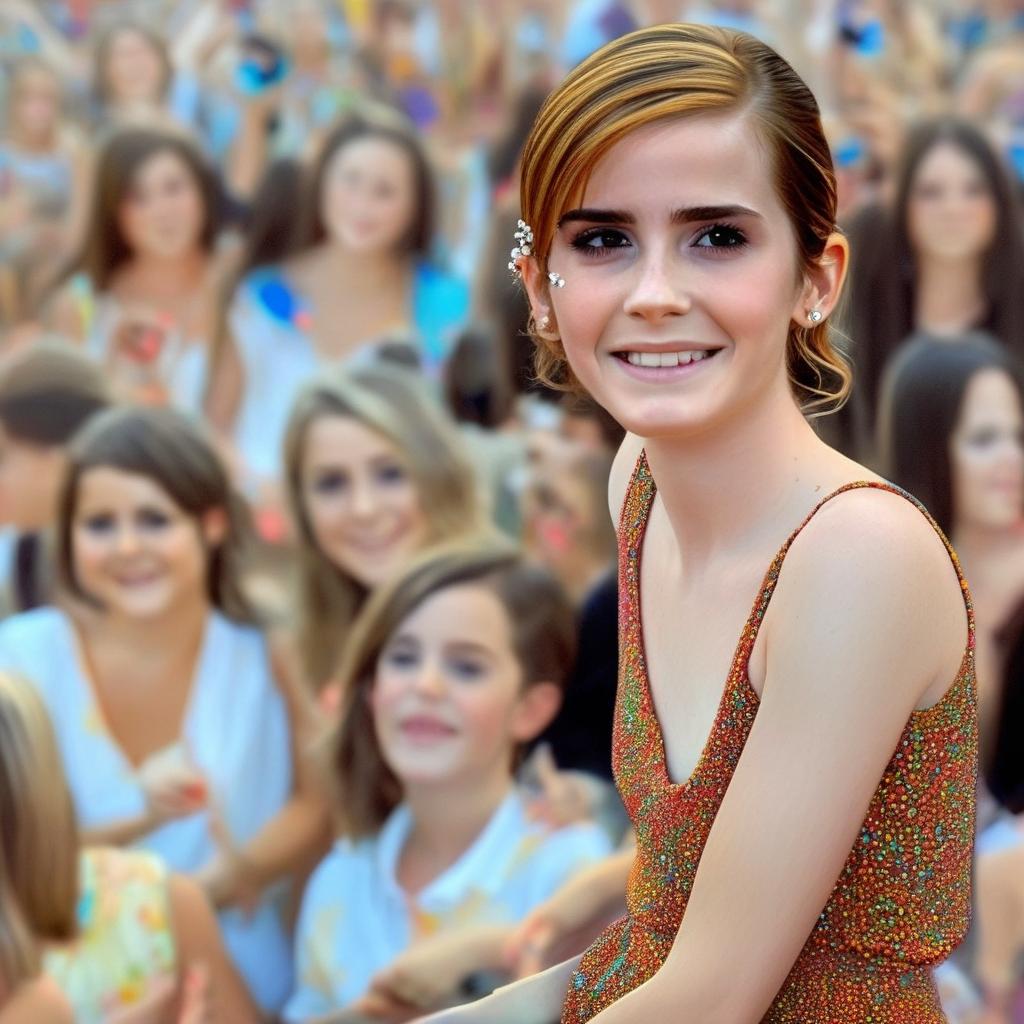} &
        \includegraphics[width=0.16\textwidth,height=0.16\textwidth]{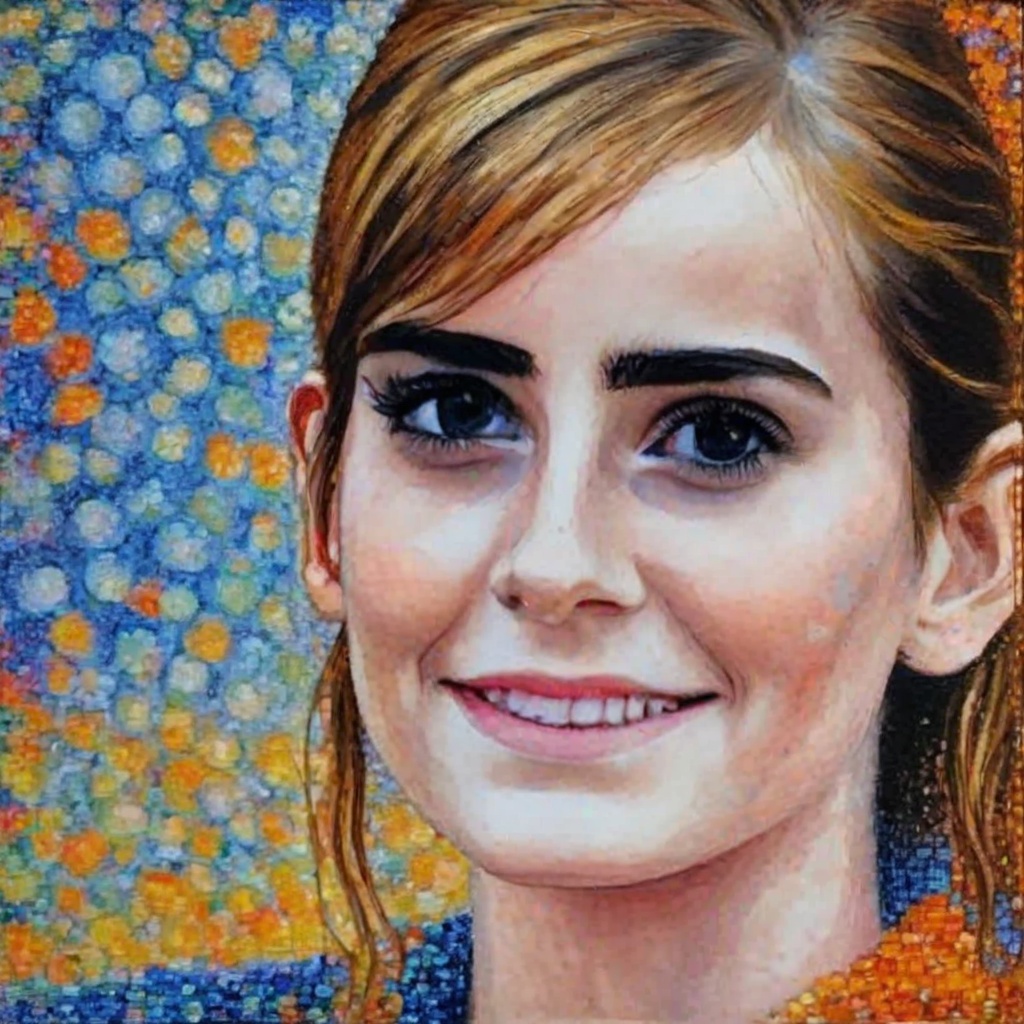} &
        
        \raisebox{0.065\textwidth}{\rotatebox[origin=t]{-90}{\scalebox{0.9}{\begin{tabular}{c@{}c@{}c@{}} Pointillism\end{tabular}}}} \\

    \end{tabular}
    
    }
    \caption{Comparisons against prior and concurrent face-personalization encoders on celebrity data. IP-A ($1.0$) and ($0.5$) represent the IP-Adapter results with a scale of $1.0$ and $0.5$, respectively. IP-A ($1.0$) serves as the backbone which we fine-tune.}
    \label{fig:celeb_comp}
\end{figure*}

In \cref{fig:celeb_comp} we show additional comparisons against the baseline methods, using celebrity inputs. Most baselines succeed in preserving the identities of all celebrities. The baseline IP-Adapter~\cite{ye2023ipadapter} variants still struggle with stylization prompts, and show background leak (though the latter can be fixed through appropriate background masking). Our results show good identity preservation while providing high editability.

\subsection{Comparisons on InstantID paper results}

\begin{figure*}[!th]

    \includegraphics[width=0.98\textwidth]{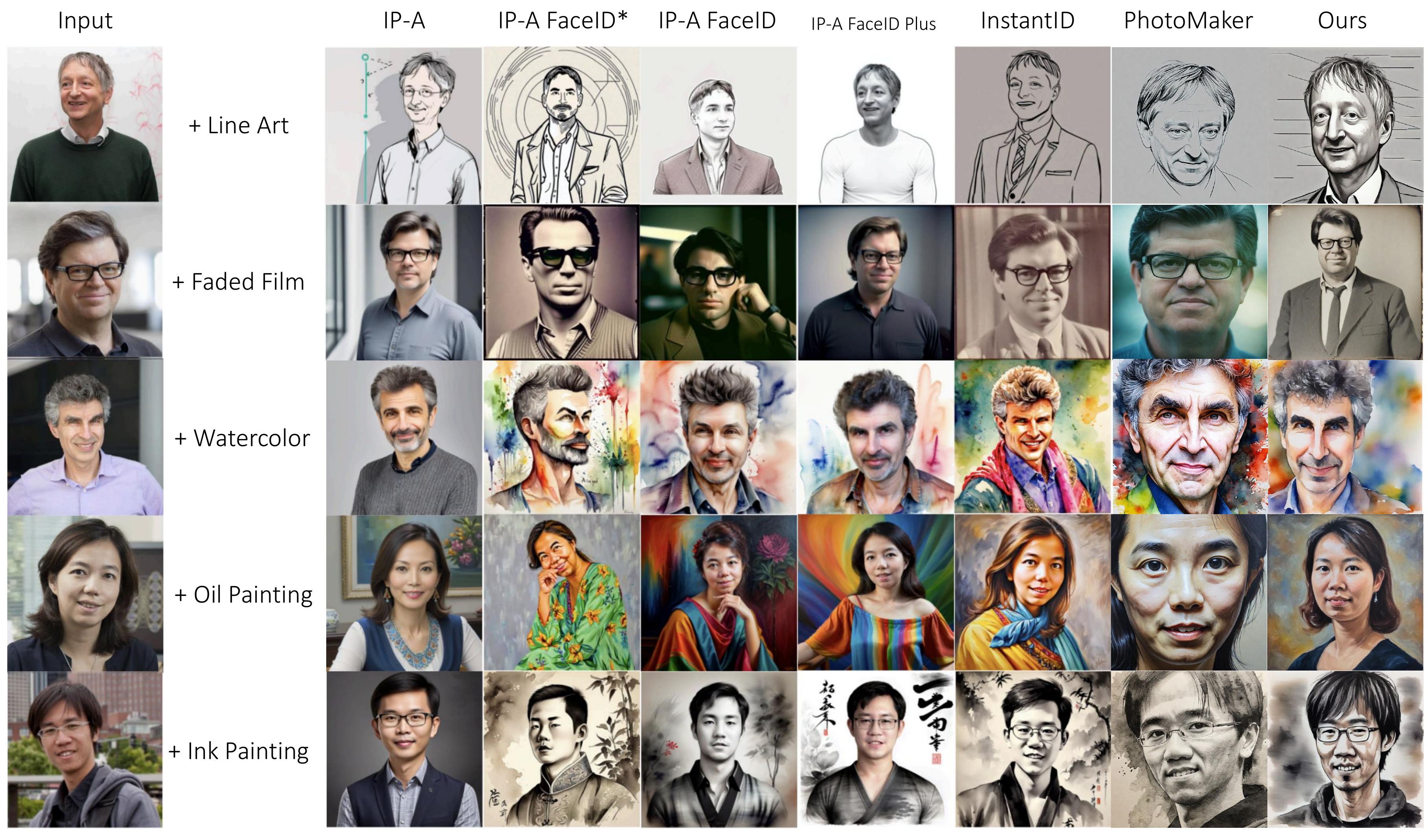}

    \caption{We expand Fig. 5 of the InstantID~\cite{wang2024instantid} paper with PhotoMaker~\cite{li2023photomaker} and our own results. Here, we keep the column terminology employed by the original InstantID paper. Hence, IP-A refers to IP-Adapter-SDXL, IP-A FaceID* is the experimental version of IP-Adapter-SDXL-FaceID. IP-A FaceID is the IP-Adapter-SD1.5-FaceID model, and IP-A FaceID Plus is the IP-Adapter-SD1.5-FaceID-Plus model. Note that the last two models are based on SD1.5 and not on SDXL.}\label{fig:iid_comp}

\end{figure*}

In \cref{fig:iid_comp} we expand Fig. 5 of the InstantID paper~\cite{wang2024instantid} with the results of PhotoMaker~\cite{li2023photomaker} and our own method. 
Here too, our method shows improvement over the prior art. While we were unable to verify this due to the LAION~\cite{schuhmann2021laion} dataset being withdrawn, a search through \url{https://haveibeentrained.com/} indicates that some of these individuals (e.g., Yann Lecun) may be included dozens of times in InstantID's reported training set. IP-Adapter did not report their training set for their Face ID versions. However, their diminished performance on these individuals hints that they did not observe them.

\section{Maintaining model alignment}

\begin{table}
\caption{Comparison of different alignment preservation approaches.}\label{tab:alignment_comp}
\begin{tabular}{l : c c : c c}
& \multicolumn{2}{c}{FFHQ-5000} & \multicolumn{2}{c}{Unsplash-50} \\
& ID $\uparrow$ & CLIP-T $\uparrow$ & ID $\uparrow$ & CLIP-T $\uparrow$ \\
\toprule
LoRA Scaling & 0.345 & 26.33 & 0.308 & 26.79 \\
LCM Loss & 0.324 & 26.63 & 0.280 & 27.05 \\
SDS & 0.267 & 27.03 & 0.232 & 27.16 \\
No Alignment & 0.246 & 27.46 & 0.220 & 27.48 \\
\end{tabular}
\end{table}

\begin{table}
\caption{Comparison of real and synthetic data generation options}\label{tab:dataset_comp}
\begin{tabular}{l : c c : c c}
& \multicolumn{2}{c}{FFHQ-5000} & \multicolumn{2}{c}{Unsplash-50} \\
& ID $\uparrow$ & CLIP-T $\uparrow$ & ID $\uparrow$ & CLIP-T $\uparrow$ \\
\toprule
SDXL Turbo & 0.345 & 26.33 & 0.308 & 26.79 \\
Synthetic Celebs & 0.334 & 26.37 & 0.299 & 26.81 \\
ConsiStory & 0.358 & 24.87 & 0.303 & 25.44 \\
CelebA & 0.345 & 25.81 & 0.294 & 26.14 \\
\end{tabular}
\end{table}

In the core paper, we note that applying the LCM-lookhead loss naively over extended training leads to a breaking of the alignment between the LCM~\cite{luo2023lcmlora} and non-LCM models. To avoid this, we investigated three options for preserving model alignment: (1) Applying a score distillation sampling (SDS) loss~\cite{poole2023dreamfusion} to the LCM-model outputs, where we use the non-LCM model to estimate the score function. (2) Adding the standard Consistency Model~\cite{song2023consistency} loss to the objective of the LCM-path's outputs, and (3) scaling the LCM-LoRA weights during training. Finally, we also investigate the results when avoiding any alignment-preserving mechanism.

The results are provided in \cref{tab:alignment_comp}. Notably, without using any alignment-preserving objective, our method shows only mild improvement over not using an identity loss at all. Introducing the SDS objective improves matters, but still under-performs the alternatives, including simply passing the identity loss through the single-step DDPM~\cite{ho2020denoising} approximation. Finally, using a consistency loss or the LoRA scaling approach leads to the best alignment preservation and hence the best downstream performance. Of the two, we settled with LoRA scaling because it provides improved results, is much simpler to implement in practice, and has negligible impact on compute requirements. 

\section{Choice of synthetic data}

Here we report evaluation results for models trained on different datasets or synthetic data generation options. Specifically, we consider: (1) Synthetic identities created by leveraging SDXL Turbo's~\cite{sauer2023adversarial} mode collapse. (2) Celebrities created using SDXL's~\cite{podell2024sdxl} prior knowledge. (3) ConsiStory~\cite{tewel2024trainingfree}, a training-free consistent subject generation method built on SDXL, and (4) the CelebA dataset~\cite{liu2015faceattributes}.
The results on FFHQ~\cite{karras2019style} and Unsplash-50 are shown in ~\cref{tab:dataset_comp}.

Notably, the CelebA dataset and the ConsiStory results contain only photo-realistic images (the latter because of its limitation in changing styles across the batch). This leads to diminished prompt-alignment, showing the importance of training on data for which the encoder is not pushed to adhere to a fixed output style.

\section{Limitations}

\begin{figure*}[t]
    \centering
    \begin{tabular}{c c c}
         Accessory preservation & Unintended stylization & Rare concepts \\
         \includegraphics[width=0.32\textwidth,height=0.16\textwidth]{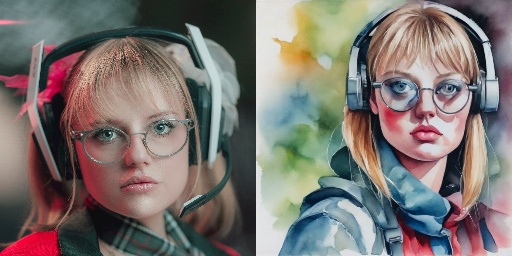} & \includegraphics[width=0.32\textwidth,height=0.16\textwidth]{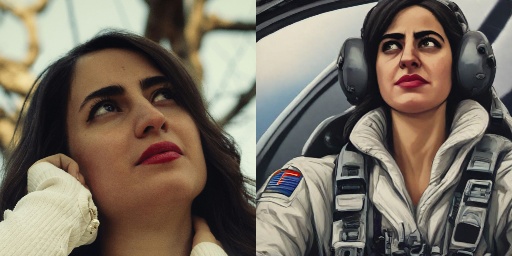} & \includegraphics[width=0.32\textwidth,height=0.16\textwidth]{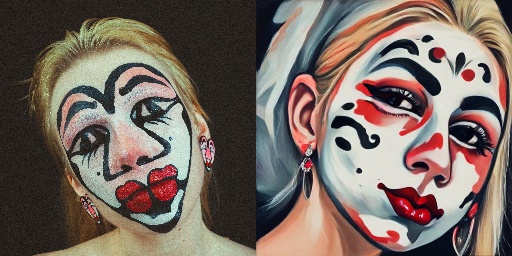} \\ 
         ``Watercolor Painting" & ``Jet Pilot" & ``Digital Art"
    \end{tabular}
    \caption{Limitations: \textbf{(left)} Our model captures accessories such as hats or headphones as part of the character. However, it does not accurately reproduce them in novel prompts. \textbf{(middle)} Our model may add stylization to output images even when not prompted for it. \textbf{(right)} The model fails to preserve tail concepts, such as excessive makeup.}\label{fig:limitations}
\end{figure*}

As an encoder, our model must learn to generalize from its training data. This places limits on its ability to adapt to concepts which were sufficiently rare (or entirely unseen) during training, such as unusual makeup (\cref{fig:limitations}, right). 

Additionally, our model posses unique limitations that may not be shared by prior art. Our training data contains a higher portion of non photo-realistic samples. Hence, the model may default to stylized results more often than the baseline (\cref{fig:limitations}, middle).

Finally, the model attempts to capture non-identity data from the image, such as accessories. Hence, providing it with a photo of a person wearing headphones will drive it to generate more photos with headphones. However, the model struggles to preserve the exact details of such accessories (\cref{fig:limitations}, left). 

\section{Additional implementation details}

\subsection{Classifier Free Guidance details}

When using the full version of the model with attention-based key and value expansion, we find it useful to modify the classifier free guidance (CFG, ~\cite{ho2021classifier}) formulation to:

\begin{align}
    \epsilon = \epsilon_{uncond} &+ s_{no\mathrm{-}kv} \cdot \left(\epsilon_{no\mathrm{-}kv} - \epsilon_{uncond}\right) \nonumber \\ &+ s_{full} \cdot \left(\epsilon_{full} - \epsilon_{uncond}\right)  \\ &+ s_{kv} \cdot \left(\epsilon_{kv} - \epsilon_{uncond}\right),  \nonumber
\end{align}
where $\epsilon_{no\mathrm{-}kv}$ is the denoiser's prediction when the keys and values are not extended (\ie using $K_{z_{r,t}}^{l}$, $V_{z_{r,t}}^{l}$), and $\epsilon_{kv}$ is the denoiser's output when these keys and values are used, but all other conditioning codes (text, IP-Adapter) are set to their null value. $\epsilon_{full}$ includes all conditioning inputs, and $\epsilon_{uncond}$ includes null conditions for all. The scale parameters were set empirically to $s_{no\mathrm{-}kv} = 3.0$, $s_{full} = 2.0$, $s_{kv} = 2.0$.

To facilitate CFG, we follow IP-Adapter and continue dropping the adapter and text conditions in $10\%$ of iterations. For the expanded attention path, we draw inspiration from ConsiStory~\cite{tewel2023key} and randomly drop $5\%$ of keys and values at every iteration.
        
\subsection{Additional parameters}

When expanding the cross attention mechanism, we follow prior art~\cite{alaluf2023crossimage,tewel2024trainingfree} and enable FreeU~\cite{si2023freeu} at inference time. We use the following parameters: $b1=1.1$, $b2=1.1$, $s1=0.9$, $s2=0.2$. Without FreeU, blur or low resolution artifacts are more often inherited from the conditioning image.

\section{Evaluation details}

For all Unsplash-50 evaluations and figures in the core paper, we generated a single image of each combination of identity and prompt. In the quantitative evaluations, we compute the automatic metrics over the entire set. For the user study, we randomly sampled an identity and prompt combo for each question. For qualitative figures, here and in the core paper, we manually selected 14 of these identities. 

For FFHQ evaluations, generating images with all prompts for every identity would take weeks. Hence, we randomly sampled one prompt from our set for each identity, and calculated the metrics over all $5,000$ image-prompt pairs. The same image-prompt pairs were used for all baselines and ablations.

We used the following prompts, to ensure coverage of both background changes and re-contextualization (which prior methods handle well), and style modifications (which they struggle with):

\begin{itemize}
\item ``A photo of a face"
\item ``A pencil drawing of a face"
\item ``A face riding a bicycle"
\item ``A face as a Pixar character"
\item ``A face as a samurai in medieval japan"
\item ``An oil painting of a face"
\item ``A painting of a face in the style of Van Gogh"
\item ``A sculpture of a face"
\item ``A pop figure of a face"
\item ``A face on a billboard in times square"
\item ``A face in an astronaut suit"
\item ``A face piloting a fighter jet"
\item ``A face dressed as a superhero"
\item ``A face in papercraft style"
\item ``A digitial art painting of a face"
\item ``A cubism painting of a face"
\item ``A garden gnome of a face"
\item ``A grainy old time photo of a face"
\end{itemize}

For methods which require specific keywords in the prompt we replaced the subject word ``face" with an appropriate reference to the keyword (e.g. ``face img" for PhotoMaker).

\bibliographystyle{ACM-Reference-Format}
\bibliography{supplementary}